\documentclass[twocolumn]{svjour3}          
\smartqed  
\usepackage[usenames]{color}
\usepackage[numbers]{natbib}
\usepackage[table]{xcolor}
\usepackage{amssymb}
\usepackage{makecell}
\usepackage{multirow}
\usepackage{enumitem}
\usepackage{rotating}
\usepackage{array}
\usepackage{times}
\usepackage{graphicx}
 \usepackage{psfrag}
\usepackage{subfigure}
\usepackage{rotating}
\usepackage{url}
\usepackage{rotating, graphicx}
\usepackage{amsmath}
\usepackage{booktabs}
\usepackage{lscape}
\usepackage{verbatim}
\usepackage{geometry}
\usepackage[misc]{ifsym}
\papertype{generic article}
\usepackage{breakurl}
\usepackage{graphicx}
\usepackage[pagebackref=true,breaklinks=true,linkcolor=red,anchorcolor=blue, citecolor=green,letterpaper=true,colorlinks,bookmarks=false]{hyperref}
\definecolor{DarkBlue}{rgb}{0.0,0.08,0.6}
\definecolor{DarkRed}{rgb}{0.6,0.00,0.08}
\definecolor{DarkGreen}{rgb}{0.0,0.6,0.08}
\definecolor{LightBlue}{rgb}{0.88,0.92,0.95}
\definecolor{Orange}{rgb}{1,0.75,0}

\begin{document}

\title{Deep Learning for Generic Object Detection: A Survey}

\subtitle{}

\author{Li Liu $^{1,2}$  \and
        Wanli Ouyang $^{3}$ \and
        Xiaogang Wang $^{4}$ \and \\
        Paul Fieguth $^{5}$ \and
        Jie Chen $^{2}$ \and
        Xinwang Liu $^{1}$ \and
        Matti Pietik\"{a}inen $^{2}$
}

\authorrunning{Li Liu \emph{et al.}} 

\institute{ $\textrm{\Letter}$ Li Liu (li.liu@oulu.fi) \\
        Wanli Ouyang (wanli.ouyang@sydney.edu.au) \\
        Xiaogang Wang (xgwang@ee.cuhk.edu.hk) \\
        Paul Fieguth (pfieguth@uwaterloo.ca) \\
        Jie Chen (jie.chen@oulu.fi) \\
        Xinwang Liu (xinwangliu@nudt.edu.cn) \\
        Matti Pietik\"{a}inen (matti.pietikainen@oulu.fi) \\
       1 National University of Defense Technology, China \\
       2 University of Oulu, Finland \\
       3 University of Sydney, Australia \\
       4 Chinese University of Hong Kong, China \\
       5 University of Waterloo, Canada \\
 }

 \date{Received: 12 September 2018 }

\maketitle

\begin{abstract}
Object detection, one of the most fundamental and challenging problems in computer vision, seeks to locate object instances from a large number of predefined categories in natural images. Deep learning techniques have emerged as a powerful strategy for learning feature representations directly from data and have led to remarkable breakthroughs in the field of generic object detection. Given this period of rapid evolution, the goal of this paper is to provide a comprehensive survey of the recent achievements in this field brought about by deep learning techniques. More than 300 research contributions are included in this survey, covering many aspects of generic object detection:  detection frameworks, object feature representation, object proposal generation, context modeling, training strategies, and evaluation metrics.  We finish the survey by identifying promising directions for future research.


\end{abstract}

\section{Introduction}
\label{sec:intro}

\begin {figure}[!t]
\centering
\includegraphics[width=0.48\textwidth]{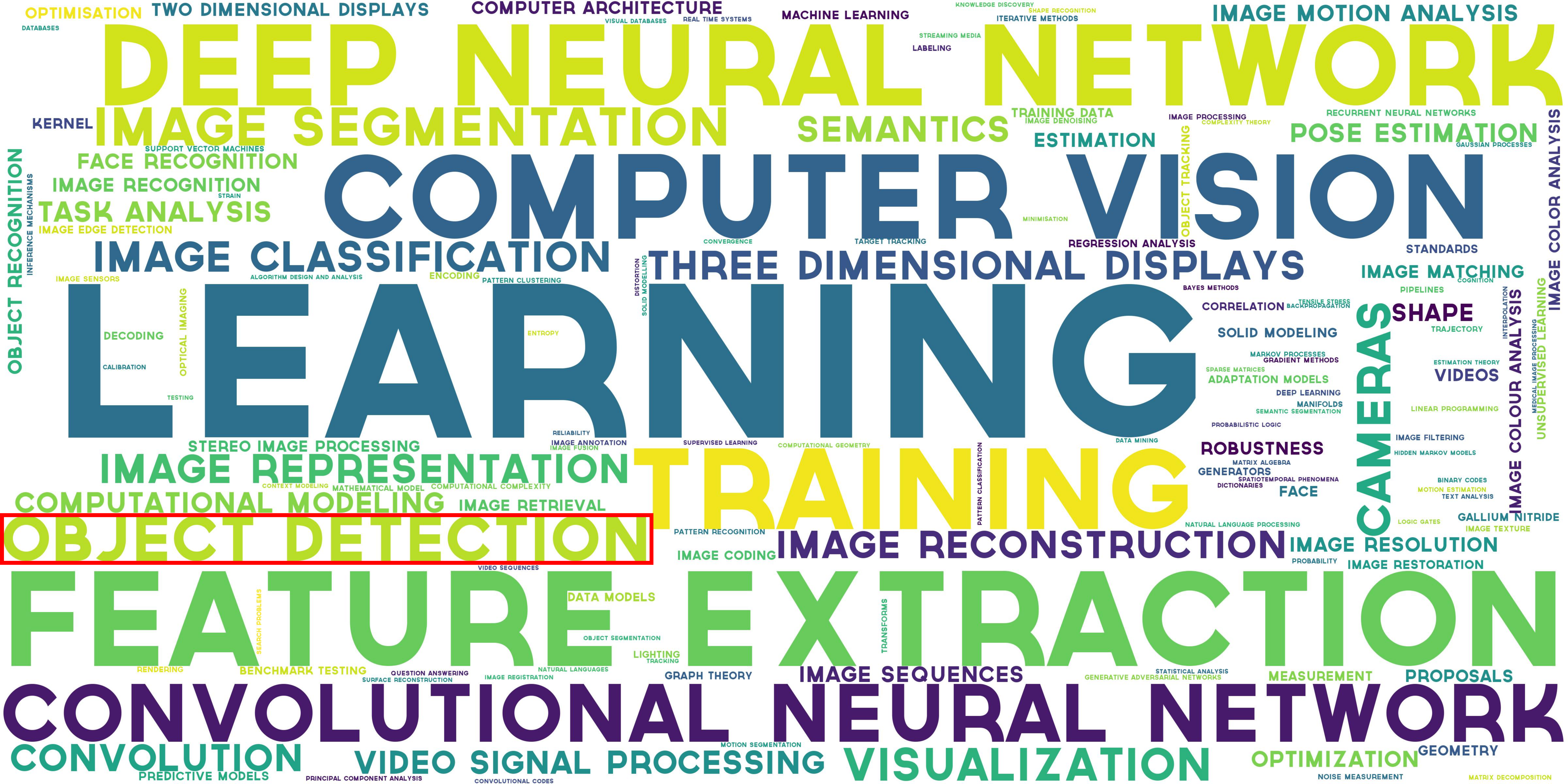}
\caption{Most frequent keywords in ICCV and CVPR conference papers from 2016 to 2018. The size of each word is proportional to the frequency of that keyword. We can see that object detection has received significant attention in recent years.}
\label{Fig:conferencekeywords}
\end {figure}

As a longstanding, fundamental and challenging problem in computer vision, object detection (illustrated in Fig.~\ref{Fig:conferencekeywords}) has been an active area of research for several decades \cite{Fischler1973}. The goal of object detection is to determine whether there are any instances of objects from given categories (such as humans, cars, bicycles, dogs or cats) in an image and, if present, to return the spatial location and extent of each object instance (\emph{e.g.,} via a bounding box \cite{Everingham2010,Russakovsky2015}). As the cornerstone of image understanding and computer vision, object detection forms the basis for solving complex or high level vision tasks such as segmentation, scene understanding, object tracking, image captioning, event detection, and activity recognition. Object detection supports a wide range of applications, including robot vision, consumer electronics, security, autonomous driving, human computer interaction, content based image retrieval, intelligent video surveillance, and augmented reality.

\begin {figure}[!t]
\centering
\includegraphics[width=0.5\textwidth]{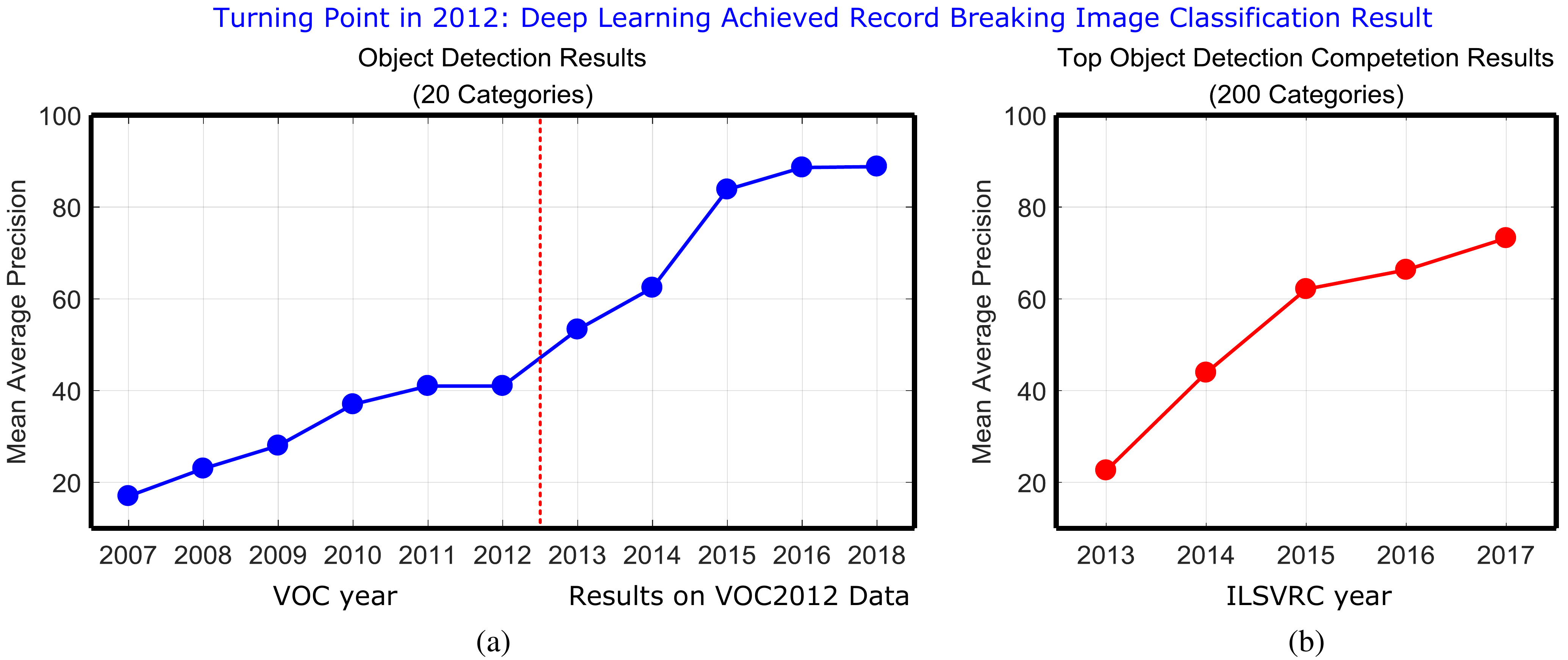}
\caption{An overview of recent object detection performance: We can observe a
significant improvement in performance (measured as mean average precision) since the arrival of deep learning in 2012. (a) Detection results of winning entries in the VOC2007-2012 competitions, and (b) Top object detection competition results in ILSVRC2013-2017 (results in both panels use only the provided training data).}
\label{fig:GODResultsStatistics}
\end {figure}

Recently, deep learning techniques \cite{Hinton2006Reducing,LeCun15} have emerged as powerful methods for learning feature representations automatically from data.  In particular, these techniques have provided major improvements in object detection, as illustrated in Fig.~\ref{fig:GODResultsStatistics}.

\begin {figure}[!t]
\centering
\includegraphics[width=0.45\textwidth]{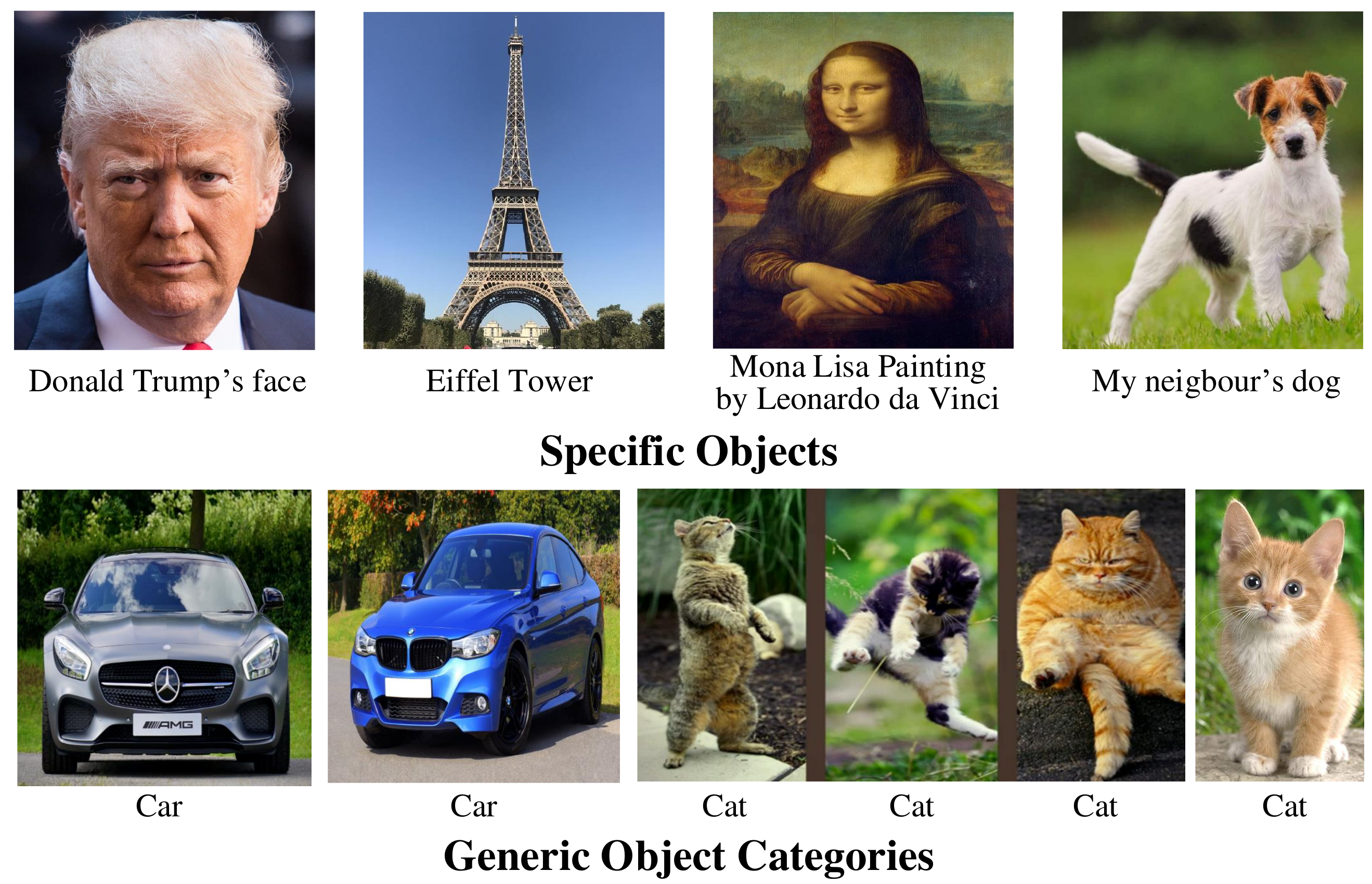}
\caption{Object detection includes localizing instances of a {\em particular} object (top), as well as generalizing to detecting object {\em categories} in general (bottom). This survey focuses on recent advances for the latter problem of generic object detection.}
\label{fig:ObjectInstancevsCategory}
\end {figure}

As illustrated in Fig.~\ref{fig:ObjectInstancevsCategory}, object detection can be grouped into one of two types \cite{Grauman2011Visual,Zhang13}:
detection of specific instances versus the detection of broad categories. The first type aims to detect instances of a particular object (such as Donald Trump's face, the Eiffel Tower, or a neighbor's dog), essentially a matching problem. The goal of the second type is to detect (usually previously unseen) instances of some predefined object categories (for example humans, cars, bicycles, and dogs). Historically, much of the effort in the field of object detection has focused on the detection of a single category (typically faces and pedestrians) or a few specific categories. In contrast, over the past several years, the research community has started moving towards the more challenging goal of building general purpose object detection systems where the breadth of object detection ability rivals that of humans.

\begin {figure*}[!t]
\centering
\includegraphics[width=0.95\textwidth]{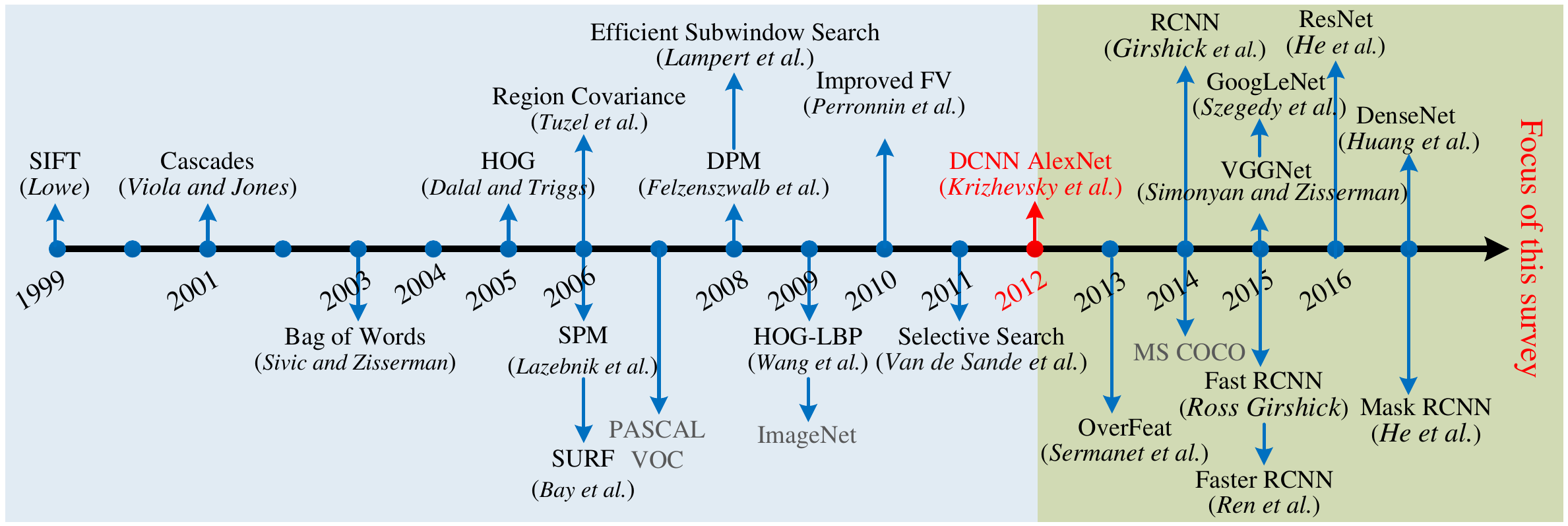}
\caption{Milestones of object detection and recognition, including feature representations \cite{Csurka2004,Dalal2005HOG,He2016ResNet,Krizhevsky2012,Lazebnik2006SPM,Lowe1999Object,Lowe2004,
Perronnin2010,Simonyan2014VGG,Sivic2003,GoogLeNet2015,Viola2001,HOGLBP2009}, detection frameworks \cite{Felzenszwalb2010b,Girshick2014RCNN,OverFeat2014,Uijlings2013b,Viola2001}, and datasets  \cite{Everingham2010,Lin2014,Russakovsky2015}.
The time period up to 2012 is dominated by handcrafted features, a transition took place in 2012 with the development of DCNNs for image classification by Krizhevsky \emph{et al.} \cite{Krizhevsky2012}, with methods after 2012 dominated by related deep networks.  Mostof the listed methods are highly cited and won a major ICCV or CVPR prize. See Section~\ref{Sec:Progress} for details.}
\label{fig:milestones}
\end {figure*}

In 2012, Krizhevsky \emph{et al.} \cite{Krizhevsky2012} proposed a Deep Convolutional Neural Network (DCNN) called AlexNet which achieved record breaking image classification accuracy in the Large Scale Visual Recognition Challenge (ILSVRC) \cite{Russakovsky2015}. Since that time, the research focus in most aspects of computer vision has been specifically on deep learning methods, indeed including the domain of generic object detection \cite{Girshick2014RCNN,He2014SPP,Girshick2015FRCNN,OverFeat2014,Ren2016a}.  Although  tremendous progress has been achieved, illustrated in Fig.~\ref{fig:GODResultsStatistics},  we are unaware of comprehensive surveys of this subject over the past five years. Given the exceptionally rapid rate of progress,
this article attempts to track recent advances and summarize their achievements in order to gain a clearer picture of the current panorama in generic object detection.

\begin{table*}[!t]
\caption {Summary of related object detection surveys since 2000.}\label{Tab:Surveys}
\centering
\renewcommand{\arraystretch}{1.2}
\setlength\arrayrulewidth{0.2mm}
\setlength\tabcolsep{2pt}
\resizebox*{16cm}{!}{
\begin{tabular}{!{\vrule width1.2bp}c|p{6cm}<{\centering}|c|c|c|p{9cm}<{\centering}!{\vrule width1.2bp}}
\Xhline{1pt}
\footnotesize No.   & \footnotesize Survey Title  & \footnotesize Ref.	 & \footnotesize Year & \footnotesize Venue	 & \footnotesize Content  \\
\hline
\raisebox{-1.5ex}[0pt]{\footnotesize 1 }& \footnotesize Monocular Pedestrian Detection: Survey and Experiments  &\raisebox{-1.5ex}[0pt]{ \footnotesize \cite{Enzweiler2009Monocular} 	}
&\raisebox{-1.5ex}[0pt]{ \footnotesize 2009}
& \raisebox{-1.5ex}[0pt]{\footnotesize PAMI}	& \footnotesize An evaluation of three pedestrian detectors \\
\hline
\raisebox{-1.5ex}[0pt]{\footnotesize 2 }& \footnotesize 	 Survey of Pedestrian Detection for Advanced Driver Assistance Systems & \raisebox{-1.5ex}[0pt]{\footnotesize \cite{Geronimo2010Survey}}	 &\raisebox{-1.5ex}[0pt]{ \footnotesize  2010 }
& \raisebox{-1.5ex}[0pt]{\footnotesize PAMI }& \footnotesize \raisebox{-1.5ex}[0pt]{	 A survey of pedestrian detection for advanced driver assistance systems} \\
\hline
\raisebox{-1.5ex}[0pt]{\footnotesize 3} & \footnotesize	 Pedestrian Detection: An Evaluation of the State of The Art	 & \raisebox{-1.5ex}[0pt]{\footnotesize \cite{Dollar2012Pedestrian} }
 & \raisebox{-1.5ex}[0pt]{\footnotesize	2012	}& \raisebox{-1.5ex}[0pt]{\footnotesize PAMI }
 & \footnotesize	 A thorough and detailed evaluation of detectors in monocular images  \\
 \hline
 \footnotesize 4 & \footnotesize 	Detecting Faces in Images: A Survey	 & \footnotesize \cite{Yang2002b} & \footnotesize 2002 & \footnotesize PAMI & \footnotesize	 First survey of face detection from a single image   \\
\hline
\raisebox{-1.5ex}[0pt]{\footnotesize 5	} & \footnotesize A Survey on Face Detection in the
Wild: Past, Present and Future	& \raisebox{-1.5ex}[0pt]{\footnotesize \cite{Zafeiriou2015}}
 & \raisebox{-1.5ex}[0pt]{\footnotesize 2015 }& \raisebox{-1.5ex}[0pt]{ \footnotesize CVIU}& \raisebox{-1.5ex}[0pt]{ \footnotesize A survey of face detection in the wild since 2000} \\
\hline
\raisebox{-1.5ex}[0pt]{\footnotesize 6	}& \raisebox{-1.5ex}[0pt]{
\footnotesize On Road Vehicle Detection: A Review}	 &\raisebox{-1.5ex}[0pt]{ \footnotesize \cite{Sun2006Road} }&\raisebox{-1.5ex}[0pt]{ \footnotesize	2006	}&
\raisebox{-1.5ex}[0pt]{ \footnotesize PAMI}	& \footnotesize A review of vision based on-road vehicle detection systems  \\
 \hline
 \footnotesize 7	 & \footnotesize Text Detection and Recognition in Imagery: A Survey & \footnotesize \cite{Ye2015Text} & \footnotesize 2015	& \footnotesize PAMI	& \footnotesize A survey of text detection and
recognition in color imagery \\
 \hline
\raisebox{-1.5ex}[0pt]{ \footnotesize 8	 }&\raisebox{-1.5ex}[0pt]{
 \footnotesize Toward Category Level Object Recognition} & \raisebox{-1.5ex}[0pt]{\footnotesize \cite{Ponce2007Toward}} &\raisebox{-1.5ex}[0pt]{ \footnotesize 2007}
 &\raisebox{-1.5ex}[0pt]{ \footnotesize	Book  }	& \footnotesize Representative papers on object categorization, detection,
and segmentation  \\
 \hline
\raisebox{-1.5ex}[0pt]{ \footnotesize	9 }& \footnotesize The Evolution of Object Categorization and the Challenge of Image Abstraction	& \raisebox{-1.5ex}[0pt]{\footnotesize \cite{Dickinson2009}}
  &\raisebox{-1.5ex}[0pt]{ \footnotesize	 2009}&\raisebox{-1.5ex}[0pt]{ \footnotesize	 Book }  & \raisebox{-1.5ex}[0pt]{ \footnotesize	A trace of the evolution of object categorization over four decades}  \\
  \hline
\raisebox{-1.5ex}[0pt]{\footnotesize 10} &
\footnotesize 	Context based Object Categorization: A Critical Survey &\raisebox{-1.5ex}[0pt]{ \footnotesize \cite{Galleguillos2010}}	 &\raisebox{-1.5ex}[0pt]{ \footnotesize	2010}
& \raisebox{-1.5ex}[0pt]{\footnotesize 	CVIU}	& \footnotesize A review of contextual information
for object categorization  \\
 \hline
 \footnotesize 11	& \footnotesize 50 Years of Object Recognition: Directions
Forward	& \footnotesize \cite{Andreopoulos13} & \footnotesize	 2013& \footnotesize 	 CVIU	 & \footnotesize A review of the evolution of object recognition
systems over five decades  \\
 \hline
\raisebox{-1.5ex}[0pt]{ \footnotesize 12}	& \raisebox{-1.5ex}[0pt]{
\footnotesize Visual Object Recognition} &\raisebox{-1.5ex}[0pt]{
 \footnotesize \cite{Grauman2011Visual} } &\raisebox{-1.5ex}[0pt]{ \footnotesize2011}&\raisebox{-1.5ex}[0pt]{ \footnotesize Tutorial}
 	& \footnotesize  Instance and category object recognition techniques  \\
 \hline
 \footnotesize 13	& \footnotesize Object Class Detection: A Survey	& \footnotesize \cite{Zhang13} & \footnotesize	 2013	& \footnotesize ACM CS & \footnotesize	 Survey of generic object detection methods before 2011  \\
  \hline
\raisebox{-1.5ex}[0pt]{ \footnotesize 14}	& \footnotesize Feature Representation for Statistical
Learning based Object Detection: A Review	 &\raisebox{-1.5ex}[0pt]{ \footnotesize \cite{Li2015Feature}}
 &\raisebox{-1.5ex}[0pt]{ \footnotesize	2015} & \raisebox{-1.5ex}[0pt]{ \footnotesize	 PR} & \footnotesize	 Feature representation methods in statistical learning
based object detection, including handcrafted and deep learning based features  \\
 \hline
 \footnotesize 15	& \footnotesize Salient Object Detection: A Survey		 & \footnotesize \cite{Borji14} & \footnotesize2014	& \footnotesize arXiv	& \footnotesize A survey for salient object detection \\
  \hline
\raisebox{-1.5ex}[0pt]{ \footnotesize 16}& \footnotesize	 Representation Learning: A Review and New Perspectives	 & \raisebox{-1.5ex}[0pt]{ \footnotesize \cite{Bengio13Feature}}
&\raisebox{-1.5ex}[0pt]{ \footnotesize	2013}	 &\raisebox{-1.5ex}[0pt]{ \footnotesize PAMI}	 & \footnotesize Unsupervised feature learning and deep learning, probabilistic models, autoencoders, manifold learning, and deep networks  \\
 \hline
 \footnotesize 17 & \footnotesize	Deep Learning		 & \footnotesize \cite{LeCun15} & \footnotesize2015& \footnotesize	Nature	& \footnotesize An introduction to deep learning and applications  \\
  \hline
\raisebox{-1.5ex}[0pt]{ \footnotesize 18 }&
\footnotesize	A Survey on Deep Learning in Medical Image Analysis& \raisebox{-1.5ex}[0pt]{\footnotesize \cite{Litjens2017} }&\raisebox{-1.5ex}[0pt]{ \footnotesize	 2017}
 &\raisebox{-1.5ex}[0pt]{ \footnotesize 	MIA}	& \footnotesize A survey of deep learning for image classification, object detection, segmentation and registration in medical image analysis  \\
  \hline
\raisebox{-1.5ex}[0pt]{ \footnotesize 19} & \footnotesize	 Recent Advances in Convolutional Neural
Networks&\raisebox{-1.5ex}[0pt]{ \footnotesize \cite{Gu2015Recent}}
  &\raisebox{-1.5ex}[0pt]{ \footnotesize 2017}& \raisebox{-1.5ex}[0pt]{\footnotesize PR}& \footnotesize A broad survey of the recent advances in CNN and its applications in computer vision, speech and natural language processing  \\
 \hline
\footnotesize 20	& \footnotesize Tutorial: Tools for Efficient Object Detection	& \footnotesize $-$ & \footnotesize	 2015& \footnotesize 	 ICCV15& \footnotesize 	A short course for object detection only covering recent milestones \\
 \hline
\raisebox{-1.5ex}[0pt]{\footnotesize 21 }&\raisebox{-1.5ex}[0pt]{
 \footnotesize	Tutorial: Deep Learning for Objects and Scenes}	 &\raisebox{-1.5ex}[0pt]{
  \footnotesize $-$} & \raisebox{-1.5ex}[0pt]{ \footnotesize 	 2017}&\raisebox{-1.5ex}[0pt]{ \footnotesize	 CVPR17}
	& \footnotesize A high level summary of recent work on deep learning for visual recognition of objects and scenes  \\
 \hline
\raisebox{-1.5ex}[0pt]{\footnotesize 22} &\raisebox{-1.5ex}[0pt]{
 \footnotesize Tutorial: Instance Level Recognition}&\raisebox{-1.5ex}[0pt]{ \footnotesize $-$ }& \raisebox{-1.5ex}[0pt]{ \footnotesize	2017} &\raisebox{-1.5ex}[0pt]{ \footnotesize	 ICCV17}
& \footnotesize	A short course of recent advances on instance level recognition, including object detection, instance segmentation and human pose prediction \\
 \hline
\raisebox{-1.5ex}[0pt]{\footnotesize 23}&\raisebox{-1.5ex}[0pt]{
 \footnotesize Tutorial: Visual Recognition and Beyond} &\raisebox{-1.5ex}[0pt]{ \footnotesize $-$} & \raisebox{-1.5ex}[0pt]{\footnotesize	 2018}&\raisebox{-1.5ex}[0pt]{ \footnotesize	 CVPR18 }& \footnotesize	 A tutorial on methods and principles behind image classification, object detection, instance segmentation, and semantic segmentation. \\
 \hline
\footnotesize \textbf{24}& \footnotesize	 \textbf{Deep Learning for Generic Object Detection} & \footnotesize \textbf{Ours} & \footnotesize \textbf{2019}	& \footnotesize \textbf{VISI}	 & \footnotesize \textbf{A comprehensive survey of deep learning for generic object detection} \\
\Xhline{1pt}
\end{tabular}
}
\end{table*}

\subsection{Comparison with Previous Reviews}

Many notable object detection surveys have been published, as summarized in Table~\ref{Tab:Surveys}. These include many excellent surveys on the problem of {\em specific} object detection, such as pedestrian detection \cite{Enzweiler2009Monocular,Geronimo2010Survey,Dollar2012Pedestrian}, face detection \cite{Yang2002b,Zafeiriou2015}, vehicle detection \cite{Sun2006Road} and text detection \cite{Ye2015Text}.
There are comparatively few recent surveys focusing directly on the problem of generic object detection, except for the work by Zhang \emph{et al.} \cite{Zhang13} who conducted a survey on the topic of object class detection. However, the research reviewed in \cite{Grauman2011Visual}, \cite{Andreopoulos13} and \cite{Zhang13} is mostly pre-2012, and therefore prior to the recent striking success and dominance of deep learning and related methods.

Deep learning allows computational models to learn fantastically complex, subtle, and abstract representations, driving significant progress in a broad range of problems such as visual recognition, object detection, speech recognition, natural language processing, medical image analysis, drug discovery and genomics. Among different types of deep neural
networks, DCNNs \cite{LeCun1998Gradient,Krizhevsky2012,LeCun15} have brought about breakthroughs in processing images, video, speech and audio.  To be sure, there have been many published surveys on deep learning, including that of Bengio \emph{et al.} \cite{Bengio13Feature},  LeCun \emph{et al.} \cite{LeCun15}, Litjens \emph{et al.} \cite{Litjens2017}, Gu \emph{et al.} \cite{Gu2015Recent}, and more recently in tutorials at ICCV and CVPR.

In contrast, although many deep learning based methods have been proposed for object detection, we are unaware of any comprehensive recent survey.  A thorough review and summary of existing work is essential for further progress in object detection, particularly for researchers wishing to enter the field.  Since our focus is on {\em generic} object detection, the extensive work on DCNNs for {\em specific} object detection, such as face detection \cite{Li2015CasecadeCNN,Zhang2016Joint,Hu2017Finding}, pedestrian detection \cite{Zhang2016faster,Hosang2015taking}, vehicle detection \cite{Zhou2016dave} and traffic sign detection \cite{Zhu2016traffic} will not be considered.

\begin {figure}[!t]
\centering
\includegraphics[width=0.45\textwidth]{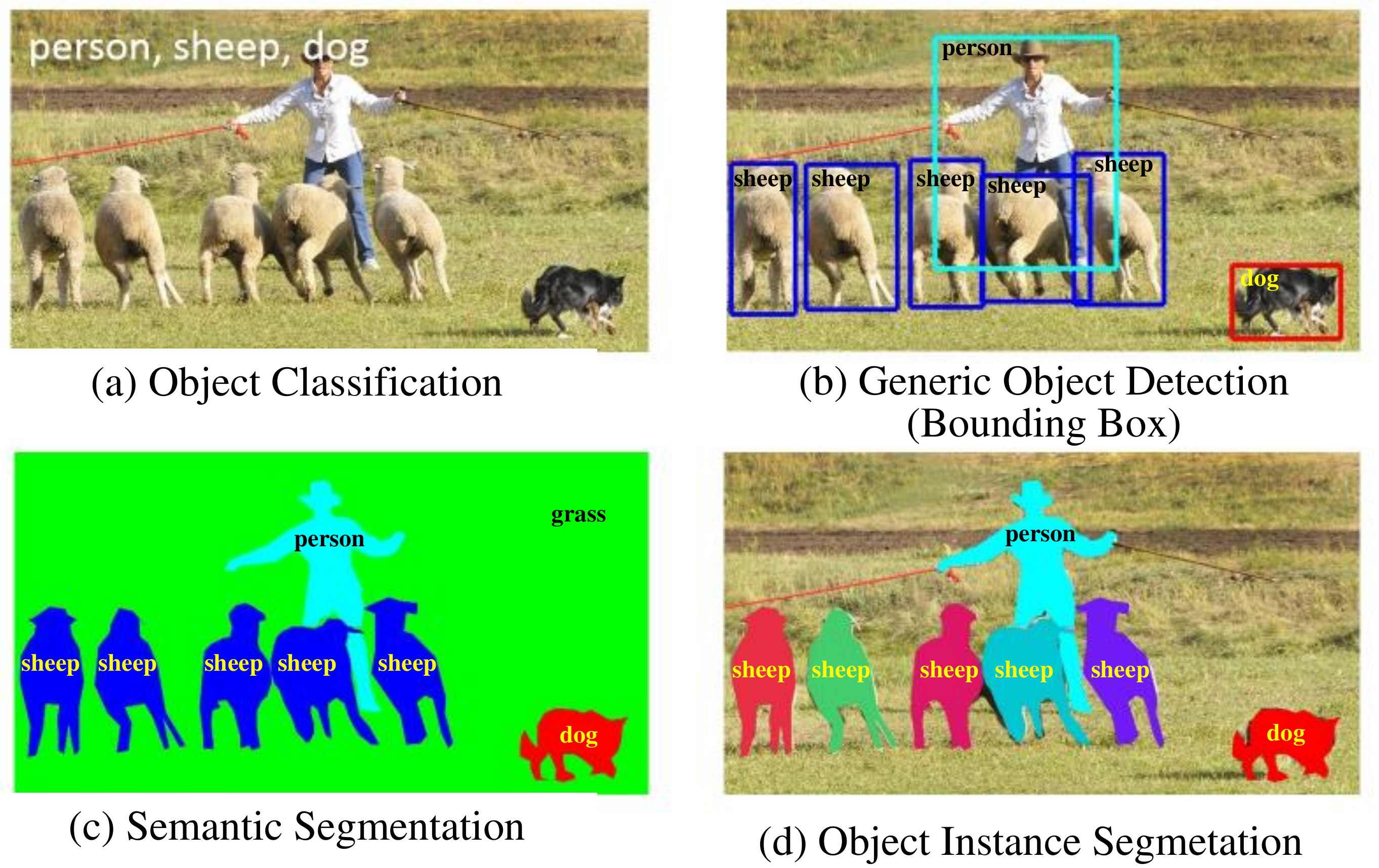}
\caption{Recognition problems related to generic object detection: (a) Image level object classification, (b) Bounding box level generic object detection, (c) Pixel-wise semantic segmentation, (d) Instance level semantic segmentation.}
\label{Fig:TheProblem}
\end {figure}

\subsection{Scope}

The number of papers on generic object detection based on deep learning is
breathtaking. There are so many, in fact, that compiling any comprehensive
review of the state of the art is beyond the scope of any reasonable length paper.  As a result, it is necessary to establish selection criteria, in such a way that we have limited our focus to top journal and conference papers. Due to these limitations, we sincerely apologize to those authors whose works are not included in this paper. For surveys of work on related topics, readers are referred to the articles in Table~\ref{Tab:Surveys}.
This survey focuses on major progress of the last five years, and we restrict our attention to still pictures, leaving the important subject of video
object detection as a topic for separate consideration in the future.

The main goal of this paper is to offer a comprehensive survey of deep learning based generic object detection techniques, and to present some degree of taxonomy, a high level perspective and organization, primarily on the basis of popular datasets,  evaluation metrics, context modeling, and detection proposal methods.  The intention is that our categorization be helpful for readers to have an accessible understanding of similarities and differences between a wide variety of strategies. The proposed taxonomy gives researchers
a framework to understand current research and to identify
open challenges for future research.

The remainder of this paper is organized as follows. Related
background and the progress
made during the last two decades are summarized in Section~\ref{Sec:Background}. A brief introduction to deep learning is given in Section \ref{Sec:CNNintro}. Popular datasets and evaluation criteria are summarized in Section \ref{Sec:DataEval}.
We describe the milestone object detection frameworks in Section~\ref{Sec:Frameworks}. From Section \ref{Sec:DCNNFeatures} to Section \ref{sec:otherissue}, fundamental sub-problems and the relevant issues involved in designing object detectors
are discussed. Finally, in Section \ref{Sec:Conclusions}, we conclude the paper with an overall discussion of object detection, state-of-the- art performance, and future research directions.

\section{Generic Object Detection}
\label{Sec:Background}

\subsection{The Problem}
\label{Sec:TheProblem}
\emph{Generic object detection}, also called generic object category detection, object class detection, or object category detection \cite{Zhang13}, is defined as follows. Given an image, determine whether or not there are instances of objects from predefined categories (usually \emph{many} categories, \emph{e.g.,} 200 categories in the ILSVRC object detection challenge) and, if present, to return the spatial location and extent of each instance.  A greater emphasis is placed on detecting a broad range of natural categories, as opposed to specific object category detection where only a narrower predefined category of interest (\emph{e.g.,} faces, pedestrians, or cars) may be present.  Although thousands of objects occupy the visual world in which we live, currently the research community is primarily interested in the localization of highly structured objects (\emph{e.g.,} cars, faces, bicycles and airplanes) and articulated objects (\emph{e.g.,} humans, cows and horses) rather than unstructured scenes (such as sky, grass and cloud).

The spatial location and extent of an object can be defined coarsely using a bounding box (an axis-aligned rectangle tightly bounding the object) \cite{Everingham2010,Russakovsky2015}, a precise pixelwise segmentation mask \cite{Zhang13}, or a closed boundary \cite{Lin2014,Russell2008}, as illustrated in Fig.~\ref{Fig:TheProblem}.
To the best of our knowledge, for the evaluation of generic object detection algorithms, it is bounding boxes which are most widely used in the current literature \cite{Everingham2010,Russakovsky2015}, and therefore this is also the approach we adopt in this survey.  However, as the research community moves towards deeper scene understanding (from image level object classification to single object localization, to generic object detection, and to pixelwise object segmentation), it is anticipated that future challenges will be at the pixel level \cite{Lin2014}.

There are many problems closely related to that of generic object detection\footnote{To the best of our knowledge, there is no universal agreement in the literature on the definitions of various vision subtasks. Terms such as detection, localization, recognition, classification, categorization, verification, identification, annotation, labeling, and understanding are often differently defined \cite{Andreopoulos13}.}.
The goal of \emph{object classification} or \emph{object categorization} (Fig.~\ref{Fig:TheProblem} (a)) is to assess the presence of objects from a given set of object classes in an image; \emph{i.e.,} assigning one or more object class labels to a given image, determining the  presence without the need of location. The additional requirement to locate the instances in an image makes detection a more challenging task than classification.  The \emph{object recognition} problem denotes the more general problem of identifying/localizing all the objects present in an image, subsuming the problems of object detection and classification \cite{Everingham2010,Russakovsky2015,Opelt2006generic,Andreopoulos13}.
Generic object detection is closely related to \emph{semantic image segmentation} (Fig.~\ref{Fig:TheProblem} (c)), which aims to assign each pixel in an image to a semantic class label.
\emph{Object instance segmentation} (Fig.~\ref{Fig:TheProblem} (d)) aims to distinguish different instances of the same object class, as opposed to semantic segmentation which does not.

\begin {figure}[!t]
\centering
\includegraphics[width=0.5\textwidth]{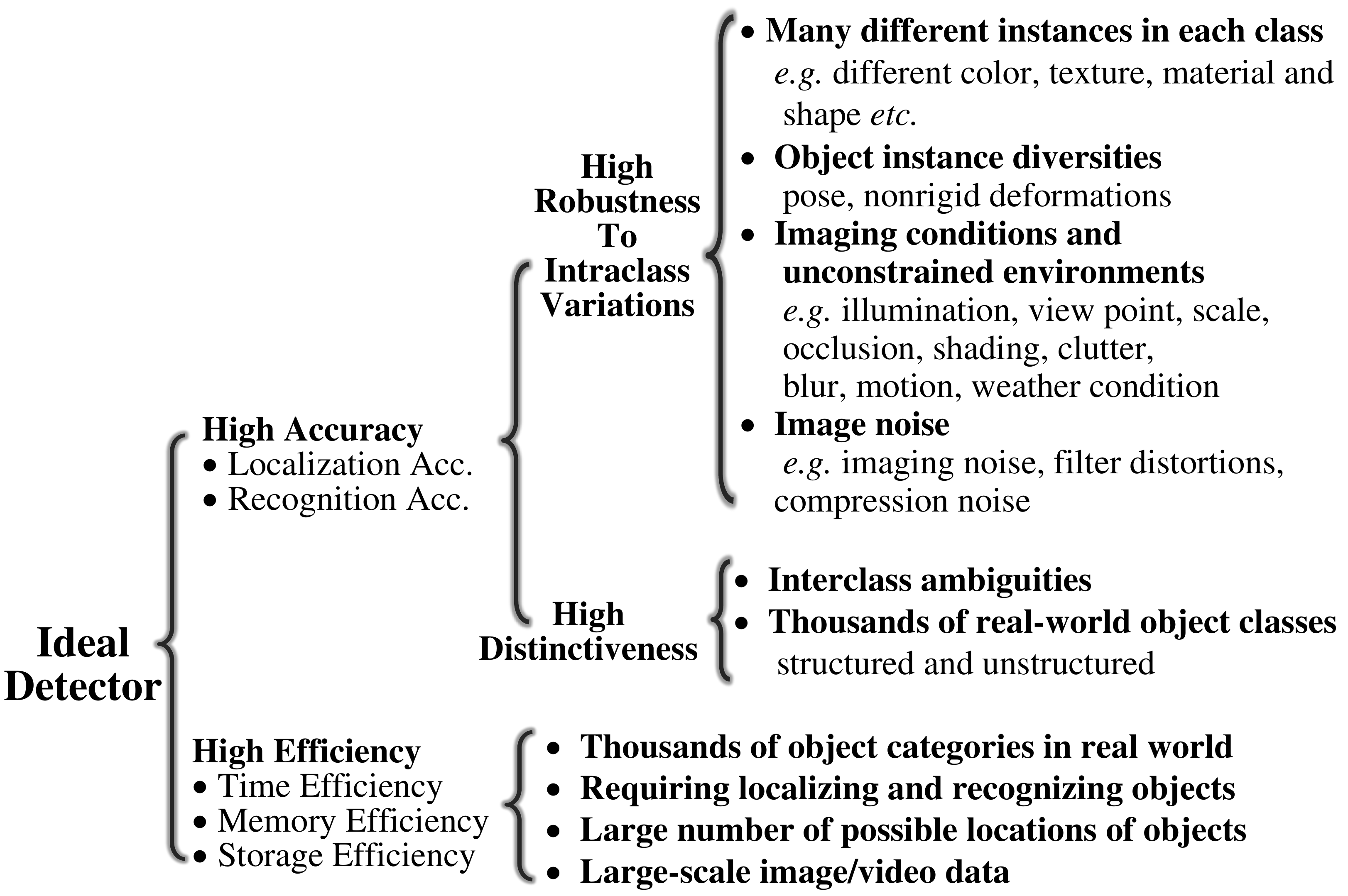}
\caption{Taxonomy of challenges in generic object detection.}
\label{Fig:challenges}
\end {figure}

\begin {figure}[!t]
\centering
\includegraphics[width=0.5\textwidth]{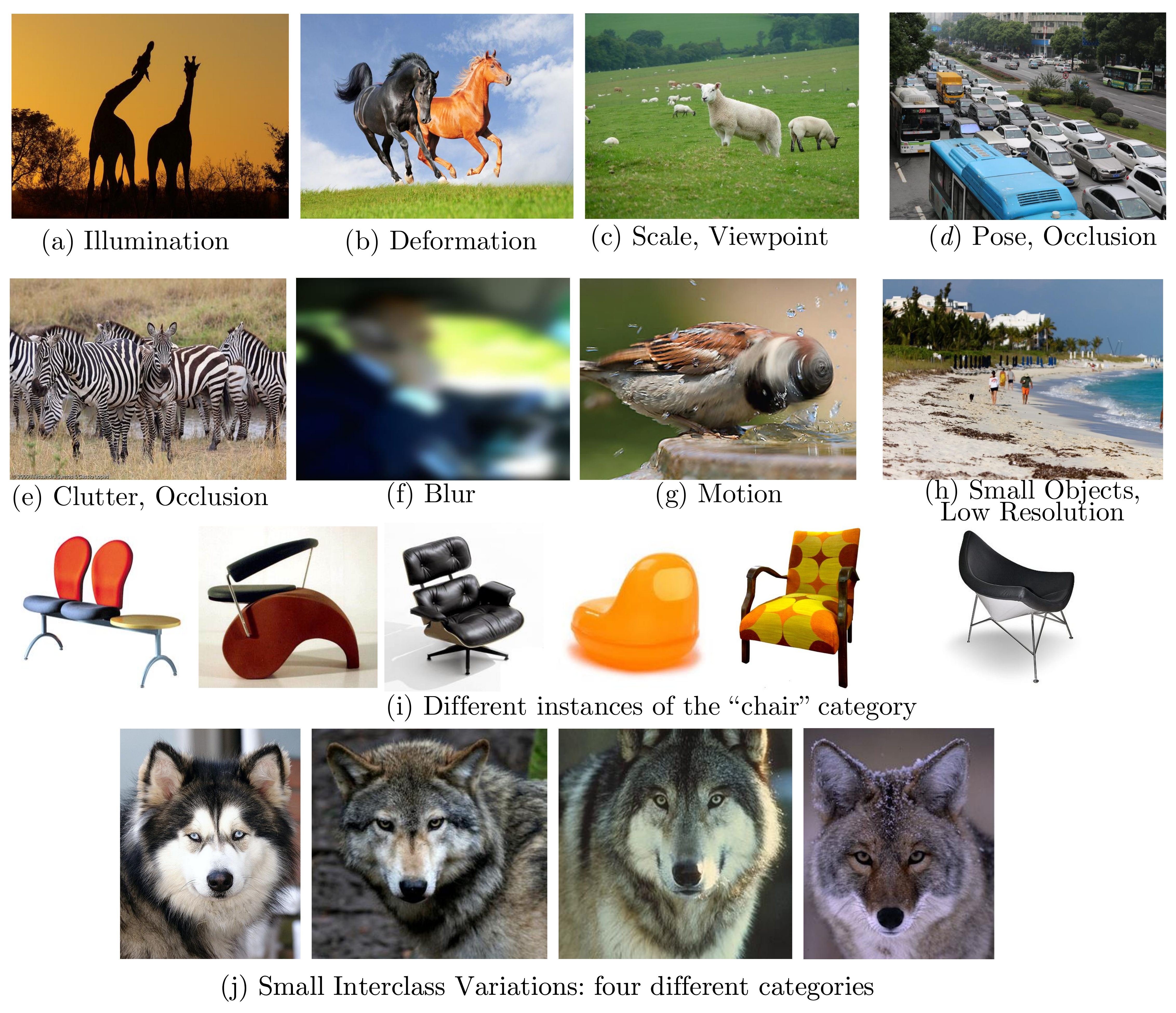}
\caption{\footnotesize{Changes in appearance of the same class with variations in imaging conditions (a-h).  There is an astonishing variation in what is meant to be a single object class (i).  In contrast, the four images in (j) appear very similar, but in fact are from four {\emph different} object classes. Most images are from ImageNet \cite{Russakovsky2015} and MS COCO \cite{Lin2014}.}}
\label{Fig:IntraInterClass}
\end {figure}

\subsection{Main Challenges}
\label{Sec:MainChallenges}
The ideal of generic object detection is to develop a general-purpose algorithm that achieves two competing goals of \emph{high quality/accuracy} and \emph{high efficiency} (Fig.~\ref{Fig:challenges}). As illustrated in Fig.~\ref{Fig:IntraInterClass}, high quality detection must accurately localize and recognize objects in images or video frames, such that the large variety of object categories in the real world can be distinguished (\emph{i.e.,} high distinctiveness), and that object instances from the same category, subject to intra-class appearance variations, can be localized and recognized (\emph{i.e.,} high robustness). High efficiency requires that the entire detection task runs in real time with acceptable memory and storage demands.

\subsubsection{Accuracy related challenges}

Challenges in detection accuracy  stem from 1) the vast range of intra-class variations and 2) the huge number of object categories.

Intra-class variations can be divided into two types: intrinsic factors and imaging conditions.
In terms of intrinsic factors, each object category can have many different object instances, possibly varying in one or more of color, texture, material, shape, and size, such as the ``chair'' category shown in Fig.~\ref{Fig:IntraInterClass} (\emph{i}). Even in a more narrowly defined class, such as human or horse, object instances can appear in different poses, subject to nonrigid deformations or with the addition of clothing.

Imaging condition variations are caused by the dramatic impacts unconstrained environments can have  on object appearance, such as lighting (dawn, day, dusk, indoors), physical location, weather conditions, cameras, backgrounds, illuminations, occlusion, and viewing distances. All of these conditions produce significant variations in object appearance, such as illumination, pose,  scale, occlusion, clutter, shading, blur and motion, with examples illustrated in Fig.~\ref{Fig:IntraInterClass} (\emph{a}-\emph{h}).  Further challenges may be added by digitization artifacts, noise corruption, poor resolution, and filtering distortions.

In addition to \emph{intra}class variations, the large number of object categories, on the order of $10^4-10^5$, demands great discrimination power from the detector to distinguish between subtly different \emph{inter}class variations, as illustrated in Fig.~\ref{Fig:IntraInterClass} (j). In practice, current detectors focus mainly on structured object categories, such as the 20, 200 and 91 object classes in PASCAL VOC \cite{Everingham2010}, ILSVRC \cite{Russakovsky2015} and MS COCO \cite{Lin2014} respectively. Clearly, the number of object categories under consideration in existing benchmark datasets is much smaller than can be recognized by humans.

\subsubsection{Efficiency and scalability related challenges}

The prevalence of social media networks and mobile/wearable devices has led to increasing demands for analyzing visual data.  However, mobile/wearable devices have limited computational capabilities and storage space, making efficient object detection critical.

The efficiency challenges stem from the need to localize and recognize, computational complexity growing with the (possibly large) number of object categories, and with the (possibly very large) number of locations and scales within a single image, such as the examples in Fig.~\ref{Fig:IntraInterClass} (c, d).

A further challenge is that of scalability:  A detector should be able to handle previously unseen objects, unknown situations, and high data rates.  As the number of images and the number of categories continue to grow, it may become impossible to annotate them manually, forcing a reliance on weakly supervised strategies.

\begin {figure*}[!t]
\centering
\includegraphics[width=0.45\textwidth]{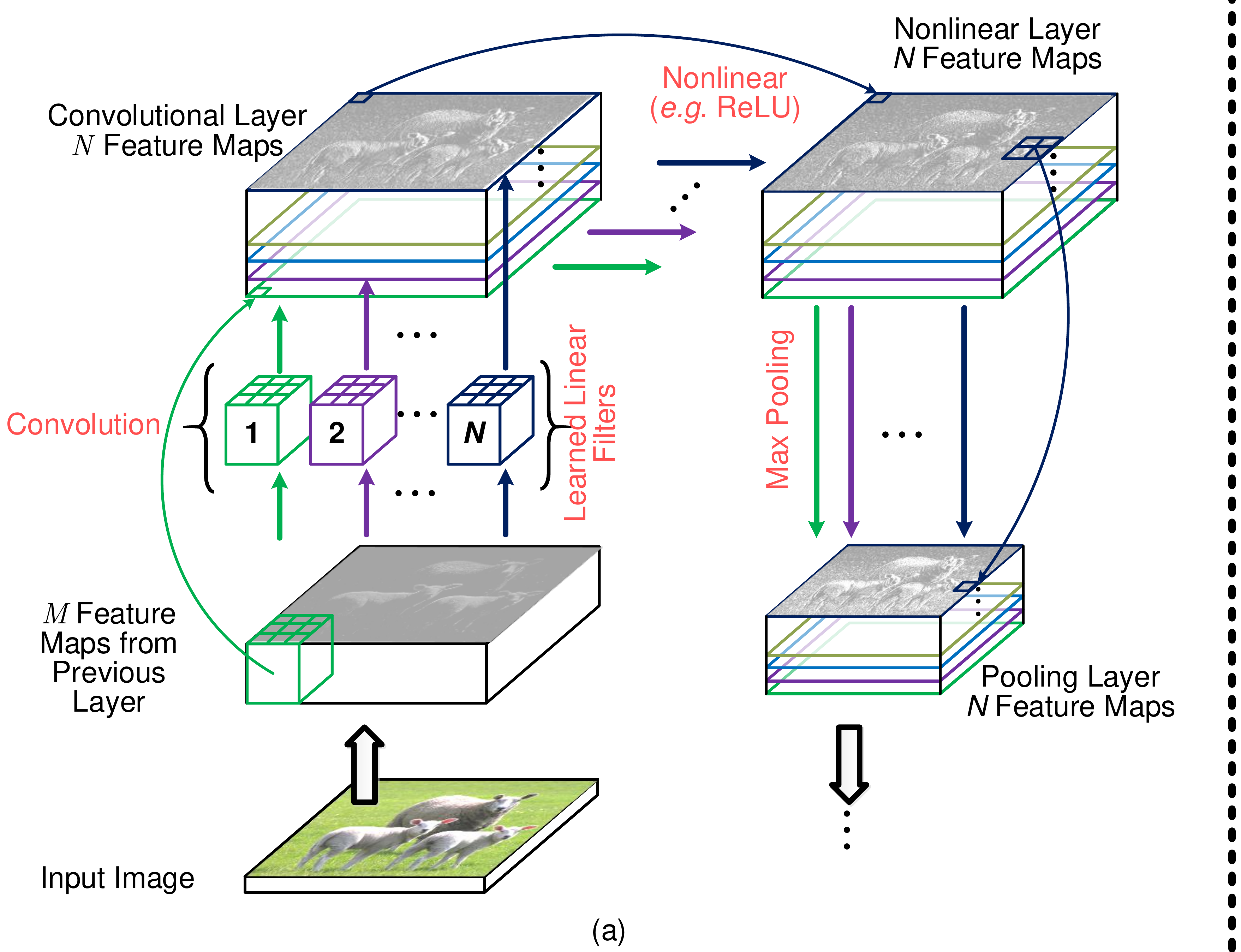}
\includegraphics[width=0.45\textwidth]{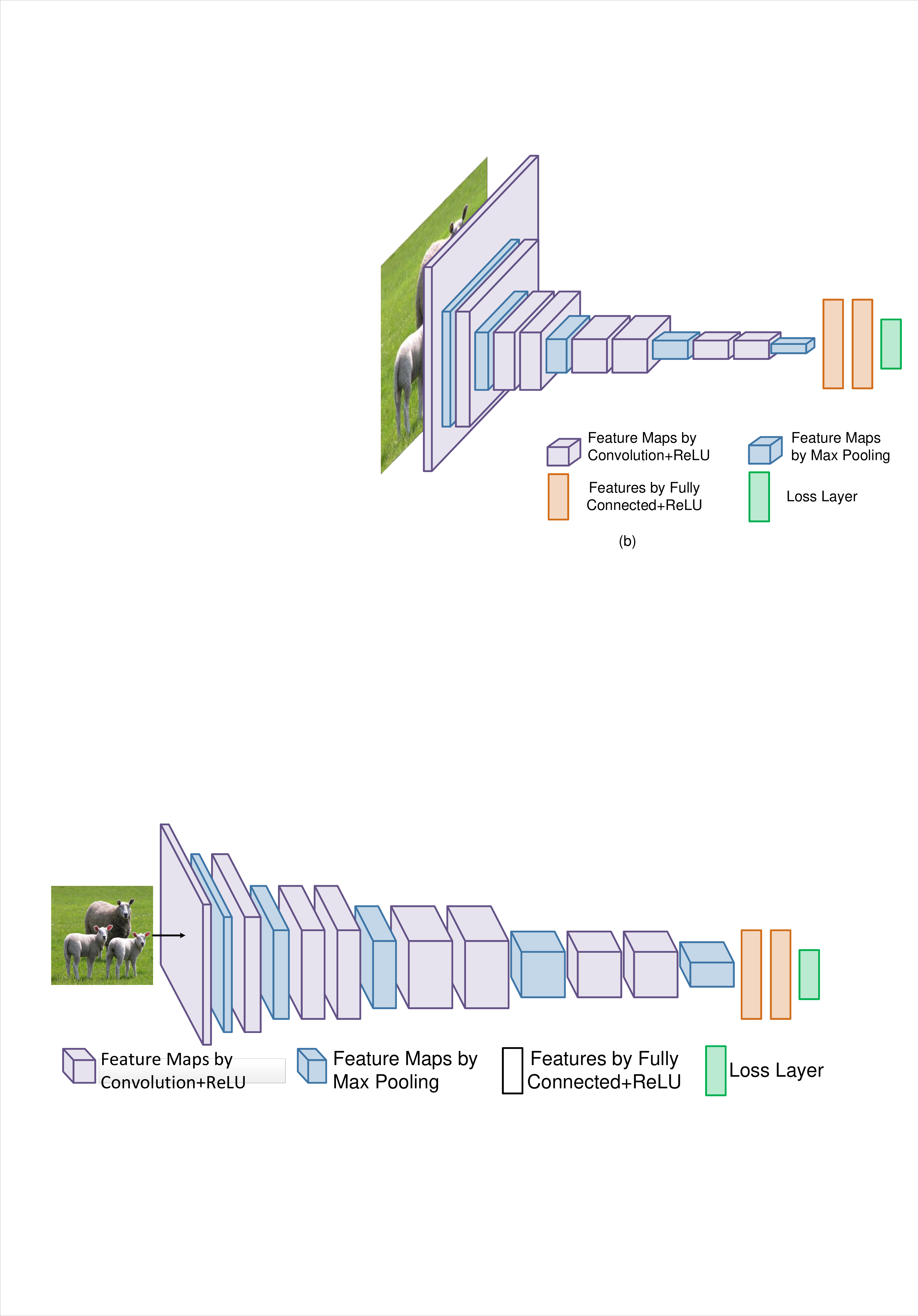}
\caption{\textcolor{black}{(a) Illustration of three operations that are repeatedly applied by a typical CNN:
Convolution with a number of linear filters;
Nonlinearities (\emph{e.g.} ReLU);
and Local pooling (\emph{e.g.} Max Pooling). The $M$ feature maps from a previous layer are convolved with $N$ different filters (here shown as size $3\times3\times M$), using a stride of 1. The resulting $N$ feature maps are then passed through a nonlinear function (\emph{e.g.} ReLU), and pooled (\emph{e.g.} taking a maximum over $2\times2$ regions) to give $N$ feature
maps at a reduced resolution. (b) Illustration of the architecture of VGGNet \cite{Simonyan2014VGG}, a typical CNN with 11 weight layers. An image with 3 color channels is presented as
the input.  The network has 8 convolutional layers, 3 fully connected layers, 5 max pooling layers and a softmax classification layer.  The last three fully connected layers take features from the top convolutional layer as input in vector form. The final layer is a $C$-way softmax function, $C$ being the number of classes. The whole network can be learned from labeled training data by optimizing an objective function (\emph{e.g.} mean squared error or cross entropy
loss) via Stochastic Gradient Descent.}}
\label{fig:ConvReLuMax}
\end {figure*}

\subsection{Progress in the Past Two Decades}
\label{Sec:Progress}

Early research on object recognition was based on template matching techniques and simple part-based models \cite{Fischler1973}, focusing on specific objects whose spatial layouts are roughly rigid, such as faces. Before 1990 the leading paradigm of object recognition was based on geometric representations~\cite{Mundy2006Object,Ponce2007Toward}, with the focus later moving away from geometry and prior models towards the use of statistical classifiers (such as Neural Networks \cite{Rowley1998}, SVM \cite{Osuna1997Train} and Adaboost \cite{Viola2001,Xiao2003Boosting}) based on appearance features \cite{Murase1995,Schmid1997Local}. This successful family of object detectors set the stage for most subsequent research in this field.

The milestones of object detection in more recent years are presented in Fig.~\ref{fig:milestones}, in which two main eras (SIFT\emph{ vs.} DCNN) are highlighted. The appearance features moved from global representations \cite{Murase1995Visual,Swain1991Color,Turk1991Face} to local representations that are designed to be invariant to changes in translation, scale, rotation, illumination, viewpoint and occlusion. Handcrafted local invariant features gained tremendous popularity, starting from the Scale Invariant Feature Transform (SIFT) feature \cite{Lowe1999Object}, and the progress on various visual recognition tasks was based substantially on the use of local descriptors \cite{Mikolajczyk2005} such as Haar-like features \cite{Viola2001}, SIFT \cite{Lowe2004}, Shape Contexts \cite{Belongie2002shape}, Histogram of Gradients (HOG) \cite{Dalal2005HOG} Local Binary Patterns (LBP) \cite{Ojala02}, and region covariances \cite{Tuzel2006Region}. These local features are usually aggregated by simple concatenation or feature pooling encoders such as the Bag of Visual Words approach, introduced by Sivic and Zisserman \cite{Sivic2003} and Csurka \emph{et al.} \cite{Csurka2004}, Spatial Pyramid Matching (SPM) of BoW models \cite{Lazebnik2006SPM}, and Fisher Vectors \cite{Perronnin2010}.

For years, the multistage hand tuned pipelines of handcrafted local descriptors and discriminative classifiers dominated a variety of domains in computer vision, including object detection, until the significant turning point in 2012 when DCNNs \cite{Krizhevsky2012} achieved their record-breaking results in image classification.

The use of CNNs for detection and localization \cite{Rowley1998} can be traced back to the 1990s, with a modest number of hidden layers used for object detection \cite{Vaillant1994,Rowley1998,Sermanet2013c}, successful in restricted domains such as
face detection. However, more recently, deeper CNNs have led to record-breaking
improvements in the detection of more general object
categories, a shift which came about when the successful application
of DCNNs in image classification \cite{Krizhevsky2012} was
transferred to object detection, resulting in the milestone Region-based CNN (RCNN) detector of Girshick
\emph{et al.} \cite{Girshick2014RCNN}.

The successes of deep detectors rely heavily on vast training data and large networks with millions or even billions of parameters.  The availability of GPUs with very high computational capability and large-scale detection datasets (such as ImageNet \cite{ImageNet2009,Russakovsky2015} and MS COCO \cite{Lin2014}) play a key role in their success.  Large datasets have allowed researchers to target more realistic and complex problems from images with large intra-class variations and inter-class similarities \cite{Lin2014,Russakovsky2015}.  However, accurate annotations are labor intensive to obtain, so detectors must consider methods that can relieve annotation difficulties or can learn with smaller training datasets.

The research community has started moving towards the challenging goal of building general purpose object detection systems whose ability to detect many object categories matches that of humans. This is a major challenge: according to cognitive scientists, human beings can identify around 3,000 entry level categories and 30,000 visual categories overall, and the number of categories distinguishable with domain expertise may be to the order of $10^5$ \cite{Biederman1987}.  Despite the remarkable progress of the past years, designing an accurate, robust, efficient detection and recognition system that approaches human-level performance on $10^4-10^5$ categories is undoubtedly an unresolved problem.

\section{A Brief Introduction to Deep Learning}
\label{Sec:CNNintro}
\textcolor{black}{Deep learning has revolutionized a wide range of machine learning tasks, from image classification and video processing to speech recognition and natural language understanding. Given this tremendously rapid evolution, there exist many recent survey papers on deep learning  \cite{Bengio13Feature,Goodfellow2016Deep,Gu2015Recent,LeCun15,
Litjens2017,Pouyanfar2018Survey,
Wu2019Comprehensive,Young2018Recent,
Zhang2018Deep,Zhou2018Graph,Zhu2017Deep}. These surveys have reviewed deep learning techniques from different perspectives \cite{Bengio13Feature,Goodfellow2016Deep,Gu2015Recent,
LeCun15,Pouyanfar2018Survey,Wu2019Comprehensive,Zhou2018Graph}, or with applications to medical image analysis \cite{Litjens2017},
natural language processing \cite{Young2018Recent}, speech recognition systems \cite{Zhang2018Deep}, and remote sensing \cite{Zhu2017Deep}.}

Convolutional Neural Networks (CNNs), the most representative models of deep learning, are able to exploit the basic properties underlying natural signals: translation invariance, local connectivity, and compositional hierarchies \cite{LeCun15}. A typical CNN, illustrated in Fig.~\ref{fig:ConvReLuMax}, has a hierarchical structure and is composed of a number of layers to learn representations of data with multiple levels of abstraction \cite{LeCun15}. We begin with a convolution
\begin{equation}
\textbf{\emph{x}}^{l-1} * \textbf{\emph{w}}^{l}
\end{equation}
between an input feature map $\textbf{\emph{x}}^{l-1}$ at a feature map from previous layer $l-1$, convolved with a 2D convolutional kernel (or filter or weights) $\textbf{\emph{w}}^{l}$.  This convolution appears over a sequence of layers, subject to a nonlinear operation $\sigma$, such that
\begin{equation}
\textbf{\emph{x}}^l_j = \sigma(\sum_{i=1}^{N^{l-1}} \textbf{\emph{x}}^{l-1}_i * \textbf{\emph{w}}^{l}_{i, j} +b^{l}_j), \label{eq:conv}
\end{equation}
with a convolution now between the $N^{l-1}$ input feature maps $\textbf{\emph{x}}^{l-1}_i$ and the corresponding kernel $\textbf{\emph{w}}^{l}_{i, j}$, plus a bias term $b^{l}_j$.  The elementwise nonlinear function $\sigma(\cdot)$ is typically a rectified linear unit (ReLU) for each element,
\begin{equation}
\sigma(x) = \max\{x, 0\}.
\end{equation}
Finally, pooling corresponds to the downsampling/upsampling of feature maps.  These three operations (convolution, nonlinearity, pooling) are illustrated in Fig. \ref{fig:ConvReLuMax} (a); CNNs having a large number of layers, a ``deep'' network, are referred to as Deep CNNs (DCNNs), with a typical DCNN architecture illustrated in Fig.~\ref{fig:ConvReLuMax} (b).

Most layers of a CNN consist of a number of
feature maps, within which each pixel acts like a neuron. Each neuron in a convolutional
layer is connected to feature maps of the previous
layer through a set of weights $\textbf{\emph{w}}_{i,j}$ (essentially a set of 2D filters). As can be seen in Fig.~\ref{fig:ConvReLuMax} (b),
where the early CNN layers are typically composed of convolutional and pooling layers, the later layers are normally fully connected.
From earlier to later layers, the input
image is repeatedly convolved, and with each layer, the receptive
field or region of support increases.
In general, the initial CNN layers extract low-level features (\emph{e.g.,} edges),
with later layers extracting more general features of increasing complexity \cite{ZeilerFergus2014,Bengio13Feature,LeCun15,Oquab2014Learning}.

DCNNs have a number of outstanding advantages:  a hierarchical structure to learn representations of data with multiple levels of abstraction, the capacity to learn very complex functions, and learning feature representations directly and automatically from data with minimal domain knowledge.  What has particularly made DCNNs successful has been the availability of large scale labeled datasets and of GPUs with very high computational capability.

Despite the great successes, known deficiencies remain.  In particular, there is an extreme need for labeled training data and a requirement of expensive computing resources, and considerable skill and experience are still needed to select appropriate learning parameters and network architectures.  Trained networks are poorly interpretable, there is a lack of robustness to degradations, and many DCNNs have shown serious vulnerability to attacks \cite{Goodfellow2015Explaining}, all of which currently limit the use of DCNNs in real-world applications.

\begin{table}[!t]
\caption{Most frequent object classes for each detection challenge. The size of each word is proportional to the frequency of that class in the training dataset.}
\centering
\setlength\tabcolsep{2pt}
\begin{tabular}{!{\vrule width1.2bp}c!{\vrule width1.2bp}c!{\vrule width1.2bp}}
\Xhline{1.2pt}
\includegraphics[width=0.24\textwidth]{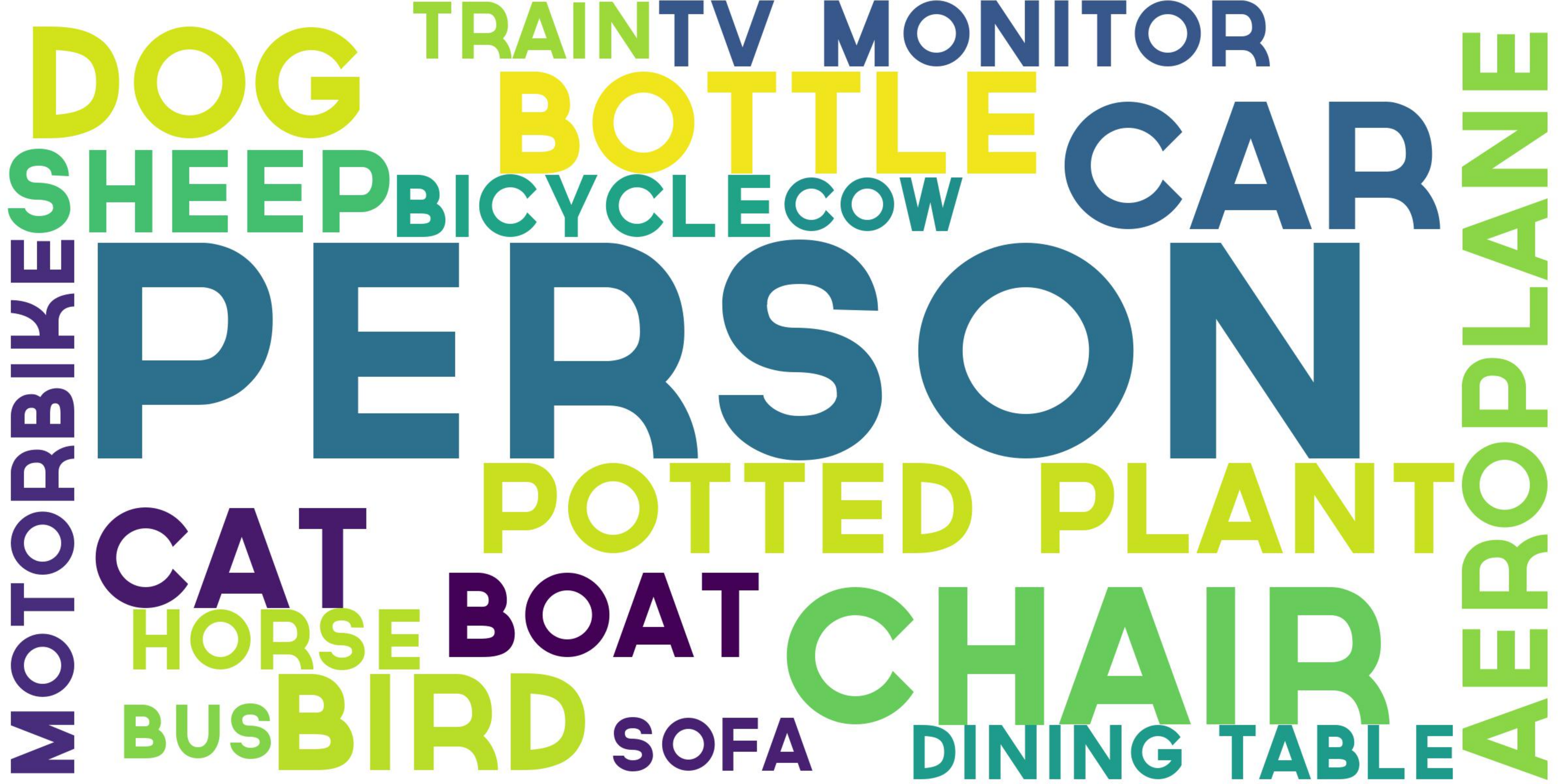}&
\includegraphics[width=0.24\textwidth]{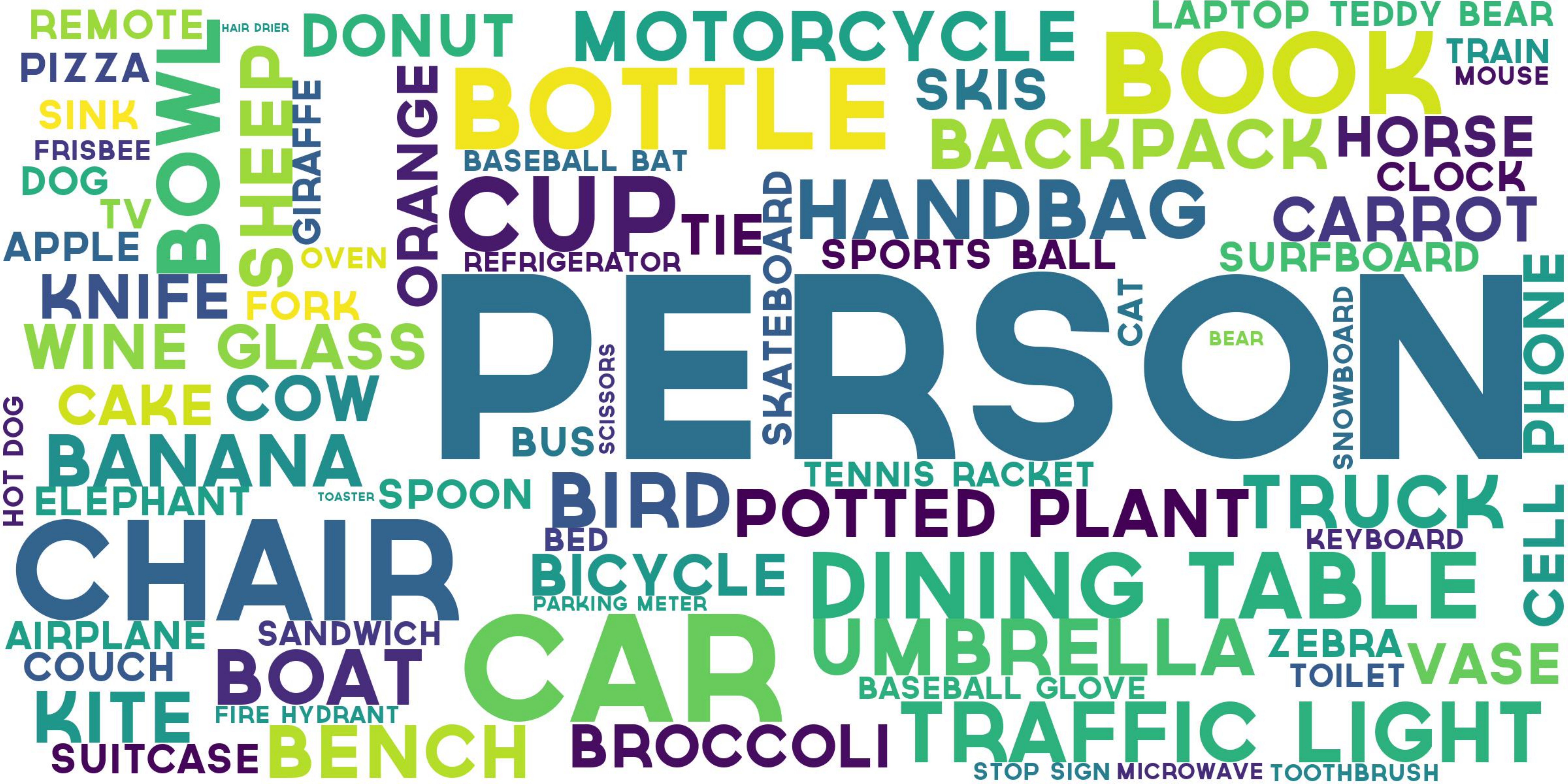}\\
\Xhline{1.2pt}
\multicolumn{1}{c}{(a) PASCAL VOC (20 Classes) }& \multicolumn{1}{c}{(b) MS COCO (80 Classes)}\\ \multicolumn{1}{c}{}  \\
\Xhline{1.2pt}
\multicolumn{2}{!{\vrule width1.2bp}c!{\vrule width1.2bp}}{\includegraphics[width=0.48\textwidth]{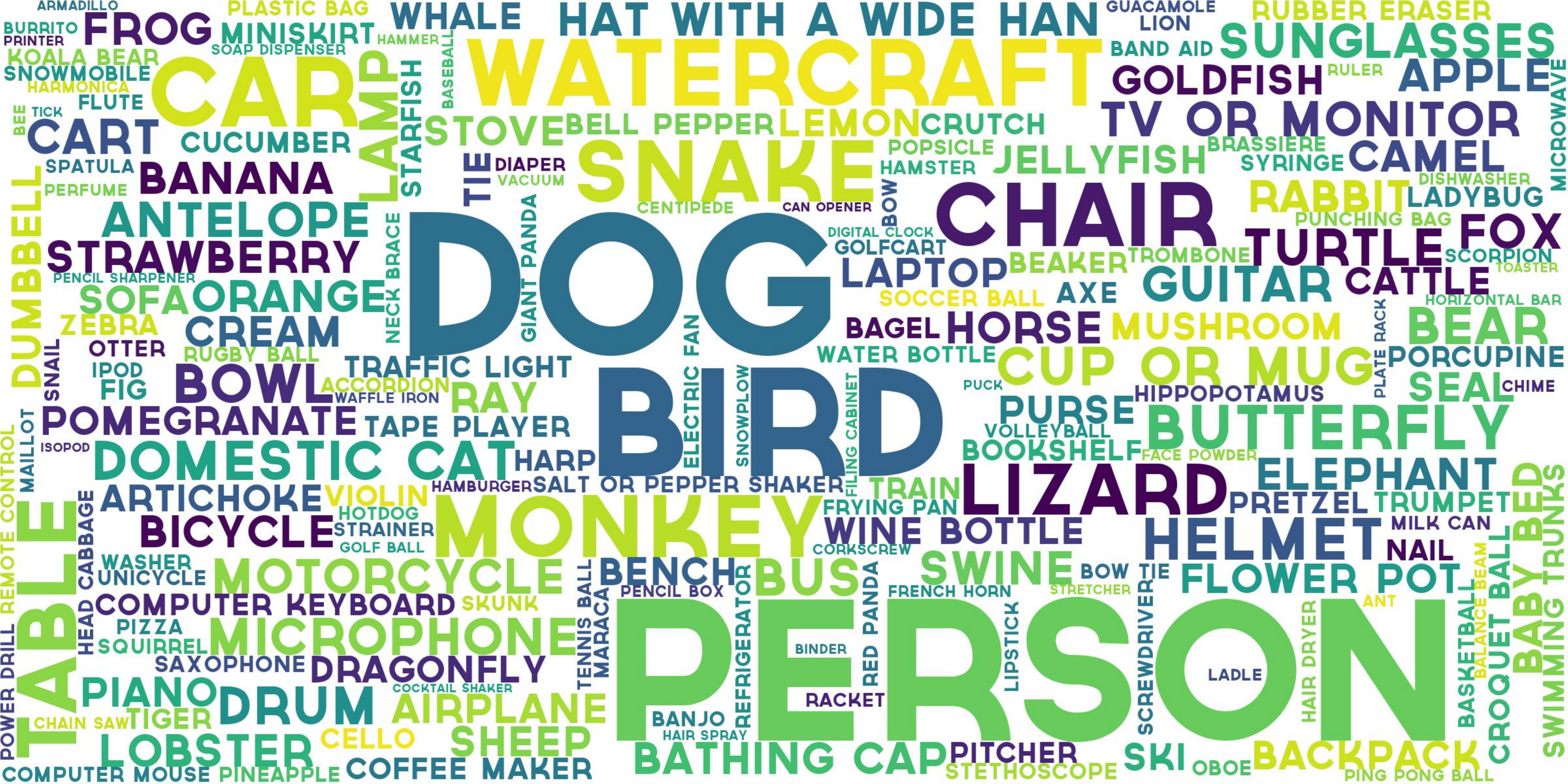}}\\
\Xhline{1.2pt}
\multicolumn{2}{c}{(c) ILSVRC (200 Classes)}\\ \multicolumn{1}{c}{} \\
\Xhline{1.2pt}
\multicolumn{2}{!{\vrule width1.2bp}c!{\vrule width1.2bp}}{\includegraphics[width=0.48\textwidth]{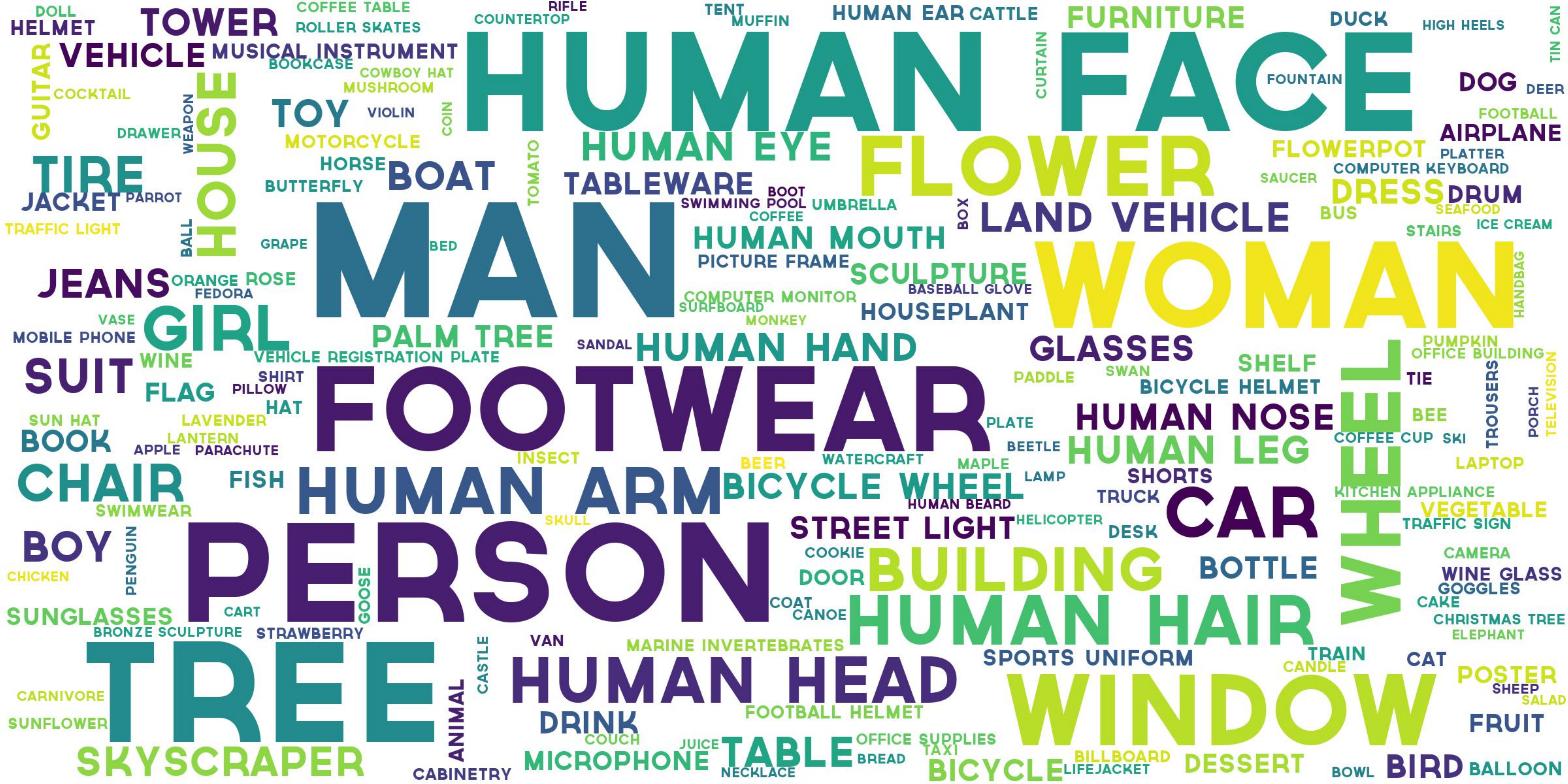}}\\
\Xhline{1.2pt}
\multicolumn{2}{c}{(d) Open Images Detection Challenge (500 Classes)}
\end{tabular}
\label{fig:ClassFrequency}
\end {table}

\section{Datasets and Performance Evaluation}
\label{Sec:DataEval}
\subsection{Datasets}
\label{sec:datasets}
Datasets have played a key role throughout the history of object recognition research, not only as a common ground for measuring and comparing the performance of competing algorithms, but also pushing the field towards increasingly complex and challenging problems. In particular, recently, deep learning techniques have brought tremendous success to many visual recognition problems, and it is the large amounts of annotated data which play a key role in their success. Access to large numbers of images on the Internet makes it possible to build comprehensive datasets in order to capture a vast richness and diversity of objects, enabling unprecedented performance in object recognition.

\begin {figure*}[!t]
\centering
\includegraphics[width=0.98\textwidth]{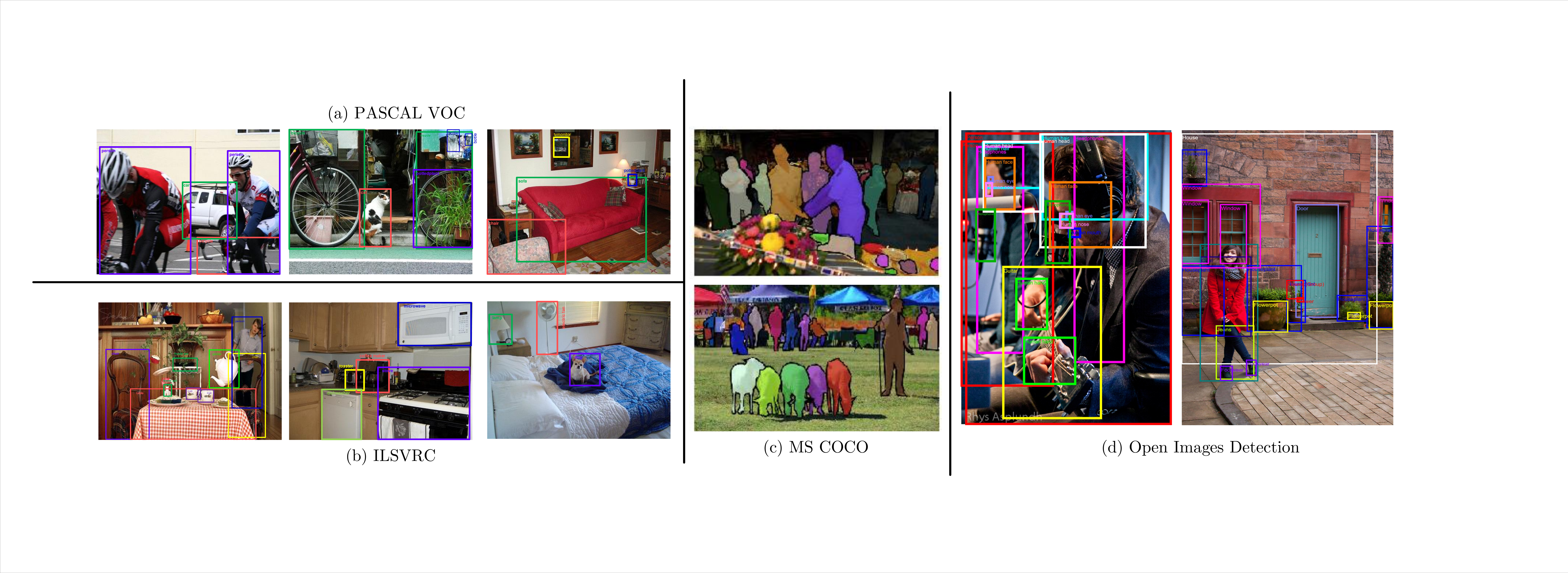}
\caption{Some example images with object annotations from PASCAL VOC, ILSVRC, MS COCO and Open Images. See Table \ref{Tab:maindatasets} for a summary of these datasets.}
\label{fig:ObjectImages}
\end {figure*}

For generic object detection, there are four famous datasets:  PASCAL VOC \cite{Everingham2010,Everingham2015}, ImageNet \cite{ImageNet2009}, MS COCO \cite{Lin2014} and Open Images \cite{Kuznetsova2018Open}. The attributes of these datasets are summarized in Table~\ref{Tab:maindatasets}, and selected sample images are shown in Fig.~\ref{fig:ObjectImages}. There are three steps to creating large-scale annotated datasets: determining the set of target object categories, collecting a diverse set of candidate images to represent the selected categories on the Internet, and annotating the collected images, typically by designing crowdsourcing strategies. Recognizing space limitations, we refer interested readers to the original papers \cite{Everingham2010,Everingham2015,Lin2014,Russakovsky2015,Kuznetsova2018Open} for detailed descriptions of these datasets in terms of construction and properties.

\begin{table*}[!t]
\caption {Popular databases for object recognition.  Example images from PASCAL VOC, ImageNet, MS COCO and Open Images are shown in Fig.~\ref{fig:ObjectImages}.}\label{Tab:maindatasets}
\centering
\renewcommand{\arraystretch}{1.2}
\setlength\arrayrulewidth{0.2mm}
\setlength\tabcolsep{1pt}
\resizebox*{18cm}{!}{
\begin{tabular}{!{\vrule width1.2bp}c|c|c|c|c|c|c|p{8cm}!{\vrule width1.2bp}}
\Xhline{1pt}
 \scriptsize   \shortstack [c] {\textbf{Dataset}\\ \textbf{Name}} & \scriptsize   \shortstack [c]
{\textbf{Total} \\ \textbf{Images}} & \scriptsize   \shortstack [c]
{\textbf{Categories}} & \scriptsize   \shortstack [c]
{\textbf{Images Per} \\ \textbf{Category}} & \scriptsize   \shortstack [c]
{\textbf{Objects Per }\\ \textbf{Image}} & \scriptsize   \shortstack [c] {\textbf{Image} \\ \textbf{Size} }
& \scriptsize   \shortstack [c] {\textbf{Started} \\  \textbf{Year}}
&\raisebox{1.3ex}[0pt]{ \scriptsize   \shortstack [c] {$\quad\quad\quad\quad\quad\quad\quad\quad\quad\quad\quad\quad$\textbf{Highlights}}} \\
\hline
\raisebox{-6.3ex}[0pt]{\scriptsize \shortstack [c] {PASCAL \\VOC \\ (2012) \cite{Everingham2015} } } &\raisebox{-5ex}[0pt]{\scriptsize   $11,540$ }& \raisebox{-5ex}[0pt]{\scriptsize $20$} & \raisebox{-5ex}[0pt]{ \scriptsize  $303\sim4087$}& \raisebox{-5ex}[0pt]{\scriptsize $2.4$}
& \raisebox{-5ex}[0pt]{\scriptsize $470\times380$}& \raisebox{-5ex}[0pt]{ \scriptsize  $2005$} & \scriptsize  Covers only 20 categories that are common in everyday life; Large number of training images; Close to real-world applications; Significantly larger intraclass variations; Objects in scene context; Multiple objects in one image; Contains many difficult samples.  \\
\hline
\raisebox{-3.3ex}[0pt]{\scriptsize ImageNet \cite{Russakovsky2015}} &\raisebox{-4.3ex}[0pt]{\shortstack [c] {14 \\  millions+} } &\raisebox{-3.3ex}[0pt]{ \scriptsize $21,841$} &\raisebox{-3.3ex}[0pt]{ \scriptsize  $-$}
&\raisebox{-3.3ex}[0pt]{ \scriptsize  $1.5$}
& \raisebox{-3.3ex}[0pt]{\scriptsize $500\times400$}&\raisebox{-3.3ex}[0pt]{ \scriptsize  $2009$} & \scriptsize  Large number of object categories; More instances and more categories of objects per image; More challenging than PASCAL VOC; Backbone of the ILSVRC challenge; Images are object-centric. \\
\hline
\raisebox{-3.3ex}[0pt]{\scriptsize MS COCO \cite{Lin2014}} & \raisebox{-3.3ex}[0pt]{\scriptsize   $328,000+$ } & \raisebox{-3.3ex}[0pt]{\scriptsize $91$} & \raisebox{-3.3ex}[0pt]{ \scriptsize  $-$}
 & \raisebox{-3.3ex}[0pt]{ \scriptsize  $7.3$}
& \raisebox{-3.3ex}[0pt]{ \scriptsize $640\times480$}& \raisebox{-3.3ex}[0pt]{\scriptsize  $2014$} & \scriptsize  Even closer to real world scenarios; Each image contains more instances of objects and richer object annotation information; Contains object segmentation notation data that is not available in the ImageNet dataset. \\
\hline
\raisebox{-1.3ex}[0pt]{\scriptsize Places \cite{Zhou2017Places}} &\raisebox{-2.3ex}[0pt]{\shortstack [c] {10 \\  millions+} } &\raisebox{-1.3ex}[0pt]{ \scriptsize $434$ }&\raisebox{-1.3ex}[0pt]{ \scriptsize  $-$}&\raisebox{-1.3ex}[0pt]{ \scriptsize  $-$}
& \raisebox{-1.3ex}[0pt]{ \scriptsize $256\times256$}&\raisebox{-1.3ex}[0pt]{ \scriptsize  $2014$} & \scriptsize  The largest labeled dataset for scene recognition; Four subsets Places365 Standard, Places365 Challenge, Places 205 and Places88 as benchmarks. \\
\hline
\raisebox{-3.3ex}[0pt]{\scriptsize Open Images \cite{Kuznetsova2018Open}} &\raisebox{-4.3ex}[0pt]{\shortstack [c] {9 \\  millions+} } &\raisebox{-3.3ex}[0pt]{ \scriptsize $6000$+ }&\raisebox{-3.3ex}[0pt]{ \scriptsize  $-$}&\raisebox{-3.3ex}[0pt]{ \scriptsize  $8.3$}
& \raisebox{-3.3ex}[0pt]{ \scriptsize varied}&\raisebox{-3.3ex}[0pt]{ \scriptsize  $2017$} & \scriptsize Annotated with image level labels, object bounding boxes and visual relationships; Open Images V5 supports large scale object detection, object instance segmentation and visual relationship detection. \\
\Xhline{1pt}
\end{tabular}
}
\end{table*}

The four datasets form the backbone of their respective detection challenges.
Each challenge consists of a publicly available dataset of images together with ground truth annotation and standardized evaluation software, and an annual competition
and corresponding workshop. Statistics for the number of images and object instances
in the training, validation and testing datasets\footnote{The annotations on the test set are not publicly
released, except for PASCAL VOC2007.} for the detection challenges are given in
Table~\ref{Tab:detdatasets}. The most frequent object classes in VOC, COCO, ILSVRC and Open Images detection datasets  are visualized in Table \ref{fig:ClassFrequency}.

\textbf{PASCAL VOC} \cite{Everingham2010,Everingham2015} is a multi-year effort devoted to the creation and maintenance of a series of benchmark datasets for classification and object detection, creating the precedent for standardized evaluation of recognition algorithms in the form of annual competitions. Starting from only four categories in 2005, the dataset has increased to 20 categories that are common in everyday life.
Since 2009, the number of images has grown every year, but with all previous images retained to allow test results to be compared from year to year. Due the availability of larger datasets like ImageNet, MS COCO and Open Images, PASCAL VOC has gradually fallen out of fashion.

\textbf{ILSVRC}, the ImageNet Large Scale Visual Recognition Challenge \cite{Russakovsky2015}, is derived from ImageNet \cite{ImageNet2009}, scaling up PASCAL VOC's goal of standardized training and evaluation of detection algorithms by more
than an order of magnitude in the number of object classes and images.  ImageNet1000, a subset of ImageNet images  with 1000 different object categories and a total of 1.2 million images, has been fixed to provide a standardized benchmark for the ILSVRC image classification challenge.

\textbf{MS COCO} is a response to the criticism of ImageNet that objects in its dataset tend
to be large and well centered, making the ImageNet dataset atypical of real-world scenarios.
To push for richer image
understanding, researchers created the MS COCO database \cite{Lin2014} containing complex everyday scenes with common objects in their natural context, closer to real life, where objects are labeled using fully-segmented instances to provide more accurate detector evaluation.  The COCO object detection challenge \cite{Lin2014} features two object detection tasks: using either bounding box output or object instance segmentation output.  COCO introduced three new challenges:
\begin{enumerate}
\item It contains objects at a wide range of scales, including a high percentage of small objects \cite{Singh2018SNIP};
\item Objects are less iconic and amid clutter or heavy occlusion;
\item The evaluation metric (see Table~\ref{Tab:Metrics}) encourages more accurate object localization.
\end{enumerate}
Just like ImageNet in its time, MS COCO has become the standard for object detection today.

\textbf{OICOD} (the Open Image Challenge Object Detection) is derived from Open Images V4 (now V5 in 2019) \cite{Kuznetsova2018Open}, currently
the largest publicly available object detection dataset. OICOD is different from previous large scale object detection datasets like ILSVRC and MS COCO, not merely in terms of the significantly increased number of classes, images, bounding box annotations and instance segmentation mask annotations,
but also regarding the annotation process. In ILSVRC and MS COCO,
instances of all classes in the dataset are exhaustively annotated, whereas for Open Images V4 a classifier was applied to each image and only those labels with sufficiently high scores were sent for human verification. Therefore in OICOD only the object instances of human-confirmed positive labels are annotated.

\begin{table*}[!t]
\caption {Statistics of commonly used object detection datasets. Object statistics for VOC challenges list the non-difficult objects used in the evaluation (all annotated objects). For the COCO challenge, prior to 2017, the test set had four splits (\emph{Dev}, \emph{Standard}, \emph{Reserve}, and \emph{Challenge}), with each having about 20K images. Starting in 2017, the test set has only the \emph{Dev} and \emph{Challenge} splits, with the other two splits removed.  Starting in 2017, the train and val sets are arranged differently, and the test set is divided into two roughly equally sized splits of about $20,000$ images each: Test Dev and Test Challenge. Note that the 2017 Test Dev/Challenge splits contain the same images as the 2015 Test Dev/Challenge splits, so results across the years are directly comparable.}\label{Tab:detdatasets}
\centering
\renewcommand{\arraystretch}{1.2}
\setlength\arrayrulewidth{0.2mm}
\setlength\tabcolsep{2pt}
\resizebox*{14cm}{!}{
\begin{tabular}{!{\vrule width1.2bp}c|c|r|r|c!{\vrule width1.2bp}r|c!{\vrule width1.2bp}r|r|c!{\vrule width1.2bp}}
\Xhline{1pt}
\multirow{2}{*}{\footnotesize Challenge} &  \multirow{2}{*}{\footnotesize \shortstack [c] {Object \\Classes}}&  \multicolumn{3}{c!{\vrule width1.2bp}}{\footnotesize Number of Images} &   \multicolumn{2}{c!{\vrule width1.2bp}}{\footnotesize \shortstack [c] {Number of Annotated Objects}} &  \multicolumn{3}{c!{\vrule width1.2bp}}{\footnotesize Summary (Train$+$Val)}  \\
\cline{3-10}
\footnotesize   & \footnotesize   & \footnotesize Train	& \footnotesize Val & \footnotesize Test	 & \footnotesize \shortstack [c] { Train} &  \footnotesize Val &  \footnotesize Images &  \footnotesize Boxes &  \footnotesize Boxes/Image\\
\Xhline{1pt}
\multicolumn{10}{!{\vrule width1.2bp}c!{\vrule width1.2bp}}{\footnotesize PASCAL VOC Object Detection Challenge}\\
\hline
\footnotesize VOC07 & \footnotesize $20$ & \footnotesize$ 2,501$	& \footnotesize $2,510	 $ & \footnotesize$4,952	 $
& \footnotesize $6,301(7,844)$ & \footnotesize $6,307(7,818)$ &  \footnotesize $5,011$ &  \footnotesize$12,608$&  \footnotesize $2.5$\\
\hline
\footnotesize VOC08 & \footnotesize $20$
& \footnotesize$2,111$	& \footnotesize$2,221$	& \footnotesize$4,133$	 & \footnotesize $5,082(6,337)	 $ 	& \footnotesize$5,281(6,347) $ &  \footnotesize $4,332$&  \footnotesize$10,364$&  \footnotesize $2.4$ \\
\hline
\footnotesize VOC09 & \footnotesize $20 $& \footnotesize$3,473$& \footnotesize	 $3,581$	 & \footnotesize $6,650	 $& \footnotesize$8,505(9,760)	 $ & \footnotesize$	 8,713(9,779)$&  \footnotesize$7,054$&  \footnotesize $17,218$&  \footnotesize $2.3$\\
\hline
\footnotesize VOC10 & \footnotesize $20$ & \footnotesize$4,998	$& \footnotesize$5,105$	 & \footnotesize$9,637$	 & \footnotesize $11,577(13,339)$ & \footnotesize $11,797(13,352)$ &  \footnotesize$10,103$&  \footnotesize $23,374$&  \footnotesize$2.4$ \\
\hline
\footnotesize VOC11 & \footnotesize $20$ & \footnotesize $5,717$	& \footnotesize $5,823$	 & \footnotesize$10,994$	& \footnotesize$13,609 (15,774)	$& \footnotesize	 $13,841(15,787) $ &  \footnotesize$11,540$&  \footnotesize$27,450$&  \footnotesize $2.4$ \\
\hline
\footnotesize VOC12 & \footnotesize $20$ & \footnotesize$5,717$& \footnotesize	 $5,823$	 & \footnotesize$10,991$& \footnotesize	 $13,609 (15,774)	$& \footnotesize	 $13,841(15,787) $&  \footnotesize $11,540$&  \footnotesize $27,450$&  \footnotesize$2.4$\\
\Xhline{1pt}
\multicolumn{10}{!{\vrule width1.2bp}c!{\vrule width1.2bp}}{\footnotesize ILSVRC Object Detection Challenge}\\
\hline
\footnotesize ILSVRC13& \footnotesize	$200$	& \footnotesize$395,909$	& \footnotesize $20,121	 $& \footnotesize$40,152$	& \footnotesize$345,854	 $& \footnotesize$55,502$&  \footnotesize$416,030$&  \footnotesize $401,356$&  \footnotesize $1.0$\\
\hline
\footnotesize ILSVRC14	& \footnotesize$200	$& \footnotesize$456,567$	 & \footnotesize$20,121$	& \footnotesize$40,152$& \footnotesize$	 478,807	 $& \footnotesize$55,502$&  \footnotesize $476,668$&  \footnotesize$534,309$&  \footnotesize$1.1$\\
\hline
\footnotesize ILSVRC15& \footnotesize	$200$& \footnotesize	 $456,567$	& \footnotesize$20,121$& \footnotesize	 $51,294$	& \footnotesize$478,807$& \footnotesize	 $55,502$&  \footnotesize$476,668$&  \footnotesize $534,309$ &  \footnotesize$1.1$\\
\hline
\footnotesize ILSVRC16	& \footnotesize$200$& \footnotesize$	 456,567$& \footnotesize$	 20,121$& \footnotesize	$60,000$& \footnotesize$	 478,807$& \footnotesize	$55,502$&  \footnotesize$476,668$&  \footnotesize $534,309$ &  \footnotesize$1.1$\\
\hline
\footnotesize ILSVRC17	& \footnotesize$ 200$& \footnotesize	 $456,567	 $& \footnotesize$ 20,121$& \footnotesize$	65,500	$& \footnotesize$478,807$	 & \footnotesize$55,502$&  \footnotesize$476,668$&  \footnotesize $534,309$&  \footnotesize$1.1$\\
\Xhline{1pt}
\multicolumn{10}{!{\vrule width1.2bp}c!{\vrule width1.2bp}}{\footnotesize MS COCO Object Detection Challenge}\\
\hline
\footnotesize MS COCO15 & \footnotesize	$80 $& \footnotesize	 $82,783$	& \footnotesize $40,504	 $ & \footnotesize$81,434$	 & \footnotesize $604,907$ & \footnotesize$ 291,875$&  \footnotesize$123,287$&  \footnotesize $896,782$&  \footnotesize$7.3$\\
\hline
\footnotesize MS COCO16	& \footnotesize $80$ & \footnotesize$ 82,783$	& \footnotesize $40,504$	 & \footnotesize $81,434$ & \footnotesize$ 604,907 $& \footnotesize $291,875$&  \footnotesize$123,287$&  \footnotesize$896,782$&  \footnotesize $7.3$ \\
\hline
\footnotesize MS COCO17 & \footnotesize$ 80$ & \footnotesize$118,287$ & \footnotesize$ 5,000$& \footnotesize$40,670$& \footnotesize$860,001$& \footnotesize$36,781$ &  \footnotesize$123,287$&  \footnotesize $896,782$&  \footnotesize $7.3$ \\
\hline
\footnotesize MS COCO18 & \footnotesize$ 80$ & \footnotesize$118,287$ & \footnotesize$ 5,000$& \footnotesize$40,670$& \footnotesize$860,001$& \footnotesize$36,781$&  \footnotesize$123,287$&  \footnotesize $896,782$&  \footnotesize $7.3$ \\
\Xhline{1pt}
\multicolumn{10}{!{\vrule width1.2bp}c!{\vrule width1.2bp}}{\footnotesize Open Images Challenge Object Detection (OICOD) (Based on Open Images V4 \cite{Kuznetsova2018Open})}\\
\hline
\footnotesize OICOD18 & \footnotesize	$500$& \footnotesize	 $1,643,042$	& \footnotesize $100,000$ & \footnotesize$99,999$ & \footnotesize $11,498,734$  & \footnotesize $696,410$ &  \footnotesize $1,743,042$&  \footnotesize $12,195,144$&  \footnotesize $7.0$\\
\Xhline{1pt}
\end{tabular}
}
\end{table*}

\subsection{Evaluation Criteria}
\label{sec:EvaluationCriteria}
There are three criteria for evaluating the performance
of detection algorithms: detection speed in Frames Per Second (FPS), precision, and recall.
The most commonly used metric is \emph{Average Precision} (AP), derived from precision and recall.
AP is usually evaluated in a category specific manner, \emph{i.e.}, computed for each object
category separately. To compare performance over all object categories, the \emph{mean AP} (mAP) averaged over all object categories is adopted as the final measure of performance\footnote{In object detection challenges, such as PASCAL VOC and ILSVRC, the winning entry of
each object category is that with the highest AP score, and the winner of the challenge is the team that
wins on the most object categories. The mAP is also used as the measure of a team's performance, and is
justified since the ranking of teams by mAP was always the same as the ranking by the number of object categories won \cite{Russakovsky2015}.}. More details on these metrics can be found
in \cite{Everingham2010,Everingham2015,Russakovsky2015,Hoiem2012}.

\begin {figure}[!t]
\centering
\includegraphics[width=0.49\textwidth]{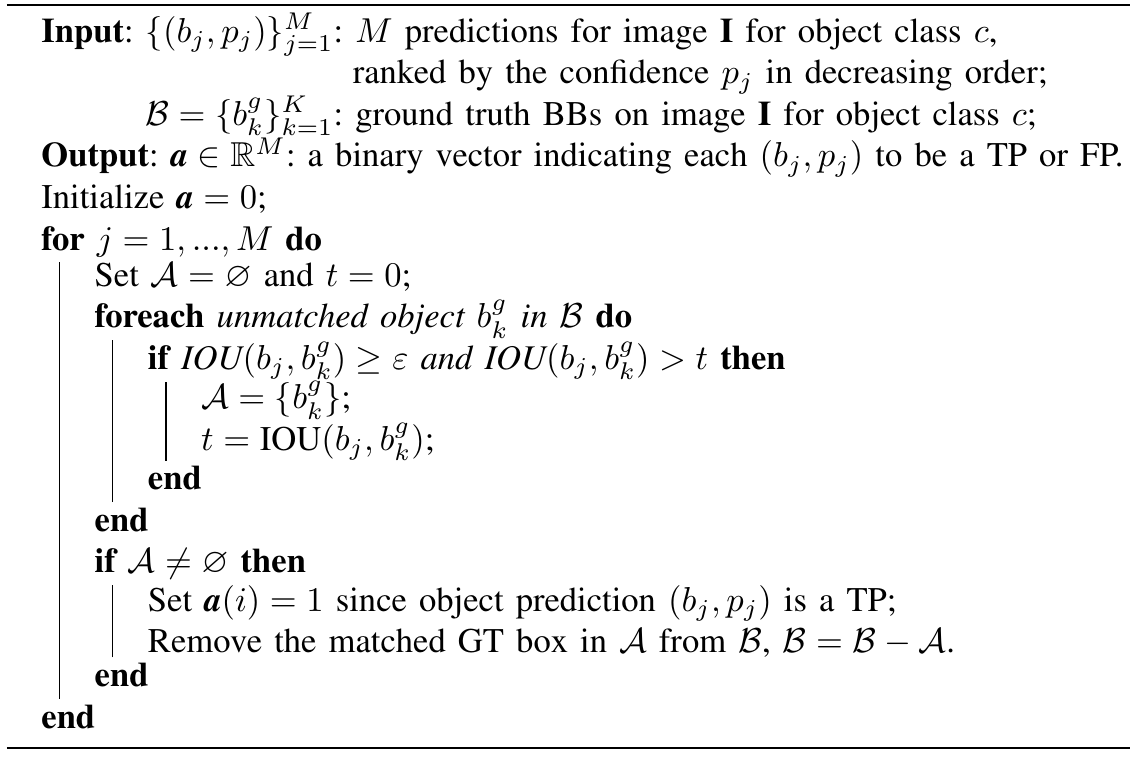}
\caption{The algorithm for determining TPs and FPs by greedily matching object detection results to ground truth boxes.}
\label{fig:Algorithm1}
\end {figure}

The standard outputs of a detector applied to a testing image $\textbf{I}$ are the predicted detections
$\{(b_j,c_j,p_j)\}_j$, indexed by object $j$, of Bounding Box (BB) $b_j$, predicted category $c_j$, and confidence $p_j$. A predicted detection $(b,c,p)$
is regarded as a True Positive (TP) if
\begin{itemize}
\renewcommand{\labelitemi}{$\bullet$}
  \item The predicted category $c$ equals the ground truth label $c_g$.
  \item The overlap ratio IOU (Intersection Over Union) \cite{Everingham2010,Russakovsky2015}
\begin{equation}\label{eqn:IOU}
    \textrm{IOU}(b,b^g)=\frac{area(b\cap b^g)}{area(b\cup b^g)},
\end{equation}
between the predicted BB $b$ and the ground truth $b^g$ is not smaller than a predefined threshold $\varepsilon$, where $\cap$ and $cup$ denote intersection and union, respectively.
A typical value of $\varepsilon$ is 0.5.
\end{itemize}
Otherwise, it is considered as a False Positive (FP). The confidence level $p$
is usually compared with some threshold $\beta$ to determine whether the predicted class label
$c$ is accepted.

AP is computed separately for each of the object classes, based on \emph{Precision} and \emph{Recall}.
For a given object class $c$ and a testing image $\textbf{I}_i$, let
$\{(b_{ij},p_{ij})\}_{j=1}^M$ denote the detections returned by a detector, ranked by confidence $p_{ij}$ in decreasing order.  Each detection $(b_{ij},p_{ij})$ is either a TP or an FP, which can be determined via the algorithm\footnote{It is worth noting that for a given threshold $\beta$, multiple detections of the same object in an image are not considered as all correct detections, and only the detection with the highest confidence level is considered as
a TP and the rest as FPs.} in Fig.~\ref{fig:Algorithm1}. Based on the TP and FP detections,
the precision $P(\beta)$ and recall $R(\beta)$ \cite{Everingham2010} can be computed as a function of the confidence threshold $\beta$,
so by varying the confidence threshold different pairs $(P,R)$ can be obtained, in principle allowing precision to be regarded as a function of recall, \emph{i.e.} $P(R)$,
from which the Average Precision (AP) \cite{Everingham2010,Russakovsky2015} can be found.

Since the introduction of MS COCO, more attention has been placed on the accuracy of the bounding box location. Instead of using a fixed IOU
threshold, MS COCO introduces a few metrics (summarized in Table~\ref{Tab:Metrics}) for characterizing the performance of an object detector. For instance, in contrast to the traditional mAP computed at a single IoU of $0.5$, $AP_{coco}$ is averaged across all object categories and multiple IOU values from $0.5$ to $0.95$ in steps of $0.05$.  Because $41\%$ of the objects in MS COCO are small and $24\%$ are large, metrics $AP_{coco}^{small}$, $AP_{coco}^{medium}$ and $AP_{coco}^{large}$ are also introduced. Finally, Table~\ref{Tab:Metrics} summarizes the main metrics used in the PASCAL, ILSVRC and MS COCO object detection challenges, with metric modifications for the Open Images challenges proposed in \cite{Kuznetsova2018Open}.

\begin{table}[!t]
\caption {Summary of commonly used metrics for evaluating object detectors.}\label{Tab:Metrics}
\centering
\renewcommand{\arraystretch}{1.2}
\setlength\arrayrulewidth{0.2mm}
\setlength\tabcolsep{1pt}
\resizebox*{9cm}{!}{
\begin{tabular}{!{\vrule width1.2bp}c|c|l|l!{\vrule width1.2bp}}
\Xhline{1pt}
\footnotesize Metric  & \footnotesize Meaning  &  \multicolumn{2}{c!{\vrule width1.2bp}}{\footnotesize Definition and Description} \\
\hline
\raisebox{1ex}[0pt]{\footnotesize TP}  & \footnotesize \shortstack [c] {True \\ Positive}  &  \multicolumn{2}{l!{\vrule width1.2bp}}{\raisebox{1ex}[0pt]{\footnotesize A true positive detection, per Fig.~\ref{fig:Algorithm1}.}} \\
\hline
\raisebox{1ex}[0pt]{\footnotesize FP}  & \footnotesize \shortstack [c] {False \\ Positive}  &  \multicolumn{2}{l!{\vrule width1.2bp}}{\raisebox{1ex}[0pt]{\footnotesize A false positive detection, per Fig.~\ref{fig:Algorithm1}.}} \\
\hline
\raisebox{1ex}[0pt]{\footnotesize $\beta$ } & \footnotesize \shortstack [c] {Confidence \\ Threshold}  &  \multicolumn{2}{l!{\vrule width1.2bp}}{\raisebox{1ex}[0pt]{\footnotesize A confidence threshold for computing $P(\beta)$ and $R(\beta)$.}} \\
\hline
\multirow{3}{*}{\footnotesize $\varepsilon$}  & \multirow{3}{*}{\footnotesize \shortstack [c] {IOU \\ Threshold}}   & \footnotesize VOC & \footnotesize Typically around $0.5$ \\
\cline{3-4}
& \footnotesize & \footnotesize ILSVRC & \footnotesize $\min(0.5,\frac{wh}{(w+10)(h+10)})$; $w\times h$ is the size of a GT box. \\
\cline{3-4}
& \footnotesize  & \footnotesize MS COCO & \footnotesize Ten IOU thresholds $\varepsilon\in\{0.5:0.05:0.95\}$  \\
\hline
\footnotesize $P(\beta)$  & \raisebox{1.3ex}[0pt]{\footnotesize \shortstack [c] {Precision}}  &  \multicolumn{2}{l!{\vrule width1.2bp}}{\footnotesize \shortstack [l] {The fraction of correct detections out of the total detections returned \\ by the detector with confidence of at least $\beta$.}} \\
\hline
\footnotesize $R(\beta)$  & \raisebox{1.3ex}[0pt]{\footnotesize \shortstack [c] {Recall}}&  \multicolumn{2}{l!{\vrule width1.2bp}}{\footnotesize \shortstack [l] {The fraction of all $N_c$ objects detected by the detector having a \\ confidence of at least $\beta$.}} \\
\hline
\raisebox{1ex}[0pt]{\footnotesize AP } &  \footnotesize \shortstack [c] {Average \\ Precision} &  \multicolumn{2}{l!{\vrule width1.2bp}}{\footnotesize \shortstack [l] {Computed over the different levels of recall achieved by varying \\ the confidence $\beta$.}} \\
\hline
\multirow{8}{*}{\footnotesize mAP}  & \multirow{8}{*}{\footnotesize \shortstack [c] {mean \\Average\\Precision}}   & \footnotesize VOC & \footnotesize AP at a single IOU and averaged over all classes. \\
\cline{3-4}
& \footnotesize & \footnotesize ILSVRC & \footnotesize AP at a modified IOU and averaged over all classes. \\
\cline{3-4}
& \footnotesize & \multirow{6}{*}{\footnotesize  MS COCO}& \footnotesize $\bullet AP_{coco}$: mAP averaged over ten IOUs: $\{0.5:0.05:0.95\}$;\\
& \footnotesize& \footnotesize& \footnotesize$\bullet$  $AP^{\textrm{IOU}=0.5}_{coco}$: mAP at IOU=0.50 (PASCAL VOC metric);\\
& \footnotesize& \footnotesize& \footnotesize$\bullet$  $AP^{\textrm{IOU}=0.75}_{coco}$: mAP at IOU=0.75 (strict metric);\\
& \footnotesize& \footnotesize& \footnotesize$\bullet$  $AP^{\textrm{small}}_{coco}$: mAP for small objects of area smaller than $32^2$;\\
& \footnotesize& \footnotesize& \footnotesize$\bullet$  $AP^{\textrm{medium}}_{coco}$: mAP for objects of area between $32^2$ and $96^2$;\\
& \footnotesize& \footnotesize& \footnotesize$\bullet$  $AP^{\textrm{large}}_{coco}$: mAP for large objects of area bigger than $96^2$; \\
\hline
\raisebox{1ex}[0pt]{\footnotesize AR}  & \footnotesize \shortstack [c] {Average \\ Recall}  &  \multicolumn{2}{c!{\vrule width1.2bp}}{\footnotesize \shortstack [l] {The maximum recall given a fixed number of detections per image, \\ averaged over all categories and IOU thresholds.}} \\
\hline
\multirow{6}{*}{\footnotesize AR}  & \multirow{6}{*}{\footnotesize \shortstack [c] {Average\\Recall}}   & \multirow{6}{*}{\footnotesize  MS COCO}& \footnotesize $\bullet AR^{\textrm{max}=1}_{coco}$: AR given 1 detection per image;\\
& \footnotesize& \footnotesize& \footnotesize$\bullet$  $AR^{\textrm{max}=10}_{coco}$: AR given 10 detection per image;\\
& \footnotesize& \footnotesize& \footnotesize$\bullet$  $AR^{\textrm{max}=100}_{coco}$: AR given 100 detection per image;\\
& \footnotesize& \footnotesize& \footnotesize$\bullet$  $AR^{\textrm{small}}_{coco}$: AR for small objects of area smaller than $32^2$;\\
& \footnotesize& \footnotesize& \footnotesize$\bullet$  $AR^{\textrm{medium}}_{coco}$: AR for objects of area between $32^2$ and $96^2$;\\
& \footnotesize& \footnotesize& \footnotesize$\bullet$  $AR^{\textrm{large}}_{coco}$: AR for large objects of area bigger than $96^2$; \\
\Xhline{1pt}
\end{tabular}
}
\end{table}

\section{Detection Frameworks}
\label{Sec:Frameworks}

There has been steady progress in object feature representations and classifiers for recognition, as evidenced by the  dramatic change from handcrafted features \cite{Viola2001,Dalal2005HOG,Felzenszwalb08CVPR,
Harzallah2009Combining,Vedaldi09Multiple} to learned DCNN
features \cite{Girshick2014RCNN,Ouyang2015deepid,Girshick2015FRCNN,
Ren2015NIPS,Dai2016RFCN}.  In contrast, in terms of localization, the basic ``sliding window''  strategy \cite{Dalal2005HOG,Felzenszwalb2010b,Felzenszwalb08CVPR}
remains mainstream, although with some efforts to avoid exhaustive search \cite{lampert2008beyond,Uijlings2013b}. However, the number of windows is large and grows
quadratically with the number of image pixels, and the need to search over multiple scales and aspect ratios
further increases the search space. Therefore, the design of efficient and effective detection frameworks plays a key role in reducing this computational cost. Commonly adopted strategies include cascading, sharing feature computation, and reducing per-window computation.

\begin {figure*}[!t]
\centering
\includegraphics[width=0.9\textwidth]{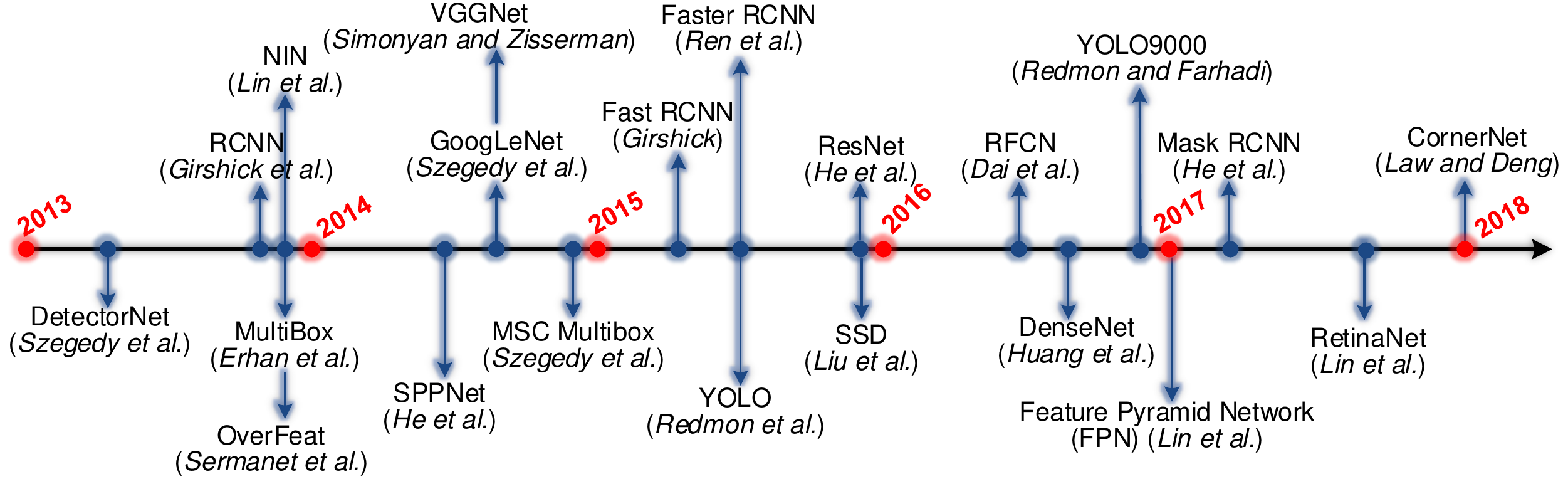}
\caption{Milestones in generic object detection.}
\label{fig:MilestonesAfter2014}
\end {figure*}

This section reviews detection frameworks, listed in Fig.~\ref{fig:MilestonesAfter2014} and Table~\ref{Tab:Detectors}, the milestone approaches appearing since deep learning entered the field, organized into two main categories:
 \begin{enumerate}
 \item [a.] Two stage detection frameworks, which include a preprocessing step for generating object proposals;
 \item [b.] One stage detection frameworks, or region proposal free frameworks, having a single proposed method which does not separate the process of the detection proposal.
 \end{enumerate}
Sections~\ref{Sec:DCNNFeatures} through~\ref{sec:otherissue} will discuss fundamental sub-problems involved in  detection frameworks in greater detail, including DCNN features, detection proposals, and context modeling.

\subsection{Region Based (Two Stage) Frameworks}
\label{Sec:RegionBased}
In a region-based framework, category-independent region proposals\footnote{Object proposals, also called region proposals or detection proposals, are a set of candidate regions or bounding boxes in an image that may potentially contain an object. \cite{Chavali2016,Hosang2016}} are generated from an image, CNN \cite{Krizhevsky2012} features are extracted from these regions, and then category-specific classifiers are used
to determine the category labels of the proposals. As can be observed from Fig.~\ref{fig:MilestonesAfter2014}, DetectorNet \cite{Szegedy2013Deep}, OverFeat \cite{OverFeat2014}, MultiBox \cite{MultiBox1} and RCNN \cite{Girshick2014RCNN} independently and almost simultaneously proposed using CNNs for generic object detection.

\begin {figure}[!t]
\centering
\includegraphics[width=0.5\textwidth]{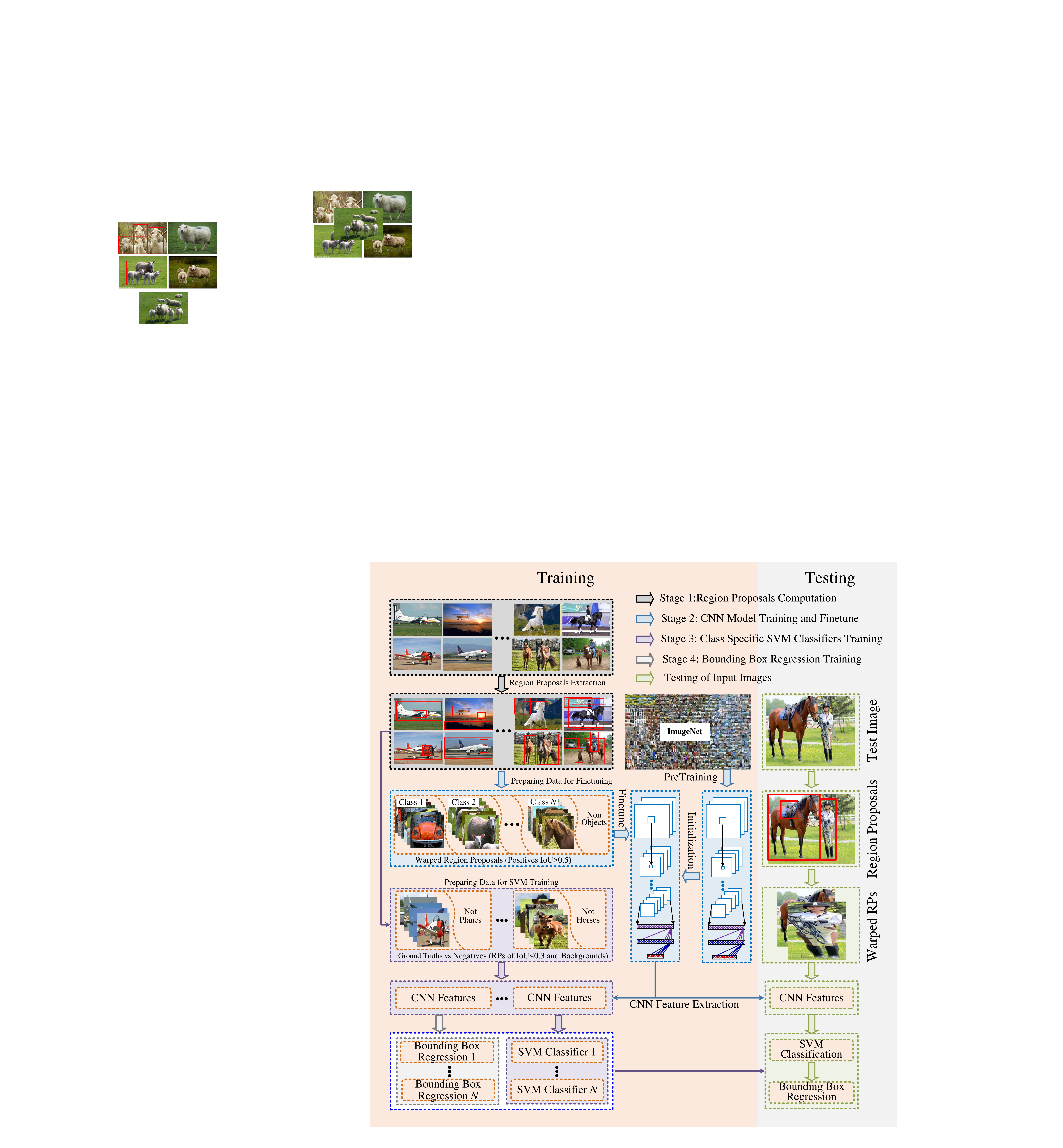}
\caption{Illustration of the RCNN detection framework \cite{Girshick2014RCNN,Girshick2016TPAMI}.}
\label{fig:RegionBased}
\end {figure}

\begin {figure}[!t]
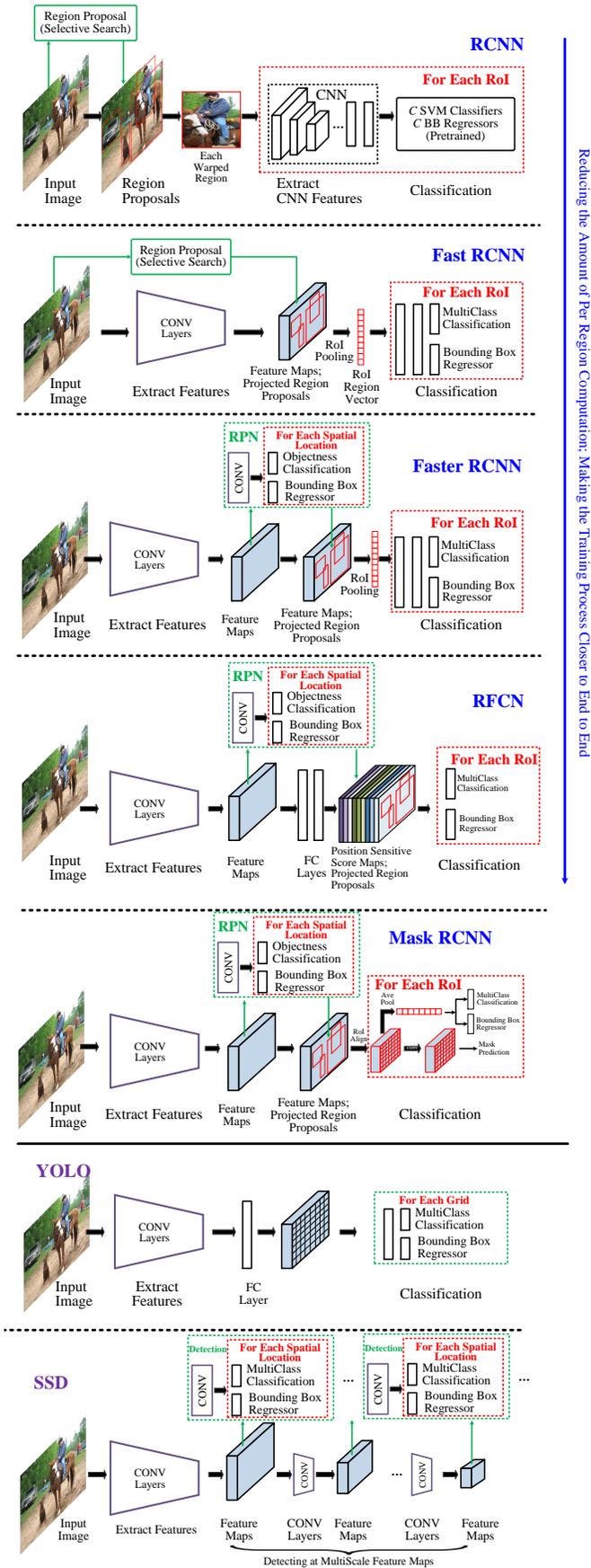

\centering
\includegraphics[width=0.46\textwidth]{RegionVsUnified1.pdf}
\includegraphics[width=0.46\textwidth]{RegionVsUnified2.pdf}
\caption{High level diagrams of the leading frameworks for generic object detection. The properties of these methods are summarized in Table \ref{Tab:Detectors}.}
\label{Fig:RegionVsUnified}
\end {figure}

\textbf{RCNN} \cite{Girshick2014RCNN}: Inspired by the breakthrough image classification results obtained by CNNs and the success of the selective search in region proposal for handcrafted features \cite{Uijlings2013b}, Girshick \emph{et al.} were among the first to explore CNNs for generic object detection and developed RCNN \cite{Girshick2014RCNN,Girshick2016TPAMI}, which integrates AlexNet \cite{Krizhevsky2012} with a region proposal selective search \cite{Uijlings2013b}.  As illustrated in detail in Fig.~\ref{fig:RegionBased}, training an RCNN framework consists of multistage pipelines:
\begin{enumerate}
\item \emph{Region proposal computation:} Class agnostic region proposals, which are candidate regions that might contain objects, are obtained via a selective search \cite{Uijlings2013b}.
\item \emph{CNN model finetuning:} Region proposals, which are cropped  from the image and warped into the same size, are used as the input for fine-tuning a CNN model pre-trained using a large-scale dataset such as ImageNet. At this stage, all region proposals with $\geqslant0.5$ IOU \footnote{Please refer to Section \ref{sec:EvaluationCriteria} for the definition of IOU.} overlap with a ground truth box are defined as positives for that ground truth box's class and the rest as negatives.
\item \emph{Class specific SVM classifiers training:} A set of class-specific linear SVM classifiers are trained using fixed length features extracted with CNN, replacing the softmax classifier learned by fine-tuning. For training SVM classifiers, positive examples are
defined to be the ground truth boxes for each class. A region proposal with less than 0.3 IOU overlap with all ground truth instances of a class is negative for that
class. Note that the positive and negative examples defined for training the SVM classifiers are different from those for fine-tuning the CNN.
\item \emph{Class specific bounding box regressor training:} Bounding box regression is learned for each object class with CNN features.
\end{enumerate}
In spite of achieving high object detection quality, RCNN has notable drawbacks \cite{Girshick2015FRCNN}:
\begin{enumerate}
\item Training is a multistage pipeline, slow and hard to optimize because each individual stage must be trained separately.
\item For SVM classifier and bounding box regressor training, it is expensive in both disk space and time, because CNN features need to be extracted from each object proposal in each image, posing great challenges for large scale detection, particularly with very deep networks, such as
VGG16 \cite{Simonyan2014VGG}.
\item Testing is slow, since CNN features are extracted per object proposal in each test image, without shared computation.
\end{enumerate}
All of these drawbacks have motivated successive innovations, leading to a number of improved detection frameworks such as SPPNet, Fast RCNN, Faster RCNN \emph{etc}., as follows.

\textbf{SPPNet} \cite{He2014SPP}: During testing, CNN feature extraction is the main bottleneck of the RCNN detection pipeline, which requires the extraction of CNN features from thousands of warped region proposals per image.  As a result, He \emph{et al.} \cite{He2014SPP} introduced  traditional spatial pyramid pooling (SPP) \cite{Grauman2005pyramid,Lazebnik2006SPM} into CNN architectures.
Since  convolutional layers accept inputs of arbitrary sizes, the requirement of fixed-sized images in CNNs is due only to the Fully Connected (FC) layers, therefore He \emph{et al.} added an SPP layer on top of the last convolutional (CONV) layer to obtain features of fixed length for the FC layers. With this SPPNet, RCNN obtains a significant speedup without sacrificing any detection quality, because it only needs to run the convolutional layers {\emph once} on the entire test image to generate fixed-length features for region proposals of arbitrary size. While SPPNet accelerates RCNN evaluation by orders of magnitude, it does not result in a comparable speedup of the detector training.  Moreover, fine-tuning in SPPNet \cite{He2014SPP} is unable to update the convolutional layers before the SPP layer, which limits the accuracy of very deep networks.

\textbf{Fast RCNN} \cite{Girshick2015FRCNN}:
Girshick proposed Fast RCNN \cite{Girshick2015FRCNN} that addresses some of the disadvantages of RCNN and SPPNet, while improving on their detection speed and quality.  As illustrated in Fig.~\ref{Fig:RegionVsUnified}, Fast RCNN enables end-to-end detector training by developing a streamlined training process that
simultaneously learns a softmax classifier and class-specific bounding
box regression, rather than separately training a softmax
classifier, SVMs, and Bounding Box Regressors (BBRs) as in RCNN/SPPNet.
Fast RCNN employs the idea of sharing the computation of convolution
across region proposals, and adds a Region of Interest (RoI) pooling layer
between the last CONV layer and the first FC layer to extract a fixed-length
feature for each region proposal.
Essentially, RoI pooling uses warping at the feature level to approximate warping at the image level. The features after the RoI pooling layer are fed into a sequence of FC layers that finally branch into two sibling output layers: softmax probabilities for object category prediction, and class-specific bounding box regression offsets for proposal refinement. Compared to RCNN/SPPNet, Fast RCNN improves the efficiency considerably -- typically 3 times faster in training and 10 times faster in testing. Thus there is higher detection quality, a single training process that updates all network layers, and no storage required for feature caching.

\textbf{Faster RCNN} \cite{Ren2015NIPS,Ren2016a}:
Although Fast RCNN significantly sped up the detection process,
it still relies on external region proposals, whose computation
is exposed as the new speed bottleneck in Fast RCNN.
Recent work has shown that CNNs have a remarkable ability to
localize objects in CONV layers \cite{Zhoubolei2014,Zhou2016learning,
Cinbis2017,Oquab2015object,Hariharan2016}, an ability which is weakened in the
FC layers. Therefore, the selective search can be replaced by a CNN in producing region proposals.
The Faster RCNN framework proposed by Ren \emph{et al.} \cite{Ren2015NIPS,Ren2016a}
offered an efficient and accurate Region Proposal Network (RPN) for
generating region proposals. They utilize the same backbone network, using features from the last shared convolutional
layer to accomplish the task of RPN for region proposal and Fast RCNN for region classification, as
shown in Fig.~\ref{Fig:RegionVsUnified}.

RPN first initializes $k$ reference boxes (\emph{i.e.} the so called \emph{anchors}) of different scales and aspect ratios at each CONV feature map location. The anchor {\emph positions} are image content independent, but the feature vectors themselves, extracted from anchors, are image content dependent. Each anchor is mapped to a lower dimensional vector, which is fed into two sibling FC layers --- an object category classification layer and a box regression layer.  In contrast to detection in Fast RCNN, the features used for regression in RPN are of the same shape as the
anchor box, thus $k$ anchors lead to $k$ regressors.  RPN shares CONV features with Fast RCNN, thus enabling highly efficient region proposal computation. RPN is, in fact, a kind of Fully Convolutional Network (FCN) \cite{FCNCVPR2015,FCNTPAMI}; Faster RCNN is thus a purely CNN based framework without using handcrafted features.

For the VGG16 model \cite{Simonyan2014VGG}, Faster RCNN can test at 5 FPS (including all stages) on a GPU, while achieving state-of-the-art object detection accuracy on PASCAL VOC 2007 using 300 proposals per image. The initial Faster RCNN in \cite{Ren2015NIPS} contains several alternating training stages, later simplified in \cite{Ren2016a}.

Concurrent with the development of Faster RCNN, Lenc and Vedaldi \cite{Lenc2015} challenged the role of region proposal generation methods such as selective search, studied the role of region proposal generation in CNN based detectors, and found that CNNs contain sufficient geometric information for accurate object detection in the CONV rather than FC layers. They showed the possibility of building integrated, simpler, and faster object detectors that rely exclusively on CNNs, removing region proposal generation methods such as selective search.

{\textbf{RFCN (Region based Fully Convolutional Network)}}:  While Faster RCNN is an order of magnitude faster than Fast RCNN, the fact that the region-wise sub-network still needs to be applied per RoI (several hundred RoIs per image) led Dai \emph{et al.} \cite{Dai2016RFCN} to propose the RFCN detector which is \emph{fully convolutional} (no hidden FC layers) with almost all computations shared over the entire image. As shown in Fig.~\ref{Fig:RegionVsUnified}, RFCN differs from Faster RCNN only in the RoI sub-network. In Faster RCNN, the computation after the RoI pooling layer cannot be shared, so Dai \emph{et al.} \cite{Dai2016RFCN} proposed using all CONV layers to construct a shared RoI sub-network, and RoI crops are taken from the last layer of CONV features prior to prediction. However, Dai \emph{et al.} \cite{Dai2016RFCN} found that this naive design turns out to have considerably inferior detection accuracy, conjectured to be that deeper CONV layers are more sensitive to category semantics, and less sensitive to translation, whereas object detection needs localization representations that respect translation invariance. Based on this observation, Dai \emph{et al.} \cite{Dai2016RFCN} constructed a set of position-sensitive score maps by using a bank of specialized CONV layers as the FCN output, on top of which a position-sensitive RoI pooling layer is added. They showed that RFCN with ResNet101 \cite{He2016ResNet} could achieve comparable accuracy to Faster RCNN, often at faster running times.

\textbf{Mask RCNN}:  He \emph{et al.} \cite{MaskRCNN2017} proposed Mask RCNN to tackle pixelwise object instance segmentation by extending Faster RCNN. Mask RCNN adopts the same two stage pipeline, with an identical first stage (RPN), but in the second stage, in parallel to predicting the class and box offset, Mask RCNN adds a branch which outputs a binary mask for each RoI. The new branch is a Fully Convolutional Network (FCN) \cite{FCNCVPR2015,FCNTPAMI} on top of a CNN feature map. In order to avoid the misalignments caused by the original RoI pooling (RoIPool) layer, a RoIAlign layer was proposed to preserve the pixel level spatial correspondence. With a backbone network ResNeXt101-FPN \cite{Xie2016Aggregated,FPN2016}, Mask RCNN achieved top results for the COCO object instance segmentation and bounding box
object detection. It is simple to train, generalizes well, and adds only a small overhead to Faster RCNN, running at 5 FPS \cite{MaskRCNN2017}.

\textbf{Chained Cascade Network and Cascade RCNN}:  The essence of cascade \cite{Felzenszwalb2010Cascade,Bourdev2005Robust,Li2004Floatboost} is to learn more discriminative classifiers by using multistage
classifiers, such that early stages discard a large number
of easy negative samples so that later stages
can focus on handling more difficult examples.  Two-stage object detection can be considered as a cascade, the first  detector removing large amounts of background, and the second stage classifying the remaining regions. Recently, end-to-end learning of more than two cascaded classifiers and DCNNs for generic object detection were proposed in the Chained Cascade Network \cite{Ouyang2017Chained}, extended in Cascade RCNN \cite{CascadeRCNN2018}, and more recently applied for simultaneous object detection and instance segmentation \cite{Chen2019Hybrid}, winning the COCO 2018 Detection Challenge.

\textbf{Light Head RCNN}:
In order to further increase the detection speed of RFCN \cite{Dai2016RFCN},
Li \emph{et al.} \cite{Li2018Light} proposed Light Head RCNN, making the head of the detection network as light as possible to reduce the RoI computation.  In particular, Li \emph{et al.} \cite{Li2018Light} applied a convolution to produce thin feature maps with small channel numbers (\emph{e.g.,} 490 channels for COCO) and a
cheap RCNN sub-network, leading to an excellent trade-off of speed and accuracy.

\subsection{Unified (One Stage) Frameworks}
\label{Sec:Unified}
The region-based pipeline strategies of Section~\ref{Sec:RegionBased}
have dominated
since RCNN \cite{Girshick2014RCNN}, such that the leading results on popular benchmark
datasets are all based on Faster RCNN \cite{Ren2015NIPS}.
Nevertheless, region-based approaches are computationally
expensive for current mobile/wearable devices, which have limited storage and computational capability,
therefore instead of trying to optimize the individual components of a complex region-based pipeline, researchers have begun to develop \emph{unified} detection strategies.

Unified pipelines refer to architectures that directly predict class probabilities and bounding box offsets from full images with a single feed-forward CNN in a monolithic setting that does not involve region proposal generation or post classification / feature resampling, encapsulating all computation in a single network. Since the whole pipeline is a single network, it can be optimized end-to-end directly on detection performance.

\textbf{DetectorNet}: Szegedy \emph{et al.} \cite{Szegedy2013Deep} were among the first to explore CNNs for object detection. DetectorNet formulated object detection a regression problem to object bounding box masks.  They use AlexNet \cite{Krizhevsky2012} and replace the final softmax classifier layer with a regression layer. Given an image window, they use one network to predict foreground pixels over a coarse grid, as well as four additional networks to predict the object's top, bottom, left and right halves. A grouping process then converts the predicted masks into
detected bounding boxes. The network needs to be trained per object type and mask type, and does not scale to multiple classes. DetectorNet must take many crops of the image, and run multiple networks for each part on every crop, thus making it slow.

\begin {figure}[!t]
\centering
\includegraphics[width=0.5\textwidth]{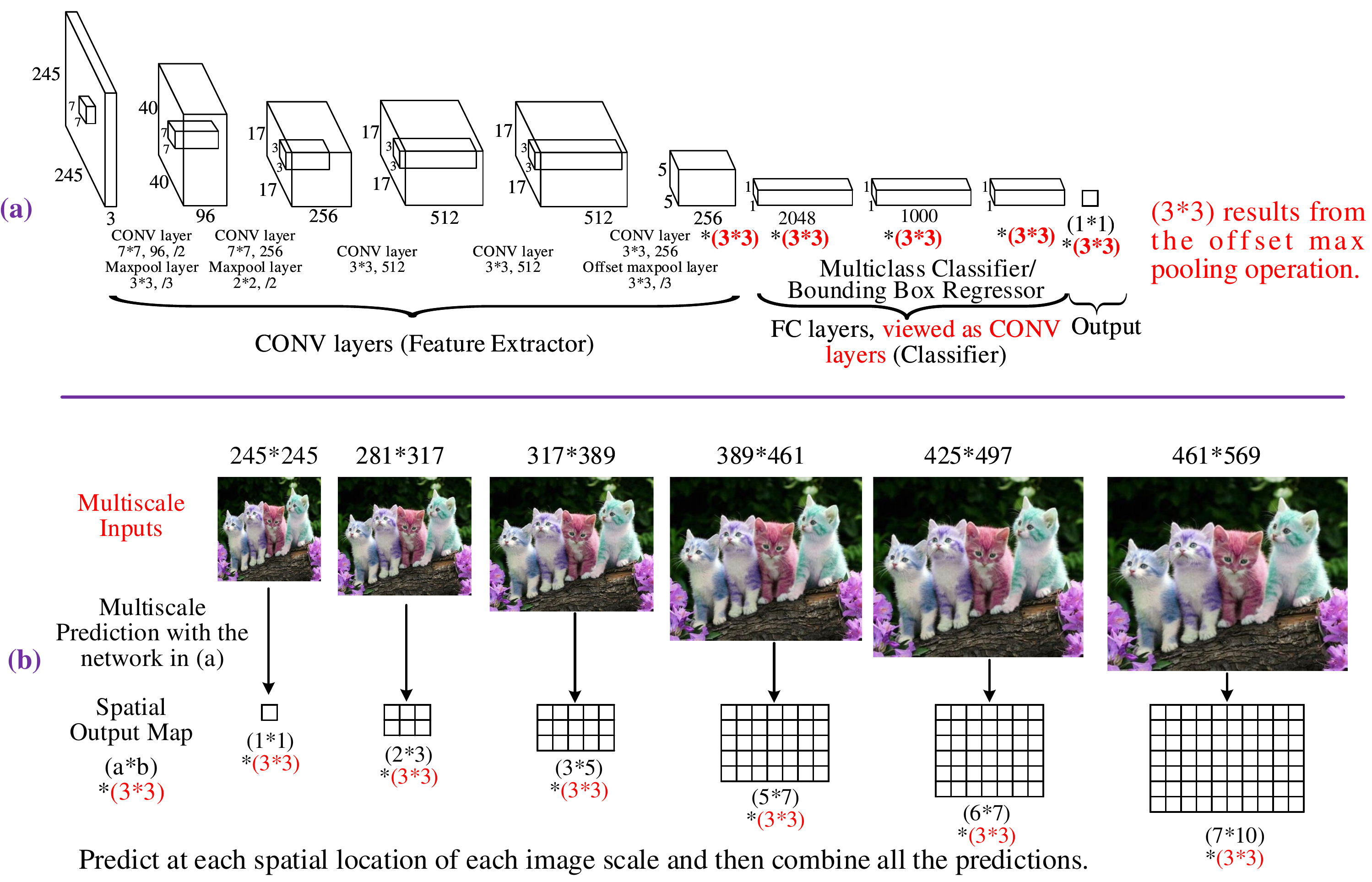}
\caption{Illustration of the OverFeat \cite{OverFeat2014} detection framework.}
\label{fig:OverFeat}
\end {figure}
\textbf{OverFeat}, proposed by Sermanet \emph{et al.} \cite{OverFeat2014} and illustrated in Fig.~\ref{fig:OverFeat}, can be considered as one of the first single-stage object detectors based on fully convolutional deep networks.  It is one of the most influential object detection frameworks, winning the ILSVRC2013
localization and detection competition.  OverFeat performs object detection via a single forward pass through
 the fully convolutional layers in the network (\emph{i.e.} the ``Feature Extractor", shown in Fig. \ref{fig:OverFeat} (a)). The key steps of object detection at test time can be summarized as follows:
\begin{enumerate}
  \item \emph{Generate object candidates by performing object classification via a sliding window fashion on multiscale images.} OverFeat uses a CNN like AlexNet \cite{Krizhevsky2012}, which would require input images ofa  fixed size due to its fully connected layers, in order to make the sliding window approach computationally efficient, OverFeat casts the network (as shown in Fig. \ref{fig:OverFeat} (a)) into a fully convolutional network, taking inputs of any size, by viewing fully connected layers as convolutions with kernels of size $1\times1$. OverFeat leverages multiscale features to improve the overall performance by passing up to six enlarged scales of the original image through the network (as shown in Fig. \ref{fig:OverFeat} (b)), resulting in a significantly increased number of evaluated context views. For each of the multiscale inputs, the classifier outputs a grid of predictions (class and confidence).
  \item \emph{Increase the number of predictions by offset max pooling}. In order to increase resolution, OverFeat applies offset max pooling after the last CONV layer, \emph{i.e.} performing a subsampling operation at every offset, yielding many more views for voting, increasing robustness while remaining efficient.
  \item \emph{Bounding box regression.} Once an object is identified, a single bounding box regressor is applied. The classifier and the regressor share the same feature extraction (CONV) layers, only the FC layers need to be recomputed after computing the classification network.
\item \emph{Combine predictions.} OverFeat uses a greedy merge strategy to combine the individual bounding box predictions across all locations and scales.
\end{enumerate}
OverFeat has a significant speed advantage, but is less accurate than RCNN \cite{Girshick2014RCNN}, because it was difficult to train fully convolutional networks at the time.  The speed advantage derives from sharing the computation of convolution between overlapping windows in the fully convolutional network. OverFeat is similar to later frameworks such as YOLO \cite{YoLo2016} and SSD \cite{Liu2016SSD}, except that the classifier and the regressors in OverFeat are trained sequentially.

\textbf{YOLO}: Redmon \emph{et al.} \cite{YoLo2016} proposed YOLO (You Only Look Once), a unified detector casting object detection as a regression
problem from image pixels to spatially separated bounding boxes and associated class probabilities, illustrated in Fig.~\ref{Fig:RegionVsUnified}.
Since the region proposal generation stage is completely dropped,
YOLO directly predicts detections using a small set of candidate regions\footnote{YOLO uses far fewer bounding boxes, only 98
per image, compared to about 2000 from Selective Search.}.
Unlike region based approaches (\emph{e.g.} Faster RCNN) that predict detections based
on features from a local region, YOLO uses features from an entire image globally.
In particular, YOLO divides an image into an $S\times S$ grid, each predicting $C$ class probabilities,
$B$ bounding box locations, and confidence scores.
By throwing out the region proposal generation step entirely, YOLO is fast by design, running in real time at 45 FPS and Fast YOLO \cite{YoLo2016} at 155 FPS. Since YOLO sees the entire image when making predictions, it implicitly encodes contextual information about object classes, and is less likely to predict false positives in the background. YOLO makes more localization errors than Fast RCNN, resulting from the coarse division of bounding box location, scale and aspect ratio.
As discussed in \cite{YoLo2016}, YOLO may fail to localize some objects, especially small ones, possibly because of the coarse grid division, and because each grid cell can only contain one object.  It is unclear to what extent YOLO can translate to good performance
on datasets with many objects per image, such as MS COCO.

\textbf{YOLOv2 and YOLO9000}: Redmon and Farhadi \cite{YOLO9000} proposed YOLOv2, an improved version of YOLO, in which the custom GoogLeNet \cite{GoogLeNet2015} network is replaced with the simpler DarkNet19, plus batch normalization \cite{He2015delving}, removing the fully connected layers, and using good anchor boxes\footnote{Boxes of various sizes and aspect ratios that serve as object candidates.} learned via \emph{k}means and multiscale training.
YOLOv2 achieved state-of-the-art on standard detection tasks.  Redmon and Farhadi \cite{YOLO9000} also introduced YOLO9000, which can detect over 9000 object categories in real time by proposing a joint optimization method to train simultaneously on an ImageNet classification dataset and a COCO detection dataset with WordTree to combine data from multiple sources. Such joint training allows YOLO9000 to perform weakly supervised detection, \emph{i.e.} detecting object classes that do not have bounding box annotations.

\textbf{SSD}: In order to preserve real-time speed without sacrificing too much detection accuracy, Liu \emph{et al.} \cite{Liu2016SSD} proposed SSD (Single Shot Detector), faster than YOLO \cite{YoLo2016} and with an accuracy competitive with region-based detectors such as Faster RCNN \cite{Ren2015NIPS}. SSD effectively combines ideas from RPN in Faster RCNN \cite{Ren2015NIPS}, YOLO \cite{YoLo2016} and multiscale CONV features \cite{Hariharan2016} to achieve fast detection speed, while still retaining high detection quality. Like YOLO, SSD predicts a fixed number of bounding boxes and scores, followed by an NMS step to produce the final detection. The CNN network in SSD is fully convolutional, whose early layers are based on a standard architecture, such as VGG \cite{Simonyan2014VGG}, followed by several auxiliary CONV layers, progressively decreasing in size.  The information in the last layer may be too coarse spatially to allow precise localization, so SSD performs detection over multiple scales by operating on multiple CONV feature maps, each of which predicts category scores and box offsets for bounding boxes of appropriate sizes.  For a $300\times300$ input, SSD achieves $74.3\%$ mAP on the VOC2007 test at 59 FPS versus Faster RCNN 7 FPS / mAP $73.2\%$ or YOLO 45 FPS / mAP $63.4\%$.

\textbf{CornerNet:} Recently, Law \emph{et al.} \cite{Law2018CornerNet} questioned the dominant role that anchor boxes have come to play in SoA object detection frameworks \cite{Girshick2015FRCNN,MaskRCNN2017,YoLo2016,Liu2016SSD}. Law \emph{et al.} \cite{Law2018CornerNet} argue that the use of anchor boxes, especially in one stage detectors \cite{DSSD2016,LinICCV2017,Liu2016SSD,YoLo2016}, has drawbacks \cite{Law2018CornerNet,LinICCV2017} such as
causing a huge imbalance between positive and negative examples, slowing down training and introducing extra hyperparameters.  Borrowing ideas from the work on Associative Embedding in multiperson pose estimation \cite{Newell2017Associative}, Law \emph{et al.} \cite{Law2018CornerNet} proposed CornerNet by formulating bounding box object detection as detecting paired top-left and bottom-right keypoints\footnote{The idea of using keypoints for object detection appeared previously in DeNet \cite{SmithICCV2017}. }. In CornerNet, the backbone network consists of two stacked Hourglass networks \cite{Newell2016Stacked}, with a simple corner pooling approach to better
localize corners.  CornerNet achieved a $42.1\%$ AP on MS COCO, outperforming all previous one stage detectors; however, the average inference time is about 4FPS on a Titan X GPU, significantly slower than SSD \cite{Liu2016SSD} and YOLO \cite{YoLo2016}. CornerNet generates incorrect bounding boxes because it is challenging to decide which pairs of keypoints should be grouped into the same objects.  To further improve on CornerNet, Duan \emph{et al.} \cite{Duan2019CenterNet} proposed CenterNet to detect each object as a triplet of keypoints, by introducing one extra keypoint at the centre of a proposal, raising the MS COCO AP to $47.0\%$, but with an inference speed slower than CornerNet.


\section{Object Representation}
\label{Sec:DCNNFeatures}
As one of the main components in any
detector, good feature representations are of primary importance in object detection \cite{Dickinson2009,Girshick2014RCNN,Gidaris2015,Zhu2016Do}.
In the past, a great deal of effort was devoted to designing local descriptors (\emph{e.g.,} SIFT \cite{Lowe1999Object} and HOG \cite{Dalal2005HOG}) and to explore approaches (\emph{e.g.,} Bag of Words \cite{Sivic2003} and Fisher Vector \cite{Perronnin2010}) to group and abstract descriptors into higher level representations in order to allow the discriminative parts to emerge; however, these feature representation methods required careful engineering and considerable domain expertise.

In contrast, deep learning methods (especially {\em deep} CNNs) can learn powerful feature representations with multiple levels of abstraction directly from raw images \cite{Bengio13Feature,LeCun15}. As the learning procedure reduces the dependency of specific domain knowledge and complex procedures needed in traditional feature engineering \cite{Bengio13Feature,LeCun15}, the burden for feature representation has been transferred to the design of better network architectures and training procedures.

\begin {figure}[!t]
\centering
\includegraphics[width=0.45\textwidth]{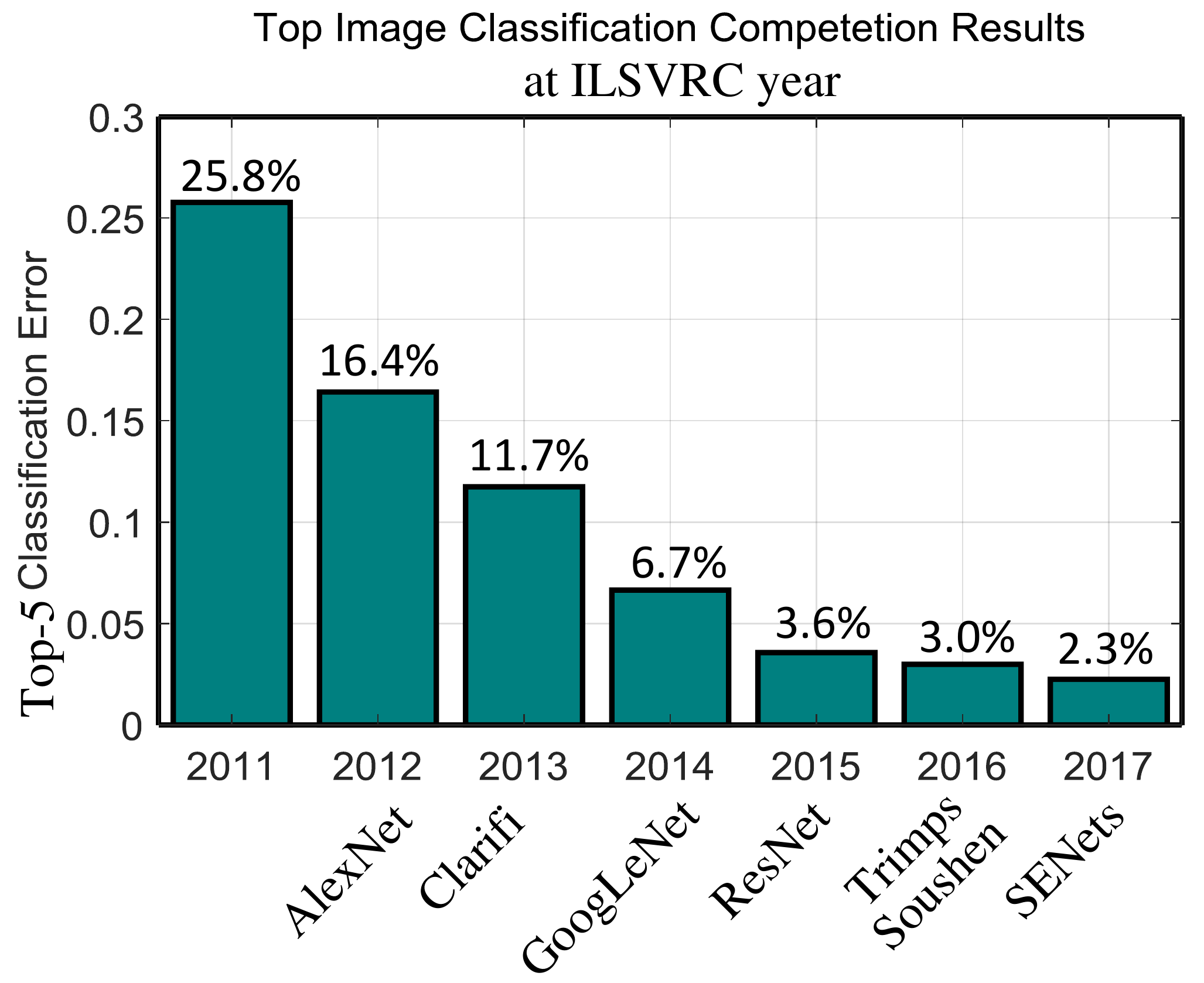}
\caption{\footnotesize{Performance of winning entries in the ILSVRC competitions from 2011 to 2017
in the image classification task.}}
\label{fig:ILSVRCclassificationResults}
\end {figure}

The leading frameworks reviewed in Section \ref{Sec:Frameworks} (RCNN \cite{Girshick2014RCNN}, Fast RCNN \cite{Girshick2015FRCNN}, Faster RCNN \cite{Ren2015NIPS}, YOLO \cite{YoLo2016}, SSD \cite{Liu2016SSD}) have persistently promoted detection accuracy and speed, in which it is generally accepted that the CNN architecture (Section \ref{Sec:PopularNetworks} and Table \ref{fig:ILSVRCclassificationResults}) plays a crucial
role.  As a result, most of the recent improvements in detection accuracy have been via research into the development of novel networks. Therefore we begin by reviewing popular CNN architectures used in Generic Object Detection, followed by a review of the effort devoted to improving object feature representations, such as developing invariant features to accommodate geometric variations in object scale, pose, viewpoint,  part deformation and performing multiscale analysis to improve object detection over a wide range of scales.
\begin{table*}[!t]
\caption {DCNN architectures that were commonly used for generic object detection. Regarding the statistics for ``\#Paras'' and ``\#Layers'', the final FC prediction layer is not taken  into consideration. ``Test Error'' column indicates the Top 5 classification test error on ImageNet1000. When ambiguous, the ``\#Paras'', ``\#Layers'', and ``Test Error'' refer to: OverFeat (accurate model), VGGNet16, ResNet101 DenseNet201 (Growth Rate 32, DenseNet-BC), ResNeXt50 (32*4d), and SE ResNet50.
}\label{Tab:dcnnarchitectures}
\centering
\renewcommand{\arraystretch}{1.2}
\setlength\arrayrulewidth{0.2mm}
\setlength\tabcolsep{2pt}
\resizebox*{18cm}{!}{
\begin{tabular}{!{\vrule width1.2bp}c|c|c|c|c|c|p{9cm}!{\vrule width1.2bp}}
\Xhline{1pt}
\footnotesize No.  & \footnotesize \shortstack [c] {DCNN \\ Architecture}  & \footnotesize \shortstack [c] {\#Paras \\ ($\times10^6$)}	& \footnotesize \shortstack [c] {\#Layers \\ (CONV+FC)} & \footnotesize \shortstack [c] {Test Error \\ (Top 5)}	& \footnotesize \shortstack [c] { \shortstack [c] {First \\ Used In}} &  \footnotesize Highlights \\
\Xhline{1pt}
\raisebox{-3.3ex}[0pt]{$1$}	& \raisebox{-3.3ex}[0pt]{\footnotesize AlexNet \cite{AlexNet2012}}
 & \raisebox{-3.3ex}[0pt]{ \footnotesize	$57$}& \raisebox{-3.3ex}[0pt]{\footnotesize $5+2 $}
  & \raisebox{-3.3ex}[0pt]{\footnotesize $15.3\%$}
 & \raisebox{-3.3ex}[0pt]{\footnotesize \cite{Girshick2014RCNN}} & \footnotesize
 The first DCNN found effective for ImageNet classification; the historical turning point from hand-crafted features to CNN;
Winning the ILSVRC2012 Image classification competition. \\
\hline
$2$ & \footnotesize	ZFNet (fast)
\cite{ZeilerFergus2014}	& \footnotesize $58$ 	& \footnotesize$ 5+2 $ &\footnotesize $14.8\%$ & \footnotesize \cite{He2014SPP}& \footnotesize	Similar to AlexNet, different in stride for convolution, filter size, and number of filters for some layers. \\
\hline
$3 $& \footnotesize	OverFeat \cite{OverFeat2014} & \footnotesize $140	 $& \footnotesize$ 6+2$ &\footnotesize $13.6\%$ & \footnotesize \cite{OverFeat2014} & \footnotesize	Similar to AlexNet, different in  stride for convolution, filter size, and number of filters for some layers. \\
\hline
\raisebox{-1.3ex}[0pt]{$4$}&\raisebox{-1.3ex}[0pt]{ \footnotesize	 VGGNet \cite{Simonyan2014VGG}}	& \raisebox{-1.3ex}[0pt]{\footnotesize$ 134$}	 & \raisebox{-1.3ex}[0pt]{\footnotesize $13+2 $} &
 \raisebox{-1.3ex}[0pt]{\footnotesize $6.8\%$}&\raisebox{-1.3ex}[0pt]{ \footnotesize \cite{Girshick2015FRCNN}}& \footnotesize Increasing network depth significantly by stacking $3\times3$ convolution filters and increasing the network depth step by step. \\
\hline
\raisebox{-2.3ex}[0pt]{$5$}&\raisebox{-2.3ex}[0pt]{ \footnotesize	 GoogLeNet
\cite{GoogLeNet2015} }&\raisebox{-2.3ex}[0pt]{ \footnotesize$ 6$} & \raisebox{-2.3ex}[0pt]{\footnotesize $	 22$ } & \raisebox{-2.3ex}[0pt]{\footnotesize $6.7\%$} &\raisebox{-2.3ex}[0pt]{ \footnotesize \cite{GoogLeNet2015} }& \footnotesize	 Use Inception module, which uses multiple branches of convolutional layers with different filter sizes and then concatenates feature maps produced by these branches. The first inclusion of bottleneck structure and global average pooling.  \\
\hline
$6$	& \footnotesize Inception v2 \cite{Ioffe2015} & \footnotesize	 $12$ & \footnotesize	 $31$ & \footnotesize $4.8\%$ & \footnotesize \cite{Howard2017MobileNets} & \footnotesize Faster training with the introduce of Batch Normalization.\\
\hline
$7$ & \footnotesize	Inception v3
\cite{ Szegedy2016a} & \footnotesize	$22$ & \footnotesize	 $47$  & \footnotesize $3.6\%$ & \footnotesize	  & \footnotesize Inclusion of separable convolution and spatial resolution reduction. \\
\hline
$8$	& \footnotesize YOLONet \cite{YoLo2016} & \footnotesize	$64	 $& \footnotesize $24+1$  & \footnotesize$-$ & \footnotesize	 \cite{YoLo2016} & \footnotesize A network inspired by GoogLeNet used in YOLO detector. \\
\hline
$9$& \footnotesize ResNet50
\cite{He2016ResNet} & \footnotesize$	23.4	$& \footnotesize $49$	  & \footnotesize $3.6\%$ & \footnotesize	 \cite{He2016ResNet} & \footnotesize With identity mapping, substantially deeper networks can be learned. \\
\cline{1-4}\cline{6-6}
$10$	& \footnotesize ResNet101
\cite{He2016ResNet} & \footnotesize	$42$	& \footnotesize $100 $  & \footnotesize (ResNets) & \footnotesize \cite{He2016ResNet} & \footnotesize
Requires fewer parameters than VGG by using the global
average pooling and bottleneck introduced in GoogLeNet. \\
\hline
\raisebox{-1.3ex}[0pt]{$11$} & \raisebox{-1.3ex}[0pt]{ \footnotesize	 InceptionResNet v1
\cite{InceptionV4} } & \raisebox{-1.3ex}[0pt]{\footnotesize	 $21$ }& \raisebox{-1.3ex}[0pt]{\footnotesize $87$}	 &\multirow{3}{*}{ \footnotesize $3.1\%$} &\raisebox{-1.3ex}[0pt]{ \footnotesize}  & \footnotesize Combination of identity mapping and Inception module, with similar computational cost of Inception v3, but faster training process. \\
\cline{1-4}\cline{6-7}
\raisebox{-1.3ex}[0pt]{$12$} &\raisebox{-1.3ex}[0pt]{ \footnotesize	 InceptionResNet v2
\cite{InceptionV4}} &\raisebox{-1.3ex}[0pt]{ \footnotesize	 $30$ }& \raisebox{-1.3ex}[0pt]{\footnotesize	 $95$ }& \raisebox{-0.3ex}[0pt]{ \footnotesize (Ensemble)	 } &\raisebox{-1.3ex}[0pt]{ \footnotesize \cite{Huang2016Speed}} &  \footnotesize A costlier residual version of Inception, with significantly improved recognition performance.  \\
\cline{1-4}\cline{6-7}
\raisebox{-1.3ex}[0pt]{$13 $}&\raisebox{-1.3ex}[0pt]{ \footnotesize	 Inception v4
\cite{InceptionV4}}	& \raisebox{-1.3ex}[0pt]{\footnotesize $41$} &\raisebox{-1.3ex}[0pt]{ \footnotesize $75$	 } & \raisebox{-1.3ex}[0pt]{\footnotesize} & \footnotesize & \footnotesize	 An Inception variant without residual connections, with roughly the same recognition performance as InceptionResNet v2, but significantly slower. \\
\hline
\raisebox{-1.3ex}[0pt]{$14$} &\raisebox{-1.3ex}[0pt]{ \footnotesize	 ResNeXt
\cite{ Xie2016Aggregated}}	 &\raisebox{-1.3ex}[0pt]{ \footnotesize $23 $}&\raisebox{-1.3ex}[0pt]{ \footnotesize	 $49	$} &\raisebox{-1.3ex}[0pt]{ \footnotesize	 $3.0\%$} &\raisebox{-1.3ex}[0pt]{ \footnotesize \cite{Xie2016Aggregated}}& \footnotesize	 Repeating a building block that aggregates a set of transformations with the same topology. \\
\hline
\raisebox{-3.3ex}[0pt]{$15$} & \raisebox{-3.3ex}[0pt]{\footnotesize	 DenseNet201
\cite{Huang2016Densely}} &\raisebox{-3.3ex}[0pt]{ \footnotesize	 $18$}	 & \raisebox{-3.3ex}[0pt]{ \footnotesize$ 200$ }  &\raisebox{-3.3ex}[0pt]{ \footnotesize	 $-$} & \raisebox{-3.3ex}[0pt]{\footnotesize \cite{Zhou2018Scale}} & \footnotesize Concatenate each layer with every other layer in a feed forward fashion. Alleviate the vanishing gradient problem, encourage feature reuse, reduction in number of parameters.\\
\hline
$16$ & \footnotesize	DarkNet
\cite{YOLO9000} & \footnotesize	$20$	& \footnotesize $19$	& \footnotesize $-$& \footnotesize	 \cite{YOLO9000} & \footnotesize Similar to VGGNet, but with significantly fewer parameters. \\
\hline
\raisebox{-1.3ex}[0pt]{$17$} &\raisebox{-1.3ex}[0pt]{ \footnotesize MobileNet \cite{Howard2017MobileNets}} &\raisebox{-1.3ex}[0pt]{ \footnotesize 	$3.2$  }& \raisebox{-1.3ex}[0pt]{\footnotesize	 $27+1$ }& \raisebox{-1.3ex}[0pt]{\footnotesize	 $-$ } & \raisebox{-1.3ex}[0pt]{ \footnotesize \cite{Howard2017MobileNets}} & \footnotesize Light weight deep CNNs using depth-wise separable convolutions. \\
\hline
\raisebox{-3.3ex}[0pt]{$18$}&\raisebox{-3.3ex}[0pt]{ \footnotesize SE ResNet \cite{ Hu2018Squeeze} }& \raisebox{-3.3ex}[0pt]{\footnotesize	$26$} &\raisebox{-3.3ex}[0pt]{ \footnotesize	 $50$}
 &\raisebox{-3.3ex}[0pt]{ \shortstack [c] {$2.3\%$ \\ (SENets)} }& \raisebox{-3.3ex}[0pt]{\footnotesize \cite{ Hu2018Squeeze}	}& \footnotesize Channel-wise attention by a novel block called \emph{Squeeze and Excitation}. Complementary to existing backbone CNNs.  \\
\Xhline{1pt}
\end{tabular}
}
\end{table*}

\subsection{Popular CNN Architectures}
\label{Sec:PopularNetworks}
CNN architectures (Section \ref{Sec:CNNintro}) serve as
network backbones used in the detection frameworks of Section~\ref{Sec:Frameworks}. Representative frameworks include AlexNet \cite{AlexNet2012}, ZFNet \cite{ZeilerFergus2014} VGGNet \cite{Simonyan2014VGG}, GoogLeNet \cite{GoogLeNet2015}, Inception series \cite{Ioffe2015,Szegedy2016a,InceptionV4}, ResNet \cite{He2016ResNet}, DenseNet \cite{Huang2016Densely} and SENet \cite{ Hu2018Squeeze}, summarized in Table \ref{Tab:dcnnarchitectures}, and where the improvement over time is seen in Fig.~\ref{fig:ILSVRCclassificationResults}.
A further review of recent CNN advances can be found in \cite{Gu2015Recent}.

The trend in architecture evolution is for greater depth:  AlexNet has 8 layers, VGGNet 16 layers, more recently ResNet and DenseNet both surpassed the 100 layer mark, and it was VGGNet \cite{Simonyan2014VGG} and GoogLeNet \cite{GoogLeNet2015} which showed that increasing depth can improve the representational power.  As can be observed from Table~\ref{Tab:dcnnarchitectures}, networks such as AlexNet, OverFeat, ZFNet and VGGNet have an enormous number of parameters, despite being only a few layers deep, since a large fraction of the parameters come from the FC layers. Newer networks like Inception, ResNet, and DenseNet, although having a great depth, actually have far fewer parameters by avoiding the use of FC layers.

With the use of Inception modules \cite{GoogLeNet2015} in carefully designed topologies, the number of parameters of GoogLeNet is dramatically reduced, compared to AlexNet, ZFNet or VGGNet.  Similarly, ResNet demonstrated the effectiveness of skip connections for learning extremely deep networks with hundreds of layers, winning the ILSVRC 2015 classification task. Inspired by ResNet \cite{He2016ResNet}, InceptionResNets \cite{InceptionV4} combined the Inception networks with shortcut connections, on the basis that shortcut connections can significantly accelerate network training. Extending ResNets, Huang \emph{et al.} \cite{Huang2016Densely} proposed DenseNets, which are built from dense blocksconnecting each layer
to every other layer in a feedforward fashion, leading to compelling
advantages such as parameter efficiency, implicit deep supervision\footnote{DenseNets perform deep supervision in an implicit way, \emph{i.e.} individual layers receive additional supervision from other layers through the shorter connections. The benefits of deep supervision have previously
been demonstrated in Deeply Supervised Nets (DSN) \cite{Lee2015Deeply}.}, and feature reuse.
Recently, Hu \emph{et al.} \cite{He2016ResNet} proposed Squeeze and Excitation (SE) blocks, which can be combined with existing deep architectures to boost their performance at minimal additional computational
cost, adaptively recalibrating channel-wise feature responses by explicitly modeling
the interdependencies between convolutional feature channels, and which led to winning the ILSVRC 2017 classification task. Research on CNN architectures remains active, with emerging networks such as Hourglass \cite{Law2018CornerNet}, Dilated Residual Networks \cite{Yu2017Dilated}, Xception \cite{Chollet2017Xception}, DetNet \cite{Li2018DetNet}, Dual Path Networks (DPN) \cite{Chen2017Dual}, FishNet \cite{Sun2018Fishnet}, and GLoRe \cite{Chen2019Graph}.

The training of a CNN requires a large-scale labeled dataset with intraclass diversity. Unlike image classification, detection requires localizing (possibly many) objects from an image. It has been shown \cite{Ouyang2016} that pretraining a deep model with a large scale dataset having object level annotations (such as ImageNet), instead of only the image level annotations, improves the detection performance. However, collecting bounding box labels is expensive, especially for hundreds of thousands of categories.  A common scenario is for a CNN to be pretrained on a large dataset (usually with a large number of visual categories) with image-level labels; the pretrained CNN can then be applied to a small dataset, directly, as a generic feature extractor \cite{Razavian2014,Azizpour2016,Donahue2014DeCAF,Yosinski2014Transferable}, which can support a wider range of visual recognition tasks. For detection, the pre-trained network is typically fine-tuned\footnote{Fine-tuning is done by initializing a network with weights
optimized for a large labeled dataset like ImageNet. and then updating the network's weights using the target-task training set.} on a given detection dataset \cite{Donahue2014DeCAF,Girshick2014RCNN,Girshick2016TPAMI}.
Several large scale image classification datasets are used for CNN pre-training, among them ImageNet1000  \cite{ImageNet2009,Russakovsky2015} with 1.2 million images of 1000 object categories, Places \cite{Zhou2017Places}, which is much larger than ImageNet1000 but with fewer classes, a recent Places-Imagenet hybrid \cite{Zhou2017Places}, or  JFT300M  \cite{Hinton2015Distilling,Sun2017Revisiting}.

Pretrained CNNs without fine-tuning were explored for object classification and detection in \cite{Donahue2014DeCAF,Girshick2016TPAMI,Agrawal2014}, where it was shown that detection accuracies are different for features extracted from different layers; for example, for AlexNet pre-trained on ImageNet, FC6 / FC7 / Pool5 are in descending order of detection accuracy \cite{Donahue2014DeCAF,Girshick2016TPAMI}.  Fine-tuning a pre-trained network can increase detection performance significantly \cite{Girshick2014RCNN,Girshick2016TPAMI}, although in the case of AlexNet, the fine-tuning performance boost was shown to be much larger for FC6 / FC7 than for
Pool5, suggesting that Pool5 features are more general.  Furthermore, the relationship between the source and target datasets plays a critical role, for example that ImageNet based CNN features show better performance  for object detection than for human action  \cite{Zhoubolei2014,Azizpour2016}.

\begin{table*}[!t]
\caption {Summary of properties of representative methods in improving DCNN feature representations for generic object detection. Details for Groups (1), (2), and (3) are provided in Section \ref{Sec:EnhanceFeatures}.  Abbreviations: Selective Search (SS), EdgeBoxes (EB), InceptionResNet (IRN). \emph{Conv-Deconv} denotes the use of upsampling and
convolutional layers with lateral connections to supplement the standard backbone network. Detection results on VOC07, VOC12 and COCO were reported with mAP@IoU=0.5, and the additional COCO results are computed as the average of mAP for IoU thresholds from 0.5 to 0.95. Training data: ``07''$\leftarrow$VOC2007 trainval; ``07T''$\leftarrow$VOC2007 trainval and test; ``12''$\leftarrow$VOC2012 trainval; CO$\leftarrow$ COCO trainval. The COCO detection results were reported with COCO2015 Test-Dev, except for MPN \cite{Zagoruyko2016} which reported with COCO2015 Test-Standard.}\label{Tab:EnhanceFeatures}
\centering
\renewcommand{\arraystretch}{1.2}
\setlength\arrayrulewidth{0.2mm}
\setlength\tabcolsep{1pt}
\resizebox*{18.5cm}{!}{
\begin{tabular}{!{\vrule width1.2bp}c|c|c|c|c|c|c|c|c|c|p{8cm}<{\centering}!{\vrule width1.2bp}}
\Xhline{1.5pt}
\footnotesize  & \footnotesize Detector & \footnotesize Region  & \footnotesize Backbone & \footnotesize Pipelined & \multicolumn{3}{c|}{mAP@IoU=0.5} & \footnotesize mAP & \footnotesize Published  &  \footnotesize  \\
 \cline{6-9}
\footnotesize Group & \footnotesize  Name & \footnotesize Proposal & \footnotesize  DCNN & \footnotesize  Used & \footnotesize  VOC07 & \footnotesize  VOC12 & \footnotesize  COCO & \footnotesize  COCO & \footnotesize   In  &  \footnotesize Highlights \\
\Xhline{1.5pt}
 \footnotesize \multirow{3}{*}{\rotatebox{90}{\scriptsize \shortstack [c] {\textbf{(1) Single detection }\\ \textbf{with multilayer features}$\quad$} }} 	 &\footnotesize \raisebox{-3.5ex}[0pt]{ION \cite{Bell2016ION}} & \footnotesize \raisebox{-4.5ex}[0pt]{ \shortstack [c] {SS+EB\\MCG+RPN} }& \footnotesize	 \raisebox{-3.5ex}[0pt]{VGG16} & \footnotesize	 \raisebox{-4.5ex}[0pt]{ \shortstack [c] {Fast \\ RCNN}}	& \footnotesize \raisebox{-4.5ex}[0pt]{ \shortstack [c] {$79.4$\\(07+12)}} & \footnotesize	 \raisebox{-4.5ex}[0pt]{\shortstack [c] {$76.4$\\(07+12)}} & \footnotesize \raisebox{-3.5ex}[0pt]{$55.7$}& \footnotesize \raisebox{-3.5ex}[0pt]{$33.1$} & \footnotesize \raisebox{-3.5ex}[0pt]{CVPR16}& \footnotesize  Use features from multiple layers; use spatial recurrent neural networks for modeling contextual information; the Best Student Entry and the $3^{\textrm{rd}}$ overall in the COCO detection challenge 2015. \\
 \cline{2-11}
& \footnotesize	\raisebox{-2ex}[0pt]{HyperNet \cite{HyperNet2016}}	 & \footnotesize\raisebox{-2ex}[0pt]{ RPN }& \footnotesize \raisebox{-2ex}[0pt]{	 VGG16} & \footnotesize
\raisebox{-2.5ex}[0pt]{ \shortstack [c] {Faster \\ RCNN}} & \footnotesize \raisebox{-2.5ex}[0pt]{ \shortstack [c] {$76.3$\\(07+12)}} & \footnotesize	 \raisebox{-2.5ex}[0pt]{ \shortstack [c] {$71.4$\\(07T+12)}} & \footnotesize	 \raisebox{-2ex}[0pt]{$-$}& \footnotesize	 \raisebox{-2ex}[0pt]{$-$} & \footnotesize \raisebox{-2ex}[0pt]{CVPR16	} & \footnotesize Use features from multiple layers
for both region proposal and region classification. \\
 \cline{2-11}
& \footnotesize	\raisebox{-3ex}[0pt]{ PVANet \cite{PVANET2016}} & \footnotesize	 \raisebox{-3ex}[0pt]{RPN} & \footnotesize \raisebox{-3ex}[0pt]{PVANet} & \footnotesize	 \raisebox{-3.5ex}[0pt]{ \shortstack [c] {Faster \\ RCNN}} & \footnotesize	 \raisebox{-3.5ex}[0pt]{ \shortstack [c] {$\textbf{84.9}$\\(07+12+CO)}}& \footnotesize \raisebox{-3.5ex}[0pt]{\shortstack [c] { $\textbf{84.2}$\\(07T+12+CO)}}& \footnotesize	 \raisebox{-3ex}[0pt]{ $-$}& \footnotesize	 \raisebox{-3ex}[0pt]{ $-$} & \footnotesize	 \raisebox{-3ex}[0pt]{NIPSW16}	 & \footnotesize Deep but lightweight; Combine ideas from concatenated ReLU \cite{Shang2016Understanding}, Inception \cite{GoogLeNet2015}, and HyperNet \cite{HyperNet2016}.  \\
\Xhline{1.5pt}
\footnotesize \multirow{5}{*}{\rotatebox{90}{\scriptsize \shortstack [c] {\textbf{(2) Detection at multiple layers}$\quad\quad\quad\quad\quad\quad$} }} & \footnotesize \raisebox{-5ex}[0pt]{SDP+CRC \cite{Yang2016Exploit}} & \footnotesize \raisebox{-5ex}[0pt]{EB}	 & \footnotesize \raisebox{-5ex}[0pt]{VGG16 }& \footnotesize \raisebox{-6ex}[0pt]{ \shortstack [c] {Fast \\ RCNN}}	& \footnotesize \raisebox{-6ex}[0pt]{ \shortstack [c] { $69.4$\\(07)}}& \footnotesize	 \raisebox{-5ex}[0pt]{$-$}	 & \footnotesize\raisebox{-5ex}[0pt]{ $-$}& \footnotesize\raisebox{-5ex}[0pt]{ $-$} & \footnotesize	 \raisebox{-5ex}[0pt]{CVPR16} & \footnotesize Use features in multiple layers to reject easy negatives via CRC, and then classify remaining proposals using SDP.  \\
 \cline{2-11}
 & \footnotesize \raisebox{-2ex}[0pt]{MSCNN \cite{MSCNN2016}} & \footnotesize \raisebox{-2ex}[0pt]{RPN} & \footnotesize	 \raisebox{-2ex}[0pt]{VGG }& \footnotesize	 \raisebox{-2.5ex}[0pt]{ \shortstack [c] {Faster \\ RCNN}} & \multicolumn{4}{c|}{\raisebox{-2ex}[0pt]{\footnotesize Only Tested on KITTI}}	& \footnotesize\raisebox{-2ex}[0pt]{ ECCV16 }	& \footnotesize Region proposal and classification are performed at multiple layers; includes feature upsampling; end to end learning. \\
 \cline{2-11}
 & \footnotesize \raisebox{-5ex}[0pt]{MPN \cite{Zagoruyko2016}} & \footnotesize \raisebox{-5ex}[0pt]{SharpMask \cite{Pinheiro2016}} & \footnotesize
	\raisebox{-5ex}[0pt]{VGG16} & \footnotesize  \raisebox{-6ex}[0pt]{ \shortstack [c] {Fast \\ RCNN}} & \footnotesize	 \raisebox{-5ex}[0pt]{$-$} & \footnotesize \raisebox{-5ex}[0pt]{$-$}& \footnotesize \raisebox{-5ex}[0pt]{$51.9$}	& \footnotesize \raisebox{-5ex}[0pt]{$33.2$}  & \footnotesize 	 \raisebox{-5ex}[0pt]{BMVC16} & \footnotesize	 Concatenate features from different convolutional layers and features of different contextual regions; loss function for multiple overlap thresholds; ranked $2^{\textrm{nd}}$ in both the COCO15 detection and segmentation challenges. \\
 \cline{2-11}
	 & \footnotesize\raisebox{-2ex}[0pt]{ DSOD \cite{ShenICCV2017}} & \footnotesize \raisebox{-2ex}[0pt]{ Free} & \footnotesize \raisebox{-2ex}[0pt]{DenseNet} & \footnotesize \raisebox{-2ex}[0pt]{SSD} & \footnotesize \raisebox{-3ex}[0pt]{ \shortstack [c] { $77.7$\\(07+12)}}
& \footnotesize \raisebox{-3ex}[0pt]{ \shortstack [c] { $72.2$\\(07T+12)}}& \footnotesize	 \raisebox{-2ex}[0pt]{ $47.3$ } & \footnotesize	\raisebox{-2ex}[0pt]{ $29.3$ } & \footnotesize \raisebox{-2ex}[0pt]{ICCV17} & \footnotesize Concatenate feature sequentially, like DenseNet. Train from scratch on the target dataset without pre-training. \\
 \cline{2-11}
 & \footnotesize \raisebox{-2ex}[0pt]{RFBNet \cite{Liu2017Receptive}} & \footnotesize	 \raisebox{-2ex}[0pt]{Free}	 & \footnotesize\raisebox{-2ex}[0pt]{ VGG16 }& \footnotesize	 \raisebox{-2ex}[0pt]{SSD} & \footnotesize\raisebox{-3ex}[0pt]{ \shortstack [c] { $82.2$\\(07+12)}}& \footnotesize	 \raisebox{-3ex}[0pt]{\shortstack [c] { $81.2$\\(07T+12)}} & \footnotesize \raisebox{-2ex}[0pt]{	 $55.7$}  & \footnotesize \raisebox{-2ex}[0pt]{	$34.4$} & \footnotesize \raisebox{-2ex}[0pt]{ ECCV18} & \footnotesize  Propose a multi-branch convolutional block similar to Inception \cite{GoogLeNet2015}, but using dilated convolution. \\
\Xhline{1.5pt}
\footnotesize \multirow{8}{*}{\rotatebox{90}{\scriptsize  \textbf{(3) Combination of (1) and (2) $\quad\quad\quad\quad\quad\quad\quad\quad\quad\quad\quad\quad$}  }}  & \footnotesize \raisebox{-2ex}[0pt]{DSSD \cite{DSSD2016}} & \footnotesize\raisebox{-2ex}[0pt]{ Free} & \footnotesize \raisebox{-2ex}[0pt]{ResNet101} & \footnotesize	 \raisebox{-2ex}[0pt]{SSD }& \footnotesize \raisebox{-3ex}[0pt]{ \shortstack [c] { $81.5$\\(07+12)}} & \footnotesize	 \raisebox{-3ex}[0pt]{ \shortstack [c] { $80.0$\\(07T+12)}}& \footnotesize	 \raisebox{-2ex}[0pt]{$53.3$} & \footnotesize	 \raisebox{-2ex}[0pt]{$33.2$} & \footnotesize	 \raisebox{-2ex}[0pt]{2017 }	& \footnotesize \raisebox{-2ex}[0pt]{Use Conv-Deconv, as shown in Fig. \ref{fig:MultiLayerCombine} (c1, c2).} \\
 \cline{2-11}
& \footnotesize\raisebox{-2ex}[0pt]{ FPN \cite{FPN2016}} & \footnotesize \raisebox{-2ex}[0pt]{RPN}	 & \footnotesize \raisebox{-2ex}[0pt]{ResNet101} & \footnotesize \raisebox{-3ex}[0pt]{ \shortstack [c] {Faster \\ RCNN}}	& \footnotesize\raisebox{-2ex}[0pt]{	 $-$}& \footnotesize	 \raisebox{-2ex}[0pt]{$-$}& \footnotesize\raisebox{-2ex}[0pt]{ $59.1$} & \footnotesize \raisebox{-2ex}[0pt]{$36.2$}& \footnotesize \raisebox{-2ex}[0pt]{CVPR17} & \footnotesize Use Conv-Deconv, as shown in Fig. \ref{fig:MultiLayerCombine} (a1, a2); Widely used in detectors. \\
  \cline{2-11}
 & \footnotesize  \raisebox{-2ex}[0pt]{	TDM \cite{Shrivastava2017} }& \footnotesize	 \raisebox{-2ex}[0pt]{RPN} & \footnotesize	 \raisebox{-3ex}[0pt]{ \shortstack [c] {ResNet101\\
VGG16} }& \footnotesize	\raisebox{-3ex}[0pt]{ \shortstack [c] {Faster \\ RCNN}} & \footnotesize	 \raisebox{-2ex}[0pt]{$-$}	& \footnotesize  \raisebox{-2ex}[0pt]{$-$} & \footnotesize  \raisebox{-2ex}[0pt]{ $57.7$} & \footnotesize  \raisebox{-2ex}[0pt]{$36.8$}	 & \footnotesize  \raisebox{-2ex}[0pt]{CVPR17} & \footnotesize  Use Conv-Deconv, as shown in Fig. \ref{fig:MultiLayerCombine} (b2). \\
 \cline{2-11}
 & \footnotesize  \raisebox{-2ex}[0pt]{RON  \cite{Kong2017ron}} & \footnotesize	 \raisebox{-2ex}[0pt]{RPN} & \footnotesize	 \raisebox{-2ex}[0pt]{VGG16}	 & \footnotesize \raisebox{-3ex}[0pt]{ \shortstack [c] {Faster \\ RCNN}}	 & \footnotesize \raisebox{-3ex}[0pt]{ \shortstack [c] { $81.3$\\(07+12+CO)}}
& \footnotesize	\raisebox{-3ex}[0pt]{ \shortstack [c] { $80.7$\\(07T+12+CO)}} & \footnotesize \raisebox{-2ex}[0pt]{$49.5$} & \footnotesize	 \raisebox{-2ex}[0pt]{$27.4$} & \footnotesize	 \raisebox{-2ex}[0pt]{CVPR17} & \footnotesize	 Use Conv-deconv, as shown in Fig. \ref{fig:MultiLayerCombine} (d2); Add the objectness prior to significantly reduce object search space.  \\
 \cline{2-11}
 & \footnotesize  \raisebox{-3ex}[0pt]{ZIP \cite{Hongyang2018Zoom}} & \footnotesize	 \raisebox{-3ex}[0pt]{RPN} & \footnotesize	 \raisebox{-3ex}[0pt]{Inceptionv2}	 & \footnotesize \raisebox{-4ex}[0pt]{ \shortstack [c] {Faster \\ RCNN}}	 & \footnotesize \raisebox{-4ex}[0pt]{ \shortstack [c] {$79.8$\\ (07+12)}}
& \footnotesize	\raisebox{-3ex}[0pt]{$-$} & \footnotesize \raisebox{-3ex}[0pt]{$-$} & \footnotesize\raisebox{-3ex}[0pt]{	$-$ }& \footnotesize\raisebox{-3ex}[0pt]{	IJCV18 }  & \footnotesize	 Use Conv-Deconv, as shown in Fig. \ref{fig:MultiLayerCombine} (f1). Propose a map attention decision (MAD) unit for features from different layers.\\
 \cline{2-11}
  & \footnotesize \raisebox{-2ex}[0pt]{STDN \cite{Zhou2018Scale} } & \footnotesize	 \raisebox{-2ex}[0pt]{Free} & \footnotesize \raisebox{-2ex}[0pt]{DenseNet169}
& \footnotesize \raisebox{-2ex}[0pt]{SSD}	& \footnotesize  \raisebox{-3ex}[0pt]{ \shortstack [c] {$80.9$\\(07+12)}} & \footnotesize\raisebox{-2ex}[0pt]{	$-$ }& \footnotesize \raisebox{-2ex}[0pt]{$51.0$ }& \footnotesize	 \raisebox{-2ex}[0pt]{$31.8$ }& \footnotesize	 \raisebox{-2ex}[0pt]{ CVPR18}  & \footnotesize	
A new scale transfer module, which resizes features of different scales to the same scale in parallel. \\
 \cline{2-11}
 & \footnotesize \raisebox{-2ex}[0pt]{RefineDet \cite{Zhang2018Single}} & \footnotesize \raisebox{-2ex}[0pt]{RPN} & \footnotesize \raisebox{-3ex}[0pt]{\shortstack [c] {VGG16\\ResNet101}} & \footnotesize \raisebox{-3ex}[0pt]{ \shortstack [c] {Faster \\ RCNN}} & \footnotesize \raisebox{-3ex}[0pt]{ \shortstack [c] {	$83.8$\\(07+12)}} & \footnotesize	 \raisebox{-3ex}[0pt]{ \shortstack [c] {$83.5$\\(07T+12)}}	 & \footnotesize\raisebox{-2ex}[0pt]{ $ 62.9$} & \footnotesize \raisebox{-2ex}[0pt]{$41.8$} & \footnotesize \raisebox{-2ex}[0pt]{CVPR18} & \footnotesize Use cascade to obtain better and less anchors. Use Conv-deconv, as shown in Fig. \ref{fig:MultiLayerCombine} (e2) to improve features. \\
   \cline{2-11}
 & \footnotesize \raisebox{-5ex}[0pt]{PANet \cite{Liu2018Path} }& \footnotesize \raisebox{-5ex}[0pt]{RPN} & \footnotesize \raisebox{-6.5ex}[0pt]{\shortstack [c] {ResNeXt101\\+FPN}}  & \footnotesize \raisebox{-5ex}[0pt]{Mask RCNN} & \footnotesize \raisebox{-5ex}[0pt]{ $-$ } & \footnotesize \raisebox{-5ex}[0pt]{  $-$ }& \footnotesize\raisebox{-5ex}[0pt]{ $\textbf{67.2}$ } & \footnotesize \raisebox{-5ex}[0pt]{$\textbf{47.4}$} & \footnotesize\raisebox{-5ex}[0pt]{ CVPR18}& \footnotesize Shown in Fig. \ref{fig:MultiLayerCombine} (g). Based on FPN, add another bottom-up path to pass  information between lower and topmost layers; adaptive feature pooling. Ranked $1^{st}$ and $2^{nd}$ in COCO 2017 tasks. \\
    \cline{2-11}
 & \footnotesize 	\raisebox{-1ex}[0pt]{DetNet \cite{Li2018DetNet}}& \footnotesize \raisebox{-1ex}[0pt]{RPN} & \footnotesize \raisebox{-1ex}[0pt]{DetNet59+FPN}  & \footnotesize \raisebox{-1ex}[0pt]{Faster RCNN} & \footnotesize \raisebox{-1ex}[0pt]{ $-$	} & \footnotesize \raisebox{-1ex}[0pt]{ $-$	}& \footnotesize\raisebox{-1ex}[0pt]{ $61.7$	} & \footnotesize \raisebox{-1ex}[0pt]{$40.2$} & \footnotesize\raisebox{-1ex}[0pt]{ ECCV18}& \footnotesize Introduces dilated convolution into the ResNet backbone to maintain high resolution in deeper layers; Shown in Fig. \ref{fig:MultiLayerCombine} (i).  \\
    \cline{2-11}
 & \footnotesize 	\raisebox{-2ex}[0pt]{FPR \cite{Kong2018Deep} }& \footnotesize \raisebox{-2ex}[0pt]{$-$} & \footnotesize \raisebox{-3ex}[0pt]{\shortstack [c] {VGG16\\ ResNet101}}  & \footnotesize \raisebox{-2ex}[0pt]{SSD} & \footnotesize \raisebox{-3ex}[0pt]{ \shortstack [c] { $82.4$\\(07+12)}} & \footnotesize \raisebox{-3ex}[0pt]{ \shortstack [c] { $81.1$\\(07T+12)}}& \footnotesize\raisebox{-2ex}[0pt]{$54.3$} & \footnotesize \raisebox{-2ex}[0pt]{ $34.6$} & \footnotesize\raisebox{-2ex}[0pt]{ ECCV18}& \footnotesize  Fuse task oriented features across different spatial locations and scales, globally and locally; Shown in Fig. \ref{fig:MultiLayerCombine} (h).\\
    \cline{2-11}
 & \footnotesize 	\raisebox{-5ex}[0pt]{M2Det \cite{Zhao2019M2Det}}& \footnotesize \raisebox{-5ex}[0pt]{$-$} & \footnotesize \raisebox{-5ex}[0pt]{SSD}  & \footnotesize \raisebox{-6.5ex}[0pt]{\shortstack [c] {VGG16\\ ResNet101}} & \footnotesize \raisebox{-5ex}[0pt]{ $-$	}  & \footnotesize \raisebox{-5ex}[0pt]{ $-$	} & \footnotesize\raisebox{-5ex}[0pt]{ 	$64.6$} & \footnotesize \raisebox{-5ex}[0pt]{$44.2$} & \footnotesize\raisebox{-5ex}[0pt]{ AAAI19}& \footnotesize Shown in Fig. \ref{fig:MultiLayerCombine} (j), newly designed top down path to learn a set of multilevel features, recombined
to construct a feature pyramid for object detection. \\
\Xhline{1.5pt}
 \scriptsize \multirow{6}{*}{\rotatebox{90}{ \textbf{(4) Model Geometric Transforms$\quad$} }}	 & \footnotesize \raisebox{-8ex}[0pt]{DeepIDNet \cite{Ouyang2015deepid} }& \footnotesize	 \raisebox{-8ex}[0pt]{SS+
EB} & \footnotesize \raisebox{-12ex}[0pt]{\shortstack [c] {AlexNet \\ZFNet \\OverFeat \\GoogLeNet}} & \footnotesize \raisebox{-8ex}[0pt]{
	RCNN}& \footnotesize \raisebox{-9ex}[0pt]{\shortstack [c] {$69.0$	 \\(07)}} & \footnotesize \raisebox{-8ex}[0pt]{$-$}& \footnotesize\raisebox{-8ex}[0pt]{$-$}& \footnotesize	 \raisebox{-8ex}[0pt]{$25.6$}& \footnotesize	 \raisebox{-8ex}[0pt]{CVPR15} 	 & \footnotesize Introduce a deformation constrained
pooling layer, jointly learned with convolutional layers in existing DCNNs. Utilize the following modules that are not trained end to end: cascade, context modeling, model averaging, and bounding box location refinement in the multistage detection pipeline. \\
 \cline{2-11}
	& \footnotesize \raisebox{-2ex}[0pt]{DCN \cite{Dai17Deformable}} & \footnotesize \raisebox{-2ex}[0pt]{RPN} & \footnotesize \raisebox{-2.5ex}[0pt]{\shortstack [c] {ResNet101\\IRN}}  & \footnotesize\raisebox{-2ex}[0pt]{ RFCN} & \footnotesize \raisebox{-2.5ex}[0pt]{ \shortstack [c] {$82.6$	 \\(07+12)}}& \footnotesize\raisebox{-2ex}[0pt]{ $-$	 } & \footnotesize \raisebox{-2ex}[0pt]{ $58.0$} & \footnotesize\raisebox{-2ex}[0pt]{ $37.5$} & \footnotesize	 \raisebox{-2ex}[0pt]{CVPR17} & \footnotesize Design deformable convolution and deformable RoI pooling modules that can replace plain convolution in existing DCNNs. \\
 \cline{2-11}
	& \footnotesize \raisebox{-2.5ex}[0pt]{DPFCN \cite{Mordan2018End}} 	& \footnotesize \raisebox{-2.5ex}[0pt]{AttractioNet \cite{Gidaris2016Attend} }& \footnotesize \raisebox{-2.5ex}[0pt]{ ResNet} & \footnotesize \raisebox{-2.5ex}[0pt]{ RFCN} & \footnotesize \raisebox{-3.5ex}[0pt]{\shortstack [c] {$83.3$	 \\(07+12)}}& \footnotesize \raisebox{-3.5ex}[0pt]{\shortstack [c] {$81.2$\\(07T+12)}} & \footnotesize \raisebox{-2.5ex}[0pt]{$59.1$}& \footnotesize\raisebox{-2.5ex}[0pt]{$39.1$}  & \footnotesize	 \raisebox{-2.5ex}[0pt]{IJCV18}& \footnotesize 	Design a deformable part based RoI pooling layer to explicitly select discriminative regions around object proposals. \\
\Xhline{1.5pt}
\end{tabular}
}
\end{table*}

\subsection{Methods For Improving Object Representation}
\label{Sec:EnhanceFeatures}
Deep CNN based detectors such as RCNN \cite{Girshick2014RCNN}, Fast RCNN \cite{Girshick2015FRCNN}, Faster RCNN \cite{Ren2015NIPS} and YOLO \cite{YoLo2016}, typically use the deep CNN architectures listed in Table \ref{Tab:dcnnarchitectures} as the backbone network and use features from the top layer of the CNN as object representations; however, detecting objects across a large {\em range} of scales is a fundamental challenge. A classical strategy to address this issue is to run the detector over a number of scaled input images (\emph{e.g.,} an image pyramid) \cite{Felzenszwalb2010b,Girshick2014RCNN,He2014SPP}, which typically  produces
more accurate detection,  with, however, obvious limitations of inference time and memory.

\subsubsection{Handling of Object Scale Variations}
\label{sec:objectscale}
Since a CNN computes its feature hierarchy layer by layer, the sub-sampling
layers in the feature hierarchy already lead to an inherent multiscale
pyramid, producing feature maps at different spatial resolutions, but subject to   challenges \cite{Hariharan2016,FCNCVPR2015,Shrivastava2017}.  In particular, the higher layers have a large receptive field and strong semantics, and are the most robust to variations such as object pose, illumination and part deformation, but the resolution is low and the geometric details are lost. In contrast, lower layers have a small receptive field and rich geometric details, but the resolution is high and much less sensitive to semantics.  Intuitively, semantic concepts of objects can emerge in different layers, depending on the size of the objects.  So if a target object is small
it requires fine detail information in earlier layers and may very well  disappear at later layers, in principle making small object detection very challenging, for which tricks such as dilated or ``atrous'' convolution \cite{Yu2015Multiscale,Dai2016RFCN,Chen2016deeplab} have been proposed,  increasing feature resolution, but
increasing computational complexity. On
the other hand, if the target object is large, then the semantic concept
will emerge in much later layers.  A number of methods \cite{Shrivastava2017,Zhang2018Object,FPN2016,Kong2017ron} have been proposed to improve detection accuracy by exploiting multiple CNN layers, broadly falling into three types of \textbf{multiscale object detection}:
\begin{enumerate}
\item Detecting with combined features of multiple layers;
\item Detecting at multiple layers;
\item Combinations of the above two methods.
\end{enumerate}

\begin {figure}[!t]
\centering
\includegraphics[width=0.48\textwidth]{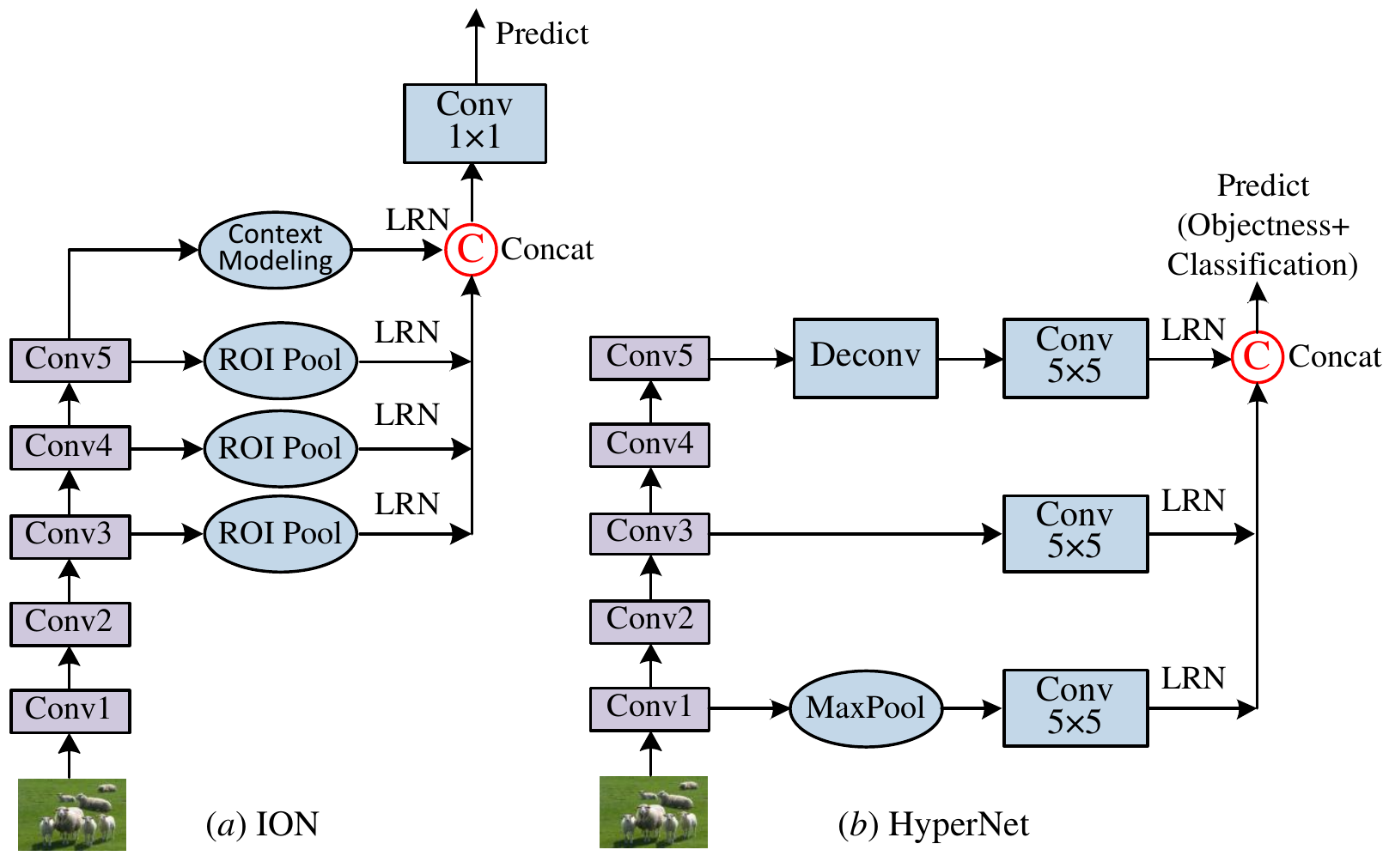}
\caption{Comparison of HyperNet and ION. LRN is Local Response Normalization, which performs a kind of ``lateral inhibition'' by normalizing over local input regions \cite{Jia2014Caffe}.}
\label{fig:HyperFeature}
\end {figure}

\textbf{(1) Detecting with combined features of multiple CNN layers:} Many approaches, including Hypercolumns \cite{Hariharan2016}, HyperNet \cite{HyperNet2016}, and ION \cite{Bell2016ION}, combine features from multiple layers
before making a prediction. Such feature combination is commonly
accomplished via concatenation, a classic neural network
idea that concatenates features from different layers, architectures
which have recently become popular for semantic segmentation
\cite{FCNCVPR2015,FCNTPAMI,Hariharan2016}. As shown in Fig.~\ref{fig:HyperFeature} (a), ION \cite{Bell2016ION} uses RoI
pooling to extract RoI features from multiple layers, and then the object proposals generated by selective search and edgeboxes are classified by using the concatenated features. HyperNet \cite{HyperNet2016}, shown in Fig.~\ref{fig:HyperFeature} (b), follows a similar idea, and integrates deep, intermediate and shallow features to generate object proposals and to predict objects via an end to end joint training strategy.  The combined feature is more descriptive, and is more beneficial for localization and classification, but at increased computational complexity.

\begin {figure*}[!t]
\centering
\includegraphics[height=0.9\textheight]{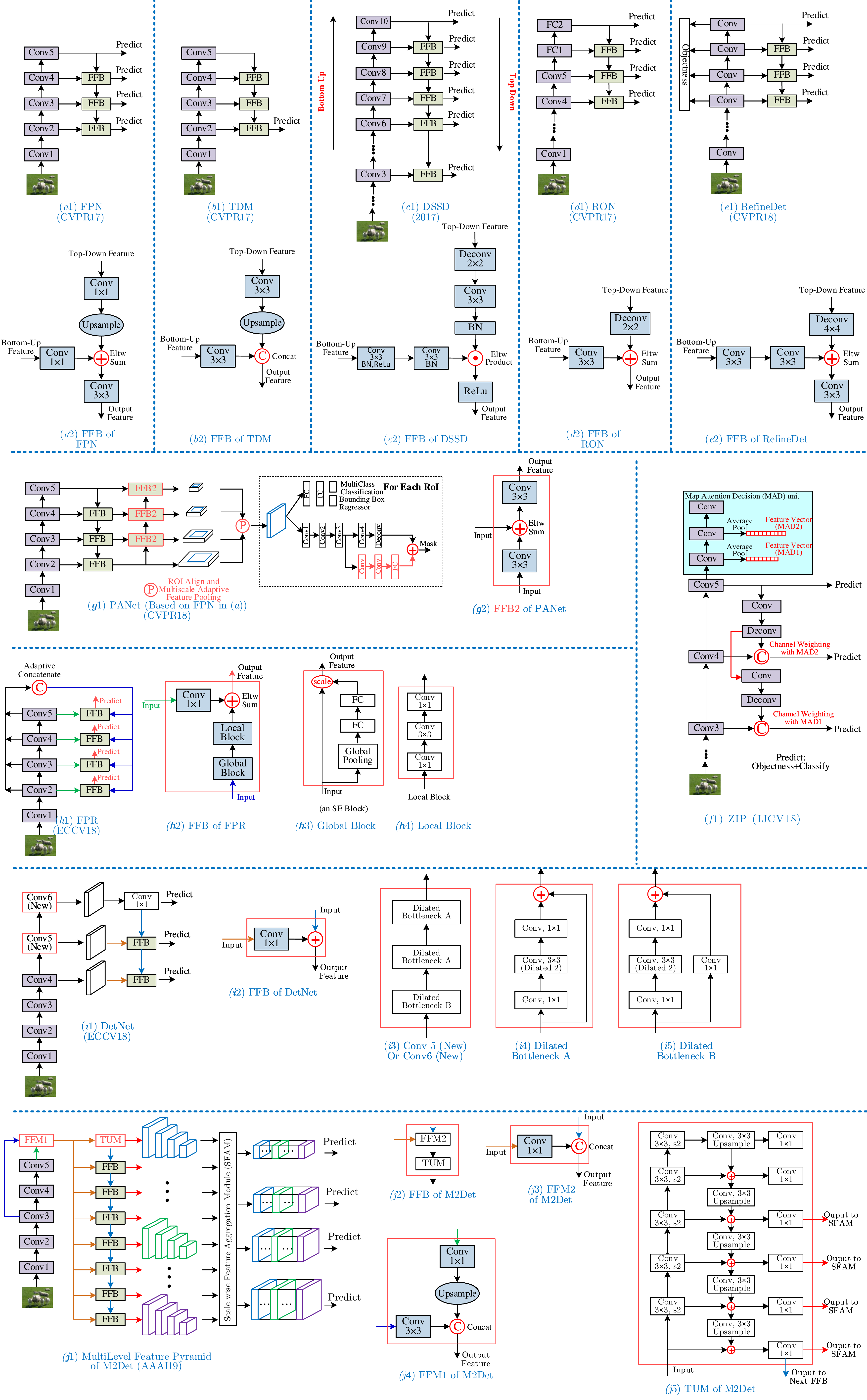}
\caption{Hourglass architectures: Conv1 to Conv5 are the main Conv blocks in backbone networks such as VGG or ResNet. The figure compares a number of Feature Fusion Blocks (FFB) commonly used in recent approaches: FPN \cite{FPN2016}, TDM \cite{Shrivastava2017}, DSSD \cite{DSSD2016}, RON \cite{Kong2017ron}, RefineDet \cite{Zhang2018Single}, ZIP \cite{Hongyang2018Zoom}, PANet \cite{Liu2018Path}, FPR \cite{Kong2018Deep}, DetNet \cite{Li2018DetNet} and M2Det \cite{Zhao2019M2Det}. FFM: Feature Fusion Module, TUM: Thinned U-shaped Module}
\label{fig:MultiLayerCombine}
\end {figure*}

\textbf{(2) Detecting at multiple CNN layers:} A number of recent
approaches improve detection by predicting objects of different
resolutions at different layers and then combining these
predictions: SSD \cite{Liu2016SSD} and MSCNN \cite{MSCNN2016}, RBFNet \cite{Liu2017Receptive}, and
DSOD \cite{ShenICCV2017}.  SSD \cite{Liu2016SSD} spreads out default boxes of different
scales to multiple layers within a CNN, and forces
each layer to focus on predicting objects of a certain scale. RFBNet \cite{Liu2017Receptive}
replaces the later convolution layers of SSD with a Receptive Field Block (RFB) to enhance the discriminability and robustness of features. The RFB is a multibranch convolutional block,
similar to the Inception block \cite{GoogLeNet2015}, but combining multiple branches with different kernels and convolution layers \cite{Chen2016deeplab}.
MSCNN \cite{MSCNN2016} applies deconvolution on multiple layers of a
CNN to increase feature map resolution before using the
layers to learn region proposals and pool features. Similar to RFBNet \cite{Liu2017Receptive}, TridentNet \cite{Li2019Scale}
constructs a parallel multibranch architecture where each branch
shares the same transformation parameters but with different
receptive fields; dilated convolution with different dilation rates are used to adapt the receptive fields for objects of different scales.

\textbf{(3) Combinations of the above two methods:}  Features from different layers are complementary to each other
and can improve detection accuracy, as shown by Hypercolumns \cite{Hariharan2016}, HyperNet \cite{HyperNet2016} and ION \cite{Bell2016ION}. On
the other hand, however, it is natural to detect objects of different scales
using features of approximately the same size, which
can be achieved by detecting large objects from downscaled feature maps
while detecting small objects from upscaled feature maps.
Therefore, in order to combine the best of both worlds, some recent works propose to detect objects at multiple layers, and the resulting features  obtained by combining features from different
layers.  This approach has been found to be effective for segmentation \cite{FCNCVPR2015,FCNTPAMI} and human pose estimation \cite{Newell2016Stacked}, has been widely exploited by both one-stage  and two-stage detectors to alleviate problems of scale
variation across object instances. Representative methods include SharpMask \cite{Pinheiro2016},
Deconvolutional Single Shot Detector (DSSD) \cite{DSSD2016}, Feature Pyramid Network (FPN) \cite{FPN2016}, Top Down Modulation (TDM)\cite{Shrivastava2017}, Reverse connection with Objectness prior Network (RON) \cite{Kong2017ron}, ZIP \cite{Hongyang2018Zoom}, Scale Transfer Detection Network (STDN) \cite{Zhou2018Scale}, RefineDet \cite{Zhang2018Single},  StairNet \cite{Woo18StairNet}, Path Aggregation Network (PANet) \cite{Liu2018Path}, Feature Pyramid Reconfiguration (FPR) \cite{Kong2018Deep}, DetNet \cite{Li2018DetNet}, Scale Aware Network (SAN) \cite{Kim2018San}, Multiscale Location aware Kernel Representation (MLKP) \cite{Wang2018Multiscale} and M2Det \cite{Zhao2019M2Det}, as shown in Table~\ref{Tab:EnhanceFeatures} and contrasted in Fig.~\ref{fig:MultiLayerCombine}.

Early works like FPN \cite{FPN2016}, DSSD \cite{DSSD2016}, TDM \cite{Shrivastava2017}, ZIP \cite{Hongyang2018Zoom}, RON \cite{Kong2017ron} and RefineDet \cite{Zhang2018Single} construct the feature pyramid according to the inherent multiscale,
pyramidal architecture of the backbone, and achieved encouraging results. As can be observed from Fig. \ref{fig:MultiLayerCombine} (a1) to (f1), these methods have very similar detection architectures which incorporate a top-down network with lateral connections to supplement the standard bottom-up, feed-forward network. Specifically, after a bottom-up pass the final high level semantic features are transmitted back by the top-down network to combine with the bottom-up features from intermediate layers after lateral processing, and the combined features are then used for detection. As can be seen from Fig.~\ref{fig:MultiLayerCombine} (a2) to (e2), the main differences lie in the design of the simple Feature Fusion Block (FFB), which handles the selection of features from different layers and the combination of multilayer features.

FPN \cite{FPN2016}
shows significant improvement as a generic feature extractor
in several applications including object detection \cite{FPN2016,LinICCV2017} and instance segmentation \cite{MaskRCNN2017}. Using FPN in a basic Faster
RCNN system achieved state-of-the-art results on the COCO detection dataset. STDN \cite{Zhou2018Scale} used DenseNet \cite{Huang2016Densely} to combine features of different layers and designed a scale transfer module to obtain feature maps
with different resolutions. The scale transfer module can be directly embedded
into DenseNet with little additional cost.

More recent work, such as PANet \cite{Liu2018Path}, FPR \cite{Kong2018Deep}, DetNet \cite{Li2018DetNet}, and M2Det \cite{Zhao2019M2Det}, as shown in Fig. \ref{fig:MultiLayerCombine} (g-j), propose to further improve on the pyramid architectures like FPN in different ways. Based on FPN, Liu \emph{et al.} designed PANet \cite{Liu2018Path} (Fig.~\ref{fig:MultiLayerCombine} (g1)) by adding another bottom-up path with clean lateral connections from  low to top levels, in order to shorten the information path and to enhance the feature pyramid. Then, an adaptive feature pooling was proposed to aggregate features from all feature levels for each proposal. In addition, in the proposal sub-network, a complementary branch capturing different views for each proposal is created to further improve mask prediction. These additional steps bring only slightly
extra computational overhead, but are effective and allowed
PANet to reach 1st place in the COCO 2017 Challenge Instance Segmentation task and 2nd place in the Object Detection task.  Kong \emph{et al.} proposed FPR \cite{Kong2018Deep} by explicitly reformulating the feature pyramid construction process (\emph{e.g.} FPN \cite{FPN2016}) as feature reconfiguration functions in a highly nonlinear but efficient way. As shown in Fig.~\ref{fig:MultiLayerCombine} (h1), instead of using a top-down path to propagate strong semantic features from the topmost layer down as in FPN, FPR first extracts features from multiple layers in the backbone network by adaptive concatenation, and then designs a more complex FFB module (Fig. \ref{fig:MultiLayerCombine} (h2)) to spread  strong semantics to all scales. Li \emph{et al.} proposed DetNet \cite{Li2018DetNet} (Fig. \ref{fig:MultiLayerCombine} (i1)) by introducing dilated convolutions to the later layers of the backbone network in order to maintain high spatial resolution
in deeper layers. Zhao \emph{et al.} \cite{Zhao2019M2Det} proposed a MultiLevel Feature Pyramid Network (MLFPN) to build more effective feature pyramids for detecting objects of different scales. As can be seen from Fig. \ref{fig:MultiLayerCombine} (j1), features from two different layers of the backbone are first fused as the base feature, after which a top-down path with lateral connections from the base feature is created to build the feature pyramid. As shown in Fig.~\ref{fig:MultiLayerCombine} (j2) and (j5), the FFB module is much more complex than those like FPN, in that  FFB involves a Thinned U-shaped Module (TUM) to generate a second pyramid structure, after which the feature maps with equivalent sizes from multiple TUMs are combined for object detection. The authors proposed M2Det by integrating MLFPN into SSD, and achieved better detection performance than other one-stage detectors.

\subsection{Handling of Other Intraclass Variations}
\label{sec:Otherchanges}

Powerful object representations should combine distinctiveness and robustness. A large amount of recent work has been devoted to handling changes in object scale, as reviewed in Section~\ref{sec:objectscale}. As discussed in Section~\ref{Sec:MainChallenges} and summarized in Fig.~\ref{Fig:challenges}, object detection still requires robustness to real-world variations other than just scale, which we group into three categories:
\begin{itemize}
\renewcommand{\labelitemi}{$\bullet$}
\item Geometric transformations,
\item Occlusions, and
\item Image degradations.
\end{itemize}
To handle these intra-class variations, the most straightforward approach is to augment the training
datasets with a sufficient amount of variations; for example, robustness to rotation could be achieved by adding rotated objects at many orientations to the training data. Robustness can frequently be learned this way, but usually at the cost of expensive
training and complex model parameters. Therefore, researchers have proposed alternative solutions to these problems.

\textbf{Handling of geometric transformations:} DCNNs are inherently limited by the lack of ability to be spatially invariant to geometric transformations of the input data \cite{Lenc2018Understanding,Liu2017Local,Chellappa2016}. The introduction of local max pooling layers has allowed DCNNs to enjoy some translation invariance, however the intermediate feature maps are not
actually invariant to large geometric transformations of the input data \cite{Lenc2018Understanding}. Therefore, many approaches have been presented to enhance robustness, aiming at learning invariant CNN representations with respect
to different types of transformations such as scale \cite{Kim2014Locally,Bruna13Invariant}, rotation \cite{Bruna13Invariant,RIFDCNN2016,Worrall2017Harmonic,Zhou2017Oriented}, or both \cite{Jaderberg2015Spatial}. One representative work is Spatial Transformer Network (STN) \cite{Jaderberg2015Spatial}, which introduces a
new learnable module to handle scaling, cropping, rotations, as well as nonrigid deformations via a global parametric transformation. STN has now
been used in rotated text detection \cite{Jaderberg2015Spatial}, rotated face
detection and generic object detection \cite{Wang2017}.

Although rotation invariance may be attractive in certain applications, such as scene text detection \cite{He2018End,Ma2018Arbitrary}, face detection \cite{Shi2018Real}, and aerial imagery \cite{Ding2018Learning,Xia2018DOTA}, there is limited generic object detection work focusing on rotation invariance because popular benchmark detection datasets (\emph{e.g.} PASCAL VOC, ImageNet, COCO) do not actually present rotated images.

Before deep learning, Deformable Part based Models (DPMs)
\cite{Felzenszwalb2010b} were successful for generic object detection, representing objects by component parts arranged in a deformable configuration. Although DPMs have been significantly outperformed by more recent object detectors, their spirit still deeply influences many recent detectors. DPM modeling is less sensitive to transformations in
object pose, viewpoint and nonrigid deformations, motivating researchers \cite{Dai17Deformable,Girshick2015DPMCNN,Mordan2018End,Ouyang2015deepid,Wan2015end}
to explicitly model object composition to improve CNN based detection.
The first attempts \cite{Girshick2015DPMCNN,Wan2015end} combined DPMs with CNNs by using deep features learned by AlexNet in DPM based detection, but without region proposals.
To enable a CNN to benefit from the built-in capability of modeling the deformations of object parts, a number of approaches were proposed, including DeepIDNet  \cite{Ouyang2015deepid}, DCN \cite{Dai17Deformable} and DPFCN \cite{Mordan2018End} (shown in Table~\ref{Tab:EnhanceFeatures}). Although  similar in spirit, deformations are computed in different ways: DeepIDNet \cite{Ouyang2016} designed a deformation constrained pooling layer to replace regular max pooling, to learn the shared visual patterns and their deformation properties across different object classes; DCN \cite{Dai17Deformable} designed a deformable convolution layer and a deformable RoI pooling layer, both of which
are based on the idea of augmenting regular grid sampling locations in feature maps; and DPFCN \cite{Mordan2018End} proposed a deformable part-based RoI pooling layer which selects discriminative parts of objects around object proposals by
simultaneously optimizing latent displacements of all parts.

\textbf{Handling of occlusions:} In real-world images, occlusions are common, resulting in information loss from object instances.
A deformable parts idea can be useful for occlusion handling, so deformable RoI Pooling \cite{Dai17Deformable,Mordan2018End,Ouyang2013Joint} and deformable convolution \cite{Dai17Deformable} have been proposed to alleviate occlusion by giving more flexibility to the typically fixed geometric structures. Wang \emph{et al.} \cite{Wang2017} propose to learn an adversarial network that generates examples with occlusions and deformations, and context may be helpful in dealing with occlusions \cite{Zhang2018Occluded}.
Despite these efforts, the occlusion problem is far from being solved; applying GANs to this problem may be a promising research direction.

\textbf{Handling of image degradations:}
Image noise is a common problem in many real-world applications. It is frequently caused by insufficient lighting, low quality cameras, image compression, or the intentional low-cost sensors on edge devices and wearable devices. While low image quality may be expected to degrade the performance of visual recognition, most current methods are evaluated in a
degradation free and clean environment, evidenced by the fact that PASCAL VOC, ImageNet, MS COCO and Open Images all focus on relatively high quality images.
To the best of our knowledge, there is so far very limited work to address this problem.

\section{Context Modeling}
\label{sec:ContextInfo}
\begin{table*}[!t]
\caption {Summary of detectors that exploit context information, with labelling details as in Table \ref{Tab:EnhanceFeatures}.}\label{Tab:ContextMethods}
\centering
\renewcommand{\arraystretch}{1.2}
\setlength\arrayrulewidth{0.2mm}
\setlength\tabcolsep{1pt}
\resizebox*{18.5cm}{!}{
\begin{tabular}{!{\vrule width1.5bp}c|c|c|c|c|c|c|c|c|p{8cm}<{\centering}!{\vrule width1.5bp}}
\Xhline{1.5pt}
\footnotesize  & \footnotesize Detector & \footnotesize Region  & \footnotesize Backbone & \footnotesize Pipelined & \multicolumn{2}{c|}{mAP@IoU=0.5} & \footnotesize mAP & \footnotesize Published  &  \footnotesize  \\
 \cline{6-8}
\footnotesize Group & \footnotesize  Name & \footnotesize Proposal & \footnotesize  DCNN & \footnotesize  Used & \footnotesize  VOC07 & \footnotesize  VOC12 & \footnotesize  COCO & \footnotesize   In  &  \footnotesize Highlights \\
\Xhline{1.5pt}
\footnotesize \multirow{4}{*}{\rotatebox{90}{\scriptsize \textbf{Global Context}
$\quad\quad\quad\quad\quad$}} &
\raisebox{-1.5ex}[0pt]{\footnotesize SegDeepM \cite{SegDeepM2015} }
&\raisebox{-1.5ex}[0pt]{ \footnotesize	SS+CMPC}
& \raisebox{-1.5ex}[0pt]{\footnotesize VGG16}
& \raisebox{-1.5ex}[0pt]{\footnotesize	RCNN}
& \raisebox{-1.5ex}[0pt]{\footnotesize	 VOC10}
&\raisebox{-1.5ex}[0pt]{ \footnotesize 	VOC12}
& \raisebox{-1.5ex}[0pt]{\footnotesize $-$}&\raisebox{-1.5ex}[0pt]{ \footnotesize CVPR15 }
& \footnotesize
Additional features extracted from an enlarged object proposal as context information.
\\
\cline{2-10}
 	& \footnotesize \raisebox{-1.5ex}[0pt]{DeepIDNet \cite{Ouyang2015deepid}} & \footnotesize	 \raisebox{-1.5ex}[0pt]{SS+EB} &\raisebox{-2.5ex}[0pt]{  \footnotesize \shortstack [c] {AlexNet\\ZFNet}} & \footnotesize	 \raisebox{-1.5ex}[0pt]{ RCNN } & \footnotesize \raisebox{-2.5ex}[0pt]{\shortstack [c] {$69.0$	 \\(07)}}	 & \footnotesize	 \raisebox{-1.5ex}[0pt]{$-$ } & \footnotesize \raisebox{-1.5ex}[0pt]{$-$}  & \footnotesize \raisebox{-1.5ex}[0pt]{CVPR15} 	& \footnotesize Use image classification scores as global contextual information to refine the detection scores of each object proposal.\\
\cline{2-10}
  & \raisebox{-1.5ex}[0pt]{\footnotesize	ION \cite{Bell2016ION} }& \raisebox{-1.5ex}[0pt]{
  \footnotesize	SS+EB}	&\raisebox{-1.5ex}[0pt]{ \footnotesize VGG16}
&\raisebox{-2.5ex}[0pt]{ \footnotesize\shortstack [c] { Fast \\RCNN}}
	& \raisebox{-1.5ex}[0pt]{\footnotesize $80.1$} & \raisebox{-1.5ex}[0pt]{\footnotesize	 $77.9$ }
&\raisebox{-1.5ex}[0pt]{ \footnotesize	$33.1$}	 &\raisebox{-1.5ex}[0pt]{ \footnotesize	 CVPR16 }
& \footnotesize	The contextual information outside the region of interest is integrated using spatial recurrent neural networks. \\
\cline{2-10}
 &  \raisebox{-1ex}[0pt]{\footnotesize	CPF \cite{ Shrivastava2016}} & \raisebox{-1ex}[0pt]{\footnotesize	 RPN}	& \raisebox{-1ex}[0pt]{\footnotesize VGG16}
&\raisebox{-2.5ex}[0pt]{ \footnotesize \shortstack [c] {	 Faster \\ RCNN} }&\raisebox{-2.5ex}[0pt]{ \footnotesize	 \shortstack [c] {$76.4$\\ (07+12)}}
& \raisebox{-2ex}[0pt]{\footnotesize \shortstack [c] { $72.6$\\ (07T+12) }}& \raisebox{-1ex}[0pt]{\footnotesize
	$-$ }&\raisebox{-1ex}[0pt]{ \footnotesize	ECCV16}	& \footnotesize
\raisebox{-1ex}[0pt]{Use semantic segmentation to provide top-down feedback. } \\
\Xhline{1.5pt}
\footnotesize \multirow{7}{*}{\rotatebox{90}{\scriptsize \textbf{Local Context}$\quad\quad\quad\quad\quad\quad\quad\quad\quad\quad\quad$}} & \footnotesize	 \raisebox{-3ex}[0pt]{MRCNN \cite{Gidaris2015}} & \footnotesize	 \raisebox{-3ex}[0pt]{SS} & \footnotesize	 \raisebox{-3ex}[0pt]{VGG16}	& \footnotesize \raisebox{-3ex}[0pt]{SPPNet} & \footnotesize \raisebox{-4ex}[0pt]{\shortstack [c] {$78.2$ \\(07+12)}}& \footnotesize \raisebox{-4ex}[0pt]{\shortstack [c] { $73.9$\\(07+12)} }& \footnotesize
	\raisebox{-3ex}[0pt]{$-$}	& \footnotesize\raisebox{-3ex}[0pt]{ ICCV15 }& \footnotesize	 
Extract features from multiple regions surrounding or inside the object proposals. Integrate the semantic segmentation-aware features. \\
\cline{2-10}
 & \raisebox{-6ex}[0pt]{\footnotesize	GBDNet
\cite{ GBDCNN2016, Zeng2017Crafting}} & \raisebox{-6ex}[0pt]{\footnotesize	 CRAFT
\cite{ CRAFT2016}} &\raisebox{-7ex}[0pt]{ \footnotesize \shortstack [c] {Inception v2\\ResNet269\\PolyNet \cite{Zhang2017PolyNet} } }&\raisebox{-7ex}[0pt]{ \footnotesize \shortstack [c] {Fast \\ RCNN} }&\raisebox{-7ex}[0pt]{ \footnotesize\shortstack [c] { $77.2$\\
(07+12)} } &\raisebox{-6ex}[0pt]{ \footnotesize $-$} &\raisebox{-6ex}[0pt]{ \footnotesize $27.0$}
&\raisebox{-6ex}[0pt]{ \footnotesize	\shortstack [c] {ECCV16\\ TPAMI18} } & \footnotesize	 
A GBDNet module to learn the relations of multiscale contextualized regions surrounding an object proposal; GBDNet passes messages among features from different context regions through convolution between neighboring support regions in two directions. \\
\cline{2-10}
 & \raisebox{-2.5ex}[0pt]{\footnotesize ACCNN\cite{Li2017Attentive}} & \raisebox{-2.5ex}[0pt]{\footnotesize
	SS }& \raisebox{-2.5ex}[0pt]{\footnotesize VGG16 }& \raisebox{-2.5ex}[0pt]{\footnotesize \shortstack [c] {Fast \\ RCNN} }&\raisebox{-2.5ex}[0pt]{ \footnotesize \shortstack [c] { $72.0$\\
(07+12) }}& \raisebox{-2.5ex}[0pt]{\footnotesize \shortstack [c] {	 $70.6$\\ (07T+12)} }&\raisebox{-5.5ex}[0pt]{ \footnotesize
	$-$ }&\raisebox{-2.5ex}[0pt]{ \footnotesize TMM17	}& \footnotesize 
Use  LSTM  to capture global context. Concatenate features from multi-scale contextual regions surrounding an object proposal. The global and local context features are concatenated for recognition. \\
\cline{2-10}
  & \raisebox{-2.5ex}[0pt]{\footnotesize	CoupleNet\cite{ ZhuICCV2017}}	 &\raisebox{-2.5ex}[0pt]{ \footnotesize RPN}	 &\raisebox{-2.5ex}[0pt]{ \footnotesize ResNet101}
& \raisebox{-2.5ex}[0pt]{\footnotesize 	RFCN	 }&\raisebox{-2.5ex}[0pt]{ \footnotesize \shortstack [c] { $\textbf{82.7}$\\(07+12)}}
&\raisebox{-2.5ex}[0pt]{  \footnotesize\shortstack [c] { $\textbf{80.4}$ \\(07T+12)} }& \raisebox{-2.5ex}[0pt]{\footnotesize
	$34.4$}&\raisebox{-2.5ex}[0pt]{ \footnotesize	ICCV17} & \footnotesize	
Concatenate features from multiscale contextual regions surrounding an object proposal. Features of different contextual regions are then combined by convolution and element-wise sum. \\
\cline{2-10}
  & \raisebox{-2.5ex}[0pt]{\footnotesize SMN \cite{ChenSpatial2017}}	 & \raisebox{-2.5ex}[0pt]{\footnotesize RPN} & \raisebox{-2.5ex}[0pt]{\footnotesize 	 VGG16	 } & \raisebox{-3.5ex}[0pt]{\footnotesize\shortstack [c] {	 Faster \\ RCNN} } &\raisebox{-3.5ex}[0pt]{ \footnotesize \shortstack [c] {$70.0$ \\(07)}} & \raisebox{-2.5ex}[0pt]{\footnotesize
	$-$} & \raisebox{-2.5ex}[0pt]{\footnotesize $-$}& \raisebox{-2.5ex}[0pt]{\footnotesize 	 ICCV17}	 & \footnotesize 
Model object-object relationships efficiently through a spatial memory network. Learn the functionality of NMS automatically. 
\\
\cline{2-10}
&\raisebox{-3.5ex}[0pt]{ \footnotesize ORN \cite{Hu2018Relation}}
& \raisebox{-3.5ex}[0pt]{\footnotesize	 RPN} &\raisebox{-5ex}[0pt]{ \footnotesize	
 \shortstack [c] {ResNet101\\+DCN}}& \raisebox{-5ex}[0pt]{\footnotesize	
  \shortstack [c] {	 Faster \\ RCNN}}&\raisebox{-3.5ex}[0pt]{ \footnotesize	 $-$}
  & \raisebox{-3.5ex}[0pt]{\footnotesize	 $-$} & \raisebox{-3.5ex}[0pt]{\footnotesize	 $\textbf{39.0}$}&\raisebox{-3.5ex}[0pt]{ \footnotesize	 CVPR18} & \footnotesize	 
Model the relations of a set of object proposals through the interactions between their appearance features and geometry. Learn the functionality of NMS automatically. 
\\
\cline{2-10}
&\raisebox{-3.5ex}[0pt]{ \footnotesize SIN \cite{Liu2018Structure}} & \raisebox{-3.5ex}[0pt]{\footnotesize RPN} &\raisebox{-3.5ex}[0pt]{ \footnotesize	 VGG16}& \raisebox{-4.5ex}[0pt]{\footnotesize	 \shortstack [c] {	 Faster \\ RCNN}}&\raisebox{-4.5ex}[0pt]{ \footnotesize	 \shortstack [c] {$76.0$\\(07+12)}} &
 \raisebox{-4.5ex}[0pt]{\footnotesize\shortstack [c] {	 $73.1$\\(07T+12)}}
 & \raisebox{-3.5ex}[0pt]{\footnotesize	 \shortstack [c] {$23.2$}}&\raisebox{-3.5ex}[0pt]{ \footnotesize	 CVPR18} & \footnotesize	
Formulate object detection as graph-structured inference, where objects are graph nodes and relationships the edges.\\
\Xhline{1.5pt}
\end{tabular}
}
\end{table*}

 \begin {figure*}[!t]
\centering
\includegraphics[width=0.9\textwidth]{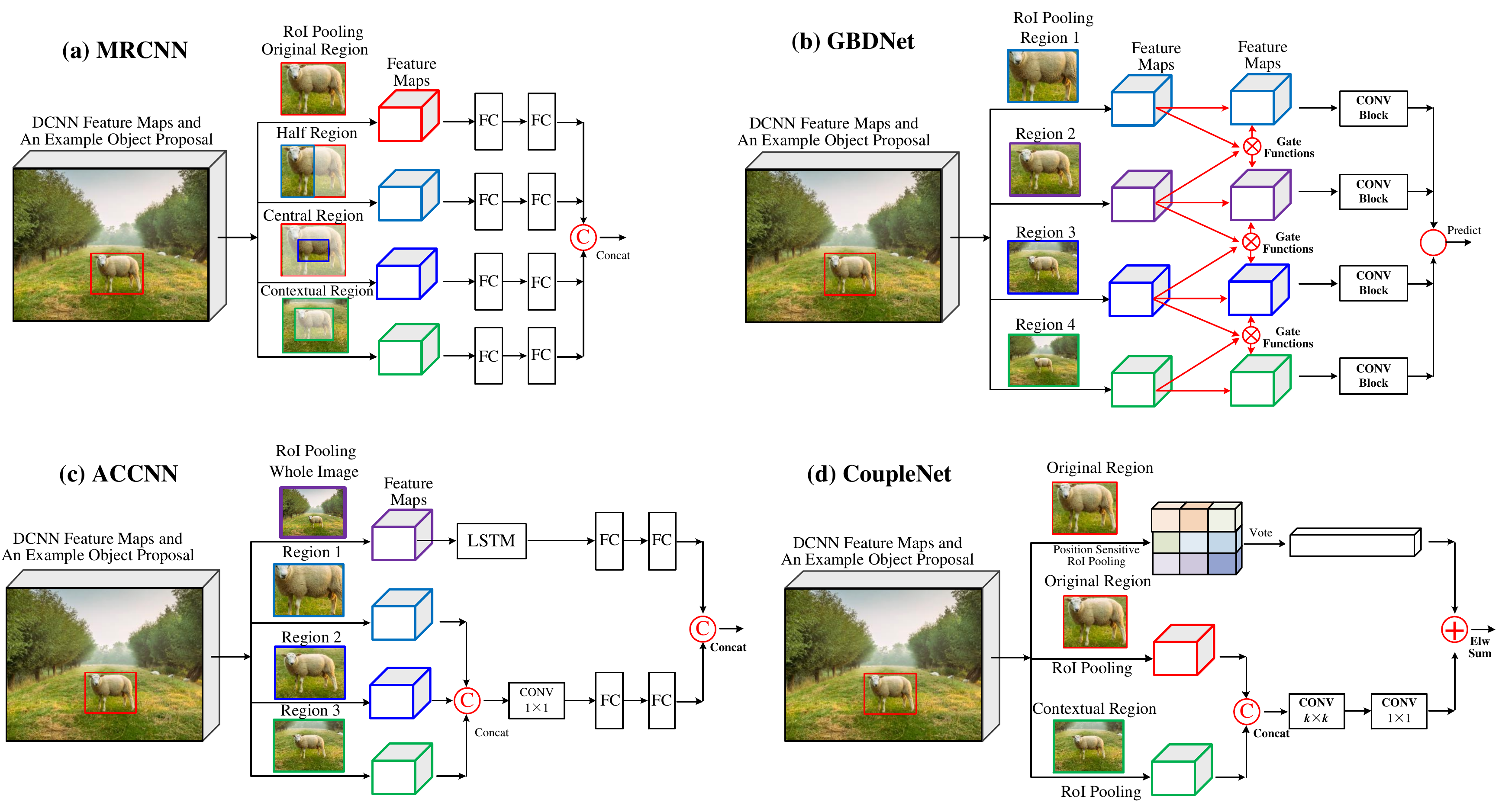}
\caption{Representative approaches that explore local surrounding contextual features:
MRCNN \cite{Gidaris2015}, GBDNet \cite{ GBDCNN2016,Zeng2017Crafting},
ACCNN \cite{Li2017Attentive} and CoupleNet \cite{ZhuICCV2017};
also see Table~\ref{Tab:ContextMethods}.}
\label{Fig:LocalContext}
\end {figure*}

In the physical world, visual objects occur in particular environments and
usually coexist with other related objects.  There is strong
psychological evidence \cite{Biederman1972Contextual,Bar2004Visual} that context plays an essential role in human object recognition, and it is recognized that a proper modeling of context
helps object detection and recognition \cite{Torralba2003,Oliva2007Role,Chen2016deeplab,
Chen2015Semantic,Divvala2009,Galleguillos2010},
especially when object appearance features are insufficient because of small object size,
object occlusion, or poor image quality. Many different types of context have been discussed  \cite{Divvala2009,Galleguillos2010}, and
can broadly be grouped into one of three categories:
\begin{enumerate}
\item Semantic context:  The likelihood of an object to be found in some scenes, but not in others;
\item Spatial context:  The likelihood of finding an object in some position and not others with respect to other objects in the scene;
\item Scale context:  Objects have a limited set of sizes relative to other objects in the scene.
\end{enumerate}
A great deal of work \cite{Chen2015c,Divvala2009,Galleguillos2010,Malisiewicz09Beyond,
 Murphy03Using,Rabinovich2007Objects,Parikh2012} preceded the prevalence of deep learning, and much of this work has yet to be explored in DCNN-based object detectors \cite{ChenSpatial2017,Hu2018Relation}.

The current state of the art in object detection \cite{Ren2015NIPS,Liu2016SSD,MaskRCNN2017}
 detects objects without explicitly exploiting any contextual information. It is broadly agreed that DCNNs
make use of contextual information implicitly \cite{ZeilerFergus2014,Zheng15Conditional}
since they learn hierarchical representations with multiple levels of abstraction. Nevertheless, there is value in exploring
contextual information explicitly in DCNN based detectors
\cite{Hu2018Relation,ChenSpatial2017,Zeng2017Crafting}, so the following
reviews recent work in exploiting contextual cues in DCNN-
based object detectors, organized into categories of {\em global} and {\em local} contexts, motivated by earlier work in \cite{Zhang13,Galleguillos2010}. Representative approaches are
summarized in Table~\ref{Tab:ContextMethods}.

\subsection{Global Context}
Global context  \cite{Zhang13,Galleguillos2010} refers to image or scene level contexts, which can serve as cues for object detection
(\emph{e.g.,} a bedroom will predict the presence of a bed).
In DeepIDNet \cite{Ouyang2015deepid}, the image classification
scores were used as contextual features, and concatenated with
the object detection scores to improve detection results.
In ION \cite{Bell2016ION}, Bell \emph{et al.} proposed to
use spatial Recurrent Neural Networks (RNNs) to explore contextual information across the entire image.  In SegDeepM \cite{SegDeepM2015}, Zhu \emph{et al.} proposed a Markov random field model that scores appearance as
well as context for each detection, and allows each candidate box to select a segment out of a large pool of object segmentation proposals and score the agreement
between them. In \cite{Shrivastava2016}, semantic segmentation was used
 as a form of contextual priming.

\subsection{Local Context}
Local context \cite{Zhang13,Galleguillos2010,Rabinovich2007Objects}
considers the relationship among locally
nearby objects, as well as the interactions  between an object and its surrounding area.  In general, modeling object relations is challenging, requiring reasoning about bounding boxes of different classes, locations, scales \emph{etc}.
Deep learning research that explicitly models object relations is quite limited,
with representative ones being Spatial Memory Network (SMN) \cite{ChenSpatial2017},
 Object Relation Network \cite{Hu2018Relation}, and Structure
Inference Network (SIN) \cite{Liu2018Structure}.  In SMN,
spatial memory essentially assembles object instances back into a
pseudo image representation that is easy to be fed into
another CNN for object relations reasoning, leading to a new sequential reasoning architecture where image
and memory are processed in parallel to obtain detections
which further update memory.
Inspired by the recent success of attention modules
in natural language processing \cite{Vaswani2017Attention}, ORN processes a set of objects simultaneously through the interaction
between their appearance feature and geometry.  It does not require additional supervision, and it is
easy to embed into existing networks, effective in
improving object recognition and duplicate removal steps
in modern object detection pipelines, giving rise to the first fully end-to-end object detector.
SIN \cite{Liu2018Structure} considered two kinds of context: scene contextual information
and object relationships within a single image. It formulates object
detection as a problem of graph inference, where the objects are treated as nodes in a graph
and relationships between objects are modeled as edges.

A wider range of methods has approached the context challenge with a simpler idea: enlarging
the detection window size to extract some form of local context. Representative approaches include
MRCNN \cite{Gidaris2015}, Gated BiDirectional CNN (GBDNet) \cite{ GBDCNN2016,Zeng2017Crafting},
Attention to Context CNN (ACCNN) \cite{Li2017Attentive}, CoupleNet \cite{ZhuICCV2017}, and
Sermanet \emph{et al.} \cite{Sermanet2013c}.
In MRCNN \cite{Gidaris2015} (Fig.~\ref{Fig:LocalContext} (a)), in addition to the features
extracted from the original object proposal at the last CONV layer of the backbone,
Gidaris and Komodakis proposed to extract features from a number of
different regions of an object proposal (half regions, border regions, central regions,
contextual region and semantically segmented regions),
in order to obtain a richer and more robust object representation.
All of these features are combined by concatenation.

Quite a number of methods, all closely related to MRCNN, have been proposed since then.  The method in \cite{Zagoruyko2016} used only four contextual regions, organized in a foveal structure, where the classifiers along multiple paths are trained jointly end-to-end.
Zeng \emph{et al.}
proposed GBDNet \cite{GBDCNN2016,Zeng2017Crafting} (Fig.~\ref{Fig:LocalContext} (b))
to extract features from multiscale contextualized regions
surrounding an object proposal to improve detection performance.
In contrast to the somewhat naive approach of learning CNN features
for each region separately and then concatenating them, GBDNet passes messages
among features from different contextual regions.
Noting that message passing is not always
helpful, but dependent on individual samples,
Zeng \emph{et al.} \cite{GBDCNN2016} used gated functions
to control message transmission. Li \emph{et al.} \cite{Li2017Attentive}
presented ACCNN (Fig.~\ref{Fig:LocalContext} (c)) to utilize both global and local contextual information:  the global context was captured using a Multiscale Local Contextualized (MLC) subnetwork, which recurrently generates an attention map
for an input image to highlight promising contextual locations; local context adopted a method similar to that of MRCNN \cite{Gidaris2015}.
As shown in Fig.~\ref{Fig:LocalContext} (d), CoupleNet \cite{ZhuICCV2017} is conceptually similar to ACCNN \cite{Li2017Attentive}, but built upon RFCN \cite{Dai2016RFCN}, which captures object information with position sensitive RoI pooling, CoupleNet added a branch to encode the global context with RoI pooling.

\section{Detection Proposal Methods}
\label{Sec:DetectionProposal}
An object can be located at any position and scale in
an image. During the heyday of handcrafted feature
descriptors (SIFT \cite{Lowe2004}, HOG \cite{Dalal2005HOG} and
LBP \cite{Ojala02}), the most successful methods for object detection (\emph{e.g.} DPM \cite{Felzenszwalb08CVPR}) used \emph{sliding window} techniques \cite{Viola2001,Dalal2005HOG,Felzenszwalb08CVPR,Harzallah2009Combining,Vedaldi09Multiple}.
However, the number of windows is huge, growing with the number of pixels in an image, and the need to search at multiple scales and aspect ratios further increases the search space\footnote{Sliding window based detection requires classifying around $10^4$-$10^5$
windows per image. The number of windows grows significantly to $10^6$-$10^7$ windows per image when considering multiple scales and aspect ratios.}. Therefore, it is computationally too expensive to apply sophisticated classifiers.

Around 2011, researchers proposed to relieve the tension between computational tractability and high detection quality by using
\emph{detection proposals}\footnote{We use the terminology \emph{detection proposals},
\emph{object proposals} and \emph{region proposals}
interchangeably.} \cite{Van2011SS,Uijlings2013b}.
Originating in the idea of \emph{objectness} proposed by \cite{Alexe2010Object},
object proposals are a set of candidate regions in an image that are likely to contain objects, and if high object recall can be achieved with a modest number of object proposals (like one hundred), significant speed-ups over the sliding window approach can be gained, allowing the use of more sophisticated classifiers.
Detection proposals are usually used as a pre-processing step, limiting the number of regions that need to be evaluated by the detector, and should have the following characteristics:
\begin{enumerate}
\item High recall, which can be achieved with only a few proposals;
\item Accurate localization, such that the proposals match the object bounding
boxes as accurately as possible; and
\item Low computational cost.
\end{enumerate}
The success of object detection based on detection proposals \cite{Van2011SS,Uijlings2013b} has attracted broad interest \cite{Carreira2012,Arbelaez2014,Alexe2012,Cheng2014bing,EdgeBox2014,Endres14Category,
Philipp14Geodesic,Manen2013Prime}.  A comprehensive review of object proposal algorithms is beyond the scope of this paper, because object proposals have applications beyond object detection \cite{Arbelaez2012Semantic,Guillaumin2014ImageNet,Xhu2017Soft}.
We refer interested readers to the recent surveys \cite{Hosang2016,Chavali2016} which provide in-depth analysis of many classical object proposal algorithms and their impact on detection performance.  Our interest here is to review object proposal methods that are based on DCNNs, output class agnostic proposals, and are related to generic object detection.

In 2014, the integration of object proposals \cite{Van2011SS,Uijlings2013b} and DCNN features \cite{Krizhevsky2012} led to the milestone RCNN \cite{Girshick2014RCNN} in generic object detection. Since then, detection proposal has quickly become a standard preprocessing step, based on the fact that all winning entries in the PASCAL VOC \cite{Everingham2010}, ILSVRC \cite{Russakovsky2015} and MS COCO \cite{Lin2014} object detection challenges since 2014 used detection proposals \cite{Girshick2014RCNN,Ouyang2015deepid,Girshick2015FRCNN,
Ren2015NIPS,Zeng2017Crafting,MaskRCNN2017}.

Among object proposal approaches based on traditional low-level cues (\emph{e.g.,} color, texture, edge and gradients), Selective Search \cite{Uijlings2013b}, MCG \cite{Arbelaez2014} and EdgeBoxes \cite{EdgeBox2014} are among the more popular. As the domain rapidly progressed,
 traditional object proposal approaches \cite{Uijlings2013b,Hosang2016,EdgeBox2014},
which were adopted as external modules independent of the detectors, became the speed bottleneck
of the detection pipeline \cite{Ren2015NIPS}. An emerging class of object proposal algorithms
\cite{MultiBox1,Ren2015NIPS,DeepBox2015,Deepproposal2015,DeepMask2015,CRAFT2016}
using DCNNs has attracted broad attention.

\begin{table*}[!t]
\caption {Summary of object proposal methods using DCNN. Blue indicates the number of object proposals. The detection results on COCO are based on mAP@IoU[0.5, 0.95], unless stated otherwise.}\label{Tab:ObjectProposals}
\centering
\renewcommand{\arraystretch}{1.2}
\setlength\arrayrulewidth{0.2mm}
\setlength\tabcolsep{1pt}
\resizebox*{18cm}{!}{
\begin{tabular}{!{\vrule width1.5bp}c|c|p{1.5cm}<{\centering}|c|c|c|c|c|c|c|c|p{9cm}<{\centering}!{\vrule width1.5bp}}
\Xhline{1.5pt}
\footnotesize $\quad\quad$  & \footnotesize Proposer & \footnotesize Backbone  & \footnotesize Detector &  \multicolumn{3}{c|}{Recall@IoU (VOC07)} 	 &\multicolumn{3}{c|}{Detection Results (mAP)}   & \footnotesize  Published &  \footnotesize   \\
\cline{5-7}\cline{8-10}
\footnotesize  & \footnotesize Name & \footnotesize Network  & \footnotesize Tested  & \footnotesize $0.5$ & \footnotesize $0.7$ & \footnotesize	$0.9$ & \footnotesize  VOC07  & \footnotesize  VOC12  & \footnotesize  COCO   & \footnotesize  In &  \footnotesize
Highlights \\
\cline{2-12}
\footnotesize \multirow{8}{*}{\rotatebox{90}{Bounding Box Object Proposal Methods$\quad\quad\quad\quad\quad\quad\quad\quad$ }} & \footnotesize \raisebox{-3ex}[0pt]{MultiBox1\cite{ MultiBox1}}	& \footnotesize	 \raisebox{-3.5ex}[0pt]{AlexNet} & \footnotesize \raisebox{-3ex}[0pt]{RCNN}& \footnotesize \raisebox{-3ex}[0pt]{$-$}& \footnotesize \raisebox{-3ex}[0pt]{$-$} & \footnotesize \raisebox{-3ex}[0pt]{$-$}	& \footnotesize \raisebox{-5.5ex}[0pt]{ \shortstack [c] {$29.0$ \\(\textcolor{blue}{10})\\ (12)}}	& \footnotesize \raisebox{-3ex}[0pt]{ $-$}	& \footnotesize \raisebox{-3ex}[0pt]{$-$}	& \footnotesize \raisebox{-3ex}[0pt]{CVPR14}	& \footnotesize  Learns a class agnostic regressor on a small set of 800 predefined anchor boxes. Do not share features for detection. \\
\cline{2-12}
  & \footnotesize \raisebox{-3.5ex}[0pt]{	DeepBox
\cite{ DeepBox2015}} & \footnotesize  \raisebox{-3.5ex}[0pt]{ VGG16 } & \footnotesize \raisebox{-3.5ex}[0pt]{\shortstack [c] {Fast \\ RCNN}}	 & \footnotesize \raisebox{-3.5ex}[0pt]{\shortstack [c] {$0.96$ \\ (\textcolor{blue}{1000})}}& \footnotesize \raisebox{-3.5ex}[0pt]{\shortstack [c] {$0.84$ \\ (\textcolor{blue}{1000})}}& \footnotesize \raisebox{-3.5ex}[0pt]{\shortstack [c] {$0.15$ \\ (\textcolor{blue}{1000})}}& \footnotesize \raisebox{-3.5ex}[0pt]{$-$}
& \footnotesize \raisebox{-3.5ex}[0pt]{ $-$} & \footnotesize \raisebox{-5.5ex}[0pt]{\shortstack [c] {$37.8$\\(\textcolor{blue}{500})\\(IoU@0.5)}	} & \footnotesize \raisebox{-3.5ex}[0pt]{ICCV15} & \footnotesize	 Use a lightweight CNN to learn to rerank proposals generated by EdgeBox. Can run at 0.26s per image. Do not share features for detection. \\
\cline{2-12}
  & \footnotesize\raisebox{-4.5ex}[0pt]{ RPN\cite{Ren2015NIPS,Ren2016a}} & \footnotesize \raisebox{-4.5ex}[0pt]{ \shortstack [c] { VGG16 } }& \footnotesize \raisebox{-4.5ex}[0pt]{ \shortstack [c] {Faster \\ RCNN}}& \footnotesize \raisebox{-7.5ex}[0pt]{\shortstack [c] {$0.97$ \\ (\textcolor{blue}{300})\\0.98\\(\textcolor{blue}{1000})}}& \footnotesize  \raisebox{-7.5ex}[0pt]{\shortstack [c] {$0.79$ \\ (\textcolor{blue}{300})\\0.84\\(\textcolor{blue}{1000})}}& \footnotesize  \raisebox{-7.5ex}[0pt]{\shortstack [c] {$0.04$ \\ (\textcolor{blue}{300})\\0.04\\(\textcolor{blue}{1000})}}& \footnotesize \raisebox{-6.5ex}[0pt]{\shortstack [c] { $73.2$ \\ (\textcolor{blue}{300})\\(07+12)}} & \footnotesize \raisebox{-6.5ex}[0pt]{\shortstack [c] { $70.4$ \\ (\textcolor{blue}{300})\\(07++12)} }& \footnotesize \raisebox{-4.5ex}[0pt]{ \shortstack [c] { $21.9$ \\ (\textcolor{blue}{300})}	}& \footnotesize\raisebox{-4.5ex}[0pt]{ NIPS15} & \footnotesize 
  The first to generate object proposals by sharing full image convolutional features with detection. 
  Most widely used object proposal method. Significant improvements in detection speed. \\
\cline{2-12}
  & \footnotesize \raisebox{-3.5ex}[0pt]{DeepProposal\cite{ Deepproposal2015}	}& \footnotesize	 \raisebox{-3.5ex}[0pt]{ VGG16	 } & \footnotesize \raisebox{-3.5ex}[0pt]{\shortstack [c] {Fast \\ RCNN}}& \footnotesize  \raisebox{-7.5ex}[0pt]{\shortstack [c] {$0.74$ \\ (\textcolor{blue}{100})\\0.92\\(\textcolor{blue}{1000})}}& \footnotesize \raisebox{-7.5ex}[0pt]{\shortstack [c] {$0.58$ \\ (\textcolor{blue}{100})\\0.80\\(\textcolor{blue}{1000})}}& \footnotesize\raisebox{-7.5ex}[0pt]{\shortstack [c] {$0.12$ \\ (\textcolor{blue}{100})\\0.16\\(\textcolor{blue}{1000})}}& \footnotesize \raisebox{-6.5ex}[0pt]{ \shortstack [c] {$53.2$ \\ (\textcolor{blue}{100})\\(07)}}	 & \footnotesize \raisebox{-3.5ex}[0pt]{ $-$}& \footnotesize \raisebox{-3.5ex}[0pt]{$-$}& \footnotesize\raisebox{-3.5ex}[0pt]{ ICCV15} & \footnotesize	 Generate proposals inside a DCNN in a multiscale manner.
  Share features with the detection network.\\
\cline{2-12}
  & \footnotesize \raisebox{-3.5ex}[0pt]{CRAFT \cite{ CRAFT2016}} & \footnotesize \raisebox{-3.5ex}[0pt]{ VGG16} & \footnotesize \raisebox{-3.5ex}[0pt]{\shortstack [c] {Faster \\ RCNN}} & \footnotesize \raisebox{-3.5ex}[0pt]{\shortstack [c] {$0.98$ \\ (\textcolor{blue}{300}) }}& \footnotesize \raisebox{-3.5ex}[0pt]{\shortstack [c] {$0.90$ \\ (\textcolor{blue}{300}) }}& \footnotesize \raisebox{-3.5ex}[0pt]{\shortstack [c] {$0.13$ \\ (\textcolor{blue}{300}) }}& \footnotesize	 \raisebox{-3.5ex}[0pt]{\shortstack [c] {$75.7$ \\ (07+12) }}& \footnotesize \raisebox{-3.5ex}[0pt]{ \shortstack [c] {71.3\\ (12)}}	 & \footnotesize \raisebox{-3.5ex}[0pt]{ $-$} & \footnotesize\raisebox{-3.5ex}[0pt]{	 CVPR16} & \footnotesize	
Introduced a classification network (\emph{i.e.} two class Fast RCNN) cascade that comes after the RPN. Not sharing features extracted for detection.\\
\cline{2-12}
  & \footnotesize\raisebox{-3ex}[0pt]{ AZNet \cite{ Lu2016Adaptive}	 }& \footnotesize
 \raisebox{-3ex}[0pt]{VGG16 }& \footnotesize	 \raisebox{-4ex}[0pt]{\shortstack [c] {Fast \\ RCNN}} & \footnotesize\raisebox{-4ex}[0pt]{ \shortstack [c] {$0.91$ \\ (\textcolor{blue}{300}) }} & \footnotesize \raisebox{-4ex}[0pt]{\shortstack [c] {$0.71$ \\ (\textcolor{blue}{300}) }}& \footnotesize \raisebox{-4ex}[0pt]{\shortstack [c] {$0.11$ \\ (\textcolor{blue}{300}) }}& \footnotesize \raisebox{-4ex}[0pt]{\shortstack [c] {$70.4$\\
(07)}	}& \footnotesize \raisebox{-3.5ex}[0pt]{$-$}& \footnotesize \raisebox{-3.5ex}[0pt]{$22.3$} & \footnotesize \raisebox{-3.5ex}[0pt]{CVPR16 }& \footnotesize	
Use coarse-to-fine search: start from large regions, then recursively search for subregions that may contain objects. Adaptively guide computational resources to focus on likely subregions. \\
\cline{2-12}
 & \footnotesize \raisebox{-3.5ex}[0pt]{ ZIP \cite{Hongyang2018Zoom}} & \footnotesize \raisebox{-3.5ex}[0pt]{Inception v2} & \footnotesize \raisebox{-3.5ex}[0pt]{ \shortstack [c] {Faster \\ RCNN}} & \footnotesize \raisebox{-5.5ex}[0pt]{\shortstack [c] {$0.85$ \\ (\textcolor{blue}{300})\\COCO }} & \footnotesize \raisebox{-5.5ex}[0pt]{\shortstack [c] {$0.74$ \\ (\textcolor{blue}{300})\\COCO }} & \footnotesize\raisebox{-5.5ex}[0pt]{\shortstack [c] {$0.35$ \\ (\textcolor{blue}{300})\\COCO }}  & \footnotesize\raisebox{-3ex}[0pt]{ \shortstack [c] {  $79.8$\\ (07+12)}}
& \footnotesize \raisebox{-3ex}[0pt]{	$-$ }& \footnotesize	 \raisebox{-3ex}[0pt]{$-$} & \footnotesize\raisebox{-3ex}[0pt]{	 IJCV18 }   & \footnotesize	
Generate proposals using conv-deconv network with multilayers; Proposed a map attention decision (MAD) unit to assign the weights for features from different layers. \\
\cline{2-12}
  & \footnotesize	 \raisebox{-3.5ex}[0pt]{DeNet\cite{SmithICCV2017}} & \footnotesize \raisebox{-3.5ex}[0pt]{ResNet101 } & \footnotesize\raisebox{-4.5ex}[0pt]{ \shortstack [c] {Fast \\ RCNN}} & \footnotesize \raisebox{-3.5ex}[0pt]{\shortstack [c] {$0.82$ \\ (\textcolor{blue}{300})}}& \footnotesize \raisebox{-3.5ex}[0pt]{\shortstack [c] {$0.74$ \\ (\textcolor{blue}{300})}}& \footnotesize\raisebox{-3.5ex}[0pt]{\shortstack [c] {$0.48$ \\ (\textcolor{blue}{300})}} & \footnotesize \raisebox{-3.5ex}[0pt]{\shortstack [c] {$77.1$ \\ (07+12)}}& \footnotesize \raisebox{-3.5ex}[0pt]{\shortstack [c] { $73.9$ \\ (07++12)} }& \footnotesize \raisebox{-3.5ex}[0pt]{$33.8$} & \footnotesize \raisebox{-3.5ex}[0pt]{ ICCV17 }& \footnotesize A lot faster than Faster RCNN; Introduces a bounding box corner
estimation for predicting object proposals efficiently
 to replace RPN; Does not require
predefined anchors.\\
\Xhline{1.5pt}
\footnotesize  & \footnotesize \shortstack [c] {Proposer\\Name} & \footnotesize \shortstack [c] {Backbone\\Network}  & \footnotesize \shortstack [c] { Detector \\ Tested}&  \multicolumn{3}{c|}{\shortstack [c] {Box Proposals \\ (AR, COCO)}} &  \multicolumn{3}{c|}{\shortstack [c] {Segment Proposals\\ (AR, COCO)}}  & \footnotesize \shortstack [c] { \shortstack [c] {Published \\ In}}  &  \footnotesize Highlights   \\
\cline{2-12}
 \footnotesize \multirow{5}{*}{\rotatebox{90}{Segment Proposal Methods $\quad\quad$ }} & \footnotesize \raisebox{-3.5ex}[0pt]{DeepMask \cite{DeepMask2015}}& \footnotesize	
\raisebox{-3.5ex}[0pt]{VGG16 }& \footnotesize \raisebox{-3.5ex}[0pt]{	 \shortstack [c] {Fast \\ RCNN}} &\multicolumn{3}{c|}{\raisebox{-3.5ex}[0pt]{$0.33$ (\textcolor{blue}{100}), $0.48(\textcolor{blue}{1000})$}}  &\multicolumn{3}{c|}{\raisebox{-3.5ex}[0pt]{$0.26$ (\textcolor{blue}{100}), $0.37(\textcolor{blue}{1000})$}}  & \footnotesize \raisebox{-3.5ex}[0pt]{ NIPS15 }& \footnotesize First to generate object mask proposals with DCNN; Slow inference time; Need segmentation annotations for training; Not sharing features with detection network; Achieved mAP of $69.9\%$ (\textcolor{blue}{500}) with Fast RCNN. \\
\cline{2-12}
 	 & \footnotesize \raisebox{-3ex}[0pt]{InstanceFCN  \cite{Dai2016Instance}}& \footnotesize
\raisebox{-3ex}[0pt]{ VGG16}& \footnotesize \raisebox{-3ex}[0pt]{$-$}&\multicolumn{3}{c|}{\raisebox{-3ex}[0pt]{$-$}}  &\multicolumn{3}{c|}{\raisebox{-3ex}[0pt]{$0.32$ (\textcolor{blue}{100}), $0.39(\textcolor{blue}{1000})$}} & \footnotesize\raisebox{-3ex}[0pt]{ECCV16}	& \footnotesize 
Combines ideas of FCN \cite{FCNCVPR2015} and DeepMask \cite{DeepMask2015}. Introduces instance sensitive score maps. Needs segmentation annotations to train the network. \\
\cline{2-12}
 	 & \footnotesize \raisebox{-3.5ex}[0pt]{SharpMask \cite{Pinheiro2016}	}& \footnotesize
\raisebox{-3.5ex}[0pt]{ MPN  \cite{Zagoruyko2016}}& \footnotesize \raisebox{-3.5ex}[0pt]{\shortstack [c] {Fast \\ RCNN}}&\multicolumn{3}{c|}{\raisebox{-3.5ex}[0pt]{$0.39$ (\textcolor{blue}{100}), $0.53(\textcolor{blue}{1000})$}}  &\multicolumn{3}{c|}{\raisebox{-3.5ex}[0pt]{$0.30$ (\textcolor{blue}{100}), $0.39(\textcolor{blue}{1000})$}}& \footnotesize\raisebox{-3.5ex}[0pt]{	 ECCV16}	 & \footnotesize Leverages features at multiple convolutional layers by introducing a top-down refinement module. Does not share features with detection network. Needs segmentation annotations for training. 
\\
\cline{2-12}
 & \footnotesize \raisebox{-3.5ex}[0pt]{FastMask\cite{ Hu2017FastMask}	}& \footnotesize \raisebox{-3.5ex}[0pt]{ResNet39} & \footnotesize \raisebox{-3.5ex}[0pt]{$-$} &\multicolumn{3}{c|}{\raisebox{-3.5ex}[0pt]{$0.43$ (\textcolor{blue}{100}), $0.57(\textcolor{blue}{1000})$}} & \multicolumn{3}{c|}{\raisebox{-3.5ex}[0pt]{$0.32$ (\textcolor{blue}{100}), $0.41(\textcolor{blue}{1000})$}} & \footnotesize \raisebox{-3.5ex}[0pt]{	 CVPR17} & \footnotesize  Generates instance segment proposals efficiently in one-shot manner similar to SSD \cite{Liu2016SSD}. Uses multiscale convolutional features. Uses segmentation annotations for training. \\
\Xhline{1.5pt}
\end{tabular}
}
\end{table*}

Recent DCNN based object proposal methods generally fall into two categories:
{\em bounding box} based and {\em object segment} based, with representative methods summarized in Table \ref{Tab:ObjectProposals}.

\begin {figure}[!t]
\centering
\includegraphics[width=0.4\textwidth]{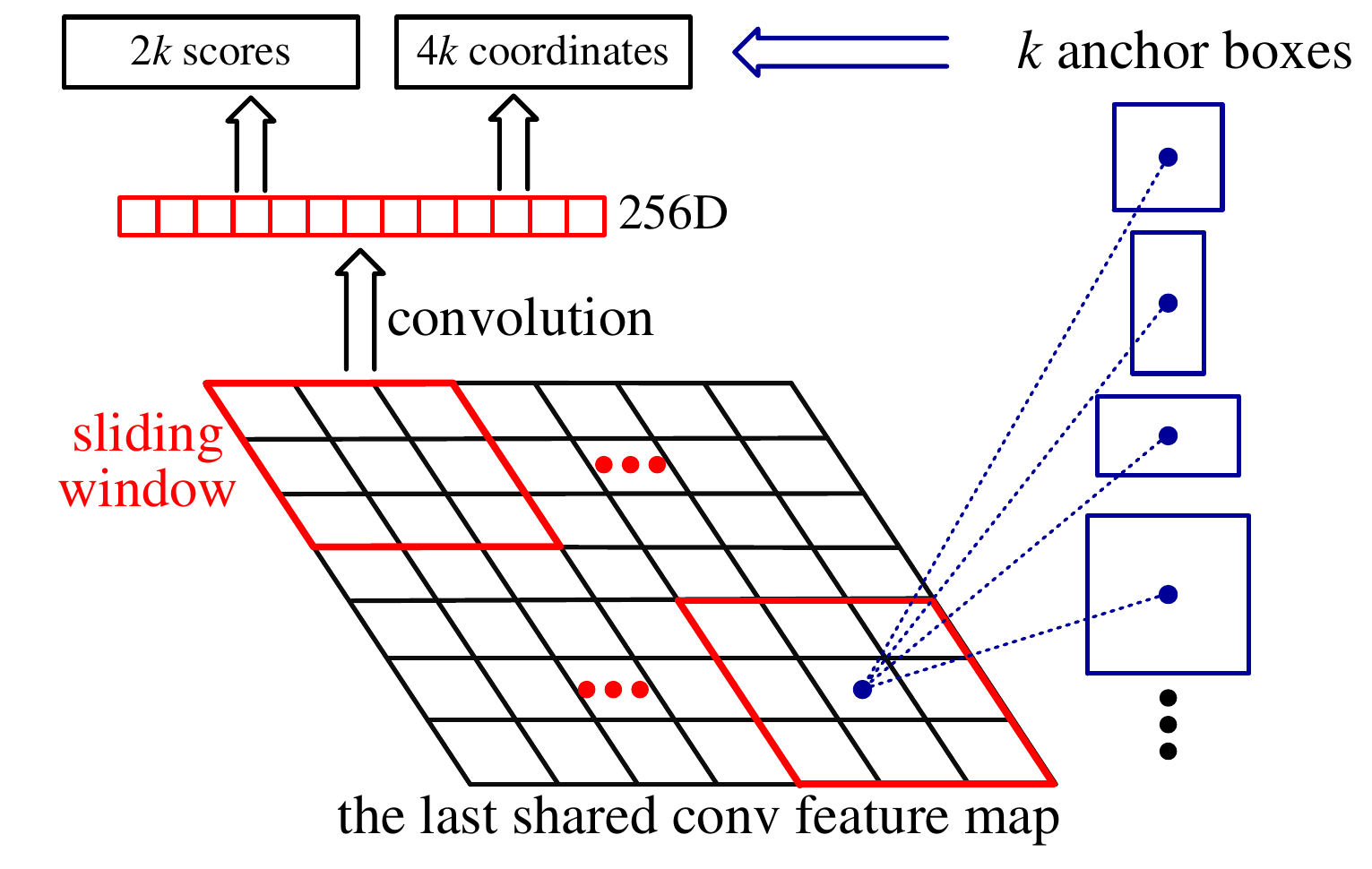}
\caption{Illustration of the Region Proposal Network (RPN) introduced in \cite{Ren2015NIPS}.}
\label{fig:anchor}
\end {figure}

\textbf{Bounding Box Proposal Methods} are best exemplified by the RPC method \cite{Ren2015NIPS} of Ren \emph{et al.}, illustrated in Fig.~\ref{fig:anchor}.
RPN predicts object proposals by sliding a small network over the feature map of the last shared CONV layer. At each sliding window location, $k$ proposals are predicted by using $k$ anchor boxes, where each anchor box\footnote{The concept of ``anchor'' first appeared in \cite{Ren2015NIPS}.} is centered at some location in the image, and is associated with a particular scale and aspect ratio.
Ren \emph{et al.} \cite{Ren2015NIPS} proposed  integrating RPN and Fast RCNN into a single network by sharing their convolutional layers, leading to Faster RCNN, the first end-to-end detection pipeline. RPN has been broadly selected as the proposal method by many state-of-the-art object detectors, as can be observed from Tables \ref{Tab:EnhanceFeatures} and \ref{Tab:ContextMethods}.

Instead of fixing {\em a priori} a set of anchors as MultiBox \cite{MultiBox1,MultiBox2} and RPN \cite{Ren2015NIPS}, Lu \emph{et al.} \cite{Lu2016Adaptive} proposed  generating anchor locations by using a recursive search strategy which can adaptively guide computational resources to focus on sub-regions likely to contain objects. Starting with the whole image,
all regions visited during the search process serve as anchors.
For any anchor region encountered during the search procedure, a
scalar zoom indicator is used to decide whether to
further partition the region, and a set of bounding
boxes with objectness scores are computed by an Adjacency and Zoom Network (AZNet), which extends RPN by adding a branch to
compute the scalar zoom indicator in parallel with the existing branch.

Further work attempts to generate object proposals
by exploiting multilayer convolutional features.
Concurrent with RPN \cite{Ren2015NIPS}, Ghodrati \emph{et al.} \cite{Deepproposal2015} proposed DeepProposal, which generates object proposals
by using a cascade of multiple convolutional features, building an inverse cascade to select the most promising object locations and to refine
their boxes in a coarse-to-fine manner. An improved variant of RPN, HyperNet \cite{HyperNet2016} designs Hyper Features which aggregate multilayer convolutional features and shares them both in generating proposals
and detecting objects via an end-to-end joint training strategy.
Yang \emph{et al.} proposed CRAFT \cite{ CRAFT2016}
which also used a cascade strategy, first
training an RPN network to generate object proposals and then using them
 to train another binary Fast RCNN network to further distinguish objects from background.
Li \emph{et al.} \cite{Hongyang2018Zoom} proposed ZIP to improve RPN by predicting object proposals with multiple
 convolutional feature maps at different network depths to integrate both low level
details and high level semantics. The backbone used in ZIP is a ``zoom out and in''
network inspired by the conv and deconv structure \cite{FCNCVPR2015}.

Finally, recent work which deserves mention includes Deepbox \cite{DeepBox2015}, which proposed a lightweight CNN to learn to rerank proposals generated by EdgeBox, and DeNet \cite{SmithICCV2017}
which introduces bounding box corner
estimation to predict object proposals efficiently
 to replace RPN in a Faster RCNN style detector.

\textbf{Object Segment Proposal Methods} \cite{DeepMask2015,Pinheiro2016}
aim to generate segment proposals that are likely to correspond to objects.
Segment proposals are more informative than bounding box proposals, and
take a step further towards object instance segmentation \cite{Hariharan2014,Dai2016Aware,Li2017Fully}. In addition, using instance segmentation supervision can improve the performance of bounding box object detection. The pioneering work of DeepMask, proposed by Pinheiro \emph{et al.} \cite{DeepMask2015},  segments proposals learnt directly from raw image data with a deep network.  Similarly to RPN,
after a number of shared convolutional layers DeepMask splits the network into two branches in order to predict a class agnostic mask and an associated objectness score. Also similar to the efficient sliding window strategy in OverFeat \cite{OverFeat2014}, the trained DeepMask network is applied in a sliding window manner to an image (and its rescaled versions) during inference.  More recently, Pinheiro \emph{et al.} \cite{Pinheiro2016} proposed SharpMask by augmenting
the DeepMask architecture with a refinement module, similar to the architectures shown in Fig.~\ref{fig:MultiLayerCombine} (b1) and (b2),
augmenting the feed-forward network with a top-down refinement process. SharpMask can efficiently integrate  spatially rich information from early features with strong semantic information encoded in later layers to generate high fidelity object masks.

Motivated by Fully Convolutional Networks (FCN) for semantic segmentation \cite{FCNCVPR2015} and DeepMask \cite{DeepMask2015}, Dai \emph{et al.} proposed InstanceFCN \cite{Dai2016Instance} to generate instance segment proposals. Similar to DeepMask, the InstanceFCN network is split into two fully convolutional branches, one to generate instance sensitive score maps, the other to
predict the objectness score. Hu \emph{et al.} proposed FastMask \cite{Hu2017FastMask} to efficiently generate instance segment proposals in a one-shot manner, similar to SSD \cite{Liu2016SSD},
in order to make use of  multiscale convolutional features. Sliding windows extracted densely from
 multiscale convolutional feature maps were input to a scale-tolerant attentional head module in order to predict segmentation masks and objectness scores. FastMask is claimed to run at 13 FPS on
$800\times600$ images.

\begin{table*}[!t]
\caption {Representative methods for training strategies and class imbalance handling. Results on COCO are reported with Test Dev. The detection results on COCO are based on mAP@IoU[0.5, 0.95].}\label{Tab:ClassImbalance}
\centering
\renewcommand{\arraystretch}{1.2}
\setlength\arrayrulewidth{0.2mm}
\setlength\tabcolsep{1pt}
\resizebox*{18.5cm}{!}{
\begin{tabular}{!{\vrule width1.5bp}c|c|c|c|c|c|c|c|p{8cm}<{\centering}!{\vrule width1.5bp}}
\Xhline{1.5pt}
\footnotesize \shortstack [c] {Detector \\ Name}  & \footnotesize \shortstack [c] {Region \\ Proposal} & \footnotesize \shortstack [c] {Backbone \\ DCNN} & \footnotesize \shortstack [c] {Pipelined \\ Used}	 & \footnotesize  \shortstack [c] {VOC07 \\ Results} & \footnotesize   \shortstack [c] {VOC12 \\ Results} & \footnotesize  \shortstack [c] {COCO \\ Results}  & \footnotesize   \shortstack [c] {Published \\ In} &  \footnotesize
Highlights \\
\Xhline{1.5pt}
	 \raisebox{-4ex}[0pt]{MegDet \cite{Peng2018MegDet}} & \footnotesize \raisebox{-4ex}[0pt]{ RPN} & \footnotesize \raisebox{-5ex}[0pt]{ \shortstack [c] { ResNet50\\+FPN }}& \footnotesize
	\raisebox{-5ex}[0pt]{ \shortstack [c] { Faster\\RCNN}} & \footnotesize  \raisebox{-4ex}[0pt]{$-$}& \footnotesize  \raisebox{-4ex}[0pt]{$-$} & \footnotesize  \raisebox{-4ex}[0pt]{$52.5$}	 & \footnotesize \raisebox{-4ex}[0pt]{CVPR18}	& \footnotesize Allow training with much larger minibatch size than before by introducing cross GPU batch normalization; Can finish the COCO training in 4 hours on 128 GPUs and achieved improved accuracy; Won COCO2017 detection challenge. \\
\hline
\raisebox{-2.5ex}[0pt]{SNIP
\cite{Singh2018sniper} }& \footnotesize	 \raisebox{-2.5ex}[0pt]{RPN	 }& \footnotesize \raisebox{-3.5ex}[0pt]{ \shortstack [c] {DPN \cite{Chen2017Dual}\\+DCN \cite{Dai17Deformable}}}	 & \footnotesize \raisebox{-2.5ex}[0pt]{RFCN} & \footnotesize	 \raisebox{-2.5ex}[0pt]{$-$ }& \footnotesize	 \raisebox{-2.5ex}[0pt]{$-$}& \footnotesize	 \raisebox{-2.5ex}[0pt]{$48.3$}
	 & \footnotesize \raisebox{-2.5ex}[0pt]{CVPR18} & \footnotesize
	 A new multiscale training scheme.
	Empirically examined the effect of up-sampling for small object detection. During training, only select objects that fit the scale of features as positive samples. \\
\hline
	\raisebox{-1ex}[0pt]{SNIPER
\cite{Singh2018sniper}} & \footnotesize	 \raisebox{-1ex}[0pt]{RPN}	& \footnotesize \raisebox{-2.5ex}[0pt]{\shortstack [c] {ResNet101\\+DCN}}	 & \footnotesize \raisebox{-2.5ex}[0pt]{\shortstack [c] {Faster \\ RCNN}}	& \footnotesize\raisebox{-1ex}[0pt]{ $-$}	 & \footnotesize \raisebox{-1ex}[0pt]{$-$}	& \footnotesize \raisebox{-1ex}[0pt]{$47.6$} & \footnotesize\raisebox{-1ex}[0pt]{
	2018	}&
	An efficient multiscale training strategy. Process context regions around ground-truth instances at the appropriate scale. \\
\hline
	\raisebox{-1.5ex}[0pt]{OHEM \cite{Shrivastava2016OHEM}	}& \footnotesize \raisebox{-1.5ex}[0pt]{SS} & \footnotesize	 \raisebox{-1.5ex}[0pt]{VGG16} & \footnotesize \raisebox{-2.5ex}[0pt]{ \shortstack [c] {	Fast \\ RCNN} }& \footnotesize\raisebox{-2.5ex}[0pt]{ \shortstack [c] {$78.9$\\
(07+12)}}& \footnotesize \raisebox{-2.5ex}[0pt]{\shortstack [c] {	 $76.3$\\
(07++12)}}& \footnotesize	 \raisebox{-1.5ex}[0pt]{$22.4$}& \footnotesize
	\raisebox{-1.5ex}[0pt]{CVPR16}& \footnotesize	
	A simple and effective Online Hard Example Mining algorithm to improve training of region based detectors. \\
\hline
	\raisebox{-1.5ex}[0pt]{ FactorNet \cite{Ouyang2016Factors}	}& \footnotesize \raisebox{-1.5ex}[0pt]{SS} & \footnotesize	 \raisebox{-1.5ex}[0pt]{GooglNet} & \footnotesize \raisebox{-2.5ex}[0pt]{ \shortstack [c] {RCNN} }& \footnotesize\raisebox{-2.5ex}[0pt]{$-$ } & \footnotesize \raisebox{-2.5ex}[0pt]{$-$ } & \raisebox{-2.5ex}[0pt]{$-$ } & \footnotesize
	\raisebox{-1.5ex}[0pt]{CVPR16}& \footnotesize Identify the  imbalance in the number of samples for different object categories; propose a divide-and-conquer feature  learning scheme.  \\
\hline
	\raisebox{-5.5ex}[0pt]{Chained Cascade \cite{CascadeRCNN2018}	}& \footnotesize \raisebox{-7ex}[0pt]{\shortstack [c] {SS \\ CRAFT} } & \footnotesize	 \raisebox{-7ex}[0pt]{\shortstack [c] {VGG \\ Inceptionv2}} & \footnotesize \raisebox{-7.5ex}[0pt]{ \shortstack [c] {Fast RCNN, \\ Faster RCNN} }& \footnotesize\raisebox{-8.5ex}[0pt]{ \shortstack [c] {$80.4$\\
(07+12) \\ (SS+VGG)}}& \footnotesize \raisebox{-5.5ex}[0pt]{$-$}& \footnotesize	 \raisebox{-5.5ex}[0pt]{$-$}& \footnotesize
	\raisebox{-5.5ex}[0pt]{ICCV17}& \footnotesize Jointly learn DCNN and multiple stages of cascaded classifiers. Boost detection accuracy on PASCAL VOC 2007 and ImageNet for both fast RCNN and Faster RCNN using different region proposal methods.  \\
\hline
	\raisebox{-5.5ex}[0pt]{Cascade RCNN \cite{CascadeRCNN2018}	}& \footnotesize \raisebox{-5.5ex}[0pt]{RPN} & \footnotesize	 \raisebox{-8.5ex}[0pt]{\shortstack [c] {VGG\\ResNet101\\+FPN}} & \footnotesize \raisebox{-6.5ex}[0pt]{ \shortstack [c] {Faster RCNN}}& \footnotesize\raisebox{-6.5ex}[0pt]{ \shortstack [c] {$-$}}& \footnotesize \raisebox{-6.5ex}[0pt]{\shortstack [c] {$-$}}& \footnotesize	 \raisebox{-5.5ex}[0pt]{$42.8$}& \footnotesize
	\raisebox{-5.5ex}[0pt]{CVPR18}& \footnotesize Jointly learn DCNN and multiple stages of cascaded classifiers, which are  learned using different localization accuracy for selecting positive samples. Stack bounding box regression at multiple stages. \\
\hline
	\raisebox{-3ex}[0pt]{RetinaNet \cite{LinICCV2017}}	& \footnotesize \raisebox{-4ex}[0pt]{$-$}	 & \footnotesize \raisebox{-4ex}[0pt]{\shortstack [c] {ResNet101\\+FPN }}& \footnotesize	 \raisebox{-3ex}[0pt]{RetinaNet }& \footnotesize	 \raisebox{-3ex}[0pt]{$-$}& \footnotesize	 \raisebox{-3ex}[0pt]{$-$}	 & \footnotesize \raisebox{-3ex}[0pt]{$39.1$} & \footnotesize
\raisebox{-3ex}[0pt]{ICCV17} & \footnotesize Propose a novel Focal Loss which focuses training on hard examples.  Handles well the problem of imbalance of positive and negative samples when training a one-stage detector.	\\
\Xhline{1.5pt}
\end{tabular}
}
\end{table*}

\section{Other Issues}
\label{sec:otherissue}
\textbf{Data Augmentation.} Performing data augmentation for learning DCNNs \cite{Chatfield2014,Girshick2015FRCNN,Girshick2014RCNN} is generally recognized to be important for visual recognition. Trivial data augmentation refers to perturbing an image by
transformations that leave the underlying category unchanged, such as
cropping, flipping, rotating, scaling, translating, color perturbations, and adding noise. By artificially enlarging the number of samples, data augmentation helps in reducing overfitting and improving generalization.
It can be used at training time, at test time, or both.
Nevertheless, it has the obvious limitation that the time required for training increases significantly. Data augmentation may synthesize completely new training images \cite{Peng2015Learning,Wang2017}, however it is hard to guarantee that the synthetic images generalize well to real ones. Some researchers \cite{Dwibedi2017Cut,Gupta2016Synthetic} proposed augmenting datasets by pasting real segmented objects into natural images; indeed, Dvornik \emph{et al.} \cite{Dvornik2018Modeling}  showed
that appropriately modeling the visual context surrounding objects is
crucial to place them in the right environment, and proposed a context model to automatically find appropriate locations on images to place new objects for data augmentation.

\textbf{Novel Training Strategies.} Detecting objects under a wide range of scale variations, especially the detection of very small objects, stands out as a key challenge.
It has been shown \cite{Huang2016Speed,Liu2016SSD}
that image resolution has a considerable impact on detection accuracy, therefore scaling is particularly commonly used in data augmentation, since higher resolutions increase the possibility of detecting small objects \cite{Huang2016Speed}.
Recently, Singh \emph{et al.} proposed advanced and efficient
data argumentation methods SNIP \cite{Singh2018SNIP}  and SNIPER \cite{Singh2018sniper}  to illustrate the scale invariance problem, as summarized in Table \ref{Tab:ClassImbalance}. Motivated by the intuitive understanding that
small and large objects are difficult to detect at smaller
and larger scales, respectively, SNIP introduces a novel training
scheme that can reduce scale variations during training, but without
reducing training samples;  SNIPER allows for efficient multiscale training, only processing context regions around ground truth objects at the appropriate scale, instead of processing a whole image pyramid.
Peng \emph{et al.} \cite{Peng2018MegDet} studied a key factor in training, the minibatch size, and proposed MegDet, a Large MiniBatch Object Detector, to enable the training with a much larger minibatch size than before (from 16 to
256). To avoid the failure of convergence and significantly speed up the training process, Peng \emph{et al.} \cite{Peng2018MegDet} proposed a learning rate policy and
Cross GPU Batch Normalization, and effectively utilized 128 GPUs, allowing MegDet to finish COCO training in 4 hours on 128 GPUs, and winning the COCO 2017 Detection
Challenge.

\begin {figure}[!t]
\centering
\includegraphics[width=0.4\textwidth]{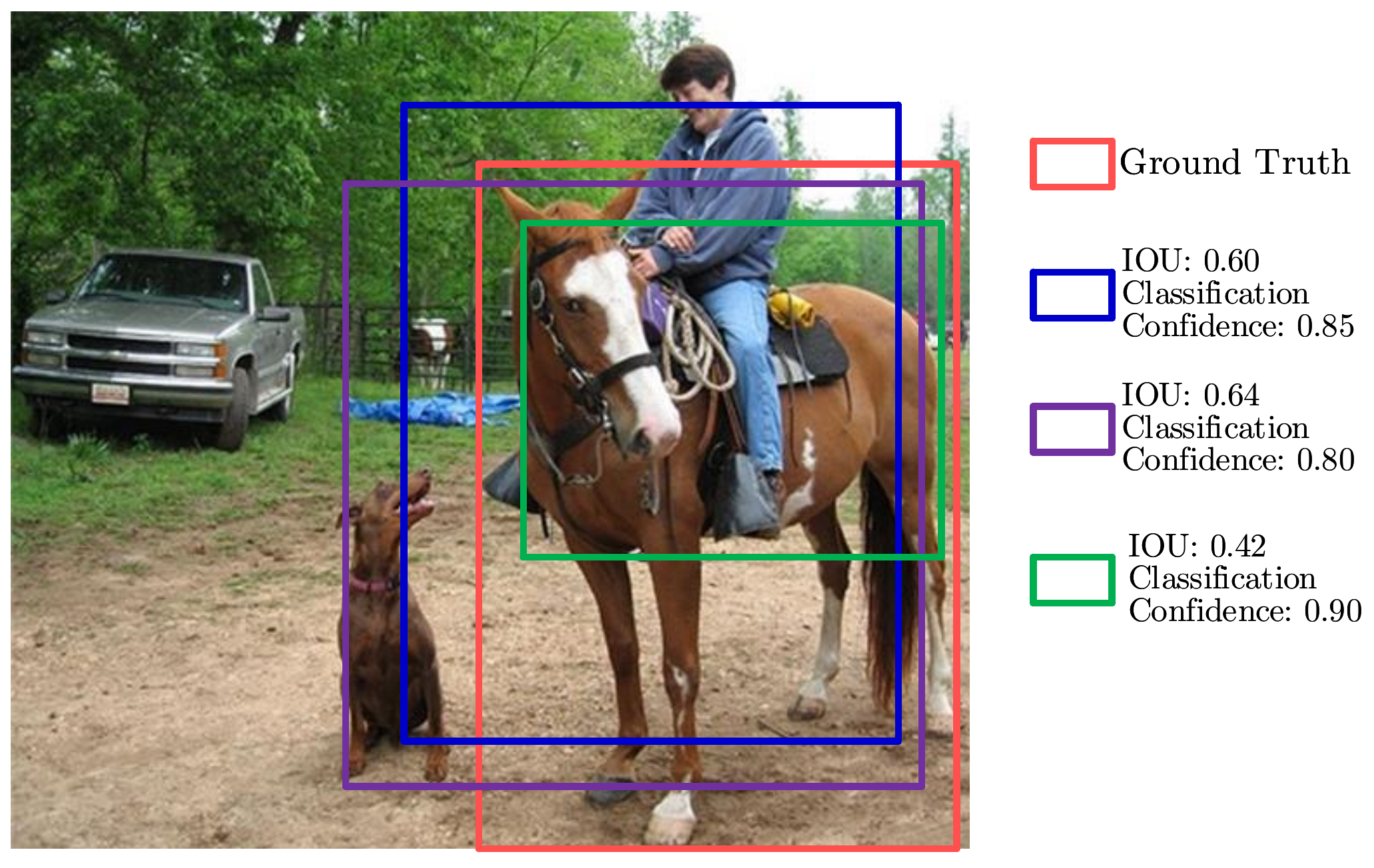}
\caption{Localization error could stem from insufficient overlap or duplicate detections. Localization error is a frequent cause of false positives.}
\label{fig:iou}
\end {figure}

\textbf{Reducing Localization Error.} In object detection, the Intersection Over Union\footnote{Please refer to Section \ref{sec:EvaluationCriteria} for more details on the definition of IOU.} (IOU) between a detected bounding box and its ground truth box is the most popular evaluation metric, and an IOU threshold (\emph{e.g.} typical value of $0.5$) is required to define positives and negatives. From Fig.~\ref{Fig:RegionVsUnified}, in most state of the art detectors \cite{Girshick2015FRCNN,Liu2016SSD,MaskRCNN2017,Ren2015NIPS,YoLo2016} object detection is formulated as a multitask learning problem, \emph{i.e.,} jointly optimizing a softmax classifier which assigns object proposals with class labels and bounding box regressors, localizing objects by maximizing IOU or other metrics between detection results and ground truth. Bounding boxes are only a crude approximation for articulated objects,  consequently background pixels are almost invariably included in a bounding box, which affects the accuracy of classification and localization. The study in \cite{Hoiem2012} shows that object localization error is one of the most influential forms of error, in addition to confusion between similar objects. Localization error could stem from insufficient overlap (smaller than the required IOU threshold, such as the green box in Fig.~\ref{fig:iou}) or duplicate detections (\emph{i.e.,} multiple overlapping detections for an object instance). Usually, some post-processing step like NonMaximum Suppression (NMS) \cite{Bodla2017Soft,Hosang2017Learning} is used for eliminating duplicate
detections. However, due to misalignments the bounding box with better localization could be suppressed during NMS, leading to poorer localization quality (such as the purple box shown in Fig.~\ref{fig:iou}). Therefore, there are quite a few methods aiming at improving detection performance by reducing localization error.

MRCNN \cite{Gidaris2015} introduces iterative bounding box regression, where an RCNN is applied several times. CRAFT \cite{CRAFT2016} and AttractioNet \cite{Gidaris2016Attend} use a multi-stage detection sub-network
to generate accurate proposals, to forward to Fast RCNN.  Cai and Vasconcelos proposed Cascade RCNN \cite{CascadeRCNN2018}, a multistage extension of RCNN, in which a sequence of detectors is trained sequentially with increasing IOU thresholds, based on the observation that the output of a detector trained with a certain IOU is a good distribution to train the detector of the next higher IOU threshold, in order to be sequentially more selective
against close false positives. This approach can be built with any RCNN-based detector, and is demonstrated to achieve consistent gains (about 2 to 4 points) independent of the baseline detector strength, at a marginal increase in computation.
There is also recent work  \cite{Jiang2018Acquisition,Rezatofighi2019Generalized,Huang2019Mask} formulating IOU directly as the optimization objective, and in proposing improved NMS results \cite{Bodla2017Soft,He2019Bounding,Hosang2017Learning,Smith2018Improving}, such as Soft NMS \cite{Bodla2017Soft}
and learning NMS \cite{Hosang2017Learning}.

\textbf{Class Imbalance Handling.} Unlike image classification, object detection has another unique problem: the serious imbalance between the number of labeled object instances
and the number of background examples (image regions
not belonging to any object class of interest). Most background examples are easy negatives, however
this imbalance can make the training very inefficient, and the large number of easy negatives tends to overwhelm the training. In the past,  this issue has typically been addressed via techniques such as
bootstrapping \cite{Sung1996Learning}. More recently, this problem has also seen some attention \cite{Li2019Gradient,LinICCV2017,Shrivastava2016OHEM}.
Because the region proposal stage rapidly filters out most background regions and proposes
a small number of object candidates, this class imbalance issue is mitigated to some extent in two-stage detectors \cite{Girshick2014RCNN,Girshick2015FRCNN,Ren2015NIPS,MaskRCNN2017}, although example mining approaches, such as Online Hard
Example Mining (OHEM) \cite{Shrivastava2016OHEM}, may be used to maintain a
reasonable balance between foreground and background. In the
case of one-stage object detectors \cite{YoLo2016,Liu2016SSD}, this imbalance is extremely serious (\emph{e.g.} 100,000 background examples to every
object). Lin \emph{et al.} \cite{LinICCV2017} proposed Focal Loss to address this by rectifying the Cross Entropy loss, such that it down-weights the
loss assigned to correctly classified examples. Li \emph{et al.} \cite{Li2019Gradient}
studied this issue from the perspective of gradient norm distribution, and proposed a Gradient Harmonizing Mechanism (GHM) to handle it.

\section{Discussion and Conclusion}
\label{Sec:Conclusions}
Generic object detection is an important and challenging problem in computer vision and has received considerable attention. Thanks to remarkable developments in deep learning techniques, the field of object detection has dramatically evolved. As a comprehensive
survey on deep learning for generic object detection, this paper has highlighted the recent achievements, provided a structural taxonomy for  methods according
to their roles in detection, summarized existing popular datasets and evaluation criteria, and discussed performance for the most representative methods. We conclude this review with a discussion of the state of the art in Section~\ref{Sec:Performance}, an overall discussion  of key issues in Section~\ref{Sec:Discussion}, and finally suggested future research directions in Section~\ref{Sec:Directions}.

\subsection{State of the Art Performance}
\label{Sec:Performance}
A large variety of detectors has appeared in the last few years, and the introduction of standard benchmarks, such as PASCAL VOC \cite{Everingham2010,Everingham2015}, ImageNet
\cite{Russakovsky2015} and COCO \cite{Lin2014}, has made it easier to compare detectors. As can be seen from our earlier discussion in Sections~\ref{Sec:Frameworks} through~\ref{sec:otherissue}, it may be misleading to compare detectors in terms of their originally reported performance (\emph{e.g.} accuracy, speed), as they can differ in fundamental / contextual respects, including the following choices:
\begin{itemize}
\renewcommand{\labelitemi}{$\bullet$}
  \item Meta detection frameworks, such as RCNN \cite{Girshick2014RCNN}, Fast RCNN \cite{Girshick2015FRCNN}, Faster RCNN \cite{Ren2015NIPS}, RFCN \cite{Dai2016RFCN}, Mask RCNN \cite{MaskRCNN2017}, YOLO \cite{YoLo2016} and SSD \cite{Liu2016SSD};
  \item Backbone networks such as VGG \cite{Simonyan2014VGG}, Inception \cite{GoogLeNet2015,Ioffe2015,Szegedy2016a}, ResNet \cite{He2016ResNet}, ResNeXt \cite{Xie2016Aggregated}, and Xception \cite{Chollet2017Xception} \emph{etc.} listed in Table \ref{Tab:dcnnarchitectures};
  \item Innovations such as multilayer feature combination \cite{FPN2016,Shrivastava2017,DSSD2016}, deformable
convolutional networks \cite{Dai17Deformable}, deformable RoI pooling \cite{Ouyang2015deepid,Dai17Deformable}, heavier heads \cite{Ren2016NOC,Peng2018MegDet}, and lighter heads \cite{Li2018Light};
\item Pretraining with datasets such as ImageNet \cite{Russakovsky2015}, COCO \cite{Lin2014}, Places \cite{Zhou2017Places}, JFT \cite{Hinton2015Distilling} and Open Images \cite{OpenImages2017};
  \item Different detection proposal methods and different numbers of object proposals;
  \item Train/test data augmentation, novel multiscale training strategies \cite{Singh2018SNIP,Singh2018sniper} \emph{etc}, and model ensembling.
\end{itemize}

Although it may be impractical to compare every recently proposed
detector, it is nevertheless valuable to integrate representative and publicly available detectors into a common platform and to compare them in a unified manner. There has been very limited work in this regard, except for Huang's study \cite{Huang2016Speed} of the three main families of detectors (Faster RCNN \cite{Ren2015NIPS}, RFCN \cite{Dai2016RFCN} and SSD \cite{Liu2016SSD}) by varying the backbone network, image resolution, and the number of box proposals.

\begin {figure}[!t]
\centering
\includegraphics[width=0.49\textwidth]{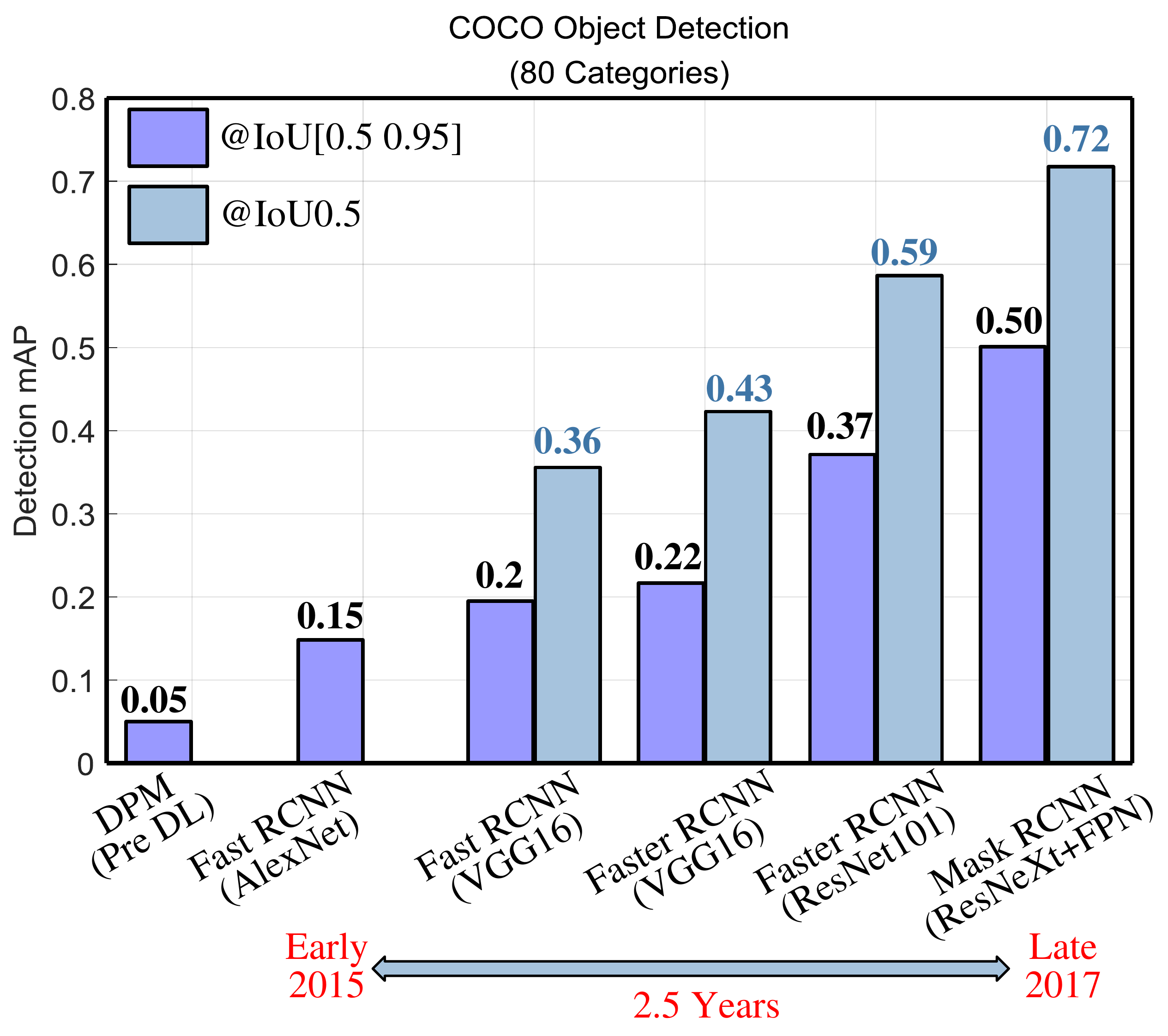}
\caption{Evolution of object detection performance on COCO (Test-Dev results). Results are quoted from \cite{Girshick2015FRCNN,MaskRCNN2017,Ren2016a}. The backbone network, the design of detection framework and the availability of good and large scale datasets are the three most important factors in detection accuracy.}
\label{fig:cocoresults}
\end {figure}

As can be seen from Tables~\ref{Tab:EnhanceFeatures}, \ref{Tab:ContextMethods}, \ref{Tab:ObjectProposals}, \ref{Tab:ClassImbalance}, \ref{Tab:Detectors}, we have summarized the best reported performance of many methods on
three widely used standard benchmarks. The results of these
methods were reported on the same test benchmark, despite their differing in one or more of the aspects listed above.

Figs.~\ref{fig:GODResultsStatistics} and~\ref{fig:cocoresults} present a very brief
overview of the state of the art, summarizing the best detection results of the PASCAL VOC, ILSVRC and MSCOCO challenges; more results can be found at detection challenge websites \cite{ILSVRCResults,COCOResults,VOCResults}. The competition winner of the open image challenge object detection task achieved $61.71\%$ mAP
in the public leader board and $58.66\%$ mAP on the private leader board, obtained by combining the detection results of several two-stage detectors including Fast RCNN \cite{Girshick2015FRCNN}, Faster RCNN \cite{Ren2015NIPS}, FPN  \cite{FPN2016}, Deformable RCNN \cite{Dai17Deformable}, and Cascade RCNN \cite{CascadeRCNN2018}. In summary, the backbone network, the detection framework, and the availability of large scale datasets are the three most important factors in detection accuracy. Ensembles of multiple models, the incorporation of context features, and data augmentation all help to achieve better accuracy.

In less than five years, since AlexNet \cite{Krizhevsky2012} was proposed, the
Top5 error on ImageNet classification \cite{Russakovsky2015} with 1000 classes has dropped from
16\% to 2\%, as shown in Fig.~\ref{fig:ILSVRCclassificationResults}.  However, the mAP of the best performing detector \cite{Peng2018MegDet} on COCO \cite{Lin2014}, trained to detect only 80 classes, is only at $73\%$, even at 0.5 IoU, illustrating how object detection is much
harder than image classification. The accuracy and robustness achieved by
the state-of-the-art detectors  far from satisfies the
requirements of real world applications, so there remains significant
room for future improvement.

\subsection{Summary and Discussion}
\label{Sec:Discussion}
With hundreds of references and many dozens of methods discussed throughout this paper, we would now like to focus on the key factors which have emerged in generic object detection based on deep learning.

\textbf{(1) Detection Frameworks: Two Stage vs. One Stage} \\
In Section~\ref{Sec:Frameworks} we identified two major categories of detection frameworks: region based (two stage) and unified (one stage):
\begin{itemize}
\renewcommand{\labelitemi}{$\bullet$}
  \item When large computational cost is allowed, two-stage detectors generally produce higher detection accuracies than one-stage, evidenced by the fact that most winning approaches used in famous detection challenges like  are predominantly based on two-stage frameworks, because their structure is more flexible and better suited for region based classification.  The most widely used frameworks are Faster RCNN \cite{Ren2015NIPS}, RFCN \cite{Dai2016RFCN} and Mask RCNN \cite{MaskRCNN2017}.
  \item It has been shown in \cite{Huang2016Speed} that the detection accuracy of one-stage SSD \cite{Liu2016SSD} is less sensitive to the quality of the backbone network than representative two-stage frameworks.
  \item  One-stage detectors like YOLO \cite{YoLo2016} and SSD \cite{Liu2016SSD} are generally faster than two-stage ones, because of avoiding preprocessing algorithms,  using lightweight backbone networks, performing prediction with fewer candidate regions, and making the classification subnetwork fully convolutional. However, two-stage detectors can run in real time with the introduction of similar techniques. In any event, whether one stage or two, the most time consuming step is the feature extractor (backbone network) \cite{Law2018CornerNet,Ren2015NIPS}.
  \item It has been shown \cite{Huang2016Speed,YoLo2016,Liu2016SSD} that one-stage frameworks like YOLO and SSD typically have much poorer performance when detecting small objects than two-stage architectures like Faster RCNN and RFCN, but are competitive in detecting large objects.
\end{itemize}
There have been many attempts to build better (faster, more accurate, or more robust) detectors by attacking each stage of the detection
framework. No matter whether one, two or multiple stages, the design of the detection framework has converged towards a number of crucial design choices:
\begin{itemize}
\renewcommand{\labelitemi}{$\bullet$}
\item A fully convolutional pipeline
\item Exploring complementary information from other correlated tasks, \emph{e.g.}, Mask RCNN \cite{MaskRCNN2017}
\item Sliding windows \cite{Ren2015NIPS}
\item Fusing information from different layers of the backbone.
\end{itemize}
The evidence from recent success of cascade for object detection \cite{CascadeRCNN2018,Cheng2018Decoupled,Cheng18Revisiting} and instance segmentation on COCO~\cite{Chen2019Hybrid} and other challenges has shown that multistage object detection could be a future framework for a speed-accuracy trade-off. A teaser investigation is being done in the 2019 WIDER Challenge~\cite{Loy2019Wider}.

\textbf{(2) Backbone Networks} \\
As discussed in Section~\ref{Sec:PopularNetworks}, backbone networks are one of the main driving forces behind the rapid improvement of detection performance, because of the key role played by discriminative object feature representation.
Generally, deeper backbones such as ResNet \cite{He2016ResNet}, ResNeXt \cite{Xie2016Aggregated}, InceptionResNet \cite{InceptionV4} perform better; however, they are computationally more expensive and require much more data and massive computing for training. Some backbones \cite{Howard2017MobileNets,SqueezeNet2016,Zhang18ShuffleNet} were proposed for focusing on speed instead, such as MobileNet \cite{Howard2017MobileNets} which has been shown to achieve VGGNet16 accuracy on ImageNet with only $\frac{1}{30}$ the computational cost and model size.
Backbone training from scratch may become possible
as more training data and better training strategies are available
\cite{Wu2018Group,Luo2019Switchable,Luo2018Towards}.

\textbf{(3) Improving the Robustness of Object Representation}\\
The variation of real world images is a key challenge in object recognition.  The variations include lighting, pose, deformations, background clutter, occlusions, blur, resolution, noise,
and camera distortions.

\textbf{(3.1) Object Scale and Small Object Size} \\
Large variations of object scale, particularly those of small objects, pose a great challenge. Here a summary and discussion on the main strategies identified in Section~\ref{Sec:EnhanceFeatures}:
\begin{itemize}
\renewcommand{\labelitemi}{$\bullet$}
  \item Using image pyramids:  They are simple and effective, helping to enlarge small
  objects and to shrink large ones.  They are  computationally expensive,  but are nevertheless commonly used during inference for better accuracy.
\item Using features from convolutional layers of different resolutions:  In early work like SSD \cite{Liu2016SSD}, predictions are performed independently, and no information from other layers is combined or merged. Now it is quite standard to combine features from different layers, e.g. in FPN \cite{FPN2016}. 
\item Using dilated convolutions \cite{Li2018DetNet,Li2019Scale}:  A simple and effective method to incorporate broader context and maintain high resolution feature maps.
\item Using anchor boxes of different scales and aspect ratios:
Drawbacks of having many parameters, and scales and aspect ratios of anchor boxes are usually heuristically determined.
\item Up-scaling:  Particularly for the detection of small objects, high-resolution networks \cite{Sun2019Deep,Sun2019High} can be developed.  It remains unclear whether super-resolution techniques improve detection accuracy or not.
\end{itemize}
Despite recent advances, the detection accuracy for small objects is still much lower than that of larger ones. Therefore, the detection of small objects remains one of the key challenges in object detection. Perhaps localization requirements need to be generalized as a function of scale, since certain applications, e.g. autonomous driving, only require the  identification of the existence of small objects within a larger region, and exact localization is not necessary.

\textbf{(3.2) Deformation, Occlusion, and other factors} \\
As discussed in Section~\ref{Sec:MainChallenges}, there are approaches to handling geometric transformation, occlusions, and deformation mainly based on
two paradigms. The first is a spatial transformer network, which uses regression to obtain a
deformation field and then  warp features according to the deformation field \cite{Dai17Deformable}.
The second is based on a deformable part-based model \cite{Felzenszwalb2010b},
which finds the maximum response to a part filter with spatial constraints taken into
consideration \cite{Ouyang2015deepid, Girshick2015DPMCNN,Wan2015end}.

Rotation invariance may be attractive in certain applications, but there are limited generic object detection work focusing on rotation invariance, because popular benchmark detection datasets (PASCAL VOC, ImageNet, COCO) do not have large variations in rotation. Occlusion handling is intensively studied in face detection and pedestrian detection, but very little work has been devoted to occlusion handling for generic object detection. In general, despite recent advances, deep networks are still limited by the lack of
robustness to a number of  variations, which significantly constrains their real-world applications.

\textbf{(4) Context Reasoning} \\
As introduced in Section \ref{sec:ContextInfo}, objects in the wild typically coexist with other objects and environments.
It has been recognized that contextual information
(object relations, global scene statistics)
helps object detection and recognition \cite{Oliva2007Role}, especially for small objects, occluded objects, and with poor image quality. There was extensive work preceding deep learning \cite{Malisiewicz09Beyond,Murphy03Using,Rabinovich2007Objects,Divvala2009,Galleguillos2010}, and also quite a few works in the era of deep learning \cite{Gidaris2015,GBDCNN2016,Zeng2017Crafting,ChenSpatial2017,Hu2018Relation}.
How to efficiently and effectively incorporate contextual information remains to be explored, possibly guided by how human vision uses context, based on scene graphs \cite{Li2017Scene}, or via the full segmentation of objects and scenes using panoptic segmentation \cite{Kirillov2018Panoptic}.

\textbf{(5) Detection Proposals} \\
Detection proposals significantly reduce search spaces. As recommended in \cite{Hosang2016}, future detection proposals will surely have to improve in repeatability, recall, localization accuracy, and speed.
Since the success of RPN \cite{Ren2015NIPS}, which integrated proposal generation and detection into a common framework, CNN based detection proposal generation methods have dominated region proposal.
It is recommended that new detection proposals should be assessed
for object detection, instead of evaluating detection proposals alone.

\textbf{(6) Other Factors} \\
As discussed in Section~\ref{sec:otherissue}, there are many other factors affecting object detection quality:  data augmentation, novel training strategies, combinations of backbone models, multiple detection frameworks, incorporating information from other related tasks, methods for reducing localization error,
handling the huge imbalance between positive and negative samples,
mining of hard negative samples, and improving loss functions.

\subsection{Research Directions}
\label{Sec:Directions}
Despite the recent tremendous progress in the field of object detection, the technology remains significantly more primitive than human vision and cannot yet satisfactorily address real-world challenges like those of Section~\ref{Sec:MainChallenges}. We see a number of long-standing challenges:
\begin{itemize}
\renewcommand{\labelitemi}{$\bullet$}
  \item Working in an open world:  being robust to any number of environmental changes, being able to evolve or adapt.
  \item Object detection under constrained conditions:  learning from weakly labeled data or few bounding box annotations, wearable devices, unseen object categories etc.
  \item Object detection in other modalities:  video, RGBD images, 3D point clouds, lidar, remotely sensed imagery \emph{etc}.
\end{itemize}
Based on these challenges, we see the following directions of future research:

\textbf{(1) Open World Learning:} The ultimate goal is to develop object detection  capable of accurately and efficiently recognizing and localizing instances in thousands or more object categories in open-world scenes, at a level competitive with the human visual system. Object detection algorithms are unable, in general, to recognize object categories outside of their training dataset, although ideally there should be the ability to recognize novel object categories \cite{Lake2015Human,Hariharan2017Low}. Current detection datasets \cite{Everingham2010,Russakovsky2015,Lin2014} contain only a few dozen to hundreds
of categories, significantly fewer than those which can be recognized by humans. New larger-scale datasets \cite{Hoffman2014lsda,Singh2018RFCN,YOLO9000} with significantly more categories will need to be developed.

\textbf{(2) Better and More Efficient Detection Frameworks:} One of the reasons for the success in generic object detection has been the development of superior detection frameworks, both region-based (RCNN \cite{Girshick2014RCNN}, Fast RCNN  \cite{Girshick2015FRCNN}, Faster RCNN \cite{Ren2015NIPS}, Mask RCNN \cite{MaskRCNN2017}) and
one-stage detectors (YOLO \cite{YoLo2016}, SSD \cite{Liu2016SSD}). Region-based detectors have higher accuracy, one-stage detectors are generally faster and simpler.
Object detectors depend heavily on the underlying backbone networks, which have been optimized for image classification, possibly causing a learning bias; learning object detectors from scratch could be helpful for new detection frameworks.

\textbf{(3) Compact and Efficient CNN Features:}  CNNs have increased remarkably in depth, from several layers (AlexNet \cite{AlexNet2012}) to hundreds
of layers (ResNet \cite{He2016ResNet}, DenseNet
\cite{Huang2016Densely}).  These networks have millions to hundreds of millions of parameters, requiring massive data and GPUs for training.  In order reduce or remove network redundancy, there has been growing research interest in designing
compact and lightweight networks \cite{Chen2017Learning,Alvarez2016Learning,CondenseNet18,Howard2017MobileNets,
Lin2017Towards,Yu2017NISP} and network acceleration
\cite{Cheng2018Model,Hubara2016Binarized,Han2016Deep,Li2017Pruning,Li2017Mimicking,Wei2018Quantization}.

\textbf{(4) Automatic Neural Architecture Search:} Deep learning bypasses  manual feature engineering which requires human experts with strong domain knowledge, however DCNNs require similarly significant expertise.  It is natural to consider automated design of detection backbone architectures, such as the recent Automated Machine Learning (AutoML) \cite{Quanming2018Taking}, which has been applied to image classification and object detection \cite{Cai2018Path,Chen2019DetNAS,Ghiasi2019NASFPN,Liu2018Progressive,
Zoph2016Neural,Zoph2018Learning}.

\textbf{(5) Object Instance Segmentation:}
For a richer and more
detailed understanding of image content, there is a need to
tackle pixel-level object instance segmentation \cite{Lin2014,MaskRCNN2017,Ronghang2018}, which can play an important role in potential
applications that require the precise boundaries of individual objects.

\textbf{(6) Weakly Supervised Detection:}
Current state-of-the-art detectors employ fully supervised models learned from labeled data with object bounding boxes or segmentation masks
\cite{Everingham2015,Lin2014,Russakovsky2015,Lin2014}. However,
fully supervised learning has serious limitations, particularly where the collection of bounding box annotations is labor intensive and where the number of images is large. Fully supervised learning
is not scalable in the absence of fully labeled training data, so it is essential
to understand how the power of CNNs can be leveraged where only weakly / partially annotated data are provided
 \cite{Bilen2016Weakly,Diba2017Weakly,Shi2017PAMI}.

\textbf{(7) Few / Zero Shot Object Detection:}
The success of deep detectors relies heavily on gargantuan
amounts of annotated training data. When the labeled data
are scarce, the performance of deep detectors frequently deteriorates and fails to
generalize well.  In contrast, humans (even children) can learn a visual concept quickly from very few given examples and can often generalize well \cite{Biederman1987Recognition,Lake2015Human,Fei2006One}. Therefore, the ability to learn from only few examples, \emph{few} shot detection, is very appealing   \cite{Chen2018LSTD,Dong2018Few,Finn2017Model, Kang2018Few,Lake2015Human,Ren2018Meta, Schwartz2019RepMet}.
Even more constrained, \emph{zero} shot object detection localizes and recognizes object classes that have never been seen\footnote{Although side information may be provided, such as a wikipedia page or an attributes vector.} before \cite{Bansa2018Zero,Demirel2018Zero,Rahman2018Zero,Rahman2018Polarity},
essential for life-long learning machines that need to intelligently and incrementally discover new
object categories.

\textbf{(8) Object Detection in Other Modalities:}
Most detectors are based on still 2D images; object detection in other modalities can be highly relevant in domains such as autonomous vehicles,
unmanned aerial vehicles, and robotics.  These modalities raise new challenges in effectively
using depth \cite{Chen20153D,Pepik2015GCPR,Xiang2014Beyond,Wu20153D}, video  \cite{Feichtenhofer17Detect,Kang2016Object}, and point clouds \cite{Qi2017PointNet,Qi2018Frustum}. 

\textbf{(9) Universal Object Detection:} Recently, there has been increasing effort in learning \emph{universal representations}, those which are effective in multiple image domains, such as natural images, videos, aerial images, and medical CT images \cite{Rebuffi2017Learning,Rebuffi2018Efficient}.  Most such research focuses on image classification, rarely targeting
object detection \cite{Wang2019Towards}, and developed detectors are usually domain specific. Object detection independent of image domain and cross-domain object detection represent important future directions.

The research field of generic object detection is still far from complete. However given the breakthroughs over the past five years we are optimistic of future developments and opportunities.

\begin{table*}[!t]
\begin{sideways}
\begin{minipage}{\textheight}
\centering
\caption {Summary of properties and performance of milestone detection frameworks for generic object detection. See Section~\ref{Sec:Frameworks} for a detailed discussion. Some architectures are illustrated in Fig.~\ref{Fig:RegionVsUnified}. The properties of the backbone DCNNs can be found in Table~\ref{Tab:dcnnarchitectures}. }\label{Tab:Detectors}
\renewcommand{\arraystretch}{1}
\setlength\arrayrulewidth{0.2mm}
\setlength\tabcolsep{1pt}
\resizebox*{!}{16cm}{
\begin{tabular}{!{\vrule width1.5bp}c|c|c|c|c|c|c|c|c|c|p{10cm}!{\vrule width1.5bp}}
 \Xhline{1.5pt}
&\scriptsize \shortstack [c] {  Detector\\Name}  & \scriptsize  RP & \scriptsize \shortstack [c] { Backbone\\DCNN} & \scriptsize \shortstack [c] {Input \\ ImgSize} & \scriptsize \shortstack [c] {VOC07 \\Results}& \scriptsize \shortstack [c] { VOC12\\Results }& \scriptsize \shortstack [c] { Speed \\(FPS) }
& \scriptsize  \shortstack [c] {Published \\ In}& \scriptsize  \shortstack [c] { Source\\Code }  & \scriptsize  Highlights and Disadvantages  \\
 \Xhline{1.5pt}
 \multirow{7}*{\hfil \rotatebox{90}{\footnotesize \textbf{Region based} (Section \ref{Sec:RegionBased})$\quad\quad\quad\quad\quad\quad\quad\quad$ }}
&\raisebox{-4.3ex}[0pt]{ \scriptsize RCNN  \cite{Girshick2014RCNN} }&\raisebox{-4.3ex}[0pt]{\scriptsize  SS}   & \raisebox{-4.3ex}[0pt]{\scriptsize AlexNet }
& \raisebox{-4.3ex}[0pt]{\scriptsize Fixed } &\raisebox{-5.3ex}[0pt]{ \scriptsize \shortstack [c] { $58.5$ \\ (07)}} & \scriptsize \raisebox{-6.3ex}[0pt]{ \scriptsize \shortstack [c] { $53.3$ \\ (12)}} &\raisebox{-4.3ex}[0pt]{ \scriptsize $<0.1$} &\raisebox{-4.3ex}[0pt]{ \scriptsize CVPR14}
 &\raisebox{-4.3ex}[0pt]{\scriptsize  \shortstack [c] {Caffe \\ Matlab}} & \scriptsize \textcolor{DarkGreen}{\textbf{Highlights:}} First to integrate CNN with RP methods; Dramatic performance improvement over previous state of the artP.
\par \textcolor{DarkRed}{\textbf{Disadvantages:}}  Multistage pipeline of sequentially-trained  (External RP computation, CNN finetuning,
 each warped RP passing through CNN, SVM and BBR training);
 Training is expensive in space and time; Testing is slow. \\
\cline{2-11}
&\raisebox{-5ex}[0pt]{ \scriptsize SPPNet  \cite{He2014SPP} }&\raisebox{-5ex}[0pt]{\scriptsize  SS  }&\raisebox{-5ex}[0pt]{ \scriptsize  ZFNet } &\raisebox{-5ex}[0pt]{ \scriptsize  Arbitrary }  & \scriptsize \raisebox{-7ex}[0pt]{ \scriptsize \shortstack [c] { $60.9$ \\ (07)}} & \scriptsize \raisebox{-7ex}[0pt]{ \scriptsize  $-$  }&\raisebox{-5ex}[0pt]{ \scriptsize $<1$}&\raisebox{-5ex}[0pt]{ \scriptsize ECCV14}
&\raisebox{-5ex}[0pt]{\scriptsize  \shortstack [c] {Caffe \\ Matlab}} & \scriptsize \textcolor{DarkGreen}{\textbf{Highlights:}} First to introduce SPP into CNN architecture; Enable convolutional feature sharing;
 Accelerate RCNN evaluation by orders of magnitude without sacrificing performance;
 Faster than OverFeat.
\par  \textcolor{DarkRed}{\textbf{Disadvantages:}}  Inherit disadvantages of RCNN; Does not result in much training speedup;
 Fine-tuning not able to update the CONV layers before SPP layer. \\
\cline{2-11}
& \raisebox{-4.5ex}[0pt]{\scriptsize Fast RCNN  \cite{Girshick2015FRCNN}}  & \raisebox{-4.5ex}[0pt]{\scriptsize SS}  & \raisebox{-7ex}[0pt]{ \scriptsize \shortstack [c] { AlexNet\\VGGM\\VGG16 }}&\raisebox{-4.5ex}[0pt]{ \scriptsize  Arbitrary }
  &  \raisebox{-7ex}[0pt]{ \scriptsize \shortstack [c] { $70.0$ \\ (VGG)\\(07+12)}} &\raisebox{-7ex}[0pt]{ \scriptsize \shortstack [c] { $68.4$ \\ (VGG)\\(07++12)}}& \raisebox{-4.5ex}[0pt]{\scriptsize $<1$ }& \raisebox{-4.5ex}[0pt]{\scriptsize ICCV15 }
  &\raisebox{-4.5ex}[0pt]{\scriptsize  \shortstack [c] {Caffe \\ Python}} & \scriptsize \textcolor{DarkGreen}{\textbf{Highlights:}} First to enable end-to-end detector training (ignoring RP generation);
 Design a RoI pooling layer; Much faster and more accurate than SPPNet; No disk storage required for feature caching.
\par \textcolor{DarkRed}{\textbf{Disadvantages:}}  External RP computation is exposed as the new bottleneck; Still too slow for real time applications. \\
\cline{2-11}
 &\raisebox{-6.5ex}[0pt]{ \scriptsize Faster RCNN  \cite{Ren2015NIPS} }&\raisebox{-6.5ex}[0pt]{\scriptsize RPN }  & \raisebox{-7ex}[0pt]{ \scriptsize \shortstack [c] { ZFnet\\VGG }}&\raisebox{-6.5ex}[0pt]{ \scriptsize  Arbitrary }
  &  \raisebox{-9ex}[0pt]{ \scriptsize \shortstack [c] { $73.2$ \\ (VGG)\\(07+12)}} &\raisebox{-9ex}[0pt]{ \scriptsize \shortstack [c] { $70.4$ \\ (VGG)\\(07++12)}}& \raisebox{-6.5ex}[0pt]{\scriptsize $<5$}& \raisebox{-6.5ex}[0pt]{\scriptsize NIPS15}
 &\raisebox{-6.5ex}[0pt]{\scriptsize  \shortstack [c] {Caffe \\ Matlab\\Python}}& \scriptsize \textcolor{DarkGreen}{\textbf{Highlights:}} Propose RPN for generating nearly cost-free and high quality RPs instead of selective search;
 Introduce translation invariant and multiscale anchor boxes as references in RPN;
 Unify RPN and Fast RCNN into a single network by sharing CONV layers;
An order of magnitude faster than Fast RCNN without performance loss;
 Can run testing at 5 FPS with VGG16.
\par  \textcolor{DarkRed}{\textbf{Disadvantages:}}  Training is complex, not a streamlined process;
 Still falls short of real time. \\
\cline{2-11}
&  \raisebox{-2ex}[0pt]{\scriptsize RCNN$\ominus$R   \cite{Lenc2015} }& \raisebox{-2ex}[0pt]{\scriptsize New} & \raisebox{-4ex}[0pt]{ \scriptsize \shortstack [c] { ZFNet\\+SPP}}&\raisebox{-2ex}[0pt]{ \scriptsize  Arbitrary }
& \raisebox{-4ex}[0pt]{ \scriptsize \shortstack [c] { $59.7$ \\(07)}}    & \raisebox{-2ex}[0pt]{ \scriptsize $-$ }&  \raisebox{-2ex}[0pt]{ \scriptsize  $<5$ }& \raisebox{-2ex}[0pt]{ \scriptsize BMVC15}
 &\raisebox{-2ex}[0pt]{\scriptsize  $-$} & \scriptsize \textcolor{DarkGreen}{\textbf{Highlights:}} Replace selective search with static RPs;
Prove the possibility of building integrated, simpler and faster detectors that rely exclusively on CNN.
\par  \textcolor{DarkRed}{\textbf{Disadvantages:}}  Falls short of real time; Decreased accuracy from poor RPs. \\
\cline{2-11}
&\raisebox{-3.5ex}[0pt]{ \scriptsize RFCN  \cite{Dai2016RFCN}  }&\raisebox{-3.5ex}[0pt]{ \scriptsize  RPN} & \raisebox{-3.5ex}[0pt]{\scriptsize ResNet101 } &\raisebox{-3.5ex}[0pt]{ \scriptsize  Arbitrary } &  \raisebox{-6.5ex}[0pt]{ \scriptsize \shortstack [c] { $80.5$\\(07+12)\\$83.6$\\(07+12+CO)}} &\raisebox{-6.5ex}[0pt]{ \scriptsize \shortstack [c] { $77.6$ \\(07++12)\\$82.0$\\(07++12+CO)}}&\raisebox{-3.5ex}[0pt]{ \scriptsize $<10$}&\raisebox{-3.5ex}[0pt]{ \scriptsize NIPS16}
 &\raisebox{-3.5ex}[0pt]{\scriptsize  \shortstack [c] {Caffe \\ Matlab}}& \scriptsize \textcolor{DarkGreen}{\textbf{Highlights:}} Fully convolutional detection network;
Design a set of position sensitive score maps using a bank of specialized CONV layers;
Faster than Faster RCNN without sacrificing much accuracy.
\par  \textcolor{DarkRed}{\textbf{Disadvantages:}}  Training is not a streamlined process;
 Still falls short of real time.  \\
\cline{2-11}
&\raisebox{-4.5ex}[0pt]{ \scriptsize Mask RCNN \cite{MaskRCNN2017}} & \raisebox{-4.5ex}[0pt]{\scriptsize RPN }& \raisebox{-5.5ex}[0pt]{\scriptsize \shortstack [c] { ResNet101 \\ ResNeXt101 } } &\raisebox{-4.5ex}[0pt]{ \scriptsize  Arbitrary }&  \multicolumn{2}{|c|}{\raisebox{-7ex}[0pt]{\scriptsize \shortstack [c] { $50.3$\\(ResNeXt101)\\ (COCO Result)}}}  & \raisebox{-4.5ex}[0pt]{\scriptsize $<5$} & \raisebox{-4.5ex}[0pt]{\scriptsize ICCV17 } &\raisebox{-4.5ex}[0pt]{\scriptsize  \shortstack [c] {Caffe \\ Matlab\\Python}}& \scriptsize  \textcolor{DarkGreen}{\textbf{Highlights:}} A simple, flexible, and effective framework for object instance segmentation; Extends Faster RCNN by adding another branch for predicting an object mask in parallel with the existing branch for BB prediction; Feature Pyramid Network (FPN) is utilized; Outstanding performance.
\par  \textcolor{DarkRed}{\textbf{Disadvantages:}} Falls short of real time applications. \\
 \Xhline{1.5pt}
 \multirow{4}{*}{\rotatebox{90}{\footnotesize \textbf{Unified } (Section \ref{Sec:Unified}) $\quad\quad\quad\quad\quad\quad$ }} & \raisebox{-5.5ex}[0pt]{\scriptsize OverFeat \cite{OverFeat2014} } & \raisebox{-5.5ex}[0pt]{ \scriptsize  $-$}
 & \raisebox{-5.5ex}[0pt]{\scriptsize AlexNet like }&\raisebox{-5.5ex}[0pt]{ \scriptsize  Arbitrary }&\raisebox{-5.5ex}[0pt]{$-$}&\raisebox{-5.5ex}[0pt]{$-$}& \raisebox{-5.5ex}[0pt]{\scriptsize $<0.1$ }& \raisebox{-5.5ex}[0pt]{\scriptsize ICLR14 } &\raisebox{-5.5ex}[0pt]{\scriptsize c++}& \scriptsize \textcolor{DarkGreen}{\textbf{Highlights:}} Convolutional feature sharing;
 Multiscale image pyramid CNN feature extraction; Won the ISLVRC2013 localization competition;
 Significantly faster than RCNN.
\par  \textcolor{DarkRed}{\textbf{Disadvantages:}}  Multi-stage pipeline sequentially trained;
 Single bounding box regressor; Cannot handle multiple object instances of the same class;
 Too slow for real time applications. \\
\cline{2-11}
 & \raisebox{-4.5ex}[0pt]{\scriptsize YOLO \cite{YoLo2016} } & \raisebox{-4.5ex}[0pt]{\scriptsize $-$ }& \raisebox{-4.5ex}[0pt]{ \scriptsize \shortstack [c] { GoogLeNet\\like }}
 &\raisebox{-4.5ex}[0pt]{ \scriptsize  Fixed }&  \raisebox{-5ex}[0pt]{ \scriptsize \shortstack [c] { $66.4$ \\(07+12)}} &\raisebox{-5ex}[0pt]{ \scriptsize \shortstack [c] { $57.9$ \\(07++12)}} &  \raisebox{-4.5ex}[0pt]{\scriptsize \shortstack [c] { $<25$\\(VGG)}} &  \raisebox{-4.5ex}[0pt]{\scriptsize CVPR16} &\raisebox{-4.5ex}[0pt]{\scriptsize  DarkNet}
 & \scriptsize \textcolor{DarkGreen}{\textbf{Highlights:}} First efficient unified detector;
 Drop RP process completely; Elegant and efficient detection framework; Significantly faster than previous detectors;  YOLO runs at 45 FPS, Fast YOLO at 155 FPS;
\par \textcolor{DarkRed}{\textbf{Disadvantages:}} Accuracy falls far behind state of the art detectors; Struggle to localize small objects. \\
\cline{2-11}
& \raisebox{-3.5ex}[0pt]{ \scriptsize YOLOv2\cite{YOLO9000}} &  \raisebox{-3.5ex}[0pt]{\scriptsize $-$ } &  \raisebox{-3.5ex}[0pt]{\scriptsize DarkNet} &\raisebox{-3.5ex}[0pt]{ \scriptsize  Fixed } &  \raisebox{-4ex}[0pt]{ \scriptsize \shortstack [c] { $78.6$ \\(07+12)}} &\raisebox{-4ex}[0pt]{ \scriptsize \shortstack [c] { $73.5$ \\(07++12)}} & \scriptsize $<50$ &  \raisebox{-3.5ex}[0pt]{ \scriptsize CVPR17} &\raisebox{-3.5ex}[0pt]{\scriptsize  DarkNet} & \scriptsize
\textcolor{DarkGreen}{\textbf{Highlights:}} Propose a faster DarkNet19; Use a number of existing strategies to improve both speed and accuracy; Achieve high accuracy and high speed; YOLO9000 can detect over 9000 object categories in real time.
\par \textcolor{DarkRed}{\textbf{Disadvantages:}} Not good at detecting small objects. \\
\cline{2-11}
&  \raisebox{-3.5ex}[0pt]{\scriptsize SSD \cite{Liu2016SSD} }& \raisebox{-3.5ex}[0pt]{ \scriptsize $-$ } & \raisebox{-3.5ex}[0pt]{ \scriptsize VGG16} &\raisebox{-3.5ex}[0pt]{ \scriptsize  Fixed }  &  \raisebox{-6.5ex}[0pt]{ \scriptsize \shortstack [c] { $76.8$\\(07+12)\\$81.5$\\(07+12+CO)}} &\raisebox{-6.5ex}[0pt]{ \scriptsize \shortstack [c] { $74.9$ \\(07++12)\\$80.0$\\(07++12+CO)}}& \raisebox{-3.5ex}[0pt]{ \scriptsize $<60$}& \raisebox{-3.5ex}[0pt]{ \scriptsize ECCV16}  &\raisebox{-3.5ex}[0pt]{\scriptsize  \shortstack [c] {Caffe \\ Python}} & \scriptsize
 \textcolor{DarkGreen}{\textbf{Highlights:}} First accurate and efficient unified detector;
 Effectively combine ideas from RPN and YOLO to perform detection at multi-scale CONV layers; Faster and significantly more accurate than YOLO; Can run at 59 FPS;
\par \textcolor{DarkRed}{\textbf{Disadvantages:}} Not good at detecting small objects. \\
 \Xhline{1.5pt}
  \multicolumn{9}{c}{$\quad$}\\
\end{tabular}
}
\par
\raggedright \small{\emph{Abbreviations in this table: Region Proposal (RP), Selective Search (SS), Region Proposal Network (RPN), RCNN$\ominus$R represents ``RCNN minus R'' and used a trivial RP method. Training data: ``07''$\leftarrow$VOC2007 trainval; ``07T''$\leftarrow$VOC2007 trainval and test; ``12''$\leftarrow$VOC2012 trainval; ``CO''$\leftarrow$COCO trainval. The ``Speed'' column roughly estimates the detection speed with a single Nvidia Titan X GPU.}}
\end{minipage}
\end{sideways}
\end{table*}
\section{Acknowledgments}
The authors would like to thank the pioneering researchers in
generic object detection and other related fields. The authors would also like to express their sincere appreciation to Professor Ji\v{r}\'{\i} Matas, the associate editor
and the anonymous reviewers for their comments and suggestions. This work has been supported by the Center for Machine Vision and Signal Analysis at the University of Oulu (Finland) and the National Natural Science Foundation of China under Grant 61872379.

\bibliographystyle{spbasic}
\footnotesize
\bibliography{lilibib}

\begin{thebibliography}{332}
\providecommand{\natexlab}[1]{#1}
\providecommand{\url}[1]{{#1}}
\providecommand{\urlprefix}{URL }
\expandafter\ifx\csname urlstyle\endcsname\relax
  \providecommand{\doi}[1]{DOI~\discretionary{}{}{}#1}\else
  \providecommand{\doi}{DOI~\discretionary{}{}{}\begingroup
  \urlstyle{rm}\Url}\fi
\providecommand{\eprint}[2][]{\url{#2}}

\bibitem[{Agrawal et~al.(2014)Agrawal, Girshick, and Malik}]{Agrawal2014}
Agrawal P., Girshick R., Malik J. (2014) Analyzing the performance of
  multilayer neural networks for object recognition. In: ECCV, pp. 329--344

\bibitem[{Alexe et~al.(2010)Alexe, Deselaers, and Ferrari}]{Alexe2010Object}
Alexe B., Deselaers T., Ferrari V. (2010) What is an object? In: CVPR, pp.
  73--80

\bibitem[{Alexe et~al.(2012)Alexe, Deselaers, and Ferrari}]{Alexe2012}
Alexe B., Deselaers T., Ferrari V. (2012) Measuring the objectness of image
  windows. IEEE TPAMI 34(11):2189--2202

\bibitem[{Alvarez and Salzmann(2016)}]{Alvarez2016Learning}
Alvarez J., Salzmann M. (2016) Learning the number of neurons in deep networks.
  In: NIPS, pp. 2270--2278

\bibitem[{Andreopoulos and Tsotsos(2013)}]{Andreopoulos13}
Andreopoulos A., Tsotsos J. (2013) 50 years of object recognition: Directions
  forward. Computer Vision and Image Understanding 117(8):827--891

\bibitem[{Arbel{\'a}ez et~al.(2012)Arbel{\'a}ez, Hariharan, Gu, Gupta, Bourdev,
  and Malik}]{Arbelaez2012Semantic}
Arbel{\'a}ez P., Hariharan B., Gu C., Gupta S., Bourdev L., Malik J. (2012)
  Semantic segmentation using regions and parts. In: CVPR, pp. 3378--3385

\bibitem[{Arbel{\'a}ez et~al.(2014)Arbel{\'a}ez, Pont-Tuset, Barron, Marques,
  and Malik}]{Arbelaez2014}
Arbel{\'a}ez P., Pont-Tuset J., Barron J., Marques F., Malik J. (2014)
  Multiscale combinatorial grouping. In: CVPR, pp. 328--335

\bibitem[{Azizpour et~al.(2016)Azizpour, Razavian, Sullivan, Maki, and
  Carlsson}]{Azizpour2016}
Azizpour H., Razavian A., Sullivan J., Maki A., Carlsson S. (2016) Factors of
  transferability for a generic convnet representation. IEEE TPAMI
  38(9):1790--1802

\bibitem[{Bansal et~al.(2018)Bansal, Sikka, Sharma, Chellappa, and
  Divakaran}]{Bansa2018Zero}
Bansal A., Sikka K., Sharma G., Chellappa R., Divakaran A. (2018) Zero shot
  object detection. In: ECCV

\bibitem[{Bar(2004)}]{Bar2004Visual}
Bar M. (2004) Visual objects in context. Nature Reviews Neuroscience
  5(8):617--629

\bibitem[{Bell et~al.(2016)Bell, Lawrence, Bala, and Girshick}]{Bell2016ION}
Bell S., Lawrence Z., Bala K., Girshick R. (2016) {Inside Outside Net}:
  Detecting objects in context with skip pooling and recurrent neural networks.
  In: CVPR, pp. 2874--2883

\bibitem[{Belongie et~al.(2002)Belongie, Malik, and
  Puzicha}]{Belongie2002shape}
Belongie S., Malik J., Puzicha J. (2002) Shape matching and object recognition
  using shape contexts. IEEE TPAMI 24(4):509--522

\bibitem[{Bengio et~al.(2013)Bengio, Courville, and Vincent}]{Bengio13Feature}
Bengio Y., Courville A., Vincent P. (2013) Representation learning: A review
  and new perspectives. IEEE TPAMI 35(8):1798--1828

\bibitem[{Biederman(1972)}]{Biederman1972Contextual}
Biederman I. (1972) Perceiving real world scenes. IJCV 177(7):77--80

\bibitem[{Biederman(1987{\natexlab{a}})}]{Biederman1987}
Biederman I. (1987{\natexlab{a}}) Recognition by components: a theory of human
  image understanding. Psychological review 94(2):115

\bibitem[{Biederman(1987{\natexlab{b}})}]{Biederman1987Recognition}
Biederman I. (1987{\natexlab{b}}) Recognition by components: a theory of human
  image understanding. Psychological review 94(2):115

\bibitem[{Bilen and Vedaldi(2016)}]{Bilen2016Weakly}
Bilen H., Vedaldi A. (2016) Weakly supervised deep detection networks. In:
  CVPR, pp. 2846--2854

\bibitem[{Bodla et~al.(2017)Bodla, Singh, Chellappa, and Davis}]{Bodla2017Soft}
Bodla N., Singh B., Chellappa R., Davis L.~S. (2017) {SoftNMS} improving object
  detection with one line of code. In: ICCV, pp. 5562--5570

\bibitem[{Borji et~al.(2014)Borji, Cheng, Jiang, and Li}]{Borji14}
Borji A., Cheng M., Jiang H., Li J. (2014) Salient object detection: A survey.
  arXiv: 14115878v1 1:1--26

\bibitem[{Bourdev and Brandt(2005)}]{Bourdev2005Robust}
Bourdev L., Brandt J. (2005) Robust object detection via soft cascade. In:
  CVPR, vol~2, pp. 236--243

\bibitem[{Bruna and Mallat(2013)}]{Bruna13Invariant}
Bruna J., Mallat S. (2013) Invariant scattering convolution networks. IEEE
  TPAMI 35(8):1872--1886

\bibitem[{Cai et~al.(2018)Cai, Yang, Zhang, Han, and Yu}]{Cai2018Path}
Cai H., Yang J., Zhang W., Han S., Yu Y. (2018) Path level network
  transformation for efficient architecture search

\bibitem[{Cai and Vasconcelos(2018)}]{CascadeRCNN2018}
Cai Z., Vasconcelos N. (2018) Cascade {RCNN}: Delving into high quality object
  detection. In: CVPR

\bibitem[{Cai et~al.(2016)Cai, Fan, Feris, and Vasconcelos}]{MSCNN2016}
Cai Z., Fan Q., Feris R., Vasconcelos N. (2016) A unified multiscale deep
  convolutional neural network for fast object detection. In: ECCV, pp.
  354--370

\bibitem[{Carreira and Sminchisescu(2012)}]{Carreira2012}
Carreira J., Sminchisescu C. (2012) {CMPC}: Automatic object segmentation using
  constrained parametric mincuts. IEEE TPAMI 34(7):1312--1328

\bibitem[{Chatfield et~al.(2014)Chatfield, Simonyan, Vedaldi, and
  Zisserman}]{Chatfield2014}
Chatfield K., Simonyan K., Vedaldi A., Zisserman A. (2014) Return of the devil
  in the details: Delving deep into convolutional nets. In: BMVC

\bibitem[{Chavali et~al.(2016)Chavali, Agrawal, Mahendru, and
  Batra}]{Chavali2016}
Chavali N., Agrawal H., Mahendru A., Batra D. (2016) Object proposal evaluation
  protocol is gameable. In: CVPR, pp. 835--844

\bibitem[{Chellappa(2016)}]{Chellappa2016}
Chellappa R. (2016) The changing fortunes of pattern recognition and computer
  vision. Image and Vision Computing 55:3--5

\bibitem[{Chen et~al.(2017{\natexlab{a}})Chen, Choi, Yu, Han, and
  Chandraker}]{Chen2017Learning}
Chen G., Choi W., Yu X., Han T., Chandraker M. (2017{\natexlab{a}}) Learning
  efficient object detection models with knowledge distillation. In: NIPS

\bibitem[{Chen et~al.(2018{\natexlab{a}})Chen, Wang, Wang, and
  Qiao}]{Chen2018LSTD}
Chen H., Wang Y., Wang G., Qiao Y. (2018{\natexlab{a}}) {LSTD}: A low shot
  transfer detector for object detection. In: AAAI

\bibitem[{Chen et~al.(2019{\natexlab{a}})Chen, Pang, Wang, Xiong, Li, Sun,
  Feng, Liu, Shi, Ouyang et~al.}]{Chen2019Hybrid}
Chen K., Pang J., Wang J., Xiong Y., Li X., Sun S., Feng W., Liu Z., Shi J.,
  Ouyang W., et~al. (2019{\natexlab{a}}) Hybrid task cascade for instance
  segmentation. In: CVPR

\bibitem[{Chen et~al.(2015{\natexlab{a}})Chen, Papandreou, Kokkinos, Murphy,
  and Yuille}]{Chen2015Semantic}
Chen L., Papandreou G., Kokkinos I., Murphy K., Yuille A. (2015{\natexlab{a}})
  Semantic image segmentation with deep convolutional nets and fully connected
  {CRFs}. In: ICLR

\bibitem[{Chen et~al.(2018{\natexlab{b}})Chen, Papandreou, Kokkinos, Murphy,
  and Yuille}]{Chen2016deeplab}
Chen L., Papandreou G., Kokkinos I., Murphy K., Yuille A. (2018{\natexlab{b}})
  {DeepLab}: Semantic image segmentation with deep convolutional nets, atrous
  convolution, and fully connected {CRFs}. IEEE TPAMI 40(4):834--848

\bibitem[{Chen et~al.(2015{\natexlab{b}})Chen, Song, Dong, Huang, Hua, and
  Yan}]{Chen2015c}
Chen Q., Song Z., Dong J., Huang Z., Hua Y., Yan S. (2015{\natexlab{b}})
  Contextualizing object detection and classification. IEEE TPAMI 37(1):13--27

\bibitem[{Chen and Gupta(2017)}]{ChenSpatial2017}
Chen X., Gupta A. (2017) Spatial memory for context reasoning in object
  detection. In: ICCV

\bibitem[{Chen et~al.(2015{\natexlab{c}})Chen, Kundu, Zhu, Berneshawi, Ma,
  Fidler, and Urtasun}]{Chen20153D}
Chen X., Kundu K., Zhu Y., Berneshawi A.~G., Ma H., Fidler S., Urtasun R.
  (2015{\natexlab{c}}) 3d object proposals for accurate object class detection.
  In: NIPS, pp. 424--432

\bibitem[{Chen et~al.(2017{\natexlab{b}})Chen, Li, Xiao, Jin, Yan, and
  Feng}]{Chen2017Dual}
Chen Y., Li J., Xiao H., Jin X., Yan S., Feng J. (2017{\natexlab{b}}) Dual path
  networks. In: NIPS, pp. 4467--4475

\bibitem[{Chen et~al.(2019{\natexlab{b}})Chen, Rohrbach, Yan, Yan, Feng, and
  Kalantidis}]{Chen2019Graph}
Chen Y., Rohrbach M., Yan Z., Yan S., Feng J., Kalantidis Y.
  (2019{\natexlab{b}}) Graph based global reasoning networks. In: CVPR

\bibitem[{Chen et~al.(2019{\natexlab{c}})Chen, Yang, Zhang, Meng, Pan, and
  Sun}]{Chen2019DetNAS}
Chen Y., Yang T., Zhang X., Meng G., Pan C., Sun J. (2019{\natexlab{c}})
  {DetNAS}: Neural architecture search on object detection. arXiv:190310979

\bibitem[{Cheng et~al.(2018{\natexlab{a}})Cheng, Wei, Shi, Feris, Xiong, and
  Huang}]{Cheng2018Decoupled}
Cheng B., Wei Y., Shi H., Feris R., Xiong J., Huang T. (2018{\natexlab{a}})
  Decoupled classification refinement: Hard false positive suppression for
  object detection. arXiv:181004002

\bibitem[{Cheng et~al.(2018{\natexlab{b}})Cheng, Wei, Shi, Feris, Xiong, and
  Huang}]{Cheng18Revisiting}
Cheng B., Wei Y., Shi H., Feris R., Xiong J., Huang T. (2018{\natexlab{b}})
  {Revisiting RCNN:} on awakening the classification power of faster {RCNN}.
  In: ECCV

\bibitem[{Cheng et~al.(2016)Cheng, Zhou, and Han}]{RIFDCNN2016}
Cheng G., Zhou P., Han J. (2016) {RIFDCNN}: Rotation invariant and fisher
  discriminative convolutional neural networks for object detection. In: CVPR,
  pp. 2884--2893

\bibitem[{Cheng et~al.(2014)Cheng, Zhang, Lin, and Torr}]{Cheng2014bing}
Cheng M., Zhang Z., Lin W., Torr P. (2014) {BING}: Binarized normed gradients
  for objectness estimation at 300fps. In: CVPR, pp. 3286--3293

\bibitem[{Cheng et~al.(2018{\natexlab{c}})Cheng, Wang, Zhou, and
  Zhang}]{Cheng2018Model}
Cheng Y., Wang D., Zhou P., Zhang T. (2018{\natexlab{c}}) Model compression and
  acceleration for deep neural networks: The principles, progress, and
  challenges. IEEE Signal Processing Magazine 35(1):126--136

\bibitem[{Chollet(2017)}]{Chollet2017Xception}
Chollet F. (2017) Xception: Deep learning with depthwise separable
  convolutions. In: CVPR, pp. 1800--1807

\bibitem[{Cinbis et~al.(2017)Cinbis, Verbeek, and Schmid}]{Cinbis2017}
Cinbis R., Verbeek J., Schmid C. (2017) Weakly supervised object localization
  with multi-fold multiple instance learning. IEEE TPAMI 39(1):189--203

\bibitem[{Csurka et~al.(2004)Csurka, Dance, Fan, Willamowski, and
  Bray}]{Csurka2004}
Csurka G., Dance C., Fan L., Willamowski J., Bray C. (2004) Visual
  categorization with bags of keypoints. In: ECCV Workshop on statistical
  learning in computer vision

\bibitem[{Dai et~al.(2016{\natexlab{a}})Dai, He, Li, Ren, and
  Sun}]{Dai2016Instance}
Dai J., He K., Li Y., Ren S., Sun J. (2016{\natexlab{a}}) Instance sensitive
  fully convolutional networks. In: ECCV, pp. 534--549

\bibitem[{Dai et~al.(2016{\natexlab{b}})Dai, He, and Sun}]{Dai2016Aware}
Dai J., He K., Sun J. (2016{\natexlab{b}}) Instance aware semantic segmentation
  via multitask network cascades. In: CVPR, pp. 3150--3158

\bibitem[{Dai et~al.(2016{\natexlab{c}})Dai, Li, He, and Sun}]{Dai2016RFCN}
Dai J., Li Y., He K., Sun J. (2016{\natexlab{c}}) {RFCN:} object detection via
  region based fully convolutional networks. In: NIPS, pp. 379--387

\bibitem[{Dai et~al.(2017)Dai, Qi, Xiong, Li, Zhang, Hu, and
  Wei}]{Dai17Deformable}
Dai J., Qi H., Xiong Y., Li Y., Zhang G., Hu H., Wei Y. (2017) Deformable
  convolutional networks. In: ICCV

\bibitem[{Dalal and Triggs(2005)}]{Dalal2005HOG}
Dalal N., Triggs B. (2005) Histograms of oriented gradients for human
  detection. In: CVPR, vol~1, pp. 886--893

\bibitem[{Demirel et~al.(2018)Demirel, Cinbis, and
  Ikizler-Cinbis}]{Demirel2018Zero}
Demirel B., Cinbis R.~G., Ikizler-Cinbis N. (2018) Zero shot object detection
  by hybrid region embedding. In: BMVC

\bibitem[{Deng et~al.(2009)Deng, Dong, Socher, Li, Li, and Li}]{ImageNet2009}
Deng J., Dong W., Socher R., Li L., Li K., Li F. (2009) {ImageNet}: A large
  scale hierarchical image database. In: CVPR, pp. 248--255

\bibitem[{Diba et~al.(2017)Diba, Sharma, Pazandeh, Pirsiavash, and {Van
  Gool}}]{Diba2017Weakly}
Diba A., Sharma V., Pazandeh A.~M., Pirsiavash H., {Van Gool} L. (2017) Weakly
  supervised cascaded convolutional networks. In: CVPR, vol~3, p.~9

\bibitem[{Dickinson et~al.(2009)Dickinson, Leonardis, Schiele, and
  Tarr}]{Dickinson2009}
Dickinson S., Leonardis A., Schiele B., Tarr M. (2009) The Evolution of Object
  Categorization and the Challenge of Image Abstraction in \emph{Object
  Categorization: Computer and Human Vision Perspectives}. Cambridge University
  Press

\bibitem[{Ding et~al.(2018)Ding, Xue, Long, Xia, and Lu}]{Ding2018Learning}
Ding J., Xue N., Long Y., Xia G., Lu Q. (2018) Learning {RoI} transformer for
  detecting oriented objects in aerial images. In: CVPR

\bibitem[{Divvala et~al.(2009)Divvala, Hoiem, Hays, Efros, and
  Hebert}]{Divvala2009}
Divvala S., Hoiem D., Hays J., Efros A., Hebert M. (2009) An empirical study of
  context in object detection. In: CVPR, pp. 1271--1278

\bibitem[{Dollar et~al.(2012)Dollar, Wojek, Schiele, and
  Perona}]{Dollar2012Pedestrian}
Dollar P., Wojek C., Schiele B., Perona P. (2012) Pedestrian detection: An
  evaluation of the state of the art. IEEE TPAMI 34(4):743--761

\bibitem[{Donahue et~al.(2014)Donahue, Jia, Vinyals, Hoffman, Zhang, Tzeng, and
  Darrell}]{Donahue2014DeCAF}
Donahue J., Jia Y., Vinyals O., Hoffman J., Zhang N., Tzeng E., Darrell T.
  (2014) {DeCAF}: A deep convolutional activation feature for generic visual
  recognition. In: ICML, vol~32, pp. 647--655

\bibitem[{Dong et~al.(2018)Dong, Zheng, Ma, Yang, and Meng}]{Dong2018Few}
Dong X., Zheng L., Ma F., Yang Y., Meng D. (2018) Few example object detection
  with model communication. IEEE TPAMI

\bibitem[{Duan et~al.(2019)Duan, Bai, Xie, Qi, Huang, and
  Tian}]{Duan2019CenterNet}
Duan K., Bai S., Xie L., Qi H., Huang Q., Tian Q. (2019) {CenterNet}: Keypoint
  triplets for object detection. arXiv preprint arXiv:190408189

\bibitem[{Dvornik et~al.(2018)Dvornik, Mairal, and
  Schmid}]{Dvornik2018Modeling}
Dvornik N., Mairal J., Schmid C. (2018) Modeling visual context is key to
  augmenting object detection datasets. In: ECCV, pp. 364--380

\bibitem[{Dwibedi et~al.(2017)Dwibedi, Misra, and Hebert}]{Dwibedi2017Cut}
Dwibedi D., Misra I., Hebert M. (2017) Cut, paste and learn: Surprisingly easy
  synthesis for instance detection. In: ICCV, pp. 1301--1310

\bibitem[{Endres and Hoiem(2010)}]{Endres14Category}
Endres I., Hoiem D. (2010) Category independent object proposals

\bibitem[{Enzweiler and Gavrila(2009)}]{Enzweiler2009Monocular}
Enzweiler M., Gavrila D.~M. (2009) Monocular pedestrian detection: Survey and
  experiments. IEEE TPAMI 31(12):2179--2195

\bibitem[{Erhan et~al.(2014)Erhan, Szegedy, Toshev, and Anguelov}]{MultiBox1}
Erhan D., Szegedy C., Toshev A., Anguelov D. (2014) Scalable object detection
  using deep neural networks. In: CVPR, pp. 2147--2154

\bibitem[{Everingham et~al.(2010)Everingham, Gool, Williams, Winn, and
  Zisserman}]{Everingham2010}
Everingham M., Gool L.~V., Williams C., Winn J., Zisserman A. (2010) The pascal
  visual object classes (voc) challenge. IJCV 88(2):303--338

\bibitem[{Everingham et~al.(2015)Everingham, Eslami, Gool, Williams, Winn, and
  Zisserman}]{Everingham2015}
Everingham M., Eslami S., Gool L.~V., Williams C., Winn J., Zisserman A. (2015)
  The pascal visual object classes challenge: A retrospective. IJCV
  111(1):98--136

\bibitem[{Feichtenhofer et~al.(2017)Feichtenhofer, Pinz, and
  Zisserman}]{Feichtenhofer17Detect}
Feichtenhofer C., Pinz A., Zisserman A. (2017) Detect to track and track to
  detect. In: ICCV, pp. 918--927

\bibitem[{FeiFei et~al.(2006)FeiFei, Fergus, and Perona}]{Fei2006One}
FeiFei L., Fergus R., Perona P. (2006) One shot learning of object categories.
  IEEE TPAMI 28(4):594--611

\bibitem[{Felzenszwalb et~al.(2008)Felzenszwalb, McAllester, and
  Ramanan}]{Felzenszwalb08CVPR}
Felzenszwalb P., McAllester D., Ramanan D. (2008) A discriminatively trained,
  multiscale, deformable part model. In: CVPR, pp. 1--8

\bibitem[{Felzenszwalb et~al.(2010{\natexlab{a}})Felzenszwalb, Girshick, and
  McAllester}]{Felzenszwalb2010Cascade}
Felzenszwalb P., Girshick R., McAllester D. (2010{\natexlab{a}}) Cascade object
  detection with deformable part models. In: CVPR, pp. 2241--2248

\bibitem[{Felzenszwalb et~al.(2010{\natexlab{b}})Felzenszwalb, Girshick,
  McAllester, and Ramanan}]{Felzenszwalb2010b}
Felzenszwalb P., Girshick R., McAllester D., Ramanan D. (2010{\natexlab{b}})
  Object detection with discriminatively trained part based models. IEEE TPAMI
  32(9):1627--1645

\bibitem[{Finn et~al.(2017)Finn, Abbeel, and Levine}]{Finn2017Model}
Finn C., Abbeel P., Levine S. (2017) Model agnostic meta learning for fast
  adaptation of deep networks. In: ICML, pp. 1126--1135

\bibitem[{Fischler and Elschlager(1973)}]{Fischler1973}
Fischler M., Elschlager R. (1973) The representation and matching of pictorial
  structures. IEEE Transactions on computers 100(1):67--92

\bibitem[{Fu et~al.(2017)Fu, Liu, Ranga, Tyagi, and Berg}]{DSSD2016}
Fu C.-Y., Liu W., Ranga A., Tyagi A., Berg A.~C. (2017) {DSSD}: Deconvolutional
  single shot detector. In: arXiv preprint arXiv:1701.06659

\bibitem[{Galleguillos and Belongie(2010)}]{Galleguillos2010}
Galleguillos C., Belongie S. (2010) Context based object categorization: A
  critical survey. Computer Vision and Image Understanding 114:712--722

\bibitem[{Geronimo et~al.(2010)Geronimo, Lopez, Sappa, and
  Graf}]{Geronimo2010Survey}
Geronimo D., Lopez A.~M., Sappa A.~D., Graf T. (2010) Survey of pedestrian
  detection for advanced driver assistance systems. IEEE TPAMI 32(7):1239--1258

\bibitem[{Ghiasi et~al.(2019)Ghiasi, Lin, Pang, and Le}]{Ghiasi2019NASFPN}
Ghiasi G., Lin T., Pang R., Le Q. (2019) {NASFPN:} learning scalable feature
  pyramid architecture for object detection. arXiv:190407392

\bibitem[{Ghodrati et~al.(2015)Ghodrati, Diba, Pedersoli, Tuytelaars, and {Van
  Gool}}]{Deepproposal2015}
Ghodrati A., Diba A., Pedersoli M., Tuytelaars T., {Van Gool} L. (2015)
  {DeepProposal}: Hunting objects by cascading deep convolutional layers. In:
  ICCV, pp. 2578--2586

\bibitem[{Gidaris and Komodakis(2015)}]{Gidaris2015}
Gidaris S., Komodakis N. (2015) Object detection via a multiregion and semantic
  segmentation aware {CNN} model. In: ICCV, pp. 1134--1142

\bibitem[{Gidaris and Komodakis(2016)}]{Gidaris2016Attend}
Gidaris S., Komodakis N. (2016) Attend refine repeat: Active box proposal
  generation via in out localization. In: BMVC

\bibitem[{Girshick(2015)}]{Girshick2015FRCNN}
Girshick R. (2015) Fast {R-CNN}. In: ICCV, pp. 1440--1448

\bibitem[{Girshick et~al.(2014)Girshick, Donahue, Darrell, and
  Malik}]{Girshick2014RCNN}
Girshick R., Donahue J., Darrell T., Malik J. (2014) Rich feature hierarchies
  for accurate object detection and semantic segmentation. In: CVPR, pp.
  580--587

\bibitem[{Girshick et~al.(2015)Girshick, Iandola, Darrell, and
  Malik}]{Girshick2015DPMCNN}
Girshick R., Iandola F., Darrell T., Malik J. (2015) Deformable part models are
  convolutional neural networks. In: CVPR, pp. 437--446

\bibitem[{Girshick et~al.(2016)Girshick, Donahue, Darrell, and
  Malik}]{Girshick2016TPAMI}
Girshick R., Donahue J., Darrell T., Malik J. (2016) Region-based convolutional
  networks for accurate object detection and segmentation. IEEE TPAMI
  38(1):142--158

\bibitem[{Goodfellow et~al.(2015)Goodfellow, Shlens, and
  Szegedy}]{Goodfellow2015Explaining}
Goodfellow I., Shlens J., Szegedy C. (2015) Explaining and harnessing
  adversarial examples. In: ICLR

\bibitem[{Goodfellow et~al.(2016)Goodfellow, Bengio, and
  Courville}]{Goodfellow2016Deep}
Goodfellow I., Bengio Y., Courville A. (2016) Deep Learning. MIT press

\bibitem[{Grauman and Darrell(2005)}]{Grauman2005pyramid}
Grauman K., Darrell T. (2005) The pyramid match kernel: Discriminative
  classification with sets of image features. In: ICCV, vol~2, pp. 1458--1465

\bibitem[{Grauman and Leibe(2011)}]{Grauman2011Visual}
Grauman K., Leibe B. (2011) Visual object recognition. Synthesis lectures on
  artificial intelligence and machine learning 5(2):1--181

\bibitem[{Gu et~al.(2017)Gu, Wang, Kuen, Ma, Shahroudy, Shuai, Liu, Wang, Wang,
  Cai, and Chen}]{Gu2015Recent}
Gu J., Wang Z., Kuen J., Ma L., Shahroudy A., Shuai B., Liu T., Wang X., Wang
  G., Cai J., Chen T. (2017) Recent advances in convolutional neural networks.
  Pattern Recognition pp. 1--24

\bibitem[{Guillaumin et~al.(2014)Guillaumin, K{\"u}ttel, and
  Ferrari}]{Guillaumin2014ImageNet}
Guillaumin M., K{\"u}ttel D., Ferrari V. (2014) Imagenet autoannotation with
  segmentation propagation. International Journal of Computer Vision
  110(3):328--348

\bibitem[{Gupta et~al.(2016)Gupta, Vedaldi, and Zisserman}]{Gupta2016Synthetic}
Gupta A., Vedaldi A., Zisserman A. (2016) Synthetic data for text localisation
  in natural images. In: CVPR, pp. 2315--2324

\bibitem[{Hariharan and Girshick(2017)}]{Hariharan2017Low}
Hariharan B., Girshick R.~B. (2017) Low shot visual recognition by shrinking
  and hallucinating features. In: ICCV, pp. 3037--3046

\bibitem[{Hariharan et~al.(2014)Hariharan, Arbel{\'a}ez, Girshick, and
  Malik}]{Hariharan2014}
Hariharan B., Arbel{\'a}ez P., Girshick R., Malik J. (2014) Simultaneous
  detection and segmentation. In: ECCV, pp. 297--312

\bibitem[{Hariharan et~al.(2016)Hariharan, Arbel\'{a}ez, Girshick, and
  Malik}]{Hariharan2016}
Hariharan B., Arbel\'{a}ez P., Girshick R., Malik J. (2016) Object instance
  segmentation and fine grained localization using hypercolumns. IEEE TPAMI

\bibitem[{Harzallah et~al.(2009)Harzallah, Jurie, and
  Schmid}]{Harzallah2009Combining}
Harzallah H., Jurie F., Schmid C. (2009) Combining efficient object
  localization and image classification. In: ICCV, pp. 237--244

\bibitem[{He et~al.(2014)He, Zhang, Ren, and Sun}]{He2014SPP}
He K., Zhang X., Ren S., Sun J. (2014) Spatial pyramid pooling in deep
  convolutional networks for visual recognition. In: ECCV, pp. 346--361

\bibitem[{He et~al.(2015)He, Zhang, Ren, and Sun}]{He2015delving}
He K., Zhang X., Ren S., Sun J. (2015) Delving deep into rectifiers: Surpassing
  human-level performance on {ImageNet} classification. In: ICCV, pp.
  1026--1034

\bibitem[{He et~al.(2016)He, Zhang, Ren, and Sun}]{He2016ResNet}
He K., Zhang X., Ren S., Sun J. (2016) Deep residual learning for image
  recognition. In: CVPR, pp. 770--778

\bibitem[{He et~al.(2017)He, Gkioxari, Doll\'{a}r, and Girshick}]{MaskRCNN2017}
He K., Gkioxari G., Doll\'{a}r P., Girshick R. (2017) {Mask RCNN}. In: ICCV

\bibitem[{He et~al.(2018)He, Tian, Huang, Shen, Qiao, and Sun}]{He2018End}
He T., Tian Z., Huang W., Shen C., Qiao Y., Sun C. (2018) An end to end
  textspotter with explicit alignment and attention. In: CVPR, pp. 5020--5029

\bibitem[{He et~al.(2019)He, Zhu, Wang, Savvides, and Zhang}]{He2019Bounding}
He Y., Zhu C., Wang J., Savvides M., Zhang X. (2019) Bounding box regression
  with uncertainty for accurate object detection. In: CVPR

\bibitem[{Hinton and Salakhutdinov(2006)}]{Hinton2006Reducing}
Hinton G., Salakhutdinov R. (2006) Reducing the dimensionality of data with
  neural networks. science 313(5786):504--507

\bibitem[{Hinton et~al.(2015)Hinton, Vinyals, and Dean}]{Hinton2015Distilling}
Hinton G., Vinyals O., Dean J. (2015) Distilling the knowledge in a neural
  network. arXiv:150302531

\bibitem[{Hoffman et~al.(2014)Hoffman, Guadarrama, Tzeng, Hu, Donahue,
  Girshick, Darrell, and Saenko}]{Hoffman2014lsda}
Hoffman J., Guadarrama S., Tzeng E.~S., Hu R., Donahue J., Girshick R., Darrell
  T., Saenko K. (2014) {LSDA:} large scale detection through adaptation. In:
  NIPS, pp. 3536--3544

\bibitem[{Hoiem et~al.(2012)Hoiem, Chodpathumwan, and Dai}]{Hoiem2012}
Hoiem D., Chodpathumwan Y., Dai Q. (2012) Diagnosing error in object detectors.
  In: ECCV, pp. 340--353

\bibitem[{Hosang et~al.(2015)Hosang, Omran, Benenson, and
  Schiele}]{Hosang2015taking}
Hosang J., Omran M., Benenson R., Schiele B. (2015) Taking a deeper look at
  pedestrians. In: Proceedings of the IEEE Conference on Computer Vision and
  Pattern Recognition, pp. 4073--4082

\bibitem[{Hosang et~al.(2016)Hosang, Benenson, Doll¨¢r, and
  Schiele}]{Hosang2016}
Hosang J., Benenson R., Doll¨¢r P., Schiele B. (2016) What makes for effective
  detection proposals? IEEE TPAMI 38(4):814--829

\bibitem[{Hosang et~al.(2017)Hosang, Benenson, and
  Schiele}]{Hosang2017Learning}
Hosang J., Benenson R., Schiele B. (2017) Learning nonmaximum suppression. In:
  ICCV

\bibitem[{Howard et~al.(2017)Howard, Zhu, Chen, Kalenichenko, Wang, Weyand,
  Andreetto, and Adam}]{Howard2017MobileNets}
Howard A., Zhu M., Chen B., Kalenichenko D., Wang W., Weyand T., Andreetto M.,
  Adam H. (2017) Mobilenets: Efficient convolutional neural networks for mobile
  vision applications. In: CVPR

\bibitem[{Hu et~al.(2017)Hu, Lan, Jiang, Cao, and Sha}]{Hu2017FastMask}
Hu H., Lan S., Jiang Y., Cao Z., Sha F. (2017) {FastMask}: Segment multiscale
  object candidates in one shot. In: CVPR, pp. 991--999

\bibitem[{Hu et~al.(2018{\natexlab{a}})Hu, Gu, Zhang, Dai, and
  Wei}]{Hu2018Relation}
Hu H., Gu J., Zhang Z., Dai J., Wei Y. (2018{\natexlab{a}}) Relation networks
  for object detection. In: CVPR

\bibitem[{Hu et~al.(2018{\natexlab{b}})Hu, Shen, and Sun}]{Hu2018Squeeze}
Hu J., Shen L., Sun G. (2018{\natexlab{b}}) Squeeze and excitation networks.
  In: CVPR

\bibitem[{Hu and Ramanan(2017)}]{Hu2017Finding}
Hu P., Ramanan D. (2017) Finding tiny faces. In: CVPR, pp. 1522--1530

\bibitem[{Hu et~al.(2018{\natexlab{c}})Hu, Doll\'{a}r, He, Darrell, and
  Girshick}]{Ronghang2018}
Hu R., Doll\'{a}r P., He K., Darrell T., Girshick R. (2018{\natexlab{c}})
  Learning to segment every thing. In: CVPR

\bibitem[{Huang et~al.(2017{\natexlab{a}})Huang, Liu, Weinberger, and {van der
  Maaten}}]{Huang2016Densely}
Huang G., Liu Z., Weinberger K.~Q., {van der Maaten} L. (2017{\natexlab{a}})
  Densely connected convolutional networks. In: CVPR

\bibitem[{Huang et~al.(2018)Huang, Liu, {van der Maaten}, and
  Weinberger}]{CondenseNet18}
Huang G., Liu S., {van der Maaten} L., Weinberger K. (2018) {CondenseNet}: An
  efficient densenet using learned group convolutions. In: CVPR

\bibitem[{Huang et~al.(2017{\natexlab{b}})Huang, Rathod, Sun, Zhu, Korattikara,
  Fathi, Fischer, Wojna, Song, Guadarrama, and Murphy}]{Huang2016Speed}
Huang J., Rathod V., Sun C., Zhu M., Korattikara A., Fathi A., Fischer I.,
  Wojna Z., Song Y., Guadarrama S., Murphy K. (2017{\natexlab{b}})
  Speed/accuracy trade offs for modern convolutional object detectors. In: CVPR

\bibitem[{Huang et~al.(2019)Huang, Huang, Gong, Huang, and
  Wang}]{Huang2019Mask}
Huang Z., Huang L., Gong Y., Huang C., Wang X. (2019) Mask scoring rcnn. In:
  CVPR

\bibitem[{Hubara et~al.(2016)Hubara, Courbariaux, Soudry, ElYaniv, and
  Bengio}]{Hubara2016Binarized}
Hubara I., Courbariaux M., Soudry D., ElYaniv R., Bengio Y. (2016) Binarized
  neural networks. In: NIPS, pp. 4107--4115

\bibitem[{Iandola et~al.(2016)Iandola, Han, Moskewicz, Ashraf, Dally, and
  Keutzer}]{SqueezeNet2016}
Iandola F., Han S., Moskewicz M., Ashraf K., Dally W., Keutzer K. (2016)
  {SqueezeNet}: Alexnet level accuracy with 50x fewer parameters and 0.5 mb
  model size. In: arXiv preprint arXiv:1602.07360

\bibitem[{ILSVRC detection challenge results(2018)}]{ILSVRCResults}
ILSVRC detection challenge results (2018)
  \url{http://www.image-net.org/challenges/LSVRC/} {\url{}}

\bibitem[{Ioffe and Szegedy(2015)}]{Ioffe2015}
Ioffe S., Szegedy C. (2015) Batch normalization: Accelerating deep network
  training by reducing internal covariate shift. In: International Conference
  on Machine Learning, pp. 448--456

\bibitem[{Jaderberg et~al.(2015)Jaderberg, Simonyan, Zisserman
  et~al.}]{Jaderberg2015Spatial}
Jaderberg M., Simonyan K., Zisserman A., et~al. (2015) Spatial transformer
  networks. In: NIPS, pp. 2017--2025

\bibitem[{Jia et~al.(2014)Jia, Shelhamer, Donahue, Karayev, Long, Girshick,
  Guadarrama, and Darrell}]{Jia2014Caffe}
Jia Y., Shelhamer E., Donahue J., Karayev S., Long J., Girshick R., Guadarrama
  S., Darrell T. (2014) Caffe: Convolutional architecture for fast feature
  embedding. In: ACM MM, pp. 675--678

\bibitem[{Jiang et~al.(2018)Jiang, Luo, Mao, Xiao, and
  Jiang}]{Jiang2018Acquisition}
Jiang B., Luo R., Mao J., Xiao T., Jiang Y. (2018) Acquisition of localization
  confidence for accurate object detection. In: ECCV, pp. 784--799

\bibitem[{Kang et~al.(2018)Kang, Liu, Wang, Yu, Feng, and
  Darrell}]{Kang2018Few}
Kang B., Liu Z., Wang X., Yu F., Feng J., Darrell T. (2018) Few shot object
  detection via feature reweighting. arXiv preprint arXiv:181201866

\bibitem[{Kang et~al.(2016)Kang, Ouyang, Li, and Wang}]{Kang2016Object}
Kang K., Ouyang W., Li H., Wang X. (2016) Object detection from video tubelets
  with convolutional neural networks. In: CVPR, pp. 817--825

\bibitem[{Kim et~al.(2014)Kim, Sharma, and Jacobs}]{Kim2014Locally}
Kim A., Sharma A., Jacobs D. (2014) Locally scale invariant convolutional
  neural networks. In: NIPS

\bibitem[{Kim et~al.(2016)Kim, Hong, Roh, Cheon, and Park}]{PVANET2016}
Kim K., Hong S., Roh B., Cheon Y., Park M. (2016) {PVANet}: Deep but
  lightweight neural networks for real time object detection. In: NIPSW

\bibitem[{Kim et~al.(2018)Kim, Kang, and Kim}]{Kim2018San}
Kim Y., Kang B.-N., Kim D. (2018) {SAN:} learning relationship between
  convolutional features for multiscale object detection. In: ECCV, pp.
  316--331

\bibitem[{Kirillov et~al.(2018)Kirillov, He, Girshick, Rother, and
  Doll{\'a}r}]{Kirillov2018Panoptic}
Kirillov A., He K., Girshick R., Rother C., Doll{\'a}r P. (2018) Panoptic
  segmentation. arXiv:180100868

\bibitem[{Kong et~al.(2016)Kong, Yao, Chen, and Sun}]{HyperNet2016}
Kong T., Yao A., Chen Y., Sun F. (2016) {HyperNet}: towards accurate region
  proposal generation and joint object detection. In: CVPR, pp. 845--853

\bibitem[{Kong et~al.(2017)Kong, Sun, Yao, Liu, Lu, and Chen}]{Kong2017ron}
Kong T., Sun F., Yao A., Liu H., Lu M., Chen Y. (2017) {RON}: Reverse
  connection with objectness prior networks for object detection. In: CVPR

\bibitem[{Kong et~al.(2018)Kong, Sun, Tan, Liu, and Huang}]{Kong2018Deep}
Kong T., Sun F., Tan C., Liu H., Huang W. (2018) Deep feature pyramid
  reconfiguration for object detection. In: ECCV, pp. 169--185

\bibitem[{Kr\"{a}henb\"{u}hl1 and Koltun(2014)}]{Philipp14Geodesic}
Kr\"{a}henb\"{u}hl1 P., Koltun V. (2014) Geodesic object proposals. In: ECCV

\bibitem[{Krasin et~al.(2017)Krasin, Duerig, Alldrin, Ferrari, {AbuElHaija},
  Kuznetsova, Rom, Uijlings, Popov, Kamali, Malloci, {PontTuset}, Veit,
  Belongie, Gomes, Gupta, Sun, Chechik, Cai, Feng, Narayanan, and
  Murphy}]{OpenImages2017}
Krasin I., Duerig T., Alldrin N., Ferrari V., {AbuElHaija} S., Kuznetsova A.,
  Rom H., Uijlings J., Popov S., Kamali S., Malloci M., {PontTuset} J., Veit
  A., Belongie S., Gomes V., Gupta A., Sun C., Chechik G., Cai D., Feng Z.,
  Narayanan D., Murphy K. (2017) {OpenImages}: A public dataset for large scale
  multilabel and multiclass image classification. Dataset available from
  https://storagegoogleapiscom/openimages/web/indexhtml

\bibitem[{Krizhevsky et~al.(2012{\natexlab{a}})Krizhevsky, Sutskever, and
  Hinton}]{Krizhevsky2012}
Krizhevsky A., Sutskever I., Hinton G. (2012{\natexlab{a}}) {ImageNet}
  classification with deep convolutional neural networks. In: NIPS, pp.
  1097--1105

\bibitem[{Krizhevsky et~al.(2012{\natexlab{b}})Krizhevsky, Sutskever, and
  Hinton}]{AlexNet2012}
Krizhevsky A., Sutskever I., Hinton G. (2012{\natexlab{b}}) {ImageNet}
  classification with deep convolutional neural networks. In: NIPS, pp.
  1097--1105

\bibitem[{Kuo et~al.(2015)Kuo, Hariharan, and Malik}]{DeepBox2015}
Kuo W., Hariharan B., Malik J. (2015) {DeepBox}: Learning objectness with
  convolutional networks. In: ICCV, pp. 2479--2487

\bibitem[{Kuznetsova et~al.(2018)Kuznetsova, Rom, Alldrin, Uijlings, Krasin,
  PontTuset, Kamali, Popov, Malloci, Duerig et~al.}]{Kuznetsova2018Open}
Kuznetsova A., Rom H., Alldrin N., Uijlings J., Krasin I., PontTuset J., Kamali
  S., Popov S., Malloci M., Duerig T., et~al. (2018) The open images dataset
  v4: Unified image classification, object detection, and visual relationship
  detection at scale. arXiv preprint arXiv:181100982

\bibitem[{Lake et~al.(2015)Lake, Salakhutdinov, and Tenenbaum}]{Lake2015Human}
Lake B., Salakhutdinov R., Tenenbaum J. (2015) Human level concept learning
  through probabilistic program induction. Science 350(6266):1332--1338

\bibitem[{Lampert et~al.(2008)Lampert, Blaschko, and
  Hofmann}]{lampert2008beyond}
Lampert C.~H., Blaschko M.~B., Hofmann T. (2008) Beyond sliding windows: Object
  localization by efficient subwindow search. In: CVPR, pp. 1--8

\bibitem[{Law and Deng(2018)}]{Law2018CornerNet}
Law H., Deng J. (2018) {CornerNet}: Detecting objects as paired keypoints. In:
  ECCV

\bibitem[{Lazebnik et~al.(2006)Lazebnik, Schmid, and Ponce}]{Lazebnik2006SPM}
Lazebnik S., Schmid C., Ponce J. (2006) Beyond bags of features: Spatial
  pyramid matching for recognizing natural scene categories. In: CVPR, vol~2,
  pp. 2169--2178

\bibitem[{LeCun et~al.(1998)LeCun, Bottou, Bengio, and
  Haffner}]{LeCun1998Gradient}
LeCun Y., Bottou L., Bengio Y., Haffner P. (1998) Gradient based learning
  applied to document recognition. Proceedings of the IEEE 86(11):2278--2324

\bibitem[{LeCun et~al.(2015)LeCun, Bengio, and Hinton}]{LeCun15}
LeCun Y., Bengio Y., Hinton G. (2015) Deep learning. Nature 521:436--444

\bibitem[{Lee et~al.(2015)Lee, Xie, Gallagher, Zhang, and Tu}]{Lee2015Deeply}
Lee C., Xie S., Gallagher P., Zhang Z., Tu Z. (2015) Deeply supervised nets.
  In: Artificial Intelligence and Statistics, pp. 562--570

\bibitem[{Lenc and Vedaldi(2015)}]{Lenc2015}
Lenc K., Vedaldi A. (2015) {R-CNN} minus {R}. In: BMVC15

\bibitem[{Lenc and Vedaldi(2018)}]{Lenc2018Understanding}
Lenc K., Vedaldi A. (2018) Understanding image representations by measuring
  their equivariance and equivalence. IJCV

\bibitem[{Li et~al.(2019{\natexlab{a}})Li, Liu, and Wang}]{Li2019Gradient}
Li B., Liu Y., Wang X. (2019{\natexlab{a}}) Gradient harmonized single stage
  detector. In: AAAI

\bibitem[{Li et~al.(2015{\natexlab{a}})Li, Lin, Shen, Brandt, and
  Hua}]{Li2015CasecadeCNN}
Li H., Lin Z., Shen X., Brandt J., Hua G. (2015{\natexlab{a}}) A convolutional
  neural network cascade for face detection. In: CVPR, pp. 5325--5334

\bibitem[{Li et~al.(2017{\natexlab{a}})Li, Kadav, Durdanovic, Samet, and
  Graf}]{Li2017Pruning}
Li H., Kadav A., Durdanovic I., Samet H., Graf H.~P. (2017{\natexlab{a}})
  Pruning filters for efficient convnets. In: ICLR

\bibitem[{Li et~al.(2018{\natexlab{a}})Li, Liu, Ouyang, and
  XiaogangWang}]{Hongyang2018Zoom}
Li H., Liu Y., Ouyang W., XiaogangWang (2018{\natexlab{a}}) Zoom out and in
  network with map attention decision for region proposal and object detection.
  IJCV

\bibitem[{Li et~al.(2017{\natexlab{b}})Li, Wei, Liang, Dong, Xu, Feng, and
  Yan}]{Li2017Attentive}
Li J., Wei Y., Liang X., Dong J., Xu T., Feng J., Yan S. (2017{\natexlab{b}})
  Attentive contexts for object detection. IEEE Transactions on Multimedia
  19(5):944--954

\bibitem[{Li et~al.(2017{\natexlab{c}})Li, Jin, and Yan}]{Li2017Mimicking}
Li Q., Jin S., Yan J. (2017{\natexlab{c}}) Mimicking very efficient network for
  object detection. In: CVPR, pp. 7341--7349

\bibitem[{Li and Zhang(2004)}]{Li2004Floatboost}
Li S.~Z., Zhang Z. (2004) Floatboost learning and statistical face detection.
  IEEE TPAMI 26(9):1112--1123

\bibitem[{Li et~al.(2015{\natexlab{b}})Li, Wang, Tian, and
  Ding}]{Li2015Feature}
Li Y., Wang S., Tian Q., Ding X. (2015{\natexlab{b}}) Feature representation
  for statistical learning based object detection: A review. Pattern
  Recognition 48(11):3542--3559

\bibitem[{Li et~al.(2017{\natexlab{d}})Li, Ouyang, Zhou, Wang, and
  Wang}]{Li2017Scene}
Li Y., Ouyang W., Zhou B., Wang K., Wang X. (2017{\natexlab{d}}) Scene graph
  generation from objects, phrases and region captions. In: ICCV, pp.
  1261--1270

\bibitem[{Li et~al.(2017{\natexlab{e}})Li, Qi, Dai, Ji, and Wei}]{Li2017Fully}
Li Y., Qi H., Dai J., Ji X., Wei Y. (2017{\natexlab{e}}) Fully convolutional
  instance aware semantic segmentation. In: CVPR, pp. 4438--4446

\bibitem[{Li et~al.(2019{\natexlab{b}})Li, Chen, Wang, and Zhang}]{Li2019Scale}
Li Y., Chen Y., Wang N., Zhang Z. (2019{\natexlab{b}}) Scale aware trident
  networks for object detection. arXiv preprint arXiv:190101892

\bibitem[{Li et~al.(2018{\natexlab{b}})Li, Peng, Yu, Zhang, Deng, and
  Sun}]{Li2018DetNet}
Li Z., Peng C., Yu G., Zhang X., Deng Y., Sun J. (2018{\natexlab{b}}) {DetNet}:
  A backbone network for object detection. In: ECCV

\bibitem[{Li et~al.(2018{\natexlab{c}})Li, Peng, Yu, Zhang, Deng, and
  Sun}]{Li2018Light}
Li Z., Peng C., Yu G., Zhang X., Deng Y., Sun J. (2018{\natexlab{c}}) Light
  head {RCNN}: In defense of two stage object detector. In: CVPR

\bibitem[{Lin et~al.(2014)Lin, Maire, Belongie, Hays, Perona, Ramanan,
  Doll{\'a}r, and Zitnick}]{Lin2014}
Lin T., Maire M., Belongie S., Hays J., Perona P., Ramanan D., Doll{\'a}r P.,
  Zitnick L. (2014) Microsoft {COCO}: Common objects in context. In: ECCV, pp.
  740--755

\bibitem[{Lin et~al.(2017{\natexlab{a}})Lin, Doll{\'a}r, Girshick, He,
  Hariharan, and Belongie}]{FPN2016}
Lin T., Doll{\'a}r P., Girshick R., He K., Hariharan B., Belongie S.
  (2017{\natexlab{a}}) Feature pyramid networks for object detection. In: CVPR

\bibitem[{Lin et~al.(2017{\natexlab{b}})Lin, Goyal, Girshick, He, and
  Doll\'{a}r}]{LinICCV2017}
Lin T., Goyal P., Girshick R., He K., Doll\'{a}r P. (2017{\natexlab{b}}) Focal
  loss for dense object detection. In: ICCV

\bibitem[{Lin et~al.(2017{\natexlab{c}})Lin, Zhao, and Pan}]{Lin2017Towards}
Lin X., Zhao C., Pan W. (2017{\natexlab{c}}) Towards accurate binary
  convolutional neural network. In: NIPS, pp. 344--352

\bibitem[{Litjens et~al.(2017)Litjens, Kooi, Bejnordi, Setio, Ciompi,
  Ghafoorian, J.~{van der Laak}, and S\'{a}nchez}]{Litjens2017}
Litjens G., Kooi T., Bejnordi B., Setio A., Ciompi F., Ghafoorian M., J.~{van
  der Laak} B.~v., S\'{a}nchez C. (2017) A survey on deep learning in medical
  image analysis. Medical Image Analysis 42:60--88

\bibitem[{Liu et~al.(2018{\natexlab{a}})Liu, Zoph, Neumann, Shlens, Hua, Li,
  FeiFei, Yuille, Huang, and Murphy}]{Liu2018Progressive}
Liu C., Zoph B., Neumann M., Shlens J., Hua W., Li L., FeiFei L., Yuille A.,
  Huang J., Murphy K. (2018{\natexlab{a}}) Progressive neural architecture
  search. In: ECCV, pp. 19--34

\bibitem[{Liu et~al.(2017)Liu, Fieguth, Guo, Wang, and
  Pietik{\"a}inen}]{Liu2017Local}
Liu L., Fieguth P., Guo Y., Wang X., Pietik{\"a}inen M. (2017) Local binary
  features for texture classification: Taxonomy and experimental study. Pattern
  Recognition 62:135--160

\bibitem[{Liu et~al.(2018{\natexlab{b}})Liu, Huang, and
  Wang}]{Liu2017Receptive}
Liu S., Huang D., Wang Y. (2018{\natexlab{b}}) Receptive field block net for
  accurate and fast object detection. In: ECCV

\bibitem[{Liu et~al.(2018{\natexlab{c}})Liu, Qi, Qin, Shi, and
  Jia}]{Liu2018Path}
Liu S., Qi L., Qin H., Shi J., Jia J. (2018{\natexlab{c}}) Path aggregation
  network for instance segmentation. In: CVPR, pp. 8759--8768

\bibitem[{Liu et~al.(2016)Liu, Anguelov, Erhan, Szegedy, Reed, Fu, and
  Berg}]{Liu2016SSD}
Liu W., Anguelov D., Erhan D., Szegedy C., Reed S., Fu C., Berg A. (2016)
  {SSD:} single shot multibox detector. In: ECCV, pp. 21--37

\bibitem[{Liu et~al.(2018{\natexlab{d}})Liu, Wang, Shan, and
  Chen}]{Liu2018Structure}
Liu Y., Wang R., Shan S., Chen X. (2018{\natexlab{d}}) {Structure Inference
  Net}: Object detection using scene level context and instance level
  relationships. In: CVPR, pp. 6985--6994

\bibitem[{Long et~al.(2015)Long, Shelhamer, and Darrell}]{FCNCVPR2015}
Long J., Shelhamer E., Darrell T. (2015) Fully convolutional networks for
  semantic segmentation. In: CVPR, pp. 3431--3440

\bibitem[{Lowe(1999)}]{Lowe1999Object}
Lowe D. (1999) Object recognition from local scale invariant features. In:
  ICCV, vol~2, pp. 1150--1157

\bibitem[{Lowe(2004)}]{Lowe2004}
Lowe D. (2004) Distinctive image features from scale-invariant keypoints. IJCV
  60(2):91--110

\bibitem[{Loy et~al.(2019)Loy, Lin, Ouyang, Xiong, Yang, Huang, Zhou, Xia, Li,
  Luo et~al.}]{Loy2019Wider}
Loy C., Lin D., Ouyang W., Xiong Y., Yang S., Huang Q., Zhou D., Xia W., Li Q.,
  Luo P., et~al. (2019) {WIDER} face and pedestrian challenge 2018: Methods and
  results. arXiv:190206854

\bibitem[{Lu et~al.(2016)Lu, Javidi, and Lazebnik}]{Lu2016Adaptive}
Lu Y., Javidi T., Lazebnik S. (2016) Adaptive object detection using adjacency
  and zoom prediction. In: CVPR, pp. 2351--2359

\bibitem[{Luo et~al.(2018)Luo, Wang, Shao, and Peng}]{Luo2018Towards}
Luo P., Wang X., Shao W., Peng Z. (2018) Towards understanding regularization
  in batch normalization. In: ICLR

\bibitem[{Luo et~al.(2019)Luo, Ren, Peng, Zhang, and Li}]{Luo2019Switchable}
Luo P., Ren J., Peng Z., Zhang R., Li J. (2019) Switchable normalization for
  learning to normalize deep representation. IEEE TPAMI

\bibitem[{Ma et~al.(2018)Ma, Shao, Ye, Wang, Wang, Zheng, and
  Xue}]{Ma2018Arbitrary}
Ma J., Shao W., Ye H., Wang L., Wang H., Zheng Y., Xue X. (2018) Arbitrary
  oriented scene text detection via rotation proposals. IEEE TMM
  20(11):3111--3122

\bibitem[{Malisiewicz and Efros(2009)}]{Malisiewicz09Beyond}
Malisiewicz T., Efros A. (2009) Beyond categories: The visual memex model for
  reasoning about object relationships. In: NIPS

\bibitem[{Manen et~al.(2013)Manen, Guillaumin, and {Van Gool}}]{Manen2013Prime}
Manen S., Guillaumin M., {Van Gool} L. (2013) Prime object proposals with
  randomized prim's algorithm. In: CVPR, pp. 2536--2543

\bibitem[{Mikolajczyk and Schmid(2005)}]{Mikolajczyk2005}
Mikolajczyk K., Schmid C. (2005) A performance evaluation of local descriptors.
  IEEE TPAMI 27(10):1615--1630

\bibitem[{Mordan et~al.(2018)Mordan, Thome, Henaff, and Cord}]{Mordan2018End}
Mordan T., Thome N., Henaff G., Cord M. (2018) End to end learning of latent
  deformable part based representations for object detection. IJCV pp. 1--21

\bibitem[{MS COCO detection leaderboard(2018)}]{COCOResults}
MS COCO detection leaderboard (2018) \url{http://cocodataset.org/\#}
  \url{detection-leaderboard}

\bibitem[{Mundy(2006)}]{Mundy2006Object}
Mundy J. (2006) Object recognition in the geometric era: A retrospective. in
  book {Toward Category Level Object Recognition} edited by J Ponce, M Hebert,
  C Schmid and A Zisserman pp. 3--28

\bibitem[{Murase and Nayar(1995{\natexlab{a}})}]{Murase1995}
Murase H., Nayar S. (1995{\natexlab{a}}) Visual learning and recognition of
  {3D} objects from appearance. IJCV 14(1):5--24

\bibitem[{Murase and Nayar(1995{\natexlab{b}})}]{Murase1995Visual}
Murase H., Nayar S. (1995{\natexlab{b}}) Visual learning and recognition of 3d
  objects from appearance. IJCV 14(1):5--24

\bibitem[{Murphy et~al.(2003)Murphy, Torralba, and Freeman}]{Murphy03Using}
Murphy K., Torralba A., Freeman W. (2003) Using the forest to see the trees: a
  graphical model relating features, objects and scenes. In: NIPS

\bibitem[{Newell et~al.(2016)Newell, Yang, and Deng}]{Newell2016Stacked}
Newell A., Yang K., Deng J. (2016) Stacked hourglass networks for human pose
  estimation. In: ECCV, pp. 483--499

\bibitem[{Newell et~al.(2017)Newell, Huang, and Deng}]{Newell2017Associative}
Newell A., Huang Z., Deng J. (2017) Associative embedding: end to end learning
  for joint detection and grouping. In: NIPS, pp. 2277--2287

\bibitem[{Ojala et~al.(2002)Ojala, Pietik\"{a}inen, and
  Maenp\"{a}\"{a}}]{Ojala02}
Ojala T., Pietik\"{a}inen M., Maenp\"{a}\"{a} T. (2002) Multiresolution
  gray-scale and rotation invariant texture classification with local binary
  patterns. IEEE TPAMI 24(7):971--987

\bibitem[{Oliva and Torralba(2007)}]{Oliva2007Role}
Oliva A., Torralba A. (2007) The role of context in object recognition. Trends
  in cognitive sciences 11(12):520--527

\bibitem[{Opelt et~al.(2006)Opelt, Pinz, Fussenegger, and
  Auer}]{Opelt2006generic}
Opelt A., Pinz A., Fussenegger M., Auer P. (2006) Generic object recognition
  with boosting. IEEE TPAMI 28(3):416--431

\bibitem[{Oquab et~al.(2014)Oquab, Bottou, Laptev, and
  Sivic}]{Oquab2014Learning}
Oquab M., Bottou L., Laptev I., Sivic J. (2014) Learning and transferring
  midlevel image representations using convolutional neural networks. In: CVPR,
  pp. 1717--1724

\bibitem[{Oquab et~al.(2015)Oquab, Bottou, Laptev, and Sivic}]{Oquab2015object}
Oquab M., Bottou L., Laptev I., Sivic J. (2015) Is object localization for
  free? weakly supervised learning with convolutional neural networks. In:
  CVPR, pp. 685--694

\bibitem[{Osuna et~al.(1997)Osuna, Freund, and Girosit}]{Osuna1997Train}
Osuna E., Freund R., Girosit F. (1997) Training support vector machines: an
  application to face detection. In: CVPR, pp. 130--136

\bibitem[{Ouyang and Wang(2013)}]{Ouyang2013Joint}
Ouyang W., Wang X. (2013) Joint deep learning for pedestrian detection. In:
  ICCV, pp. 2056--2063

\bibitem[{Ouyang et~al.(2015)Ouyang, Wang, Zeng, Qiu, Luo, Tian, Li, Yang,
  Wang, Loy et~al.}]{Ouyang2015deepid}
Ouyang W., Wang X., Zeng X., Qiu S., Luo P., Tian Y., Li H., Yang S., Wang Z.,
  Loy C.-C., et~al. (2015) {DeepIDNet}: Deformable deep convolutional neural
  networks for object detection. In: CVPR, pp. 2403--2412

\bibitem[{Ouyang et~al.(2016)Ouyang, Wang, Zhang, and Yang}]{Ouyang2016Factors}
Ouyang W., Wang X., Zhang C., Yang X. (2016) Factors in finetuning deep model
  for object detection with long tail distribution. In: CVPR, pp. 864--873

\bibitem[{Ouyang et~al.(2017{\natexlab{a}})Ouyang, Wang, Zhu, and
  Wang}]{Ouyang2017Chained}
Ouyang W., Wang K., Zhu X., Wang X. (2017{\natexlab{a}}) Chained cascade
  network for object detection. ICCV

\bibitem[{Ouyang et~al.(2017{\natexlab{b}})Ouyang, Zeng, Wang, Qiu, Luo, Tian,
  Li, Yang, Wang, Li, Wang, Yan, Loy, and Tang}]{Ouyang2016}
Ouyang W., Zeng X., Wang X., Qiu S., Luo P., Tian Y., Li H., Yang S., Wang Z.,
  Li H., Wang K., Yan J., Loy C.~C., Tang X. (2017{\natexlab{b}}) {DeepIDNet}:
  Object detection with deformable part based convolutional neural networks.
  IEEE TPAMI 39(7):1320--1334

\bibitem[{Parikh et~al.(2012)Parikh, Zitnick, and Chen}]{Parikh2012}
Parikh D., Zitnick C., Chen T. (2012) Exploring tiny images: The roles of
  appearance and contextual information for machine and human object
  recognition. IEEE TPAMI 34(10):1978--1991

\bibitem[{PASCAL VOC detection leaderboard(2018)}]{VOCResults}
PASCAL VOC detection leaderboard (2018)
  \url{http://host.robots.ox.ac.uk:8080/leaderboard/}
  {\url{main_bootstrap.php}}

\bibitem[{Peng et~al.(2018)Peng, Xiao, Li, Jiang, Zhang, Jia, Yu, and
  Sun}]{Peng2018MegDet}
Peng C., Xiao T., Li Z., Jiang Y., Zhang X., Jia K., Yu G., Sun J. (2018)
  {MegDet}: A large minibatch object detector. In: CVPR

\bibitem[{Peng et~al.(2015)Peng, Sun, Ali, and Saenko}]{Peng2015Learning}
Peng X., Sun B., Ali K., Saenko K. (2015) Learning deep object detectors from
  3d models. In: ICCV, pp. 1278--1286

\bibitem[{Pepik et~al.(2015)Pepik, Benenson, Ritschel, and
  Schiele}]{Pepik2015GCPR}
Pepik B., Benenson R., Ritschel T., Schiele B. (2015) What is holding back
  convnets for detection? In: German Conference on Pattern Recognition, pp.
  517--528

\bibitem[{Perronnin et~al.(2010)Perronnin, S{\'a}nchez, and
  Mensink}]{Perronnin2010}
Perronnin F., S{\'a}nchez J., Mensink T. (2010) Improving the fisher kernel for
  large scale image classification. In: ECCV, pp. 143--156

\bibitem[{Pinheiro et~al.(2015)Pinheiro, Collobert, and Dollar}]{DeepMask2015}
Pinheiro P., Collobert R., Dollar P. (2015) Learning to segment object
  candidates. In: NIPS, pp. 1990--1998

\bibitem[{Pinheiro et~al.(2016)Pinheiro, Lin, Collobert, and
  Doll{\'a}r}]{Pinheiro2016}
Pinheiro P., Lin T., Collobert R., Doll{\'a}r P. (2016) Learning to refine
  object segments. In: ECCV, pp. 75--91

\bibitem[{Ponce et~al.(2007)Ponce, Hebert, Schmid, and
  Zisserman}]{Ponce2007Toward}
Ponce J., Hebert M., Schmid C., Zisserman A. (2007) Toward Category Level
  Object Recognition. Springer

\bibitem[{Pouyanfar et~al.(2018)Pouyanfar, Sadiq, Yan, Tian, Tao, Reyes, Shyu,
  Chen, and Iyengar}]{Pouyanfar2018Survey}
Pouyanfar S., Sadiq S., Yan Y., Tian H., Tao Y., Reyes M.~P., Shyu M., Chen S.,
  Iyengar S. (2018) A survey on deep learning: Algorithms, techniques, and
  applications. ACM Computing Surveys 51(5):92:1--92:36

\bibitem[{Qi et~al.(2017)Qi, Su, Mo, and Guibas}]{Qi2017PointNet}
Qi C.~R., Su H., Mo K., Guibas L.~J. (2017) {PointNet}: Deep learning on point
  sets for {3D} classification and segmentation. In: CVPR, pp. 652--660

\bibitem[{Qi et~al.(2018)Qi, Liu, Wu, Su, and Guibas}]{Qi2018Frustum}
Qi C.~R., Liu W., Wu C., Su H., Guibas L.~J. (2018) Frustum pointnets for {3D}
  object detection from {RGBD} data. In: CVPR, pp. 918--927

\bibitem[{Quanming et~al.(2018)Quanming, Mengshuo, Hugo, Isabelle, Yiqi,
  Yufeng, Weiwei, Qiang, and Yang}]{Quanming2018Taking}
Quanming Y., Mengshuo W., Hugo J.~E., Isabelle G., Yiqi H., Yufeng L., Weiwei
  T., Qiang Y., Yang Y. (2018) Taking human out of learning applications: A
  survey on automated machine learning. arXiv:181013306

\bibitem[{Rabinovich et~al.(2007)Rabinovich, Vedaldi, Galleguillos, Wiewiora,
  and Belongie}]{Rabinovich2007Objects}
Rabinovich A., Vedaldi A., Galleguillos C., Wiewiora E., Belongie S. (2007)
  Objects in context. In: ICCV

\bibitem[{Rahman et~al.(2018{\natexlab{a}})Rahman, Khan, and
  Barnes}]{Rahman2018Polarity}
Rahman S., Khan S., Barnes N. (2018{\natexlab{a}}) Polarity loss for zero shot
  object detection. arXiv preprint arXiv:181108982

\bibitem[{Rahman et~al.(2018{\natexlab{b}})Rahman, Khan, and
  Porikli}]{Rahman2018Zero}
Rahman S., Khan S., Porikli F. (2018{\natexlab{b}}) Zero shot object detection:
  Learning to simultaneously recognize and localize novel concepts. In: ACCV

\bibitem[{Razavian et~al.(2014)Razavian, Azizpour, Sullivan, and
  Carlsson}]{Razavian2014}
Razavian R., Azizpour H., Sullivan J., Carlsson S. (2014) {CNN} features off
  the shelf: an astounding baseline for recognition. In: CVPR Workshops, pp.
  806--813

\bibitem[{Rebuffi et~al.(2017)Rebuffi, Bilen, and
  Vedaldi}]{Rebuffi2017Learning}
Rebuffi S., Bilen H., Vedaldi A. (2017) Learning multiple visual domains with
  residual adapters. In: Advances in Neural Information Processing Systems, pp.
  506--516

\bibitem[{Rebuffi et~al.(2018)Rebuffi, Bilen, and
  Vedaldi}]{Rebuffi2018Efficient}
Rebuffi S., Bilen H., Vedaldi A. (2018) Efficient parametrization of
  multidomain deep neural networks. In: CVPR, pp. 8119--8127

\bibitem[{Redmon and Farhadi(2017)}]{YOLO9000}
Redmon J., Farhadi A. (2017) {YOLO9000}: Better, faster, stronger. In: CVPR

\bibitem[{Redmon et~al.(2016)Redmon, Divvala, Girshick, and Farhadi}]{YoLo2016}
Redmon J., Divvala S., Girshick R., Farhadi A. (2016) You only look once:
  Unified, real time object detection. In: CVPR, pp. 779--788

\bibitem[{Ren et~al.(2018)Ren, Triantafillou, Ravi, Snell, Swersky, Tenenbaum,
  Larochelle, and Zemel}]{Ren2018Meta}
Ren M., Triantafillou E., Ravi S., Snell J., Swersky K., Tenenbaum J.~B.,
  Larochelle H., Zemel R.~S. (2018) Meta learning for semisupervised few shot
  classification. In: ICLR

\bibitem[{Ren et~al.(2015)Ren, He, Girshick, and Sun}]{Ren2015NIPS}
Ren S., He K., Girshick R., Sun J. (2015) Faster {R-CNN}: Towards real time
  object detection with region proposal networks. In: NIPS, pp. 91--99

\bibitem[{Ren et~al.(2017{\natexlab{a}})Ren, He, Girshick, and Sun}]{Ren2016a}
Ren S., He K., Girshick R., Sun J. (2017{\natexlab{a}}) {Faster RCNN}: Towards
  real time object detection with region proposal networks. IEEE TPAMI
  39(6):1137--1149

\bibitem[{Ren et~al.(2017{\natexlab{b}})Ren, He, Girshick, Zhang, and
  Sun}]{Ren2016NOC}
Ren S., He K., Girshick R., Zhang X., Sun J. (2017{\natexlab{b}}) Object
  detection networks on convolutional feature maps. IEEE TPAMI

\bibitem[{Rezatofighi et~al.(2019)Rezatofighi, Tsoi, Gwak, Sadeghian, Reid, and
  Savarese}]{Rezatofighi2019Generalized}
Rezatofighi H., Tsoi N., Gwak J., Sadeghian A., Reid I., Savarese S. (2019)
  Generalized intersection over union: A metric and a loss for bounding box
  regression. In: CVPR

\bibitem[{Rowley et~al.(1998)Rowley, Baluja, and Kanade}]{Rowley1998}
Rowley H., Baluja S., Kanade T. (1998) Neural network based face detection.
  IEEE TPAMI 20(1):23--38

\bibitem[{Russakovsky et~al.(2015)Russakovsky, Deng, Su, Krause, Satheesh, Ma,
  Huang, Karpathy, Khosla, Bernstein, Berg, and Li}]{Russakovsky2015}
Russakovsky O., Deng J., Su H., Krause J., Satheesh S., Ma S., Huang Z.,
  Karpathy A., Khosla A., Bernstein M., Berg A., Li F. (2015) {ImageNet} large
  scale visual recognition challenge. IJCV 115(3):211--252

\bibitem[{Russell et~al.(2008)Russell, Torralba, Murphy, and
  Freeman}]{Russell2008}
Russell B., Torralba A., Murphy K., Freeman W. (2008) {LabelMe}: A database and
  web based tool for image annotation. IJCV 77(1-3):157--173

\bibitem[{Schmid and Mohr(1997)}]{Schmid1997Local}
Schmid C., Mohr R. (1997) Local grayvalue invariants for image retrieval. IEEE
  TPAMI 19(5):530--535

\bibitem[{Schwartz et~al.(2019)Schwartz, Karlinsky, Shtok, Harary, Marder,
  Pankanti, Feris, Kumar, Giries, and Bronstein}]{Schwartz2019RepMet}
Schwartz E., Karlinsky L., Shtok J., Harary S., Marder M., Pankanti S., Feris
  R., Kumar A., Giries R., Bronstein A. (2019) {RepMet}: Representative based
  metric learning for classification and one shot object detection. In: CVPR

\bibitem[{Sermanet et~al.(2013)Sermanet, Kavukcuoglu, Chintala, and
  LeCun}]{Sermanet2013c}
Sermanet P., Kavukcuoglu K., Chintala S., LeCun Y. (2013) Pedestrian detection
  with unsupervised multistage feature learning. In: CVPR, pp. 3626--3633

\bibitem[{Sermanet et~al.(2014)Sermanet, Eigen, Zhang, Mathieu, Fergus, and
  LeCun}]{OverFeat2014}
Sermanet P., Eigen D., Zhang X., Mathieu M., Fergus R., LeCun Y. (2014)
  {OverFeat}: Integrated recognition, localization and detection using
  convolutional networks. In: ICLR

\bibitem[{Shang et~al.(2016)Shang, Sohn, Almeida, and
  Lee}]{Shang2016Understanding}
Shang W., Sohn K., Almeida D., Lee H. (2016) Understanding and improving
  convolutional neural networks via concatenated rectified linear units. In:
  ICML, pp. 2217--2225

\bibitem[{Shelhamer et~al.(2017)Shelhamer, Long, and Darrell}]{FCNTPAMI}
Shelhamer E., Long J., Darrell T. (2017) Fully convolutional networks for
  semantic segmentation. IEEE TPAMI

\bibitem[{Shen et~al.(2017)Shen, Liu, Li, Jiang, Chen, and Xue}]{ShenICCV2017}
Shen Z., Liu Z., Li J., Jiang Y., Chen Y., Xue X. (2017) {DSOD}: Learning
  deeply supervised object detectors from scratch. In: ICCV

\bibitem[{Shi et~al.(2018)Shi, Shan, Kan, Wu, and Chen}]{Shi2018Real}
Shi X., Shan S., Kan M., Wu S., Chen X. (2018) Real time rotation invariant
  face detection with progressive calibration networks. In: CVPR

\bibitem[{Shi et~al.(2017)Shi, Yang, Hospedales, and Xiang}]{Shi2017PAMI}
Shi Z., Yang Y., Hospedales T., Xiang T. (2017) Weakly supervised image
  annotation and segmentation with objects and attributes. IEEE TPAMI
  39(12):2525--2538

\bibitem[{Shrivastava and Gupta(2016)}]{Shrivastava2016}
Shrivastava A., Gupta A. (2016) Contextual priming and feedback for {Faster
  RCNN}. In: ECCV, pp. 330--348

\bibitem[{Shrivastava et~al.(2016)Shrivastava, Gupta, and
  Girshick}]{Shrivastava2016OHEM}
Shrivastava A., Gupta A., Girshick R. (2016) Training region based object
  detectors with online hard example mining. In: CVPR, pp. 761--769

\bibitem[{Shrivastava et~al.(2017)Shrivastava, Sukthankar, Malik, and
  Gupta}]{Shrivastava2017}
Shrivastava A., Sukthankar R., Malik J., Gupta A. (2017) Beyond skip
  connections: Top down modulation for object detection. In: CVPR

\bibitem[{Simonyan and Zisserman(2015)}]{Simonyan2014VGG}
Simonyan K., Zisserman A. (2015) Very deep convolutional networks for large
  scale image recognition. In: ICLR

\bibitem[{Singh and Davis(2018)}]{Singh2018SNIP}
Singh B., Davis L. (2018) An analysis of scale invariance in object
  detection-{SNIP}. In: CVPR

\bibitem[{Singh et~al.(2018{\natexlab{a}})Singh, Li, Sharma, and
  Davis}]{Singh2018RFCN}
Singh B., Li H., Sharma A., Davis L.~S. (2018{\natexlab{a}}) {RFCN} 3000 at
  30fps: Decoupling detection and classification. In: CVPR

\bibitem[{Singh et~al.(2018{\natexlab{b}})Singh, Najibi, and
  Davis}]{Singh2018sniper}
Singh B., Najibi M., Davis L.~S. (2018{\natexlab{b}}) {SNIPER}: Efficient
  multiscale training. arXiv:180509300

\bibitem[{Sivic and Zisserman(2003)}]{Sivic2003}
Sivic J., Zisserman A. (2003) Video google: A text retrieval approach to object
  matching in videos. In: International Conference on Computer Vision (ICCV),
  vol~2, pp. 1470--1477

\bibitem[{Song~Han(2016)}]{Han2016Deep}
Song~Han W. J.~D. Huizi~Mao (2016) {Deep Compression}: Compressing deep neural
  networks with pruning, trained quantization and huffman coding. In: ICLR

\bibitem[{Sun et~al.(2017)Sun, Shrivastava, Singh, and
  Gupta}]{Sun2017Revisiting}
Sun C., Shrivastava A., Singh S., Gupta A. (2017) Revisiting unreasonable
  effectiveness of data in deep learning era. In: ICCV, pp. 843--852

\bibitem[{Sun et~al.(2019{\natexlab{a}})Sun, Xiao, Liu, and Wang}]{Sun2019Deep}
Sun K., Xiao B., Liu D., Wang J. (2019{\natexlab{a}}) Deep high resolution
  representation learning for human pose estimation. In: CVPR

\bibitem[{Sun et~al.(2019{\natexlab{b}})Sun, Zhao, Jiang, Cheng, Xiao, Liu, Mu,
  Wang, Liu, and Wang}]{Sun2019High}
Sun K., Zhao Y., Jiang B., Cheng T., Xiao B., Liu D., Mu Y., Wang X., Liu W.,
  Wang J. (2019{\natexlab{b}}) High resolution representations for labeling
  pixels and regions. CoRR abs/1904.04514

\bibitem[{Sun et~al.(2018)Sun, Pang, Shi, Yi, and Ouyang}]{Sun2018Fishnet}
Sun S., Pang J., Shi J., Yi S., Ouyang W. (2018) {FishNet}: A versatile
  backbone for image, region, and pixel level prediction. In: NIPS, pp.
  754--764

\bibitem[{Sun et~al.(2006)Sun, Bebis, and Miller}]{Sun2006Road}
Sun Z., Bebis G., Miller R. (2006) On road vehicle detection: A review. IEEE
  TPAMI 28(5):694--711

\bibitem[{Sung et~al.(1994)Sung, , and Poggio}]{Sung1996Learning}
Sung K., , Poggio T. (1994) Learning and example selection for object and
  pattern detection. MIT AI Memo (1521)

\bibitem[{Swain and Ballard(1991)}]{Swain1991Color}
Swain M., Ballard D. (1991) Color indexing. IJCV 7(1):11--32

\bibitem[{Szegedy et~al.(2013)Szegedy, Toshev, and Erhan}]{Szegedy2013Deep}
Szegedy C., Toshev A., Erhan D. (2013) Deep neural networks for object
  detection. In: NIPS, pp. 2553--2561

\bibitem[{Szegedy et~al.(2014)Szegedy, Reed, Erhan, Anguelov, and
  Ioffe}]{MultiBox2}
Szegedy C., Reed S., Erhan D., Anguelov D., Ioffe S. (2014) Scalable, high
  quality object detection. In: arXiv preprint arXiv:1412.1441

\bibitem[{Szegedy et~al.(2015)Szegedy, Liu, Jia, Sermanet, Reed, Anguelov,
  Erhan, Vanhoucke, and Rabinovich}]{GoogLeNet2015}
Szegedy C., Liu W., Jia Y., Sermanet P., Reed S., Anguelov D., Erhan D.,
  Vanhoucke V., Rabinovich A. (2015) Going deeper with convolutions. In: CVPR,
  pp. 1--9

\bibitem[{Szegedy et~al.(2016)Szegedy, Vanhoucke, Ioffe, Shlens, and
  Wojna}]{Szegedy2016a}
Szegedy C., Vanhoucke V., Ioffe S., Shlens J., Wojna Z. (2016) Rethinking the
  inception architecture for computer vision. In: CVPR, pp. 2818--2826

\bibitem[{Szegedy et~al.(2017)Szegedy, Ioffe, Vanhoucke, and
  Alemi}]{InceptionV4}
Szegedy C., Ioffe S., Vanhoucke V., Alemi A. (2017) Inception v4, inception
  resnet and the impact of residual connections on learning. AAAI pp.
  4278--4284

\bibitem[{Torralba(2003)}]{Torralba2003}
Torralba A. (2003) Contextual priming for object detection. IJCV 53(2):169--191

\bibitem[{Turk and Pentland(1991)}]{Turk1991Face}
Turk M.~A., Pentland A. (1991) Face recognition using eigenfaces. In: CVPR, pp.
  586--591

\bibitem[{Tuzel et~al.(2006)Tuzel, Porikli, and Meer}]{Tuzel2006Region}
Tuzel O., Porikli F., Meer P. (2006) Region covariance: A fast descriptor for
  detection and classification. In: ECCV, pp. 589--600

\bibitem[{{TychsenSmith} and Petersson(2017)}]{SmithICCV2017}
{TychsenSmith} L., Petersson L. (2017) {DeNet:} scalable real time object
  detection with directed sparse sampling. In: ICCV

\bibitem[{TychsenSmith and Petersson(2018)}]{Smith2018Improving}
TychsenSmith L., Petersson L. (2018) Improving object localization with fitness
  nms and bounded iou loss. In: CVPR

\bibitem[{Uijlings et~al.(2013)Uijlings, van~de Sande, Gevers, and
  Smeulders}]{Uijlings2013b}
Uijlings J., van~de Sande K., Gevers T., Smeulders A. (2013) Selective search
  for object recognition. IJCV 104(2):154--171

\bibitem[{Vaillant et~al.(1994)Vaillant, Monrocq, and LeCun}]{Vaillant1994}
Vaillant R., Monrocq C., LeCun Y. (1994) Original approach for the localisation
  of objects in images. IEE Proceedings Vision, Image and Signal Processing
  141(4):245--250

\bibitem[{{Van de Sande} et~al.(2011){Van de Sande}, Uijlings, Gevers, and
  Smeulders}]{Van2011SS}
{Van de Sande} K., Uijlings J., Gevers T., Smeulders A. (2011) Segmentation as
  selective search for object recognition. In: ICCV, pp. 1879--1886

\bibitem[{Vaswani et~al.(2017)Vaswani, Shazeer, Parmar, Uszkoreit, Jones,
  Gomez, Kaiser, and Polosukhin}]{Vaswani2017Attention}
Vaswani A., Shazeer N., Parmar N., Uszkoreit J., Jones L., Gomez A.~N., Kaiser
  {\L}., Polosukhin I. (2017) Attention is all you need. In: NIPS, pp.
  6000--6010

\bibitem[{Vedaldi et~al.(2009)Vedaldi, Gulshan, Varma, and
  Zisserman}]{Vedaldi09Multiple}
Vedaldi A., Gulshan V., Varma M., Zisserman A. (2009) Multiple kernels for
  object detection. In: ICCV, pp. 606--613

\bibitem[{Viola and Jones(2001)}]{Viola2001}
Viola P., Jones M. (2001) Rapid object detection using a boosted cascade of
  simple features. In: CVPR, vol~1, pp. 1--8

\bibitem[{Wan et~al.(2015)Wan, Eigen, and Fergus}]{Wan2015end}
Wan L., Eigen D., Fergus R. (2015) End to end integration of a convolution
  network, deformable parts model and nonmaximum suppression. In: CVPR, pp.
  851--859

\bibitem[{Wang et~al.(2018)Wang, Wang, Gao, Li, and Zuo}]{Wang2018Multiscale}
Wang H., Wang Q., Gao M., Li P., Zuo W. (2018) Multiscale location aware kernel
  representation for object detection. In: CVPR

\bibitem[{Wang et~al.(2009)Wang, Han, and Yan}]{HOGLBP2009}
Wang X., Han T., Yan S. (2009) An {HOG-LBP} human detector with partial
  occlusion handling. In: International Conference on Computer Vision, pp.
  32--39

\bibitem[{Wang et~al.(2017)Wang, Shrivastava, and Gupta}]{Wang2017}
Wang X., Shrivastava A., Gupta A. (2017) {A Fast RCNN}: Hard positive
  generation via adversary for object detection. In: CVPR

\bibitem[{Wang et~al.(2019)Wang, Cai, Gao, and Vasconcelos}]{Wang2019Towards}
Wang X., Cai Z., Gao D., Vasconcelos N. (2019) Towards universal object
  detection by domain attention. arXiv:190404402

\bibitem[{Wei et~al.(2018)Wei, Pan, Qin, Ouyang, and Yan}]{Wei2018Quantization}
Wei Y., Pan X., Qin H., Ouyang W., Yan J. (2018) Quantization mimic: Towards
  very tiny {CNN} for object detection. In: ECCV, pp. 267--283

\bibitem[{Woo et~al.(2018)Woo, Hwang, and Kweon}]{Woo18StairNet}
Woo S., Hwang S., Kweon I. (2018) {StairNet}: Top down semantic aggregation for
  accurate one shot detection. In: WACV, pp. 1093--1102

\bibitem[{Worrall et~al.(2017)Worrall, Garbin, Turmukhambetov, and
  Brostow}]{Worrall2017Harmonic}
Worrall D.~E., Garbin S.~J., Turmukhambetov D., Brostow G.~J. (2017) Harmonic
  networks: Deep translation and rotation equivariance. In: CVPR, vol~2

\bibitem[{Wu and He(2018)}]{Wu2018Group}
Wu Y., He K. (2018) Group normalization. In: ECCV, pp. 3--19

\bibitem[{Wu et~al.(2015)Wu, Song, Khosla, Yu, Zhang, Tang, and
  Xiao}]{Wu20153D}
Wu Z., Song S., Khosla A., Yu F., Zhang L., Tang X., Xiao J. (2015) {3D
  ShapeNets}: A deep representation for volumetric shapes. In: CVPR, pp.
  1912--1920

\bibitem[{Wu et~al.(2019)Wu, Pan, Chen, Long, Zhang, and
  Yu}]{Wu2019Comprehensive}
Wu Z., Pan S., Chen F., Long G., Zhang C., Yu P.~S. (2019) A comprehensive
  survey on graph neural networks. arXiv preprint arXiv:190100596

\bibitem[{Xia et~al.(2018)Xia, Bai, Ding, Zhu, Belongie, Luo, Datcu, Pelillo,
  and Zhang}]{Xia2018DOTA}
Xia G., Bai X., Ding J., Zhu Z., Belongie S., Luo J., Datcu M., Pelillo M.,
  Zhang L. (2018) {DOTA:} a large-scale dataset for object detection in aerial
  images. In: CVPR, pp. 3974--3983

\bibitem[{Xiang et~al.(2014)Xiang, Mottaghi, and Savarese}]{Xiang2014Beyond}
Xiang Y., Mottaghi R., Savarese S. (2014) {Beyond PASCAL}: A benchmark for {3D}
  object detection in the wild. In: WACV, pp. 75--82

\bibitem[{Xiao et~al.(2003)Xiao, Zhu, and Zhang}]{Xiao2003Boosting}
Xiao R., Zhu L., Zhang H. (2003) Boosting chain learning for object detection.
  In: ICCV, pp. 709--715

\bibitem[{Xie et~al.(2017)Xie, Girshick, Doll{\'a}r, Tu, and
  He}]{Xie2016Aggregated}
Xie S., Girshick R., Doll{\'a}r P., Tu Z., He K. (2017) Aggregated residual
  transformations for deep neural networks. In: CVPR

\bibitem[{Yang et~al.(2016{\natexlab{a}})Yang, Yan, Lei, and Li}]{CRAFT2016}
Yang B., Yan J., Lei Z., Li S. (2016{\natexlab{a}}) {CRAFT} objects from
  images. In: CVPR, pp. 6043--6051

\bibitem[{Yang et~al.(2016{\natexlab{b}})Yang, Choi, and Lin}]{Yang2016Exploit}
Yang F., Choi W., Lin Y. (2016{\natexlab{b}}) Exploit all the layers: Fast and
  accurate {CNN} object detector with scale dependent pooling and cascaded
  rejection classifiers. In: CVPR, pp. 2129--2137

\bibitem[{Yang et~al.(2002)Yang, Kriegman, and Ahuja}]{Yang2002b}
Yang M., Kriegman D., Ahuja N. (2002) Detecting faces in images: A survey. IEEE
  TPAMI 24(1):34--58

\bibitem[{Ye and Doermann(2015)}]{Ye2015Text}
Ye Q., Doermann D. (2015) Text detection and recognition in imagery: A survey.
  IEEE TPAMI 37(7):1480--1500

\bibitem[{Yosinski et~al.(2014)Yosinski, Clune, Bengio, and
  Lipson}]{Yosinski2014Transferable}
Yosinski J., Clune J., Bengio Y., Lipson H. (2014) How transferable are
  features in deep neural networks? In: NIPS, pp. 3320--3328

\bibitem[{{Young} et~al.(2018){Young}, {Hazarika}, {Poria}, and
  {Cambria}}]{Young2018Recent}
{Young} T., {Hazarika} D., {Poria} S., {Cambria} E. (2018) Recent trends in
  deep learning based natural language processing. IEEE Computational
  Intelligence Magazine 13(3):55--75

\bibitem[{Yu and Koltun(2016)}]{Yu2015Multiscale}
Yu F., Koltun V. (2016) Multiscale context aggregation by dilated convolutions

\bibitem[{Yu et~al.(2017)Yu, Koltun, and Funkhouser}]{Yu2017Dilated}
Yu F., Koltun V., Funkhouser T. (2017) Dilated residual networks. In: CVPR,
  vol~2, p.~3

\bibitem[{Yu et~al.(2018)Yu, Li, Chen, Lai, and {et al.}}]{Yu2017NISP}
Yu R., Li A., Chen C., Lai J., {et al.} (2018) {NISP}: Pruning networks using
  neuron importance score propagation. CVPR

\bibitem[{Zafeiriou et~al.(2015)Zafeiriou, Zhang, and Zhang}]{Zafeiriou2015}
Zafeiriou S., Zhang C., Zhang Z. (2015) A survey on face detection in the wild:
  Past, present and future. Computer Vision and Image Understanding 138:1--24

\bibitem[{Zagoruyko et~al.(2016)Zagoruyko, Lerer, Lin, Pinheiro, Gross,
  Chintala, and Doll{\'a}r}]{Zagoruyko2016}
Zagoruyko S., Lerer A., Lin T., Pinheiro P., Gross S., Chintala S., Doll{\'a}r
  P. (2016) A multipath network for object detection. In: BMVC

\bibitem[{Zeiler and Fergus(2014)}]{ZeilerFergus2014}
Zeiler M., Fergus R. (2014) Visualizing and understanding convolutional
  networks. In: ECCV, pp. 818--833

\bibitem[{Zeng et~al.(2016)Zeng, Ouyang, Yang, Yan, and Wang}]{GBDCNN2016}
Zeng X., Ouyang W., Yang B., Yan J., Wang X. (2016) Gated bidirectional cnn for
  object detection. In: ECCV, pp. 354--369

\bibitem[{Zeng et~al.(2017)Zeng, Ouyang, Yan, Li, Xiao, Wang, Liu, Zhou, Yang,
  Wang, Zhou, and Wang}]{Zeng2017Crafting}
Zeng X., Ouyang W., Yan J., Li H., Xiao T., Wang K., Liu Y., Zhou Y., Yang B.,
  Wang Z., Zhou H., Wang X. (2017) Crafting gbdnet for object detection. IEEE
  TPAMI

\bibitem[{Zhang et~al.(2016{\natexlab{a}})Zhang, Zhang, Li, and
  Qiao}]{Zhang2016Joint}
Zhang K., Zhang Z., Li Z., Qiao Y. (2016{\natexlab{a}}) Joint face detection
  and alignment using multitask cascaded convolutional networks. IEEE SPL
  23(10):1499--1503

\bibitem[{Zhang et~al.(2016{\natexlab{b}})Zhang, Lin, Liang, and
  He}]{Zhang2016faster}
Zhang L., Lin L., Liang X., He K. (2016{\natexlab{b}}) Is faster {RCNN} doing
  well for pedestrian detection? In: ECCV, pp. 443--457

\bibitem[{Zhang et~al.(2018{\natexlab{a}})Zhang, Wen, Bian, Lei, and
  Li}]{Zhang2018Single}
Zhang S., Wen L., Bian X., Lei Z., Li S. (2018{\natexlab{a}}) Single shot
  refinement neural network for object detection. In: CVPR

\bibitem[{Zhang et~al.(2018{\natexlab{b}})Zhang, Yang, and
  Schiele}]{Zhang2018Occluded}
Zhang S., Yang J., Schiele B. (2018{\natexlab{b}}) Occluded pedestrian
  detection through guided attention in {CNNs}. In: CVPR, pp. 2056--2063

\bibitem[{Zhang et~al.(2013)Zhang, Yang, Han, Wang, and Gao}]{Zhang13}
Zhang X., Yang Y., Han Z., Wang H., Gao C. (2013) Object class detection: A
  survey. ACM Computing Surveys 46(1):10:1--10:53

\bibitem[{Zhang et~al.(2017)Zhang, Li, Change~Loy, and Lin}]{Zhang2017PolyNet}
Zhang X., Li Z., Change~Loy C., Lin D. (2017) {PolyNet:} a pursuit of
  structural diversity in very deep networks. In: CVPR, pp. 718--726

\bibitem[{Zhang et~al.(2018{\natexlab{c}})Zhang, Zhou, Lin, and
  Sun}]{Zhang18ShuffleNet}
Zhang X., Zhou X., Lin M., Sun J. (2018{\natexlab{c}}) {ShuffleNet:} an
  extremely efficient convolutional neural network for mobile devices. In: CVPR

\bibitem[{Zhang et~al.(2018{\natexlab{d}})Zhang, Geiger, Pohjalainen, Mousa,
  Jin, and Schuller}]{Zhang2018Deep}
Zhang Z., Geiger J., Pohjalainen J., Mousa A.~E., Jin W., Schuller B.
  (2018{\natexlab{d}}) Deep learning for environmentally robust speech
  recognition: An overview of recent developments. ACM Trans Intell Syst
  Technol 9(5):49:1--49:28

\bibitem[{Zhang et~al.(2018{\natexlab{e}})Zhang, Qiao, Xie, Shen, Wang, and
  Yuille}]{Zhang2018Object}
Zhang Z., Qiao S., Xie C., Shen W., Wang B., Yuille A. (2018{\natexlab{e}})
  Single shot object detection with enriched semantics. In: CVPR

\bibitem[{Zhao et~al.(2019)Zhao, Sheng, Wang, Tang, Chen, Cai, and
  Ling}]{Zhao2019M2Det}
Zhao Q., Sheng T., Wang Y., Tang Z., Chen Y., Cai L., Ling H. (2019) {M2Det}: A
  single shot object detector based on multilevel feature pyramid network. In:
  AAAI

\bibitem[{Zheng et~al.(2015)Zheng, Jayasumana, {Romera-Paredes}, Vineet, Su,
  Du, Huang, and Torr}]{Zheng15Conditional}
Zheng S., Jayasumana S., {Romera-Paredes} B., Vineet V., Su Z., Du D., Huang
  C., Torr P. (2015) Conditional random fields as recurrent neural networks.
  In: ICCV, pp. 1529--1537

\bibitem[{Zhou et~al.(2015)Zhou, Khosla, Lapedriza, Oliva, and
  Torralba}]{Zhoubolei2014}
Zhou B., Khosla A., Lapedriza A., Oliva A., Torralba A. (2015) Object detectors
  emerge in deep scene {CNNs}. In: ICLR

\bibitem[{Zhou et~al.(2016{\natexlab{a}})Zhou, Khosla, Lapedriza, Oliva, and
  Torralba}]{Zhou2016learning}
Zhou B., Khosla A., Lapedriza A., Oliva A., Torralba A. (2016{\natexlab{a}})
  Learning deep features for discriminative localization. In: CVPR, pp.
  2921--2929

\bibitem[{Zhou et~al.(2017{\natexlab{a}})Zhou, Lapedriza, Khosla, Oliva, and
  Torralba}]{Zhou2017Places}
Zhou B., Lapedriza A., Khosla A., Oliva A., Torralba A. (2017{\natexlab{a}})
  Places: A 10 million image database for scene recognition. IEEE Trans Pattern
  Analysis and Machine Intelligence

\bibitem[{Zhou et~al.(2018{\natexlab{a}})Zhou, Cui, Zhang, Yang, Liu, and
  Sun}]{Zhou2018Graph}
Zhou J., Cui G., Zhang Z., Yang C., Liu Z., Sun M. (2018{\natexlab{a}}) Graph
  neural networks: A review of methods and applications. arXiv preprint
  arXiv:181208434

\bibitem[{Zhou et~al.(2018{\natexlab{b}})Zhou, Ni, Geng, Hu, and
  Xu}]{Zhou2018Scale}
Zhou P., Ni B., Geng C., Hu J., Xu Y. (2018{\natexlab{b}}) Scale transferrable
  object detection. In: CVPR

\bibitem[{Zhou et~al.(2016{\natexlab{b}})Zhou, Liu, Shao, and
  Mellor}]{Zhou2016dave}
Zhou Y., Liu L., Shao L., Mellor M. (2016{\natexlab{b}}) {DAVE}: A unified
  framework for fast vehicle detection and annotation. In: ECCV, pp. 278--293

\bibitem[{Zhou et~al.(2017{\natexlab{b}})Zhou, Ye, Qiu, and
  Jiao}]{Zhou2017Oriented}
Zhou Y., Ye Q., Qiu Q., Jiao J. (2017{\natexlab{b}}) Oriented response
  networks. In: CVPR, pp. 4961--4970

\bibitem[{Zhu et~al.(2016{\natexlab{a}})Zhu, Vondrick, Fowlkes, and
  Ramanan}]{Zhu2016Do}
Zhu X., Vondrick C., Fowlkes C., Ramanan D. (2016{\natexlab{a}}) Do we need
  more training data? IJCV 119(1):76--92

\bibitem[{{Zhu} et~al.(2017){Zhu}, {Tuia}, {Mou}, {Xia}, {Zhang}, {Xu}, and
  {Fraundorfer}}]{Zhu2017Deep}
{Zhu} X., {Tuia} D., {Mou} L., {Xia} G., {Zhang} L., {Xu} F., {Fraundorfer} F.
  (2017) Deep learning in remote sensing: A comprehensive review and list of
  resources. IEEE Geoscience and Remote Sensing Magazine 5(4):8--36

\bibitem[{Zhu et~al.(2015)Zhu, Urtasun, Salakhutdinov, and
  Fidler}]{SegDeepM2015}
Zhu Y., Urtasun R., Salakhutdinov R., Fidler S. (2015) {SegDeepM}: Exploiting
  segmentation and context in deep neural networks for object detection. In:
  CVPR, pp. 4703--4711

\bibitem[{Zhu et~al.(2017{\natexlab{a}})Zhu, Zhao, Wang, Zhao, Wu, and
  Lu}]{ZhuICCV2017}
Zhu Y., Zhao C., Wang J., Zhao X., Wu Y., Lu H. (2017{\natexlab{a}})
  {CoupleNet}: Coupling global structure with local parts for object detection.
  In: ICCV

\bibitem[{Zhu et~al.(2017{\natexlab{b}})Zhu, Zhou, Ye, Qiu, and
  Jiao}]{Xhu2017Soft}
Zhu Y., Zhou Y., Ye Q., Qiu Q., Jiao J. (2017{\natexlab{b}}) Soft proposal
  networks for weakly supervised object localization. In: ICCV, pp. 1841--1850

\bibitem[{Zhu et~al.(2016{\natexlab{b}})Zhu, Liang, Zhang, Huang, Li, and
  Hu}]{Zhu2016traffic}
Zhu Z., Liang D., Zhang S., Huang X., Li B., Hu S. (2016{\natexlab{b}}) Traffic
  sign detection and classification in the wild. In: CVPR, pp. 2110--2118

\bibitem[{Zitnick and Doll{\'a}r(2014)}]{EdgeBox2014}
Zitnick C., Doll{\'a}r P. (2014) Edge boxes: Locating object proposals from
  edges. In: ECCV, pp. 391--405

\bibitem[{Zoph and Le(2017)}]{Zoph2016Neural}
Zoph B., Le Q. (2017) Neural architecture search with reinforcement learning

\bibitem[{Zoph et~al.(2018)Zoph, Vasudevan, Shlens, and Le}]{Zoph2018Learning}
Zoph B., Vasudevan V., Shlens J., Le Q. (2018) Learning transferable
  architectures for scalable image recognition. In: CVPR, pp. 8697--8710

\end{thebibliography}

\end{document}